\RequirePackage{fix-cm}
\documentclass[twocolumn]{svjour3}          %
\smartqed  %

\usepackage[hyphens]{url} %
\usepackage{graphicx}
\usepackage{subfig}
\usepackage{dblfloatfix} %
\usepackage[export]{adjustbox} %
\usepackage{natbib}  %
\usepackage{amssymb}
\usepackage{booktabs}  %
\usepackage{multirow}
\usepackage{hyperref}
\usepackage{xcolor,colortbl} %

\definecolor{colour1}{rgb}{1.0000, 0.3000, 0.3000}
\definecolor{colour2}{rgb}{0.4759, 0.8316, 0.0562}
\definecolor{colour3}{rgb}{0.2668, 0.6584, 0.9232}
\definecolor{colour4}{rgb}{1.0000, 0.6510, 0.3000}
\definecolor{colour5}{rgb}{0.8307, 0.4791, 0.9023}
\journalname{International Journal of Computer Vision}

\begin{document}

\title{A Comprehensive Performance Evaluation of Deformable Face Tracking ``In-the-Wild'' %
}

\titlerunning{Deformable Face Tracking}        %

\author{Grigorios G. Chrysos \and Epameinondas Antonakos$^\ast$ \and Patrick Snape$^\ast$ \and Akshay Asthana \and Stefanos Zafeiriou
}

\institute{%
$^\ast$ E. Antonakos and P. Snape contributed equally and have joint second authorship.\\\\
G. Chrysos $\cdot$ E. Antonakos $\cdot$ P. Snape $\cdot$ S. Zafeiriou \at
Department of Computing, Imperial College London, 180 Queen's Gate, London SW7 2AZ, UK \\
\email{\{g.chrysos, e.antonakos\}@imperial.ac.uk} \\
\hspace*{1.14cm} \{p.snape, s.zafeiriou\}@imperial.ac.uk\\\\
A. Asthana \at
Seeing Machines Ltd., Level 1, 11 Lonsdale St, Braddon, ACT, Australia, 2612 \\
\email{a.asthana@seeingmachines.com}
}

\date{Received: date / Accepted: date}

\maketitle

\begin{abstract}
Recently, technologies such as face detection, facial landmark localisation and face recognition and verification have matured enough to provide effective and efficient solutions for imagery captured under arbitrary conditions (referred to as ``in-the-wild''). This is partially attributed to the fact that comprehensive ``in-the-wild'' benchmarks have been developed for face detection, landmark localisation and recognition/verification. A very important technology that has not been thoroughly evaluated yet is deformable face tracking ``in-the-wild''. Until now, the performance has mainly been assessed qualitatively by visually assessing the result of a deformable face tracking technology on short videos. In this paper, we perform the first, to the best of our knowledge, thorough evaluation of state-of-the-art deformable face tracking pipelines using the recently introduced 300VW benchmark. We evaluate many different architectures focusing mainly on the task of on-line deformable face tracking. In particular, we compare the following general strategies: (a)~generic face detection plus generic facial landmark localisation, (b)~generic model free tracking plus generic facial landmark localisation, as well as (c)~hybrid approaches using state-of-the-art face detection, model free tracking and facial landmark localisation technologies. Our evaluation reveals future avenues for further research on the topic.


\end{abstract}

\section{Introduction}\label{sec:intro}
[\textcolor{red}{\textbf{PLEASE NOTE THAT THIS MANUSCRIPT HAS BEEN ACCEPTED BY IJCV. THE LATEST MANUSCRIPT CAN BE FOUND IN IBUG SITE (\url{https://ibug.doc.ic.ac.uk/media/uploads/documents/ijcv_deformable_tracking_review.pdf}) OR IN THE SPRINGER SITE/AUTHORS' SITES.}}]
The human face is arguably among the most well-studied deformable objects in the field of Computer Vision. This is due to the many roles it has in numerous applications. For example, accurate detection of faces is an essential step for tasks such as controller-free gaming, surveillance, digital photo album organization, image tagging, etc. Additionally, detection of facial features plays a crucial role for facial behaviour analysis, facial attributes analysis (e.g., gender and age recognition, etc.), facial image editing (e.g., digital make-up, etc.), surveillance, sign language recognition, lip reading, human-computer and human-robot interaction.

Due to the above applications, current research has been monopolised by the tasks of \emph{face detection}, \emph{facial landmark localisation} and \emph{face recognition or verification}. 
Firstly, face detection, despite having permeated many forms of modern technology such as digital cameras and social networking, is still a challenging problem and a popular line of research, as shown by the recent surveys of \cite{fddbTech,zhang2010survey,zafeiriou2015survey}. Although face detection on well-lit frontal facial images can be performed reliably on an embedded device, face detection on arbitrary images of people is still extremely challenging (\cite{fddbTech}). Images of faces under these unconstrained conditions are commonly referred to as ``in-the-wild'' and may include scenarios such as extreme facial pose, defocus, faces occupying a very small number of pixels or occlusions. Given the fact that face detection is still regarded as a challenging task, many generic object detection architectures such as \cite{yan2014fastest,king2015max} are either directly assessed on in-the-wild facial data, or are appropriately modified in order to explicitly perform face detection as done by \cite{zhu2012face,felzenszwalb2005pictorial}. The interested reader may refer to the most recent survey by \cite{zafeiriou2015survey} for more information on in-the-wild face detection.
The problem of localising facial landmarks that correspond to fiducial facial parts (e.g., eyes, mouth, etc.) is still extremely challenging and has only been possible to perform reliably relatively recently. Although the history of facial landmark localisation spans back many decades (\cite{cootes1995active,cootes2001active}), the ability to accurately recover facial landmarks on in-the-wild images has only become possible in recent years (\cite{matthews2004active,papandreou2008adaptive,saragih2011deformable,cao2014face}). Much of this progress can be attributed to the release of large annotated datasets of facial landmarks (\cite{sagonas2013semi,sagonas2013300,zhu2012face,le2012interactive,belhumeur2013localizing,kostinger2011annotated}) and very recently the area of facial landmark localisation has become extremely competitive with recent works including \cite{xiong2013supervised,ren2014face,kazemi2014one,zhu2015face,tzimiropoulos2015project}. For a recent evaluation of facial landmark localisation methods the interested reader may refer to the survey by \cite{wang2014facial} and to the results of the 300W competition by \cite{sagonas2015300}.
Finally, face recognition and verification are extremely popular lines of research. For the past two decades, the majority of statistical machine learning algorithms spanning from linear/non-linear subspace learning techniques (\cite{de2012least,kokiopoulou2011trace}) to Deep Convolutional Neural Networks (DCNNs) (\cite{taigman2014deepface,schroff2015facenet,parkhi2015deep}) have been applied to the problem of face recognition and verification. Recently, due to the revival of DCNNs, as well as the development of Graphics Processing Units (GPUs), remarkable face verification performance has been reported (\cite{taigman2014deepface}). The interested reader may refer to the recent survey by \cite{labeled2016learned} as well as the most popular benchmark for face verification in-the-wild in \cite{LFWTech}.

In all of the aforementioned fields, significant progress has been reported in recent years. The primary reasons behind these advances are:
\begin{itemize}
    \item \emph{The collection and annotation of large databases.} Given the abundance of facial images available primarily through the Internet via services such as Flickr, Google Images and Facebook, the collection of facial images is extremely simple. Some examples of large databases for face detection are FDDB (\cite{fddbTech}), AFW (\cite{zhu2012face}) and LFW (\cite{LFWTech}). Similar large-scale databases for facial landmark localisation include 300W (\cite{sagonas2013semi}) LFPW (\cite{belhumeur2013localizing}), AFLW (\cite{kostinger2011annotated}) and HELEN (\cite{le2012interactive}). Similarly, for face recognition there exists LFW (\cite{LFWTech}), FRVT (\cite{phillips2000feret}) and the recently introduced Janus database (IJB-A) (\cite{klare2015pushing}).
    \item \emph{The establishment of in-the-wild benchmarks and challenges} that provide a fair comparison between state of the art techniques. FDDB (\cite{fddbTech}), 300W (\cite{sagonas2013300,sagonas2015300}) and Janus (\cite{klare2015pushing}) are the most characteristic examples for face detection, facial landmark localisation and face recognition, respectively.
\end{itemize}

Contrary to face detection, facial landmark localisation and face recognition, the problem of \emph{deformable face tracking} across long-term sequences has yet to attract much attention, despite its crucial role in numerous applications. Given the fact that cameras are embedded in many common electronic devices, it is surprising that current research has not yet focused towards providing robust and accurate solutions for long-term deformable tracking. Almost all face-based applications, including facial behaviour analysis, lip reading, surveillance, human-computer and human-robot interaction etc., require accurate \emph{continuous tracking} of the facial landmarks. The facial landmarks are commonly used as input signals of higher-level methodologies to compute motion dynamics and deformations.
The performance of currently available technologies for facial deformable tracking has not been properly assessed (\cite{yacoob1996recognizing,essa1996modeling,essa1997coding,decarlo2000optical,koelstra2010dynamic,snape2015face}). This is attributed to the fact that, until recently, there was no established benchmark for the task. At ICCV 2015, the first benchmark for facial landmark tracking (so-called 300VW) was presented by \cite{shen2015first}, providing a large number of annotated videos captured in-the-wild
\footnote{\label{300VW_foot}The results and dataset of the 300VW Challenge by \cite{shen2015first} can be found at \url{http://ibug.doc.ic.ac.uk/resources/300-VW/}. This is the first facial landmark tracking challenge on challenging long-term sequences.}.
In particular, the benchmark provides 114 videos with average duration around 1 minute, split into three categories of increasing difficulty. The frames of all videos (218595 in total) were annotated by applying semi-automatic procedures, as shown in \cite{chrysos2015offline}. Five different facial tracking methodologies were evaluated in the benchmark (\cite{rajamanoharan2015multi,yang2015facial,yue2015shape,uricar2015real,xiao2015facial}) and the results are indicative of the current state-of-the-art performance.

In this paper, we make a significant step further and present the first, to the best of our knowledge, comprehensive evaluation of multiple deformable face tracking pipelines. In particular, we assess:
\begin{itemize}
    \item A pipeline which combines a generic face detection algorithm with a facial landmark localisation method. This is the most common method for facial landmark tracking. It is fairly robust since the probability of drifting is reduced due to the application of the face detector at each frame. Nevertheless, it does not exploit the dynamic characteristics of the tracked face. Many state-of-the-art face detectors as well as facial landmark localisation methodologies are evaluated in this pipeline.
    \item A pipeline which combines a model free tracking system with a facial landmark localisation method. This approach takes into account the dynamic nature of the tracked face, but is susceptible to drifting and thus losing the tracked object. We evaluate the combinations of multiple state-of-the-art model free trackers, as well as landmark localisation techniques.
    \item Hybrid pipelines that include mechanisms for detecting tracking failures and performing re-initialisation, as well as using models for ensuring robust tracking.
\end{itemize}

Summarising, the findings of our evaluation show that current face detection and model free tracking technologies are advanced enough so that even a naive combination with landmark localisation techniques is adequate to achieve state-of-the-art performance on deformable face tracking. Specifically, we experimentally show that model free tracking based pipelines are very accurate when applied on videos with moderate lighting and pose circumstances. Furthermore, the combination of state-of-the-art face detectors with landmark localisation systems demonstrates excellent performance with surprisingly high true positive rate on videos captured under arbitrary conditions (extreme lighting, pose, occlusions, etc.). Moreover, we show that hybrid approaches provide only a marginal improvement, which is not worth their complexity and computational cost. Finally, we compare these approaches with the systems that participated in the 300VW competition of \cite{shen2015first}.

The rest of the paper is organised as follows. Section~\ref{sec:related} presents a survey of the current literature on both rigid and deformable face tracking. In Section~\ref{sec:tracking}, we present the current state-of-the-art methodologies for deformable face tracking. Since, modern face tracking consists of various modules, including face detection, model free tracking and facial landmark localisation, Sections~\ref{subsec:detection},~\ref{subsec:tracking} and~\ref{subsec:alignment} briefly outline the state-of-the-art in each of these domains. Experimental results are presented in Section~\ref{sec:experiments}. Finally, in Section~\ref{sec:discussion} we discuss the challenges that still remain to be addressed, provide future research directions and draw conclusions.

\section{Related Work}\label{sec:related}
Rigid and non-rigid tracking of faces and facial features have been a very popular topic of research over the past twenty years (\cite{black1995tracking,lanitis1995unified,sobottka1996face,essa1996modeling,essa1997coding,oliver1997lafter,decarlo2000optical,jepson2003robust,matthews2004active,matthews2004template,xiao2004real,patras2004particle,kim2008face,ross2008incremental,papandreou2008adaptive,amberg2009compositional,kalal2010face,koelstra2010dynamic,tresadern2012real,tzimiropoulos2013optimization,xiong2013supervised,liwicki2013euler,smeulders2014visual,asthana2014incremental,tzimiropoulos2014gauss,li2015nus,xiong2015global,snape2015face,wu2015object,tzimiropoulos2015project}). In this section we provide an overview of face tracking spanning over the past twenty years up to the present day. In particular, we will outline the methodologies regarding rigid 2D/3D face tracking, as well as deformable 2D/3D face tracking using a monocular camera\footnote{The problem of face tracking using commodity depth cameras, which has received a lot of attention (\cite{gokturk20043d,cai20103d,weise2011realtime}), falls outside the scope of this paper.}. Finally, we outline the benchmarks for both rigid and deformable face tracking.

\subsection{Prior Art}
The first methods for rigid 2D tracking generally revolved around the use of various features or transformations and mainly explored various color-spaces for robust tracking (\cite{crowley1997multi,bradski1998real,qian1998robust,toyama1998look,jurie1999new,schwerdt2000robust,stern2002adaptive,vadakkepat2008multimodal}). The general methods of choice for tracking were Mean Shift and variations such as the Continuously Adaptive Mean Shift (Camshift) algorithm (\cite{bradski1998conputer,allen2004object}). The Mean Shift algorithm is a non-parametric technique that climbs the gradient of a probability distribution to find the nearest dominant mode (peak) (\cite{comaniciu1999mean,comaniciu2000real}). Camshift is an adaptation of the Mean Shift algorithm for object tracking. The primary difference between CamShift and Mean Shift is that the former uses continuously adaptive probability distributions (i.e., distributions that may be recomputed for each frame) while the latter is based on static distributions, which are not updated unless the target experiences significant changes in shape, size or color. Other popular methods of choice for tracking are linear and non-linear filtering techniques including Kalman filters, as well as methodologies that fall in the general category of particle filters (\cite{del1996non,gordon1993novel}), such as the popular Condensation algorithm by \cite{isard1998condensation}. Condensation is the application of Sampling Importance Resampling (SIR) estimation by \cite{gordon1993novel} to contour tracking. A recent successful 2D rigid tracker that updates the appearance model of the tracked face was proposed in \cite{ross2008incremental}. The algorithm uses incremental Principal Component Analysis (PCA) (\cite{levey2000sequential}) to learn a statistical model of the appearance in an on-line manner and contrary to other eigentrackers, such as \cite{black1998eigentracking}, it does not contain any training phase. The method in \cite{ross2008incremental} uses a variant of the Condensation algorithm to model the distribution over the object’s location as it evolves over time. The method has initiated a line of research on robust incremental object tracking including the works of \cite{liwicki2012efficient,liwicki2013euler,liwicki2012incremental,liwicki2015online}. Rigid 3D tracking has also been studied by using generic 3D models of the face (\cite{malciu2000robust,la2000fast}). For example, \cite{la2000fast} formulate the tracking task as an image registration problem in the cylindrically unwrapped texture space and \cite{sung2008pose} combine Active Appearance Models (AAMs) with a cylindrical head model for robust recovery of the global rigid motion. Currently, rigid face tracking is generally treated along the same lines as general model free object tracking (\cite{jepson2003robust,smeulders2014visual,liwicki2013euler,liwicki2012efficient,ross2008incremental,wu2015object,li2015nus}). An overview of model free object tracking is given in Section~\ref{subsec:tracking}. 

Non-rigid tracking of faces is important in many applications, spanning from facial expression analysis to motion capture for graphics and game design. Non-rigid tracking of faces can be further subdivided into tracking of certain facial landmarks (\cite{lanitis1995unified,black1995tracking,sobottka1996face,xiao2004real,matthews2004active,matthews2004template,patras2004particle,papandreou2008adaptive,amberg2009compositional,tresadern2012real,xiong2013supervised,asthana2014incremental,xiong2015global}) or tracking/estimation of dense facial motion (\cite{essa1996modeling,yacoob1996recognizing,essa1997coding,decarlo2000optical,koelstra2010dynamic,snape2015face}). The first series of model-based methods for dense facial motion tracking  were proposed by MIT Media lab in mid 1990's (\cite{essa1997coding,essa1996modeling,essa1994tracking,basu1996motion}). In particular, the method by \cite{essa1994vision} tracks facial motion using optical flow computation coupled with a geometric and a physical (muscle) model describing the facial structure. This modeling results in a time-varying spatial patterning of facial shape and a parametric representation of the independent muscle action groups which is responsible for the observed facial motions. In \cite{essa1994tracking} the physically-based face model of \cite{essa1994vision} is driven by a set of responses from a set of templates that characterise facial regions. Model generated flow has been used by the same group in \cite{basu1996motion} for motion regularisation. 3D motion estimation using sparse 3D models and optical flow estimation has also been proposed by \cite{li19933,bozdaugi19943}. Dense facial motion tracking is performed in \cite{decarlo2000optical} by solving a model-based (using a facial deformable model) least-squares optical flow problem. The constraints are relaxed by the use of a Kalman filter, which permits controlled constraint violations based on the noise present in the optical flow information, and enables optical flow and edge information to be combined more robustly and efficiently. Free-form deformations (\cite{rueckert1999nonrigid}) are used in \cite{koelstra2010dynamic} for extraction of dense facial motion for facial action unit recognition. Recently, \cite{snape2015face} proposed a statistical model of the facial flow for fast and robust dense facial motion extraction.

Arguably, the problem that has received the majority of attention is tracking of a set of sparse facial landmarks. The landmarks are either associated to a particular sparse facial model, i.e. the popular Candide facial model by \cite{li19933}, or correspond to fiducial facial regions/parts (e.g., mouth, eyes, nose etc.) (\cite{cootes2001active}). Even earlier attempts such as \cite{essa1994vision} understood the usefulness of tracking facial regions/landmarks in order to perform robust fitting of complex facial models (currently the vast majority of dense 3D facial model tracking techniques, such as \cite{wei2004real,zhang2008real,amberg2011editing}, rely on the robust tracking of a set of facial landmarks). Early approaches for tracking facial landmarks/regions included: (i)~the use of templates built around certain facial regions (\cite{essa1994vision}), (ii)~the use of facial classifiers to detect landmarks (\cite{colmenarez1999detection}) where tracking is performed using modal analysis (\cite{tao1998connected}) or (iii)~the use of face and facial region segmentation 
to detect the features where tracking is performed using block matching (\cite{sobottka1996face}). Currently, deformable face tracking has converged with the problem of facial landmark localisation on static images. That is, the methods generally rely on fitting generative or discriminative statistical models of appearance and 2D/3D sparse facial shape at each frame. Arguably, the most popular methods are generative and discriminative variations of Active Appearance Models (AAMs) and Active Shape Models (ASMs) (\cite{pighin1999resynthesizing,cootes2001active,dornaika2004fast,xiao2004real,matthews2004active,dedeouglu2007asymmetry,papandreou2008adaptive,amberg2009compositional,saragih2011deformable,xiong2013supervised,xiong2015global}). The statistical models of appearance and shape can either be generic as in \cite{cootes2001active,matthews2004active,xiong2013supervised} or incrementally updated in order to better capture the face at hand, as in \cite{sung2009adaptive,asthana2014incremental}. The vast majority of the facial landmark localisation methodologies require an initialisation provided by a face detector. More details regarding current state-of-the-art in facial landmark localisation can be found in Section~\ref{subsec:alignment}.

\begin{figure*}[!t]
    \centering
    \includegraphics[width=0.8\textwidth]{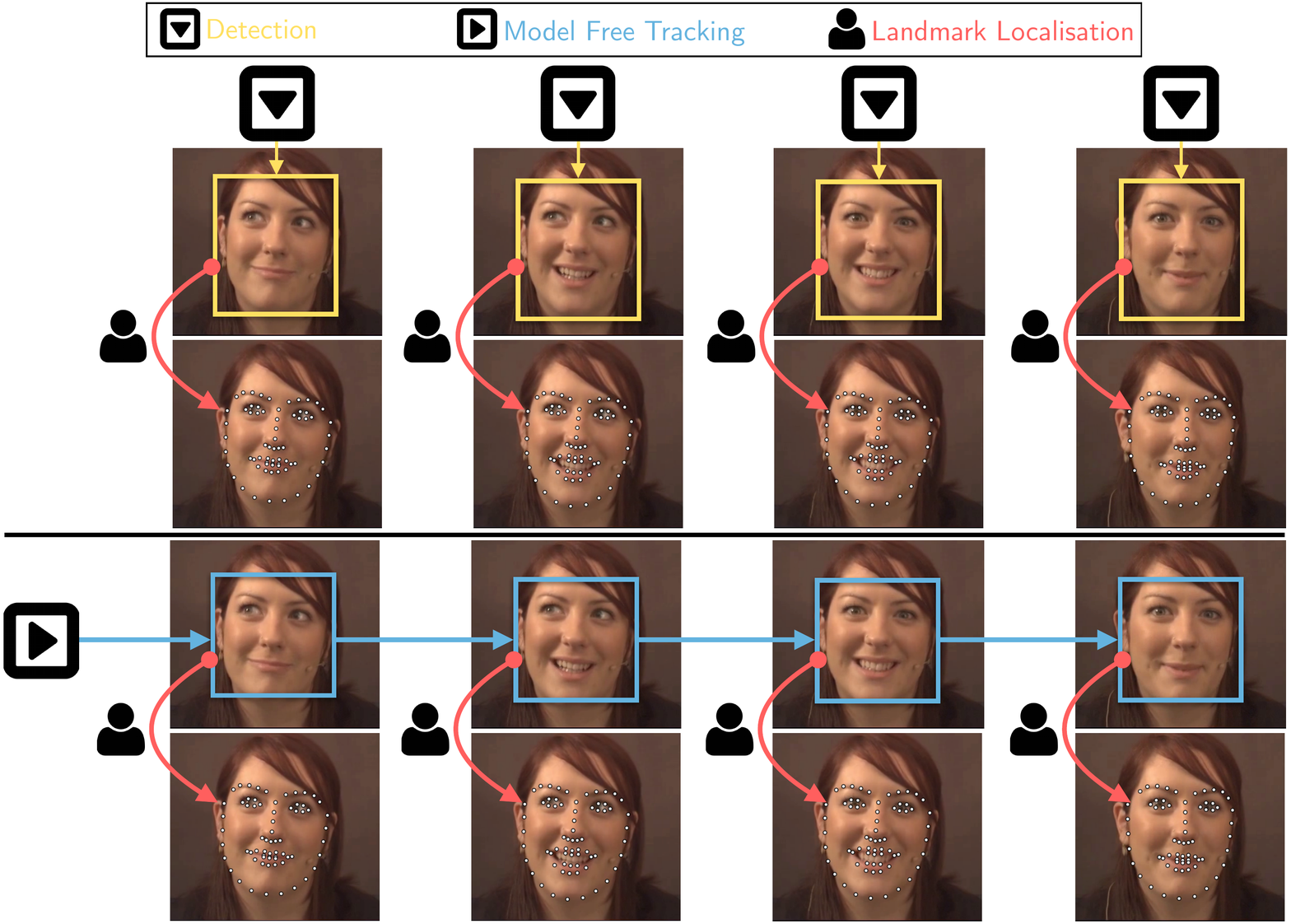}
    \caption{Overview of the standard approaches for deformable face tracking. \emph{(Top)}: Face detection is applied independently at each frame of the video followed by facial landmark localisation. \emph{(Bottom)}: Model free tracking is employed, initialised with the bounding box of the face at the first frame, followed by facial landmark localisation.}
    \label{fig:overview}
\end{figure*}

Arguably, the current practise regarding deformable face tracking includes the combination of a generic face detection and generic facial landmark localisation technique (\cite{saragih2011deformable,xiong2013supervised,xiong2015global,alabort2015unifying,asthana2015pixels}). For example, popular approaches include successive application of the face detection and facial landmark localisation procedure at each frame. Another approach  performs face detection in the first frame and then applies facial landmark localisation at each consecutive frame using the fitting result of the previous frame as initialisation. Face detection can be re-applied in case of failure. This is the approach that is used by popular packages such as \cite{asthana2014incremental}. In this paper, we thoroughly evaluate variations of the above approaches. Furthermore, we consider the use of modern model free state-of-the-art trackers for rigid 2D tracking in order to be used as initialisation for the facial landmark localisation procedure. This is pictorially described in Figure~\ref{fig:overview}.

\subsection{Face Tracking Benchmarking}
For assessing the performance of rigid 2D face tracking several short face sequences have been annotated with regards to the facial region (using a bounding box style annotation). One of the first sequences that has been annotated for this task is the so-called Dudek sequence by \cite{dudek}\footnote{The Dudek sequence has been annotated with regards to certain facial landmarks only to be used for the estimation of an affine transformation.}. Nowadays, several such sequences have been annotated and are publicly available, such as the ones by \cite{dikt_annotation,nus_pro,wu2015object}.

The performance of non-rigid dense facial tracking methodologies was usually assessed by using markers (\cite{decarlo2000optical}), simulated data (\cite{snape2015face}), visual inspection (\cite{decarlo2000optical,essa1997coding,essa1996modeling,yacoob1996recognizing,snape2015face,koelstra2010dynamic}) or indirectly by the use of the dense facial motion for certain tasks, such as expression analysis (\cite{essa1996modeling,yacoob1996recognizing,koelstra2010dynamic}). Regarding tracking of facial landmarks, up until recently, the preferred method for assessing the performance was visual inspection in a number of selected facial videos (\cite{xiong2013supervised,tresadern2012real}). Other methods were assessed on a small number of short (a few seconds in length) annotated facial videos (\cite{sagonas2014raps,asthana2014incremental}). Until recently the longest annotated facial video sequence was the so-called talking face of \cite{talking_face} which was used to evaluate many tracking methods including \cite{orozco2013hierarchical,amberg2009compositional}. The talking face video comprises of 5000 frames (around 200 seconds) taken from a video of a person engaged in a conversation. The talking face video was initially tracked using an Active Appearance Model (AAM) that had a shape model and a total of 68 landmarks are provided. The tracked landmarks were visually checked and manually corrected where necessary.

Recently, \cite{xiong2015global} introduced a benchmark for facial landmark tracking using videos from the Distracted Driver Face (DDF) and Naturalistic
Driving Study (NDS) in \cite{drivers_database}\footnote{In a private communication, the authors of \cite{xiong2015global} informed us that the annotated data, as described in the paper, will not be made publicly available (at least not in the near future).}. The DDF dataset contains 15 sequences with a total of 10,882 frames. Each sequence displays a single subject posing as the distracted driver in a stationary vehicle or indoor environment. 12 out of 15 videos were recorded with subjects sitting inside of a vehicle. Five of them were recorded during the night under infrared (IR) light and the rest were recorded during the daytime under natural lighting. The remaining three were recorded indoors. The NDS database contains 20 sub-sequences of driver faces recorded during a drive conducted between the Blacksburg, VA and Washington, DC areas (NDS is more challenging than DDF since its videos are of lower spatial and temporal resolution). Each video of the NDS database has one minute duration recorded at 15 frames per second (fps) with a $360 \times 240$ resolution. For both datasets one in every ten frames was annotated using either 49 landmarks for near-frontal faces or 31 landmarks for profile faces. The database contains many extreme facial poses (90$^o$ yaw, 50$^o$ pitch) as well as many faces under extreme lighting condition (e.g., IR). In total the dataset presented in \cite{xiong2015global} contains between 2,000 to 3,000 annotated faces (please refer to \cite{xiong2015global} for exemplar annotations).

The only existing large in-the-wild benchmark for facial landmark tracking was recently introduced by \cite{shen2015first}. The benchmark consists of 114 with varying difficulty and provides annotations generated in a semi-automatic manner (\cite{chrysos2015offline,shen2015first,tzimiropoulos2015project}). This challenge, called 300VW, is the only existing large-scale comprehensive benchmark for deformable model tracking. More details regarding the dataset of the 300VW benchmark can be found in Section~\ref{sec:exp_datasets}. The performance of the pipelines considered in this paper are compared with the participating methods of the 300VW challenge in Section~\ref{exp:competition}.

\section{Deformable Face Tracking}\label{sec:tracking}
In this paper, we focus on the problem of performing deformable face tracking across long-term sequences within unconstrained videos. The problem of tracking across long-term sequences is particularly challenging as the appearance of the face may change significantly during the sequence due to occlusions, illumination variation, motion artifacts and head pose. For the problem of deformable tracking, however, the problem is further complicated by the expectation of recovering a set of accurate fiducial points in conjunction with successfully tracking the object. As described in Section~\ref{sec:related}, current deformable facial tracking methods mainly concentrate on performing face detection per frame and then performing facial landmark localisation. However, we consider the most important metric for measuring the success of deformable face tracking as the facial landmark localisation accuracy. Given this, there are a number of strategies that could feasibly be employed in order to attempt to minimise the total facial landmark localisation error across the entire sequence. Therefore, we take advantage of current advances in face detection, model free tracking and facial landmark localisation techniques in order to perform deformable face tracking. Specifically, we investigate three strategies for deformable tracking:
\begin{enumerate}
    \item \textbf{Detection + Landmark Localisation.} Face Detection per frame, followed by facial landmark localisation initialised within the facial bounding boxes. This scenario is visualised in Figure~\ref{fig:overview} (top).
    \item \textbf{Model Free Tracking + Landmark Localisation.} Model free tracking, initialised around the interior of the face within the first frame, followed by facial landmark localisation within the tracked box. This scenario is visualised in Figure~\ref{fig:overview} (bottom).
    \item \textbf{Hybrid Systems.} Hybrid methods that attempt to improve the robustness of the placement of the bounding box for landmark localisation. Namely, we investigate methods for failure detection, trajectory smoothness and reinitialisation. Examples of such methods are pictorially demonstrated in Figures~\ref{fig:detection_failure} and~\ref{fig:tracking_failure}.
\end{enumerate}
Note that we focus on combinations of methods that provide bounding boxes of the facial region followed by landmark localisation. This is due to the fact that the current set of state-of-the-art landmark localisation methods are all local methods and require initialisation within the facial region. Although joint face detection and landmark localisation methods have been proposed (\cite{zhu2012face,chen2014joint}), they are not competitive with the most recent set of landmark localisation methods. For this reason, in this paper we focus on the combination of bounding box estimators with state-of-the-art local landmark localisation techniques.

The remainder of this Section will give a brief overview of the literature concerning face detection, model free tracking and facial landmark localisation.
\subsection{Face Detection}\label{subsec:detection}
Face detection is among the most important and popular tasks in Computer Vision and an essential step for applications such as face recognition and face analysis. Although it is one of the oldest tasks undertaken by researchers (the early works appeared about 45 years ago (\cite{sakai1972computer,fischler1973representation})), it is still an open and challenging problem. Recent advances can achieve reliable performance under moderate illumination and pose conditions, which led to the installation of simple face detection technologies in everyday devices such as digital cameras and mobile phones. However, recent benchmarks (\cite{fddbTech}) show that the detection of faces on arbitrary images is still a very challenging problem.

Since face detection has been a research topic for so many decades, the existing literature is, naturally, extremely extensive. The fact that all recent face detection surveys (\cite{hjelmaas2001face,yang2002detecting,zhang2010survey,zafeiriou2015survey}) provide different categorisations of the relative literature is indicative of the huge range of existing techniques. Consequently, herein, we only present a basic outline of the face detection literature. For an extended review, the interested reader may refer to the most recent face detection survey in \cite{zafeiriou2015survey}.

According to the most recent literature review \cite{zafeiriou2015survey}, existing methods can be separated in two major categories. The first one includes methodologies that learn a set of rigid templates, which can be further split in the following groups: (i)~boosting-based methods, (ii)~approaches that utilise SVM classifiers, (ii)~exemplar-based techniques, and (iv)~frameworks based on Neural Networks. The second major category includes deformable part models, i.e. methodologies that learn a set of templates per part as well as the deformations between them.

%%%%%%%%%%%%%%%%%%%%%
\begin{table*}[!t]
\centering
    \begin{tabular}{c@{\hspace{0.1cm}}lccl}
    \toprule
    \emph{Method} & \emph{Citation(s)} & \emph{Rigid Template} & \emph{DPM} & \emph{Implementation} \\
    \midrule[\heavyrulewidth]
    \multirow{3}{*}{DPM} & \cite{felzenszwalb2010object} & & \multirow{3}{*}{\checkmark} & \multirow{3}{*}{\url{https://github.com/menpo/ffld2}} \\
    & \cite{mathias2014face} & & & \\
    & \cite{menpo14} & & & \\
    \cmidrule(lr){1-2}\cmidrule(lr){3-5}
    SS-DPM  & \cite{zhu2012face} & & \checkmark & \url{https://www.ics.uci.edu/~xzhu/face} \\
    \cmidrule(lr){1-2}\cmidrule(lr){3-5}
    \multirow{2}{*}{SVM+HOG} & \cite{king2015max} & \multirow{2}{*}{\checkmark} & & \multirow{2}{*}{\url{https://github.com/davisking/dlib}} \\
    & \cite{king2009dlib} & & & \\
    \cmidrule(lr){1-2}\cmidrule(lr){3-5}
    \multirow{2}{*}{VJ} & \cite{viola2004robust} & \multirow{2}{*}{\checkmark} & & \multirow{2}{*}{\url{http://opencv.org}} \\
    & \cite{opencv_library} & & & \\
    \bottomrule
\end{tabular}
\caption{The set of detectors used in this paper. The table reports the short name of the method, the relevant citation(s) as well as the link to the implementation used.}
\label{tbl:detectors}
\end{table*}
%%%%%%%%%%%%%%%%%%%%%

\emph{Boosting Methods.} Boosting combines multiple ``weak'' hypotheses of moderate accuracy in order to determine a highly accurate hypothesis. The most characteristic example is Adaptive Boosting (AdaBoost) which is utilised by the most popular face detection methodology, i.e. the Viola-Jones (VJ) detector of \cite{viola2001rapid,viola2004robust}. Characteristic examples of other methods that employ variations of AdaBoost include~\cite{li2002statistical,wu2004fast,mita2005joint}. The original VJ algorithm used Haar features, however boosting (or cascade of classifiers methodologies in general) have been shown to greatly benefit from robust features (\cite{kostinger2012robust,jun2013local,li2011face,li2013learning,mathias2014face,yang2014aggregate}), such as HOG (\cite{dalal2005histograms}), SIFT (\cite{lowe1999object}), SURF (\cite{bay2008speeded}) and LBP (\cite{ojala2002multiresolution}). For example, SURF features have been successfully combined with a cascade of weak classifiers in \cite{li2011face,li2013learning}, achieving faster convergence. Additionally, \cite{jun2013local} propose robust face specific features that combine both LBP and HOG. \cite{mathias2014face} recently proposed an approach (so called HeadHunter) with state-of-the-art performance that employs various robust features with boosting. Specifically, they propose the adaptation of Integral Channel Features (ICF) (\cite{dollar2009integral}) with HOG and LUV colour channels, combined with global feature normalisation. A similar approach is followed by \cite{yang2014aggregate}, in which they combine gray-scale, RGB, HSV, LUV, gradient magnitude and histograms within a cascade of weak classifiers.

\emph{SVM Classifiers.} Maximum margin classifiers, such as Support Vector Machines (SVMs), have become popular for face detection (\cite{romdhani2001computationally,heisele2003hierarchical,ratsch2004efficient,king2015max}). Even though their detection speed was initially slow, various schemes have been proposed to speed up the process. \cite{romdhani2001computationally} propose a method that computes a reduced set of vectors from the original support vectors that are used sequentially in order to make early rejections. A similar approach is adopted by \cite{ratsch2004efficient}. A hierarchy of SVM classifiers trained on different resolutions is applied in \cite{heisele2003hierarchical}. \cite{king2015max} proposes an algorithm for efficient learning of a max-margin classifier using all the sub-windows of the training images, without applying any sub-sampling, and formulates a convex optimisation that finds the global optimum. Moreover, SVM classifiers have also been used for multi-view face detection (\cite{li2000support,wang2004multi}). For example, \cite{li2000support} first apply a face pose estimator based on Support Vector Regression (SVR), followed by an SVM face detector for each pose.

\emph{Exemplar-based Techniques.} These methods aim to match a test image against a large set of facial images. This approach is inspired by principles used in image retrieval and requires that the exemplar set covers the large appearance variation of human face. \cite{shen2013detecting} employ bag-of-word image retrieval methods to extract features from each exemplar, which creates a voting map for each exemplar that functions as a weak classifier. Thus, the final detection is performed by combining the voting maps. A similar methodology is applied in \cite{li2014efficient}, with the difference that specific exemplars are used as weak classifiers based on a boosting strategy. Recently, \cite{kumar2015visual} proposed an approach that enhances the voting procedure by using semantically related visual words as well as weighted occurrence of visual words based on their spatial distributions. 

\emph{Convolutional Neural Networks.} Another category, similar to the previous rigid template-based ones, includes the employment of Convolutional Neural Networks (CNNs) and Deep CNNs (DCNNs) (\cite{osadchy2007synergistic,zhang2014improving,ranjan2015deep,li2015convolutional,yang2015from}). \cite{osadchy2007synergistic} use a network with four convolution layers and one fully connected layer that rejects the non-face hypotheses and estimates the pose of the correct face hypothesis. \cite{zhang2014improving} propose a multi-view face detection framework by employing a multi-task DCNN for face pose estimation and landmark localization in order to obtain better features for face detection. \cite{ranjan2015deep} combine deep pyramidal features with Deformable Part Models. Recently, \cite{yang2015from} proposed a DCNN architecture that is able to discover facial parts responses from arbitrary uncropped facial images without any part supervision and report state-of-the-art performance on current face detection benchmarks.

\emph{Deformable Part Models.} DPMs (\cite{schneiderman2004object,felzenszwalb2005pictorial,felzenszwalb2010object,zhu2012face,yan2013structural,li2013probabilistic,yan2014fastest,mathias2014face,ghiasi2014occlusion,barbu2014face}) learn a patch expert for each part of an object and model the deformations between parts using spring-like connections based on a tree structure. Consequently, they perform joint facial landmark localisation and face detection. Even though they are not the best performing methods for landmark localisation, they are highly accurate for face detection in-the-wild. However, their main disadvantage is their high computational cost. Pictorial Structures (PS) (\cite{fischler1973representation,felzenszwalb2005pictorial}) are the first family of DPMs that appeared. They are generative DPMs that assume Gaussian distributions to model the appearance of each part, as well as the deformations. They became a very popular line of research after the influential work in \cite{felzenszwalb2005pictorial} that proposed a very efficient dynamic programming algorithm for finding the global optimum based on Generalized Distance Transform. Many discriminatively trained DPMs (\cite{felzenszwalb2010object,zhu2012face,yan2013structural,yan2014fastest}) appeared afterwards, which learn the patch experts and deformation parameters using discriminative classifiers, such as latent SVM.

DPMs can be further separated with respect to their training scenario into: (i) weakly supervised and (ii) strongly supervised. Weakly-supervised DPMs (\cite{felzenszwalb2010object,yan2014fastest}) are trained using only the bounding boxes of the positive examples and a set of negative examples. The most representative example is the work by \cite{felzenszwalb2010object}, which has proved to be very efficient for generic object detection. Under a strongly supervised scenario, it is assumed that a training database with images annotated with figucial landmarks is available. Several strongly supervised methods exist in the literature (\cite{felzenszwalb2005pictorial,zhu2012face,yan2013structural,ghiasi2014occlusion}). \cite{ghiasi2014occlusion} propose an hierarchical DPM that explicitly models parts' occlusions. In \cite{zhu2012face} it is shown that a strongly supervised DPM outperforms, by a large margin, a weakly supervised one. In contrast, HeadHunter by \cite{mathias2014face} shows that a weakly supervised DPM can outperform all current state-of-the-art face detection methodologies including the strongly supervised DPM of \cite{zhu2012face}. 

According to FDDB (\cite{fddbTech}), which is the most well established face detection benchmark, the currently top-performing methodology is the one by \cite{ranjan2015deep}, which combines DCNNs with a DPM. However, it is impossible to use most DCNN-based techniques, because their authors do not provide publicly available implementations and it is very complicated and time-consuming to train and fine-tune such networks. Thus, even though many DCNN-based techniques are proved to achieve state-of-the-art performance, it was not feasible to use them for deformable face tracking pipelines. Nevertheless, we employ the top performing SVM-based method for learning rigid templates (\cite{king2015max}), as well as the best weakly and strongly supervised DPM implementations of \cite{mathias2014face} and \cite{zhu2012face}. Finally, we also use the popular VJ algorithm (\cite{viola2001rapid,viola2004robust}) as a baseline face detection method. The employed face detection implementations are summarised in Table~\ref{tbl:detectors}.

%%%%%%%%%%%%%%%%
\begin{table*}[!t]
\centering
\begin{tabular}{c@{\hspace{0.1cm}}lccccl}
\toprule
\emph{Method} & \emph{Citation(s)} & \emph{D} & \emph{G} & \emph{P} & \emph{K} & \emph{Implementation} \\
\midrule[\heavyrulewidth]
CMT & \cite{Nebehay2015CVPR} & & & & \checkmark & \url{https://github.com/gnebehay/CppMT} \\
\cmidrule(lr){1-2}\cmidrule(lr){3-7}
DF & \cite{sevilla2012distribution} & & \checkmark & & & \url{http://goo.gl/YmG6W4} \\
\cmidrule(lr){1-2}\cmidrule(lr){3-7}
\multirow{2}{*}{DSST} & \cite{danelljan2014accurate} & \multirow{2}{*}{\checkmark} & & & & \multirow{2}{*}{\url{https://github.com/davisking/dlib}} \\
& \cite{king2009dlib} & & & & & \\
\cmidrule(lr){1-2}\cmidrule(lr){3-7}
FCT & \cite{zhang2014fast} & \checkmark & \checkmark & & & \url{http://goo.gl/Ujc5B0} \\
\cmidrule(lr){1-2}\cmidrule(lr){3-7}
IVT & \cite{ross2008incremental} & & \checkmark & & & \url{http://goo.gl/WtbOIX} \\
\cmidrule(lr){1-2}\cmidrule(lr){3-7}
KCF & \cite{henriques2015high} & \checkmark & & & & \url{https://github.com/joaofaro/KCFcpp} \\
\cmidrule(lr){1-2}\cmidrule(lr){3-7}
LRST & \cite{zhang2014robust} & & \checkmark & & & \url{http://goo.gl/ZC9JbQ} \\
\cmidrule(lr){1-2}\cmidrule(lr){3-7}
\multirow{2}{*}{MIL} & \cite{babenko2011mil} & \multirow{2}{*}{\checkmark} & & & & \multirow{2}{*}{\url{http://opencv.org}} \\
& \cite{opencv_library} & & & & & \\
\cmidrule(lr){1-2}\cmidrule(lr){3-7}
ORIA & \cite{wu2012online} &  & \checkmark & & & \url{https://goo.gl/RT3zNC} \\
\cmidrule(lr){1-2}\cmidrule(lr){3-7}
RPT & \cite{li2015reliable} & \checkmark & & & & \url{https://github.com/ihpdep/rpt} \\
\cmidrule(lr){1-2}\cmidrule(lr){3-7}
SPOT & \cite{zhang2014preserving} & \checkmark & & \checkmark & & \url{http://visionlab.tudelft.nl/spot} \\
\cmidrule(lr){1-2}\cmidrule(lr){3-7}
SRDCF & \cite{danelljan2015learning} & \checkmark & & & & \url{https://goo.gl/Q9d1O5} \\
\cmidrule(lr){1-2}\cmidrule(lr){3-7}
STRUCK & \cite{hare2011struck} & \checkmark & & & & \url{http://goo.gl/gLR93b} \\
\cmidrule(lr){1-2}\cmidrule(lr){3-7}
TLD & \cite{kalal2012tracking} & \checkmark & & & & \url{https://github.com/zk00006/OpenTLD} \\
\bottomrule
\end{tabular}
\caption{The set of trackers that are used in this paper. The table reports the short name of the method, the relevant citation(s) as well as the link to the implementation used. The initials stand for: (\emph{D})iscriminative, (\emph{G})enerative, (\emph{P})art-based and (\emph{K})eypoint trackers.}
\label{tbl:trackers}
\end{table*}
%%%%%%%%%%%%%%%%

\subsection{Model Free Tracking}\label{subsec:tracking}
Model free tracking is an extremely active area of research. Given the initial state (e.g., position and size of the containing box) of a target object in the first image, model free tracking attempts to estimate the states of the target in subsequent frames. Therefore, model free tracking provides an excellent method of initialising landmark localisation methods.

The literature on model free tracking is vast. For the rest of this section, we will provide an extremely brief overview of model free tracking that focuses primarily on areas that are relevant to the tracking methods we investigated in this paper. We refer the interested reader to the wealth of tracking surveys (\cite{li2013survey,smeulders2014visual,salti2012adaptive,yang2011recent}) and benchmarks (\cite{wu2013object,wu2015object,Kristan2013a,Kristan2014a,Kristan2015a,Kristan2016a,smeulders2014visual}) for more information on model free tracking methods.

\emph{Generative Trackers.} These trackers attempt to model the objects appearance directly. This includes template based methods, such as those by \cite{matthews2004template,baker2004lucas,sevilla2012distribution}, as well as parametric generative models such as \cite{balan2006adaptive,ross2008incremental,black1998eigentracking,xiao2014l2}. The work of \cite{ross2008incremental} introduces online subspace learning for tracking with a sample mean update, which allows the tracker to account for changes in illumination, viewing angle and pose of the object. The idea is to incrementally learn a low-dimensional subspace and adapt the appearance model on object changes. The update is based on an incremental principal component analysis (PCA) algorithm, however it seems to be ineffective at handling large occlusions or non-rigid movements due to its holistic model. To alleviate the partial occlusion, \cite{xiao2014l2} suggest the use of square templates along with PCA. Another popular area of generative tracking is the use of sparse representations for appearance. In \cite{mei2011robust}, a target candidate is represented by a sparse linear combination of target and trivial templates. The coefficients are extracted by solving an $\ell_1$ minimisation problem with non-negativity constraints, while the target templates are updated online. However, solving the $\ell_1$ minimisation for each particle is computationally expensive. A generalisation of this tracker is the work of \cite{zhang2012robust}, which learns the representation for all particles jointly. It additionally improves the robustness by exploiting the correlation among particles. An even further abstraction is achieved in \cite{zhang2014robust} where a low-rank sparse representation of the particles is encouraged. In \cite{zhang2014fast}, the authors generalise the low-rank constraint of \cite{zhang2014robust} and add a sparse error term in order to handle outliers. Another low-rank formulation was used by \cite{wu2012online} which is an online version of the RASL (\cite{peng2012rasl}) algorithm and attempts to jointly align the input sequence using convex optimisation.

\emph{Keypoint Trackers.} These trackers (\cite{pernici2014object,poling2014better,hare2012efficient,Nebehay2015CVPR}) attempt to use the robustness of keypoint detection methodologies like SIFT (\cite{lowe1999object}) or SURF (\cite{bay2008speeded}) in order to perform tracking. \cite{pernici2014object} collected multiple descriptors of weakly aligned keypoints over time and combined these matched keypoints in a RANSAC voting scheme. \cite{Nebehay2015CVPR} utilises keypoints to vote for the object center in each frame. A consensus-based scheme is applied for outlier detection and the votes are transformed based on the current key point arrangement to consider scale and rotation. However, keypoint methods may suffer from difficulty in capturing the global information of the tracked target by only considering the local points.

\emph{Discriminative Trackers.} These trackers attempt to explicitly model the difference between the object appearance and the background. Most commonly, these methods are named ``tracking-by-detection'' techniques as they involve classifying image regions as either part of the object or the background. In their work, \cite{grabner2006real} propose an online boosting method to select and update discriminative features which allows the system to account for minor changes in the object appearance. However, the tracker fails to model severe changes in appearance. \cite{babenko2011mil} advocate the use of a multiple instance learning boosting algorithm to mitigate the drifting problem. More recently, discriminative correlation filters (DCF) have become highly successful at tracking. The DCF is trained by performing a circular sliding window operation on the training samples. This periodic assumption enables efficient training and detection by utilizing the Fast Fourier Transform (FFT). \cite{danelljan2014accurate} learn separate correlation filters for the translation and the scale estimation. In \cite{danelljan2015learning}, the authors introduce a sparse spatial regularisation term to mitigate the artifacts at the boundaries of the circular correlation. In contrast to the linear regression commonly used to learn DCFs, \cite{henriques2015high} apply a kernel regression and propose its multi-channel extension to enable to the use of features such as HOG~\cite{dalal2005histograms}. \cite{li2015reliable} propose a new use for particle filters in order to choose reliables patches to consider part of the object. These patches are modelled using a variant of the method proposed by \cite{henriques2015high}. \cite{hare2011struck} propose the use of structured output prediction. By explicitly allowing the outputs to parametrize the needs of the tracker, an intermediate classification step is avoided.

\emph{Part-based Trackers.} These trackers attempt to implicitly model the parts of an object in order to improve tracking performance. \cite{adam2006robust} represent the object with multiple arbitrary patches. Each patch votes on potential positions and scales of the object and a robust statistic is employed to minimise the voting error. 
\cite{kalal2010forward} sample the object and the points are tracked independently in each frame by estimating optical flow. Using a forward-backward measure, the erroneous points are identified and the remaining reliable points are utilised to compute the optimal object trajectory.  
\cite{yao2013part} adapt the latent SVM of \cite{felzenszwalb2010object} for online tracking, by restricting the search in the vicinity of the location of the target object in the previous frame.
In comparison to the weakly supervised part-based model of \cite{yao2013part}, in \cite{zhang2013structure} the authors recommend an online strongly supervised part-based deformable model that learns the representation of the object and the representation of the background by training a classifier. 
\cite{wang2015tric} employ a part-based tracker by estimating a direct displacement prediction of the object. A cascade of regressors is utilised to localise the parts, while the model is updated online and the regressors are initialised by multiple motion models at each frame.

%%%%%%%%%%%%%%%%%%%%
\begin{table*}[!t]
\centering
    \begin{tabular}{c@{\hspace{0.1cm}}lccl}
    \toprule
    \emph{Method} & \emph{Citation(s)} & \emph{Discriminative} & \emph{Generative} & \emph{Implementation} \\
    \midrule[\heavyrulewidth]
    \multirow{2}{*}{AAM} & \cite{tzimiropoulos2015project} & & \multirow{2}{*}{\checkmark} & \multirow{2}{*}{\url{https://github.com/menpo/menpofit}} \\
    & \cite{menpo14} & & & \\
    \cmidrule(lr){1-2}\cmidrule(lr){3-5}
    \multirow{2}{*}{ERT} & \cite{kazemi2014one} & \multirow{2}{*}{\checkmark} & & \multirow{2}{*}{\url{https://github.com/davisking/dlib}} \\
    & \cite{king2009dlib} & & & \\
    \cmidrule(lr){1-2}\cmidrule(lr){3-5}
    CFSS & \cite{zhu2015face} & \checkmark & & \url{https://github.com/zhusz/CVPR15-CFSS} \\
    \cmidrule(lr){1-2}\cmidrule(lr){3-5}
    \multirow{2}{*}{SDM} & \cite{xiong2013supervised} & \multirow{2}{*}{\checkmark} & & \multirow{2}{*}{\url{https://github.com/menpo/menpofit}} \\
    & \cite{menpo14} & & & \\
    \bottomrule
    \end{tabular}
\caption{The landmark localisation methods employed in this paper. The table reports the short name of the method, the relevant citation(s) as well as the link to the implementation used.}
\label{tbl:alignment}
\end{table*}
%%%%%%%%%%%%%%%%%%%%

Given the wealth of available trackers, selecting appropriate trackers for deformable tracking purposes poses a difficult proposition. In order to attempt to give as broad an overview as possible, we selected a representative tracker from each of the categories described previously. Therefore, in this paper we compare against 14 trackers which are outlined in Table~\ref{tbl:trackers}. SRDCF~(\cite{danelljan2015learning}), KCF~(\cite{henriques2015high}) and DSST~(\cite{danelljan2014accurate}) are all discriminative trackers based on DCFs. They all performed well in the VOT 2015~(\cite{Kristan2015a}) challenge and DSST was the winner of VOT 2014~(\cite{Kristan2014a}). STRUCK~(\cite{hare2011struck}) is a discriminative tracker that performed very well in the Online Object Tracking benchmark~(\cite{wu2013object}). SPOT~(\cite{zhang2014preserving}) is a strong performing part based tracker, CMT~(\cite{Nebehay2015CVPR}) is a strong performing keypoint based tracker and LRST~(\cite{zhang2014robust}) and ORIA~(\cite{wu2012online}) are recent generative trackers. RPT~(\cite{li2015reliable}) is a recently proposed technique that reported state-of-the-art results on the Online Object Tracking benchmark~(\cite{wu2013object}). Finally, TLD~(\cite{kalal2012tracking}), MIL~(\cite{babenko2011mil}), FCT~(\cite{zhang2014fast}), DF~(\cite{sevilla2012distribution}) and IVT~(\cite{ross2008incremental}) were included as baseline tracking methods with publicly available implementations.

\subsection{Facial Landmark Localisation}\label{subsec:alignment}
%%%%%%%%%%%%%%%%%%%%%%%%%%%%%%%%%%%%%%%%%%%%%%%%%%%%%%%%%%%%%%%%%%%%%%
Statistical deformable models have emerged as an important research field over
the last few decades, existing at the intersection of computer vision, statistical
pattern recognition and machine learning.
Statistical deformable models aim to solve generic object alignment
in terms of localisation of fiducial points. Although deformable models can be
built for a variety of object classes, the majority of ongoing research has
focused on the task of facial alignment.
Recent large-scale challenges on facial alignment (\cite{sagonas2013semi,sagonas2013300,sagonas2015300})
are characteristic examples of the rapid progress being made in the field.

Currently, the most commonly-used and well-studied face alignment methods can be separated into two major families: (i)~\emph{discriminative} models that employ regression in a cascaded manner, and (ii)~\emph{generative} models that are iteratively optimised.

\emph{Regression-based models.} The methodologies of this category aim to
learn a regression function that regresses from the object's appearance
(e.g. commonly handcrafted features) to the target output variables
(either the landmark coordinates or the parameters of a statistical shape model).
Although the history behind using linear regression in order to tackle the
problem of face alignment spans back many years~(\cite{cootes2001active}), the
research community turned towards alternative approaches due to the lack of
sufficient data for training accurate regression functions. 
Nevertheless, recently regression-based techniques have prevailed in the field
thanks to the wealth of annotated data and effective
handcrafted features~(\cite{lowe1999object,dalal2005histograms}). Recent works
have shown that excellent performance can be achieved by employing a
cascade of regression functions~(\cite{burgos2013robust,xiong2013supervised,xiong2015global,dollar2010cascaded,xiong2013supervised,cao2014face,kazemi2014one,ren2014face,asthana2014incremental,tzimiropoulos2015project,zhu2015face}).
Regression based methods can be approximately seperated into two categories
depending on the nature of the regression function employed. Methods
that employ a linear regression such as the Supervised Descent Method (SDM) of \cite{xiong2013supervised}
tend to employ robust hand-crafted features~(\cite{xiong2013supervised,asthana2014incremental,xiong2015global,tzimiropoulos2015project,zhu2015face}).
On the other hand, methods that employ tree-based regressors such as
the Explicit Shape Regression (ESR) method of \cite{cao2014face},
tend to rely on data driven features that are optimised directly by
the regressor (\cite{burgos2013robust,cao2014face,dollar2010cascaded,kazemi2014one}).

\emph{Generative models.} The most dominant representative algorithm of this
category is, by far, the Active Appearance Model (AAM). AAMs consist of
parametric linear models of both shape and appearance of an object, typically
modelled by Principal Component Analysis (PCA). The AAM objective
function involves the minimisation of the appearance reconstruction error with
respect to the shape parameters. AAMs were initially proposed by \cite{cootes1995active,cootes2001active},
where the optimisation was performed by a single regression step between the
current image reconstruction residual and an increment to the shape parameters.
However, \cite{matthews2004active,baker2004lucas} linearised the
AAM objective function and optimised it using the Gauss-Newton algorithm.
Following this, Gauss-Newton optimisation has been the modern method for
optimising AAMs. Numerous extensions have been published, either related to
the optimisation
procedure~(\cite{papandreou2008adaptive,tzimiropoulos2013optimization,alabort2014bayesian,alabort2015unifying,tzimiropoulos2014gauss})
or the model structure~(\cite{tzimiropoulos2012generic,antonakos2014hog,tzimiropoulos2014active,antonakos2015feature,antonakos2015active}).

In recent challenges by \cite{sagonas2013300,sagonas2015300}, discriminative
methods have been shown to represent the current state-of-the-art. However,
in order to enable a fair comparison between types of methods we selected
a representative set of landmark localisation methods to compare with in this paper.
The set of landmark localisation methods used in the paper is given in Table~\ref{tbl:alignment}.
We chose to use ERT~(\cite{kazemi2014one}) as it is extremely fast and the implementation
provided by \cite{king2009dlib} is the best known implementation of a tree-based regressor.
We chose CFSS~(\cite{zhu2015face}) as it is the current state-of-the-art on the data provided
by the 300W competition of \cite{sagonas2013300}. We used the Gauss-Newton Part-based AAM of \cite{tzimiropoulos2014gauss}
as the top performing generative localisation method, as provided by the Menpo Project~(\cite{menpo14}).
Finally, we also demonstrated an SDM~(\cite{xiong2013supervised}) as implemented by \cite{menpo14}
as a baseline.

\begin{table*}[!t]
\centering
\begin{tabular}{cccccccc}
\toprule
\multirow{2}{*}{\emph{Experiment}} & \multirow{2}{*}{\emph{Section}} & \multirow{2}{*}{\emph{Tracking}} & \multirow{2}{*}{\emph{Detection}} & \emph{Landmark} & \emph{Failure} & \multirow{2}{*}{\emph{Re-initialisation}} & \emph{Kalman} \\
& & & & \emph{Localisation} & \emph{Checking} & & \emph{Smoothing} \\
\midrule[\heavyrulewidth]
1 & \ref{exp:detection}                    &                 & \checkmark       & \checkmark                   &                          &                          &                         \\
\cmidrule(lr){1-2}\cmidrule(lr){3-8}
2 & \ref{exp:detection_init_from_previous} &                 & \checkmark       & \checkmark                   &                          & \checkmark               &                         \\
\cmidrule(lr){1-2}\cmidrule(lr){3-8}
3 & \ref{exp:tracking}                     & \checkmark      &                  & \checkmark                   &                          &                          &                         \\
\cmidrule(lr){1-2}\cmidrule(lr){3-8}
4 & \ref{exp:tracking_restart}             & \checkmark      &                  & \checkmark                   & \checkmark               & \checkmark               &                         \\
\cmidrule(lr){1-2}\cmidrule(lr){3-8}
5 & \ref{exp:kalman}                       & \checkmark      & \checkmark       & \checkmark                   &                          &                          & \checkmark              \\
\cmidrule(lr){1-2}\cmidrule(lr){3-8}
6 & \ref{exp:competition} & \multicolumn{6}{l}{Comparison against state-of-the-art of 300VW competition (\cite{shen2015first}).}\\
\bottomrule
\end{tabular}
\caption{The set of experiments conducted in this paper. This table is intended as an overview of the battery of experiments that were conducted, as well as providing a reference to the relevant section.}
\label{tbl:experiments_summary}
\end{table*}
\section{Experiments}\label{sec:experiments}
In this section, details of the experimental evaluation are established. Firstly, the datasets employed for the evaluation, training and validation are introduced in Section~\ref{sec:exp_datasets}. Next, Section~\ref{sec:exp_implementation} provides details of the training procedures and of the implementations that are relevant to all experiments. Following this, in Sections~\ref{exp:detection}$-$\ref{exp:kalman}, we describe the set of experiments that were conducted in this paper, which are summarised in Table~\ref{tbl:experiments_summary}. Finally, experimental Section~\ref{exp:competition} compares the best results from the previous experiments to the winners of the 300VW competition in~\cite{shen2015first}. 

In the following sections, due to the very large amount of methodologies taken into account, we provide a summary of all the results as tables and only the top 5 methods as graphs for clarity. Please refer to the supplementary material for an extensive report of the experimental results. Additionally, we provide videos with the tracking results for the experiments of Sections~\ref{exp:detection}, \ref{exp:detection_init_from_previous} and~\ref{exp:tracking} for qualitative comparison\footnote{\label{foot:detection}In \url{https://www.youtube.com/watch?v=6bzgmsWgK20} we provide a video with the tracking results of the top methods for face detection followed by landmark localisation (Section~\ref{exp:detection}, Table~\ref{tab:exp_detection}, Figure~\ref{fig:exp_detection}) for qualitative comparison.}\textsuperscript{,}\footnote{\label{foot:detection_init_from_previous}In \url{https://www.youtube.com/watch?v=peQYzqgG2UA} we provide a video with the tracking results of the top methods for face detection followed by landmark localisation using re-initialisation in case of failure (Section~\ref{exp:detection_init_from_previous}, Table~\ref{tab:exp_detection_init_from_previous}, Figure~\ref{fig:exp_detection_init_from_previous}) for qualitative comparison.}\textsuperscript{,}\footnote{\label{foot:tracking}In \url{https://www.youtube.com/watch?v=RXo9hZAaQVQ} we provide a video with the tracking results of the top methods for model free tracking followed by landmark localisation (Section~\ref{exp:tracking}, Table~\ref{tab:exp_tracking}, Figure~\ref{fig:exp_tracking}) for qualitative comparison.}.
%
%%%%%%%%%%%%%%%%%%%%%%%%%%%%%%%%%%%%%%%%%%%%%%%%%%%%%%%%%
%%%%%%%%%%%%% EXEMPLAR FRAMES PER CATEGORY %%%%%%%%%%%%%%
%%%%%%%%%%%%%%%%%%%%%%%%%%%%%%%%%%%%%%%%%%%%%%%%%%%%%%%%%
\begin{figure*}[!t]
\centering
\subfloat[][Category 1] {
    \includegraphics[width=0.09\linewidth]{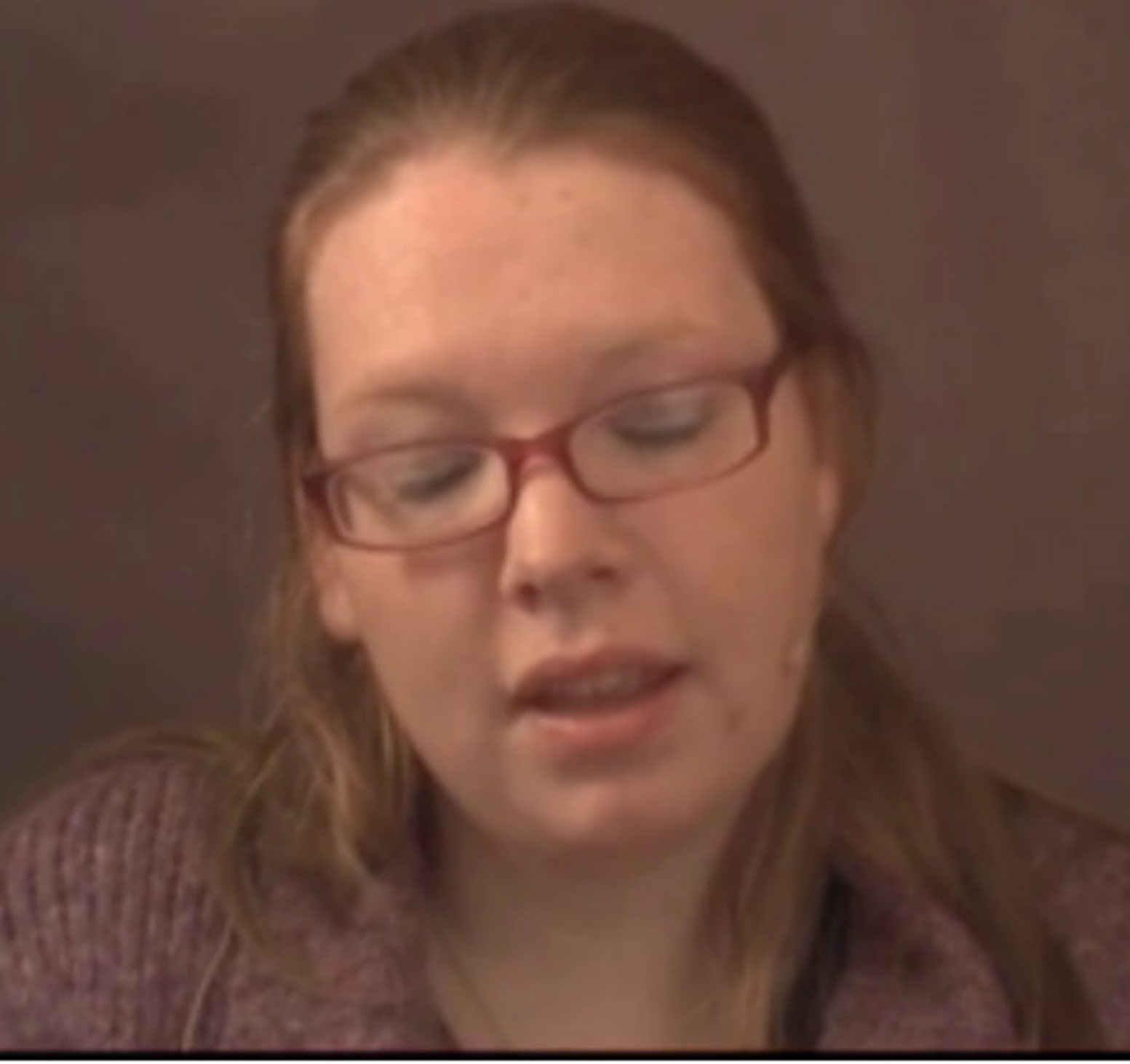}
    \includegraphics[width=0.09\linewidth]{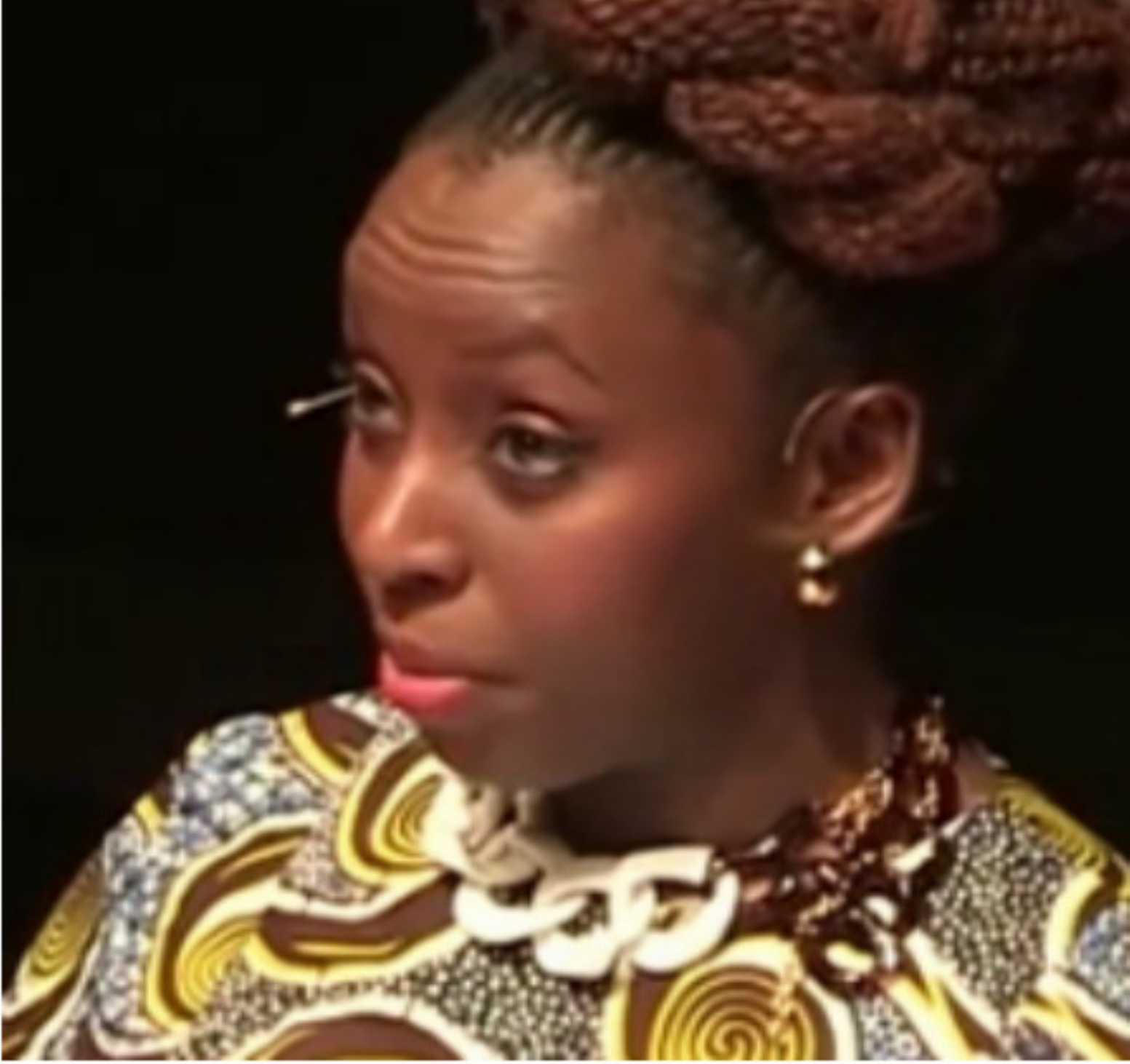}
    \includegraphics[width=0.09\linewidth]{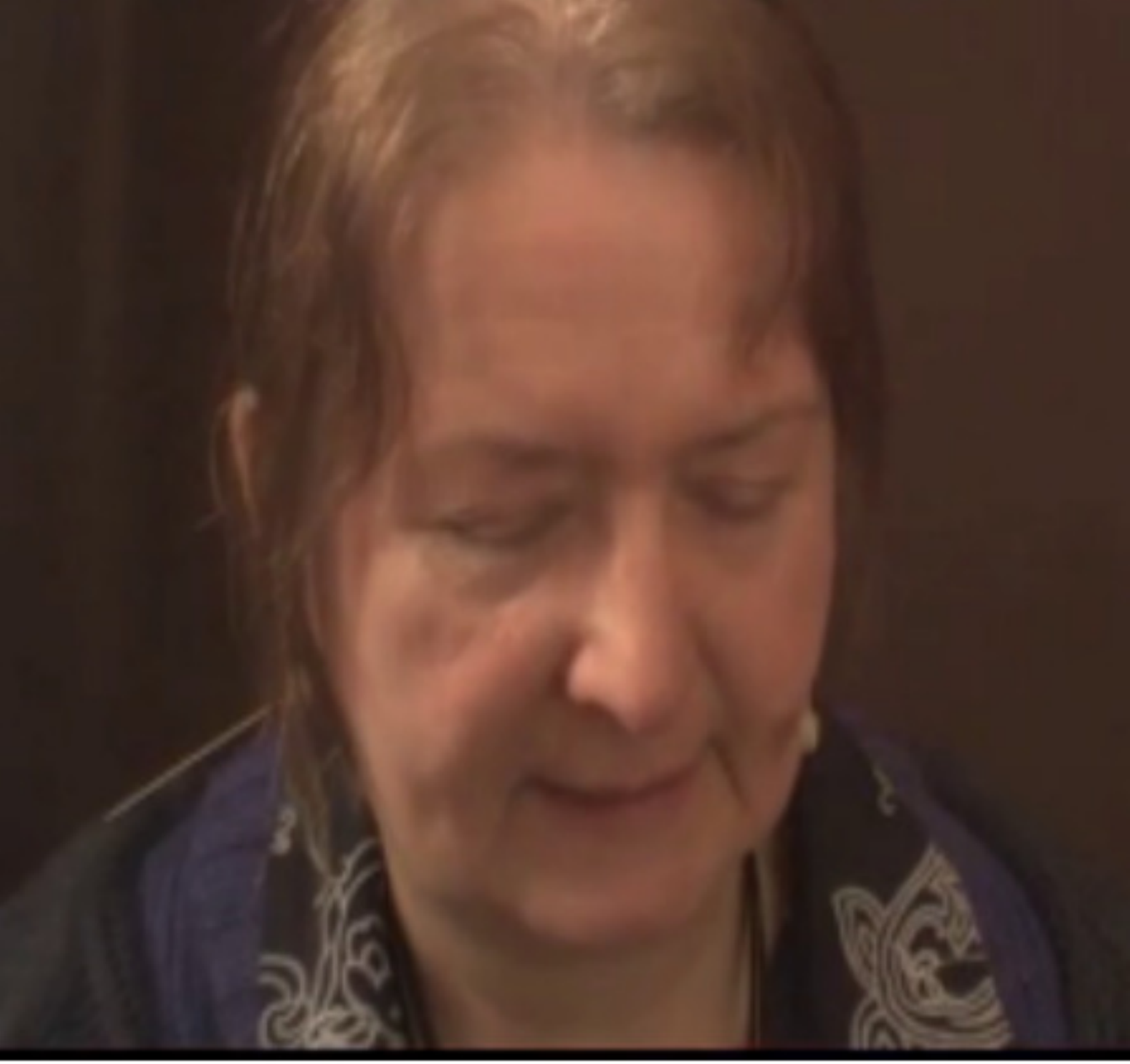}
    \includegraphics[width=0.09\linewidth]{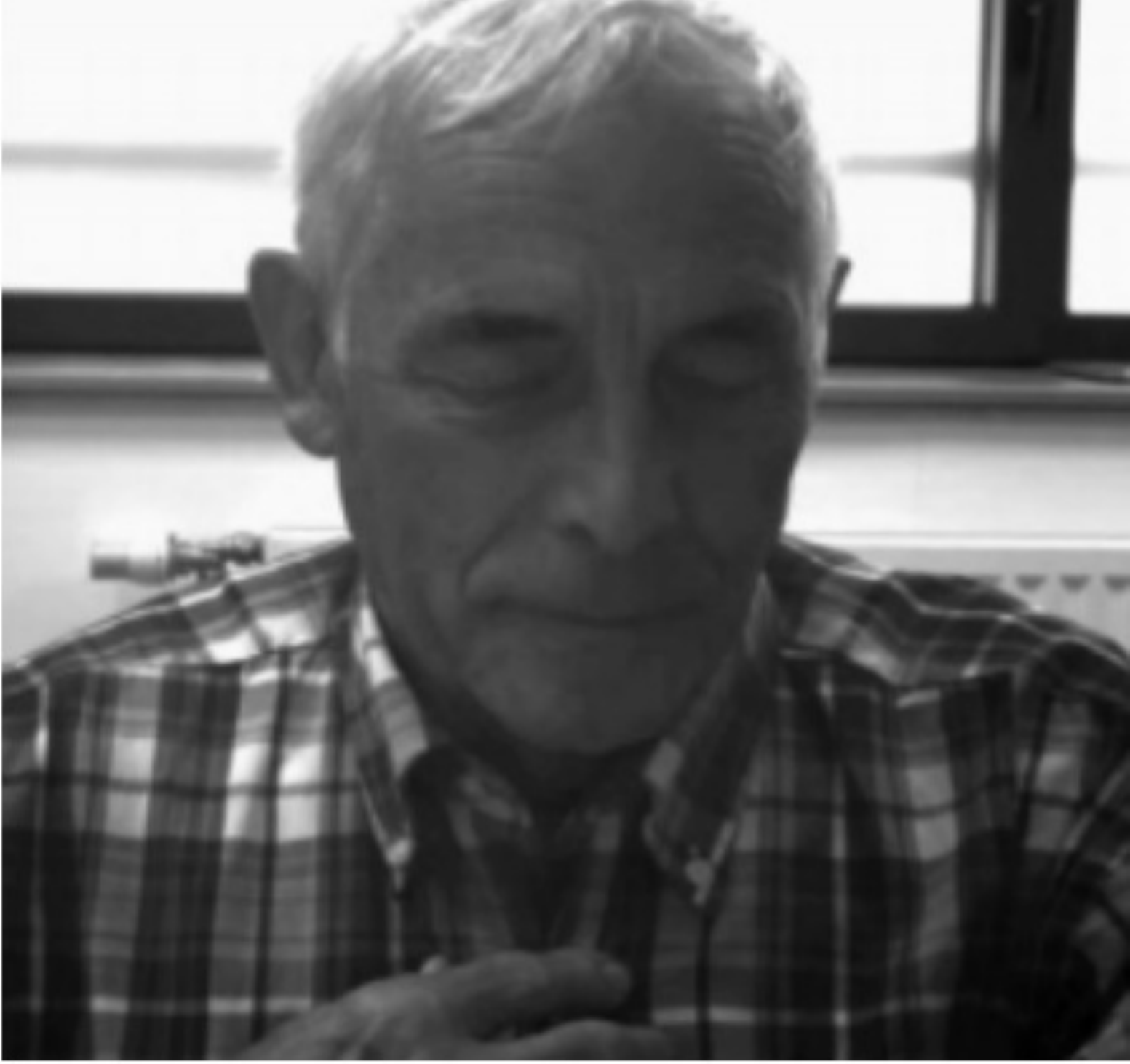}
    \includegraphics[width=0.09\linewidth]{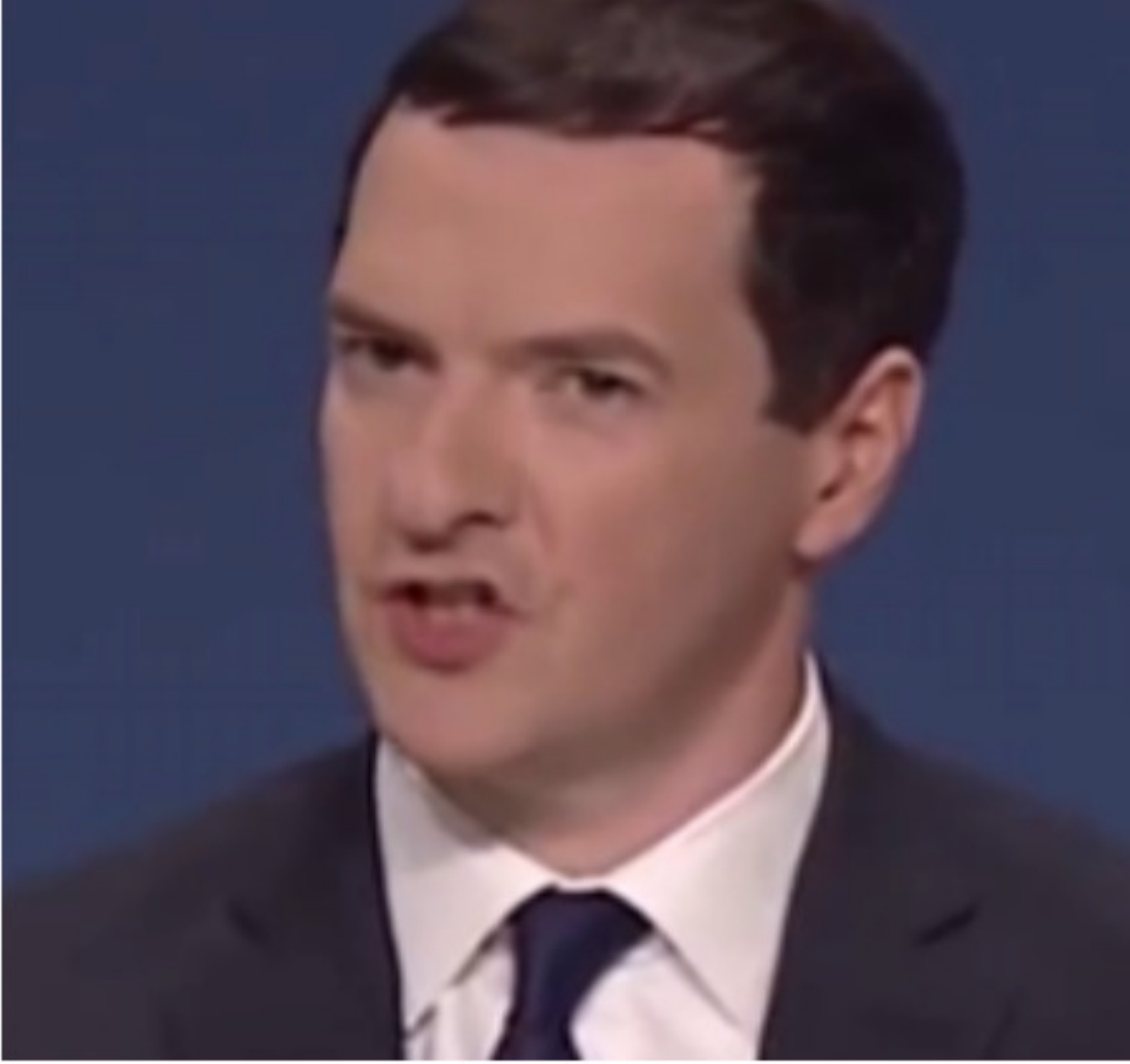}
    \includegraphics[width=0.09\linewidth]{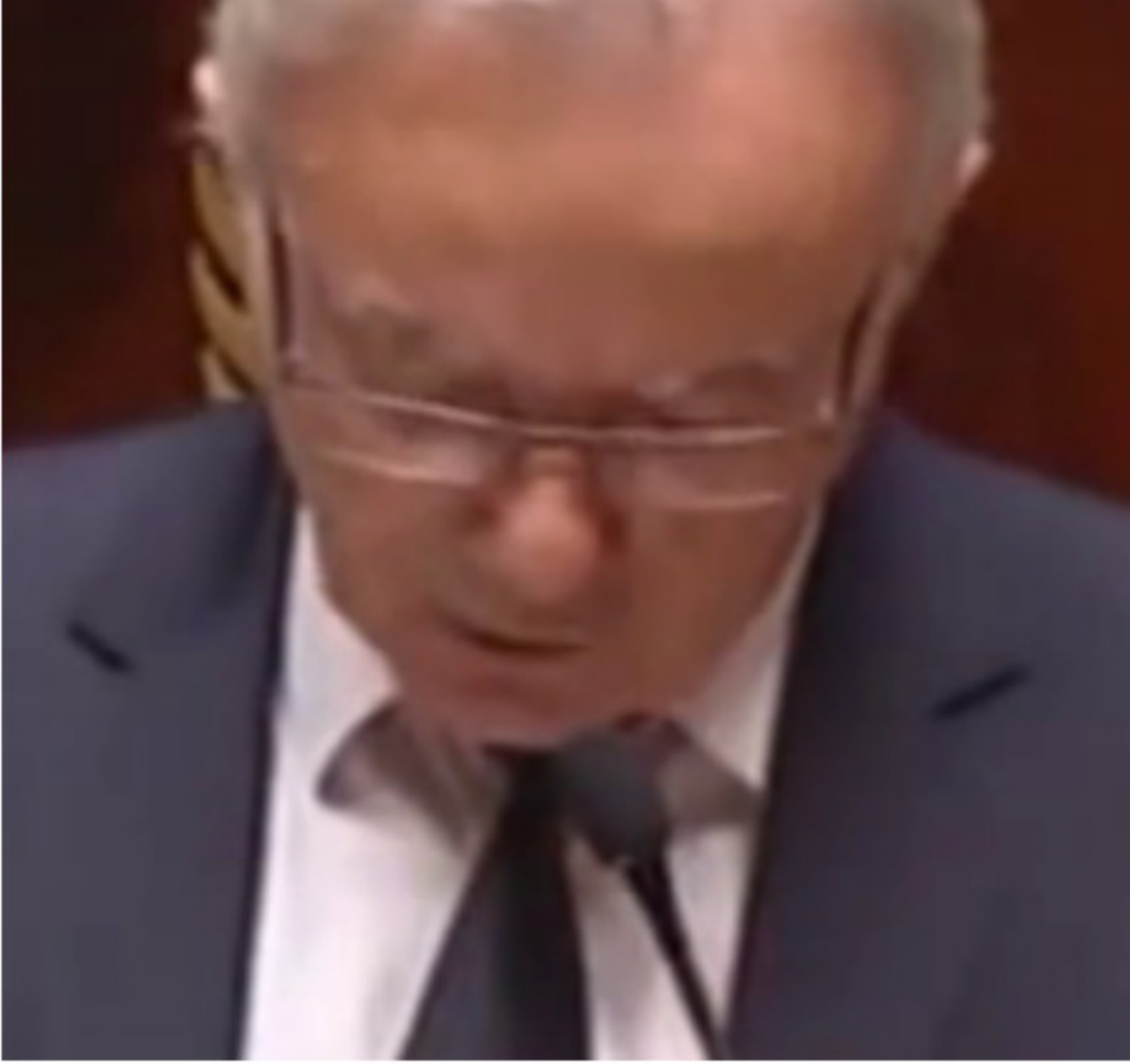}
    \includegraphics[width=0.09\linewidth]{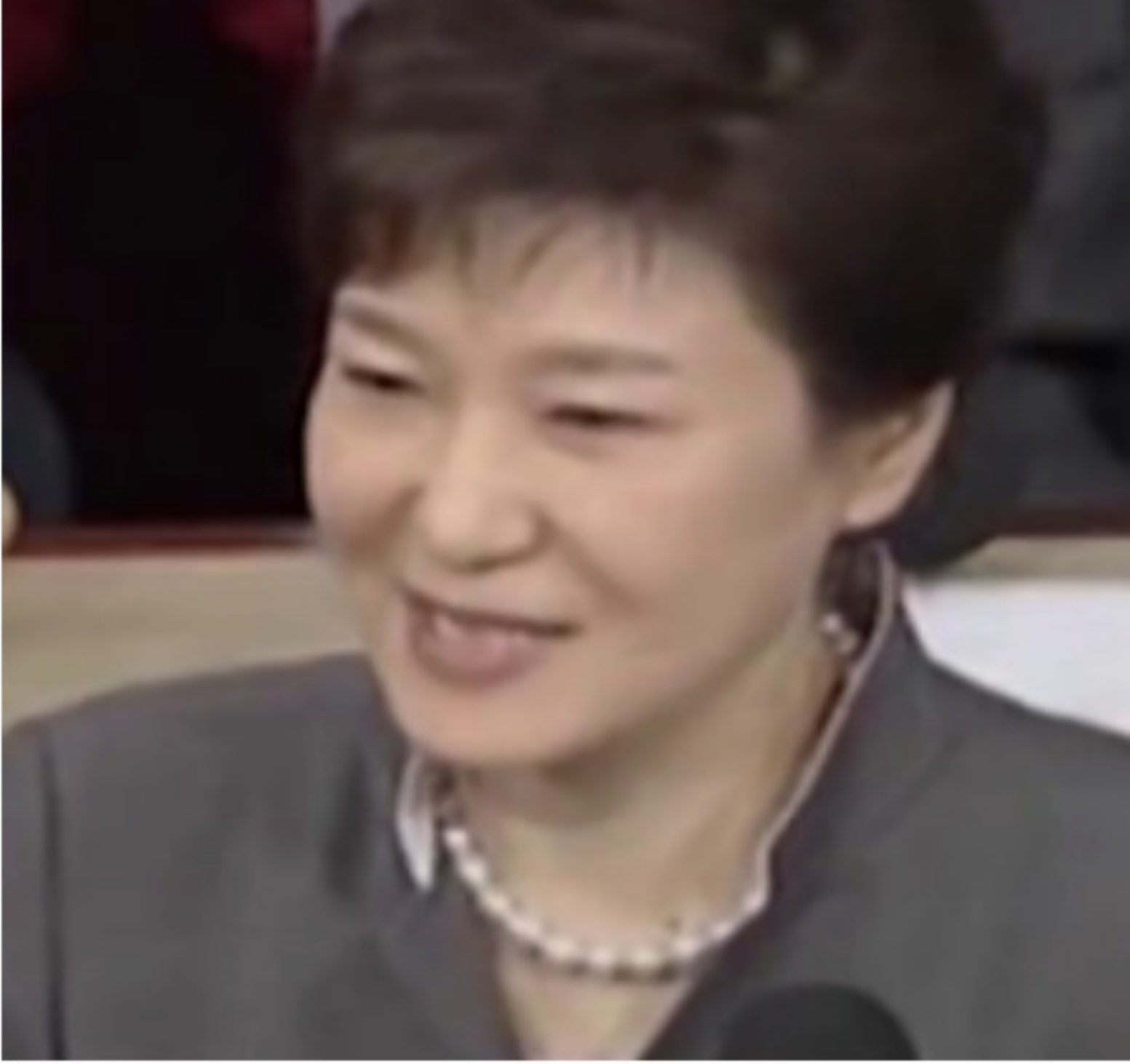}
    \includegraphics[width=0.09\linewidth]{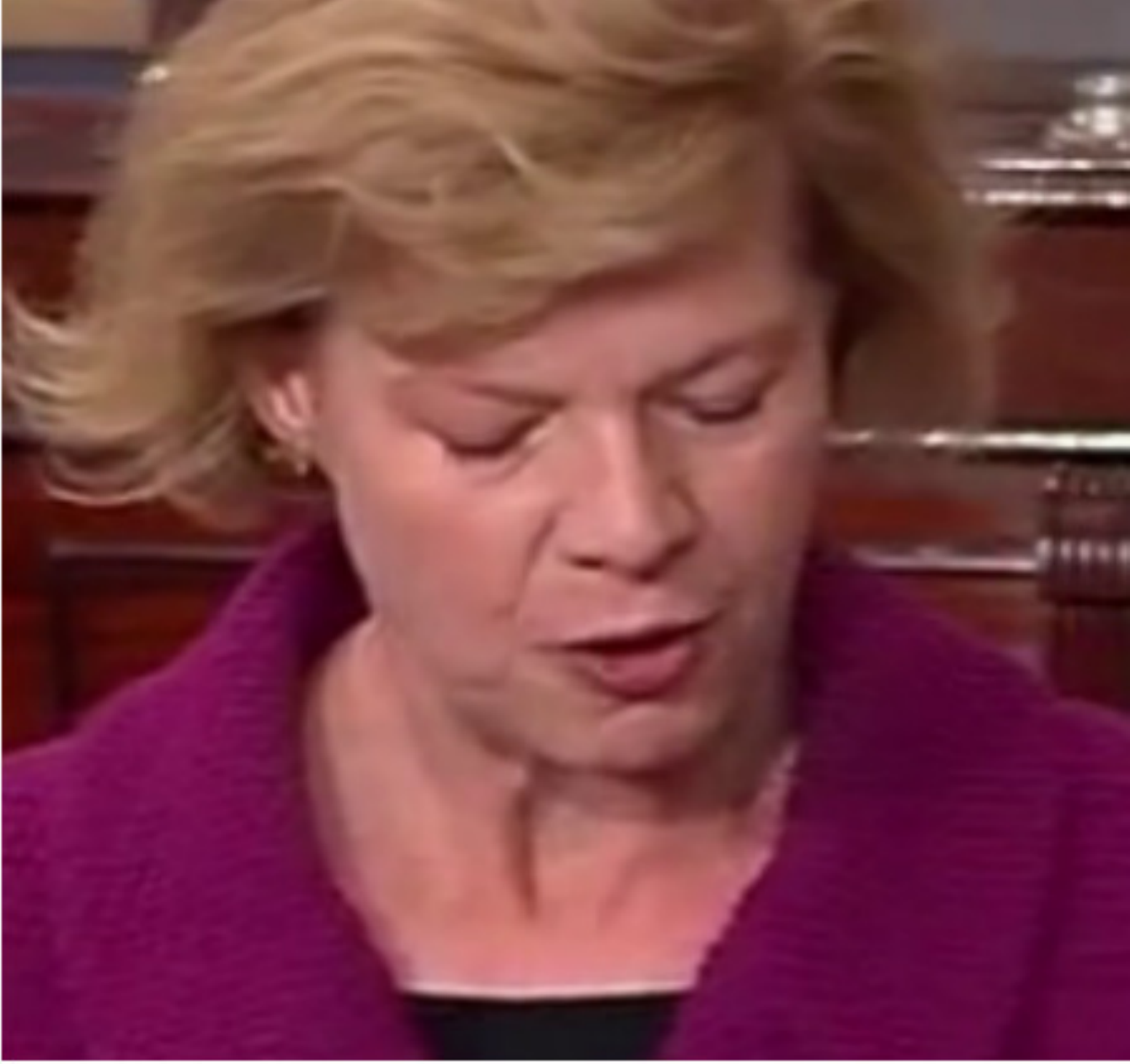}
    \includegraphics[width=0.09\linewidth]{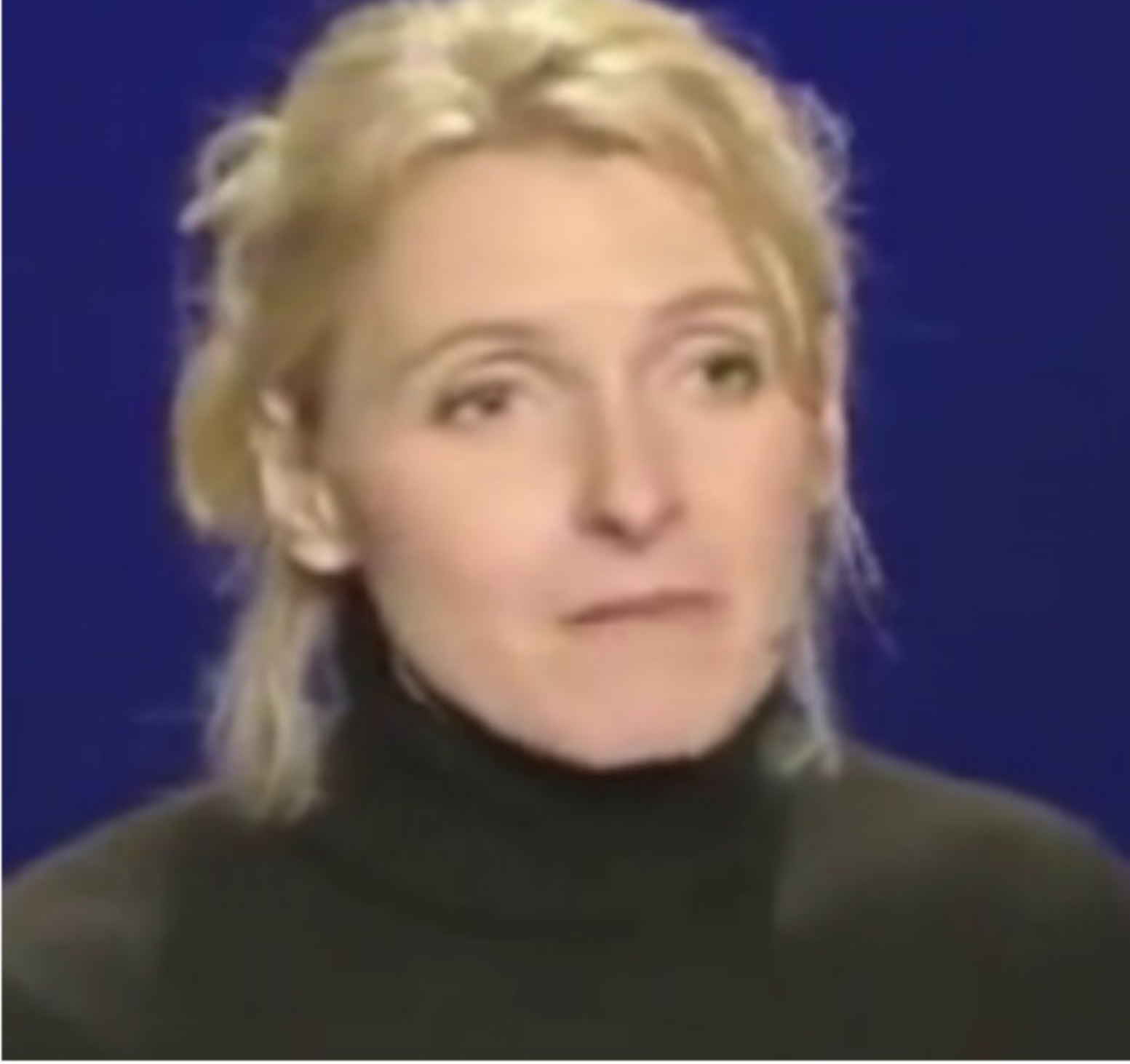}
    \includegraphics[width=0.09\linewidth]{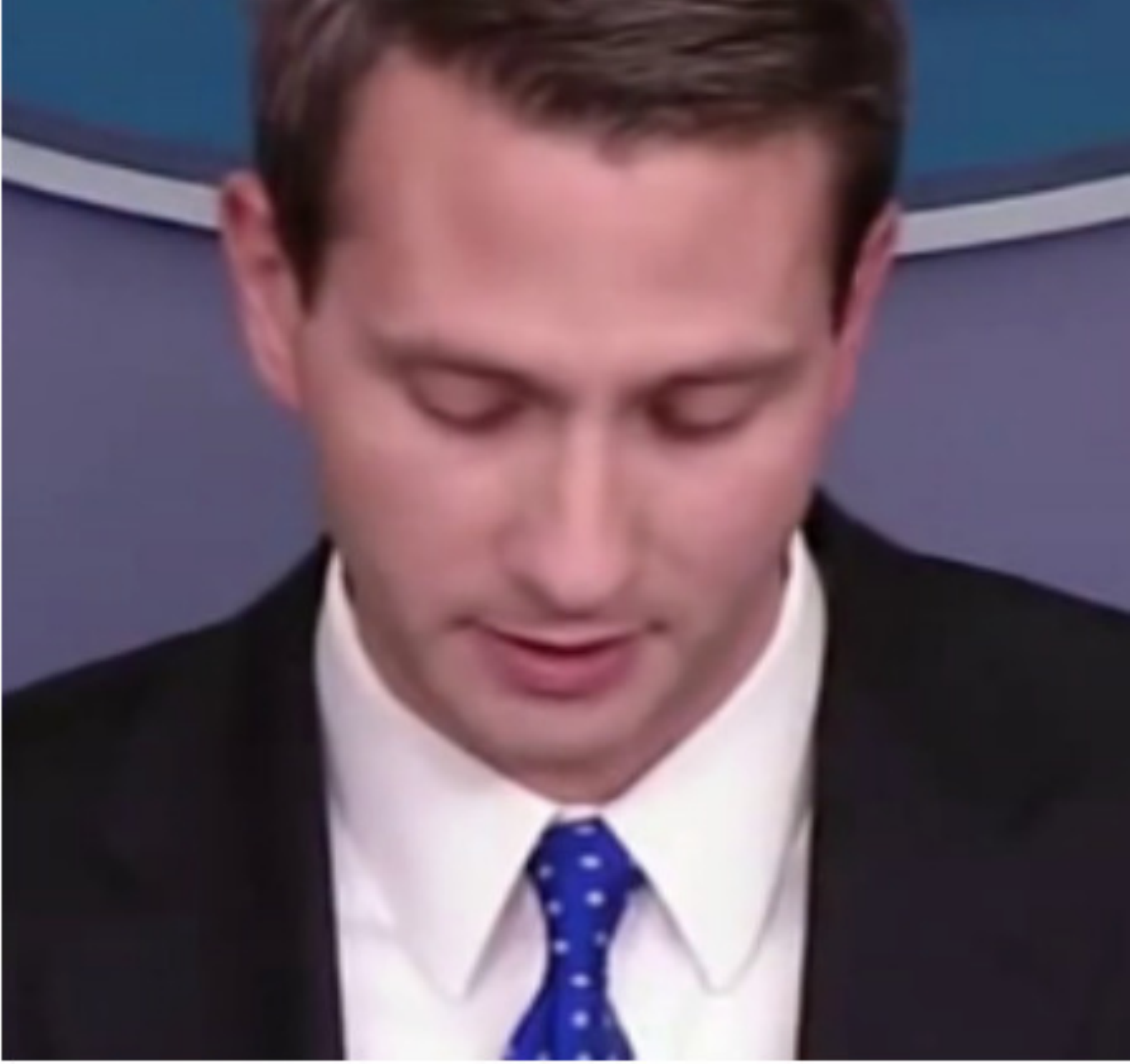}
}\\
\subfloat[][Category 2] {
    \includegraphics[width=0.09\linewidth]{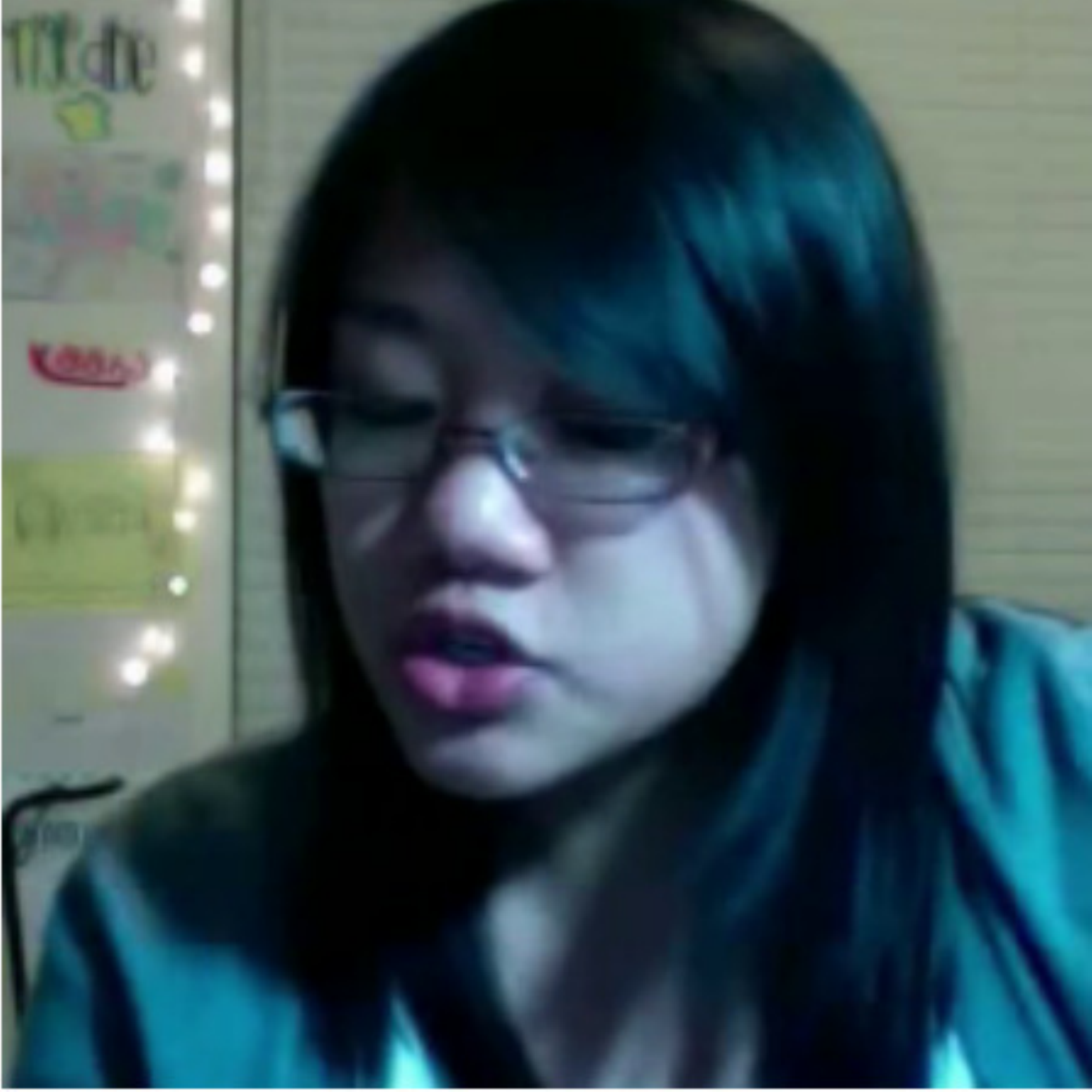}
    \includegraphics[width=0.09\linewidth]{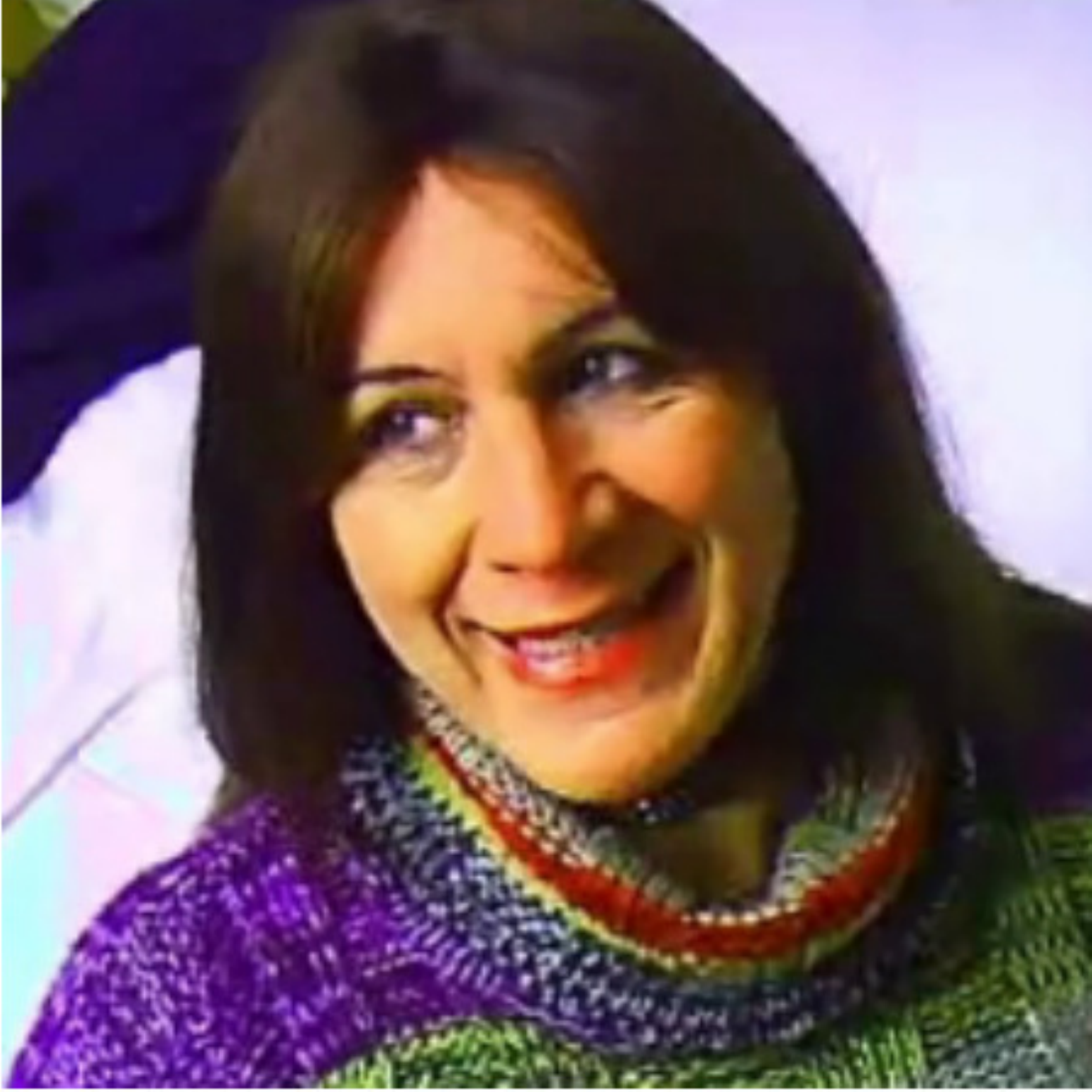}
    \includegraphics[width=0.09\linewidth]{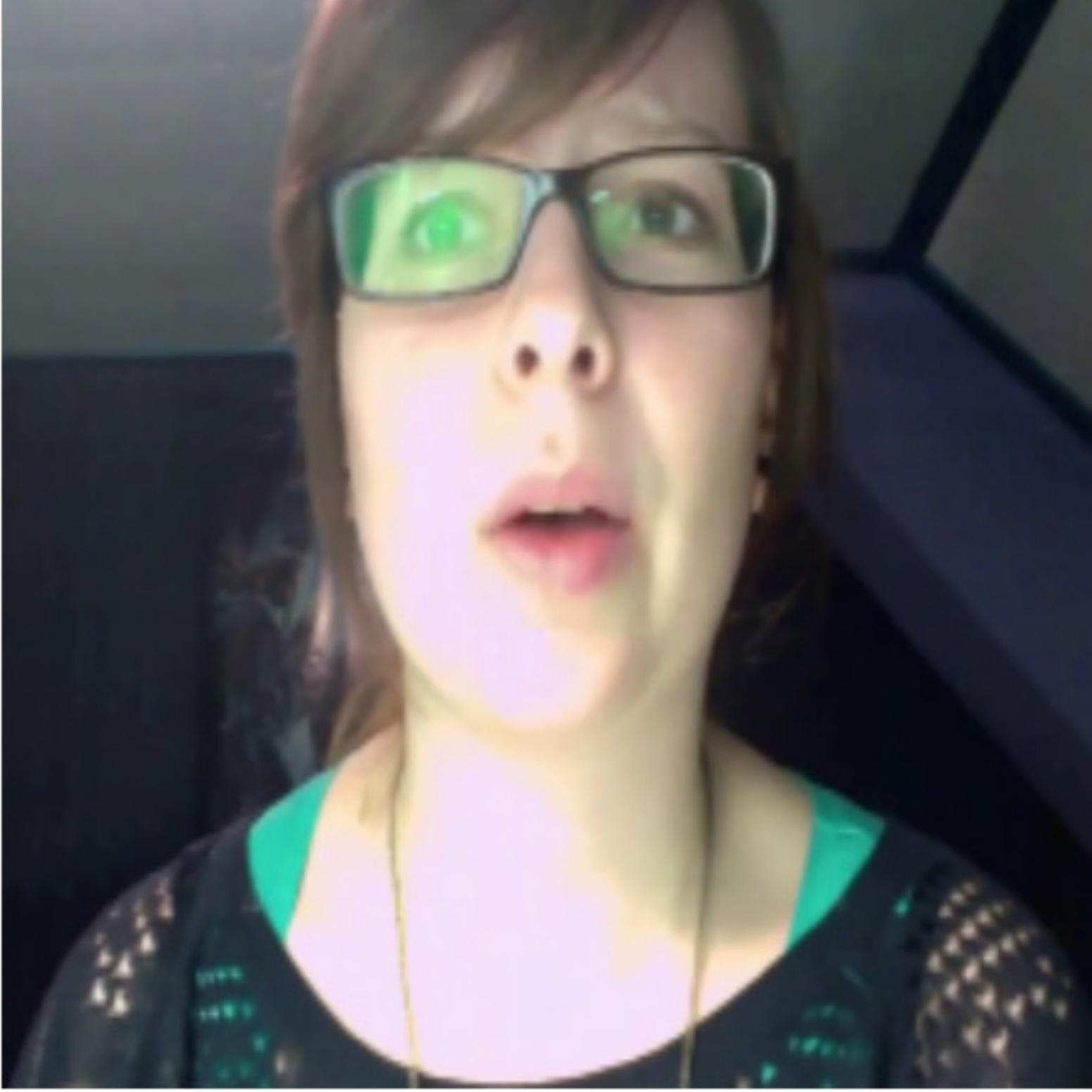}
    \includegraphics[width=0.09\linewidth]{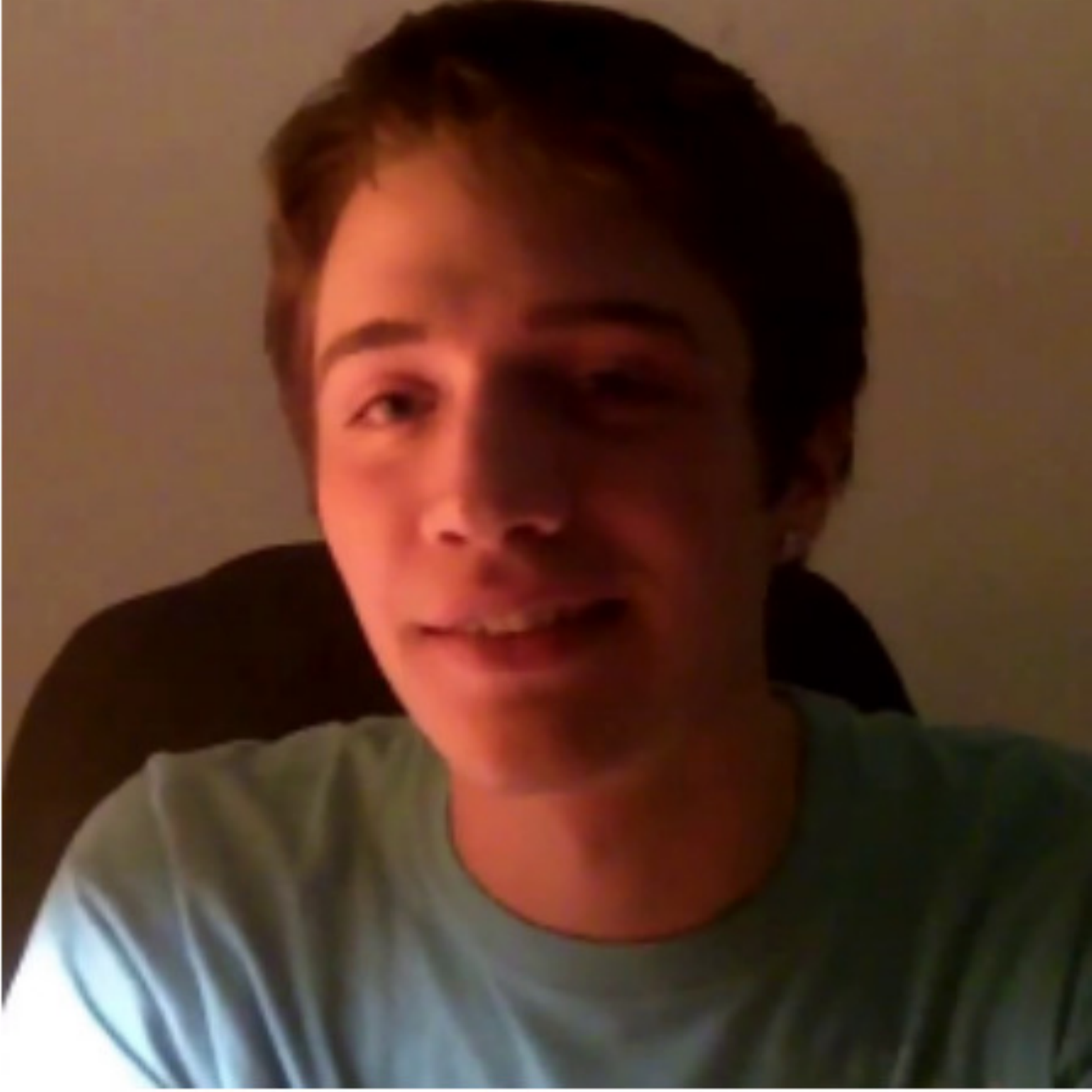}
    \includegraphics[width=0.09\linewidth]{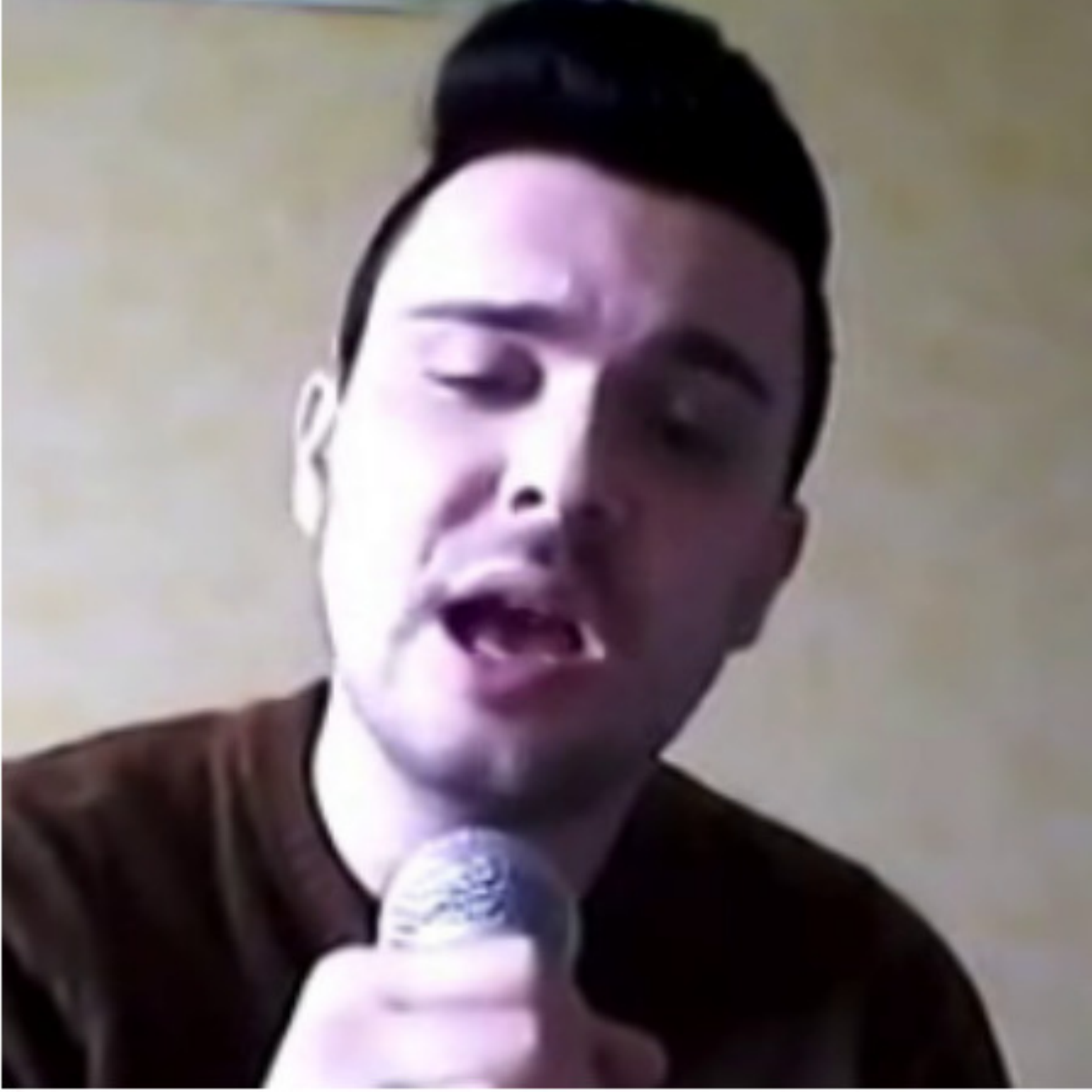}
    \includegraphics[width=0.09\linewidth]{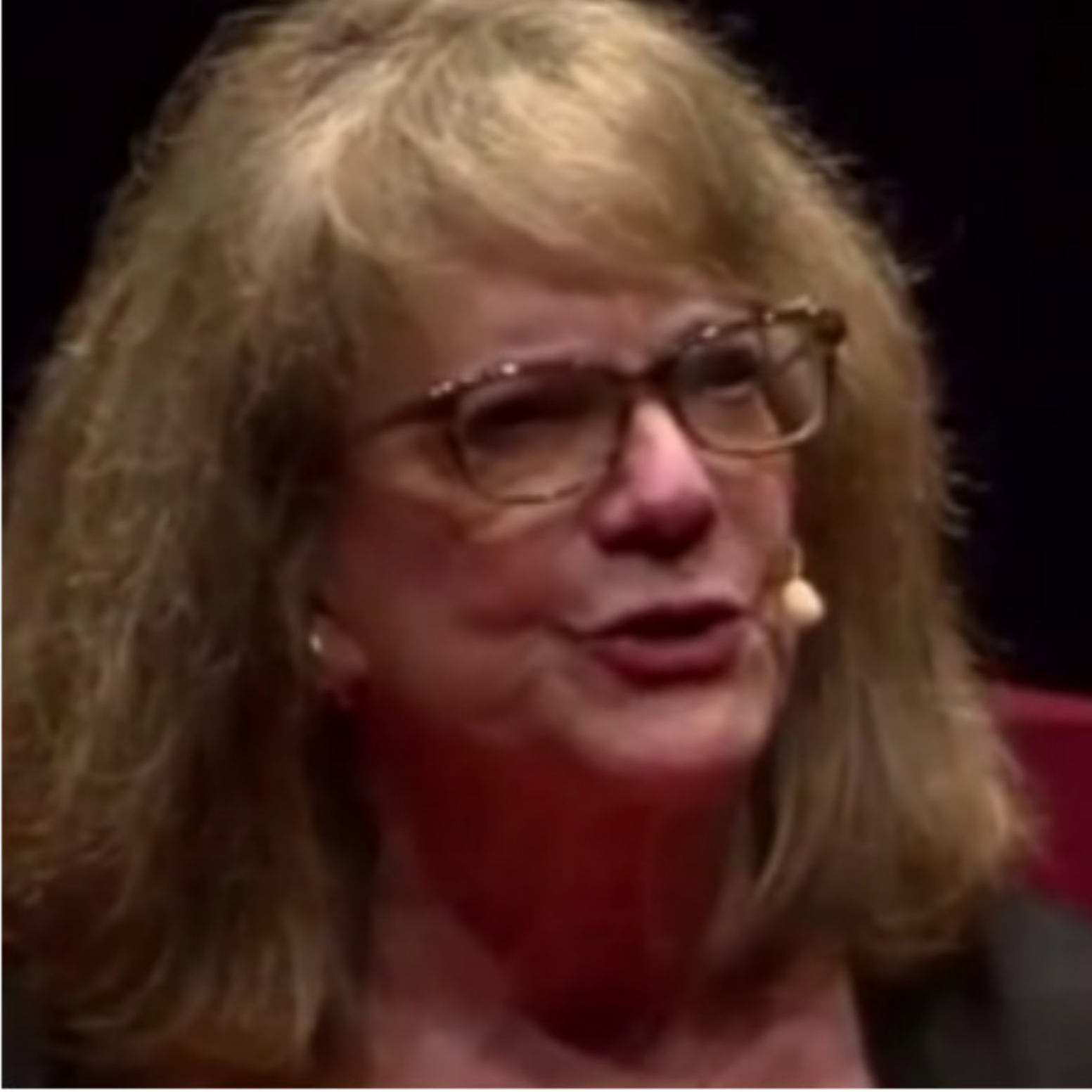}
    \includegraphics[width=0.09\linewidth]{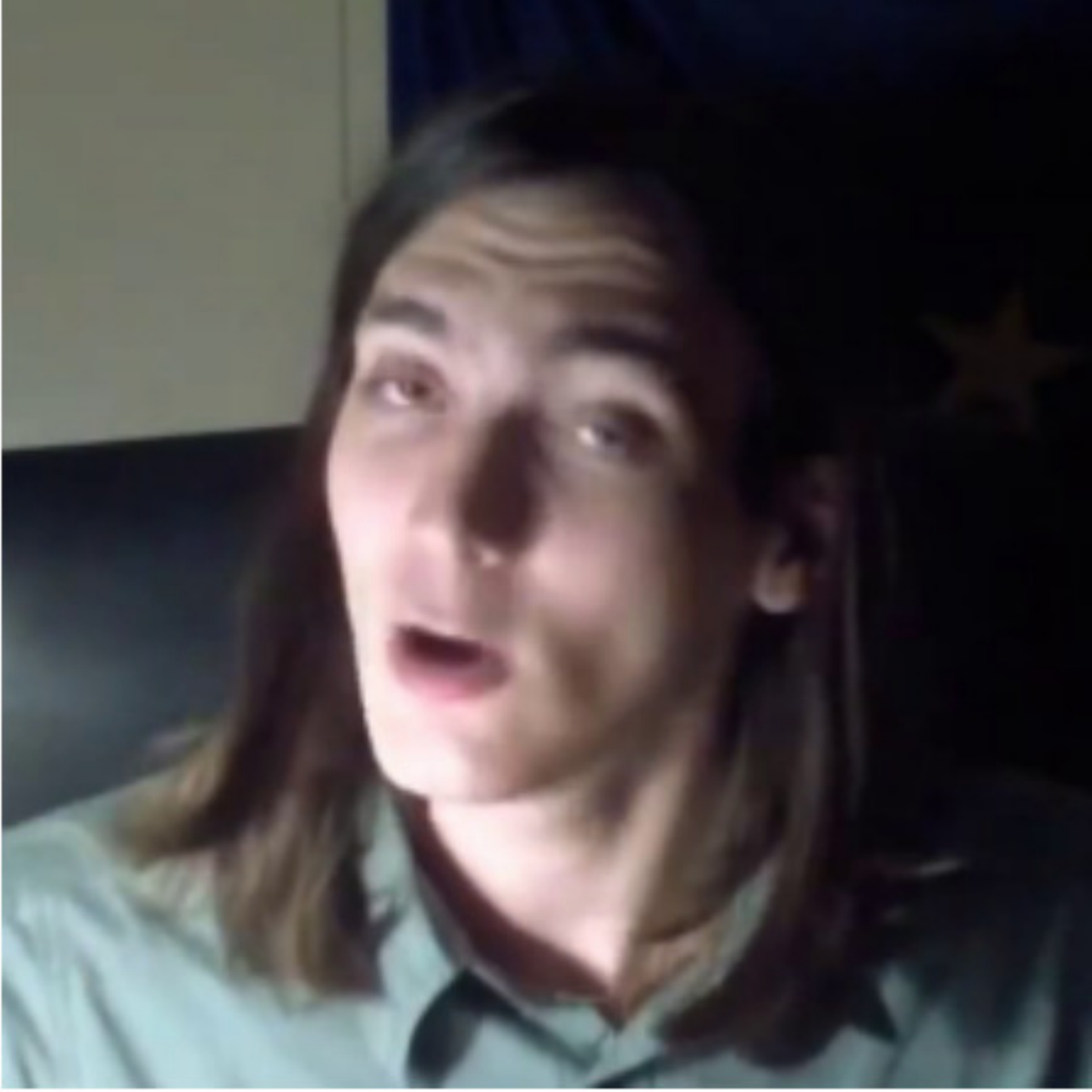}
    \includegraphics[width=0.09\linewidth]{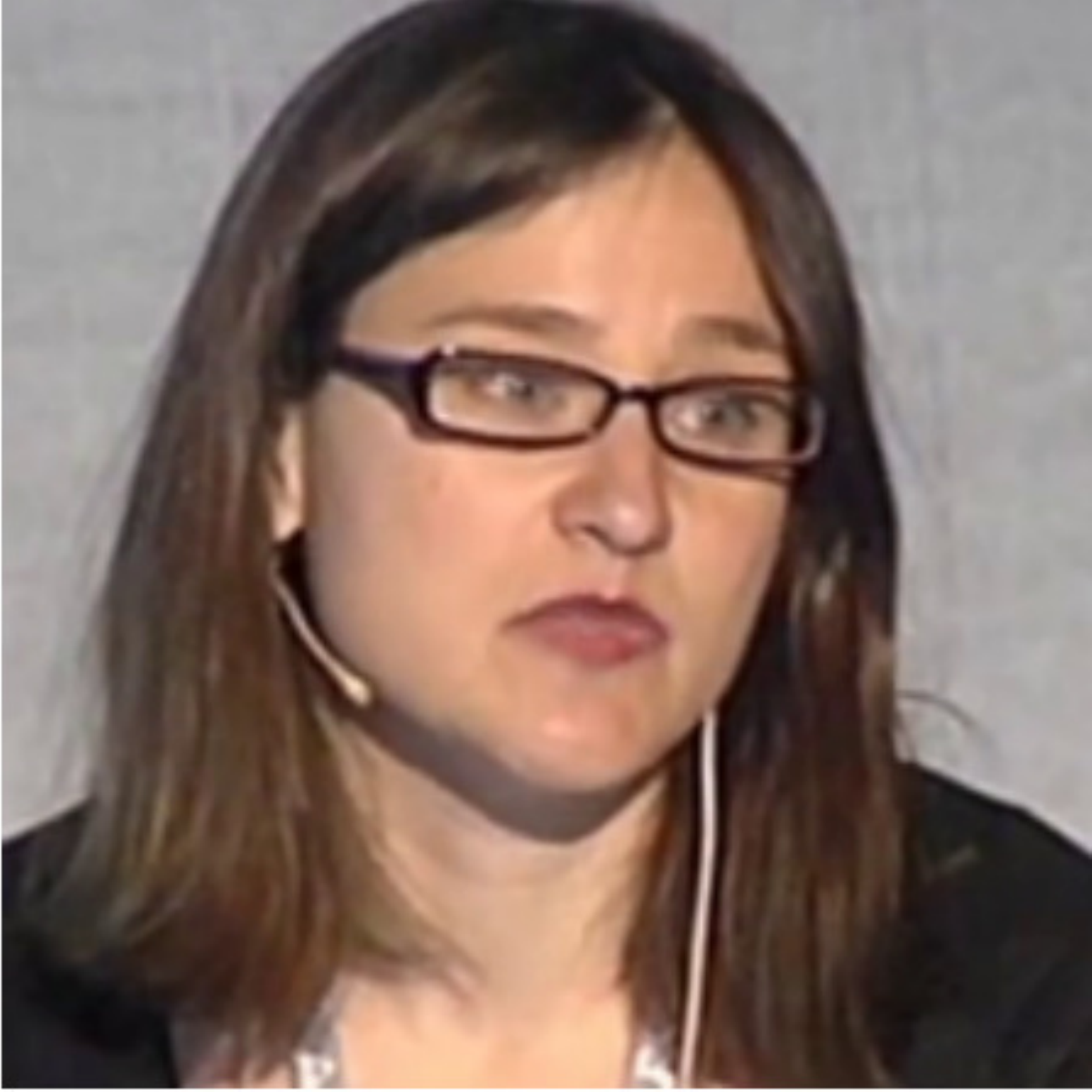}
    \includegraphics[width=0.09\linewidth]{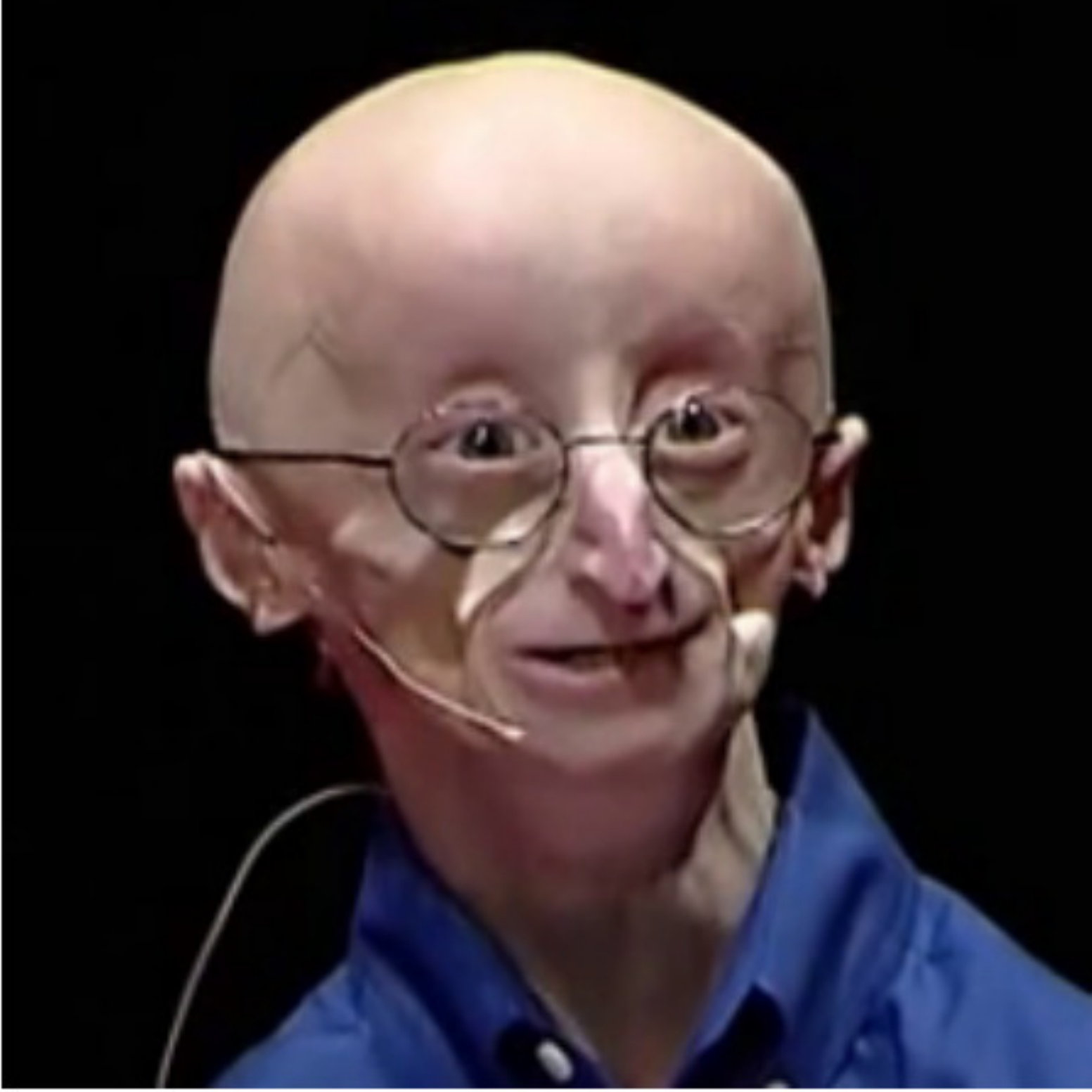}
    \includegraphics[width=0.09\linewidth]{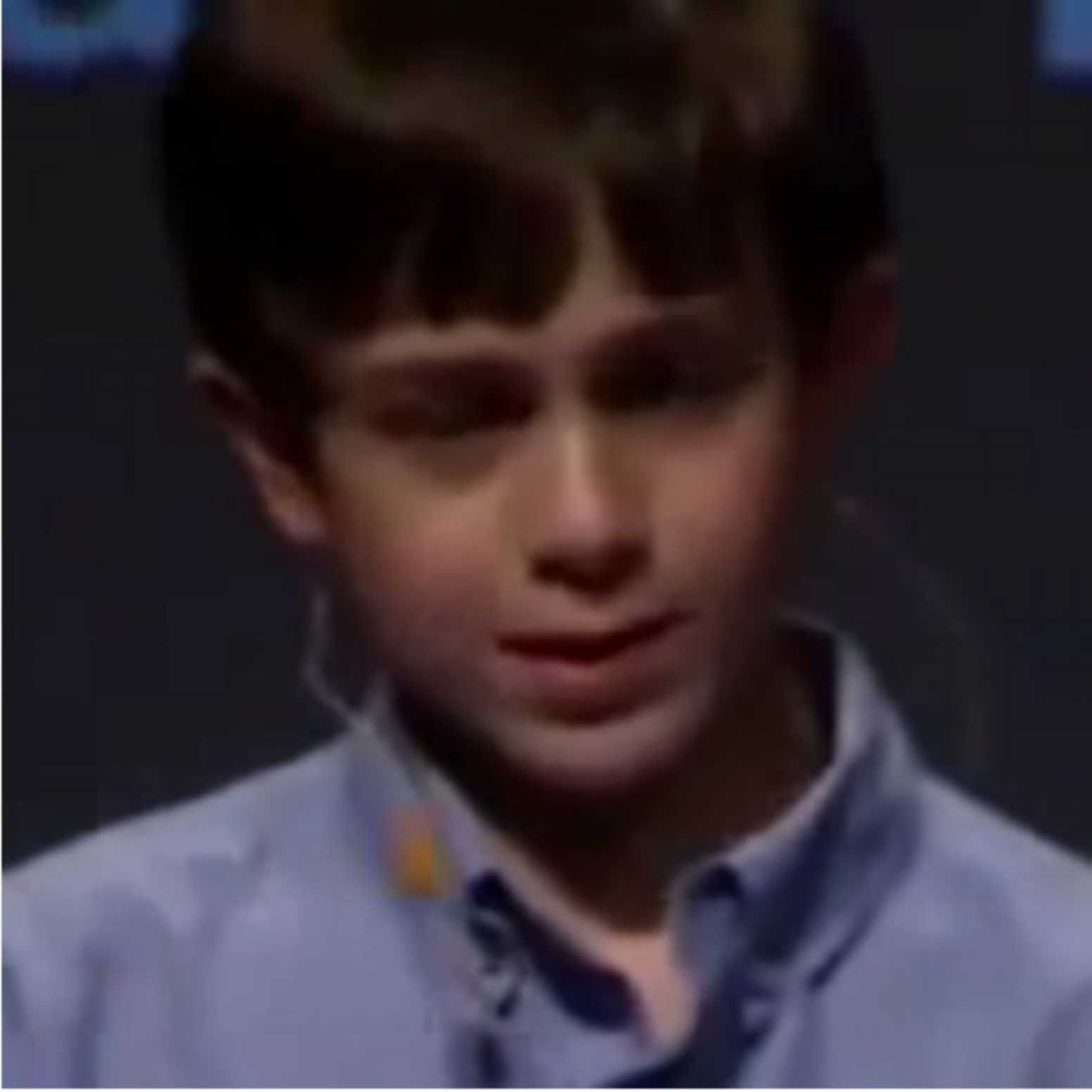}
}\\
\subfloat[][Category 3] {
    \includegraphics[width=0.09\linewidth]{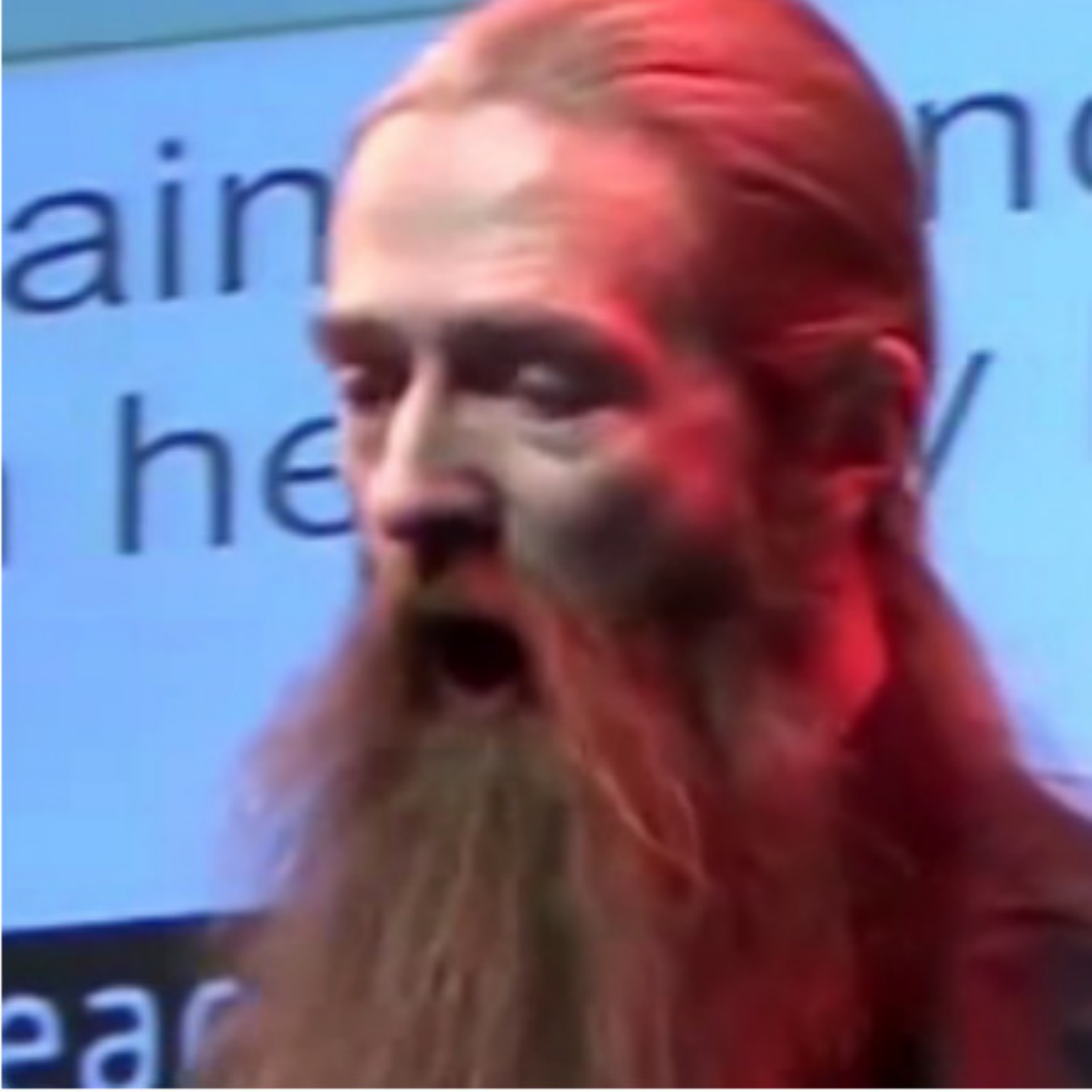}
    \includegraphics[width=0.09\linewidth]{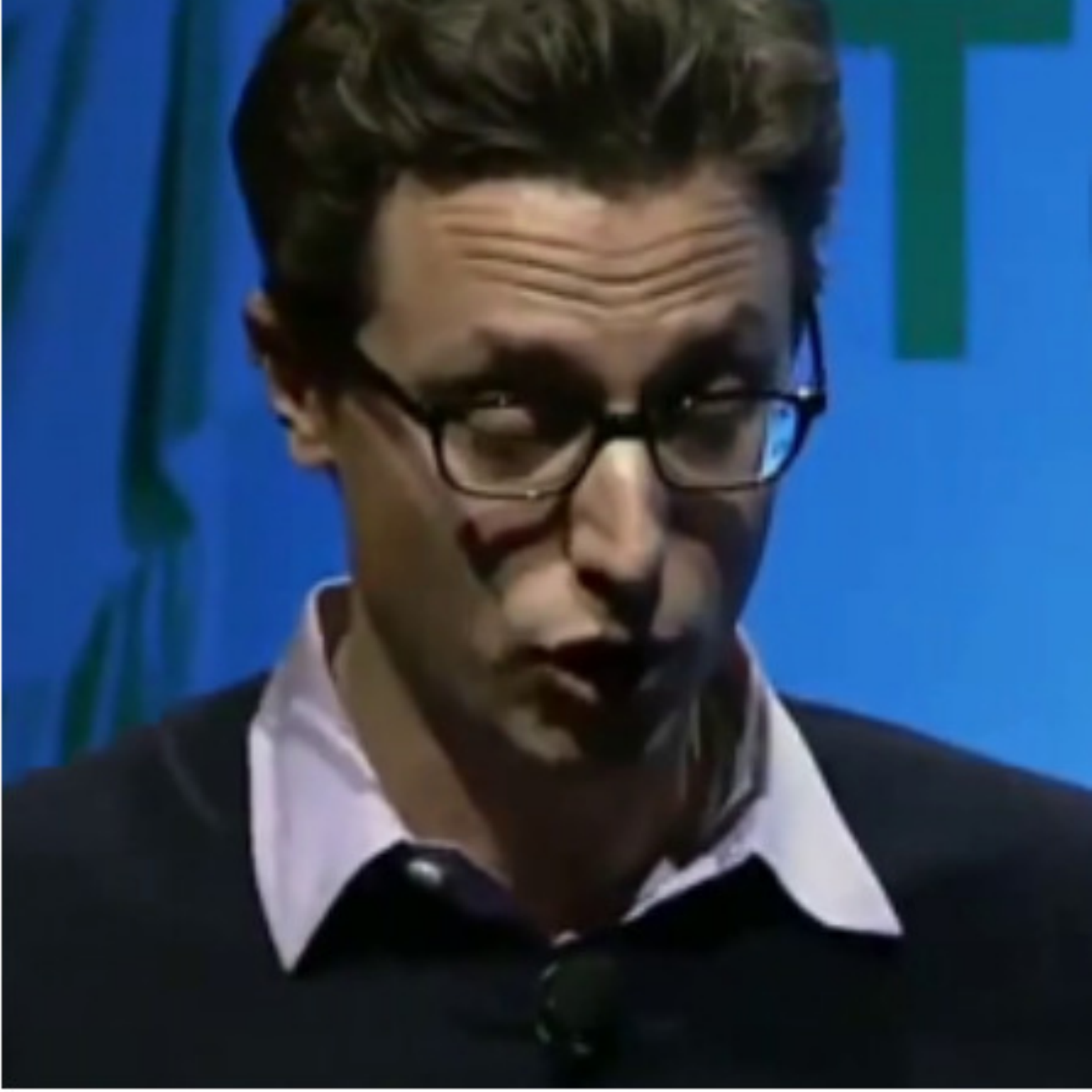}
    \includegraphics[width=0.09\linewidth]{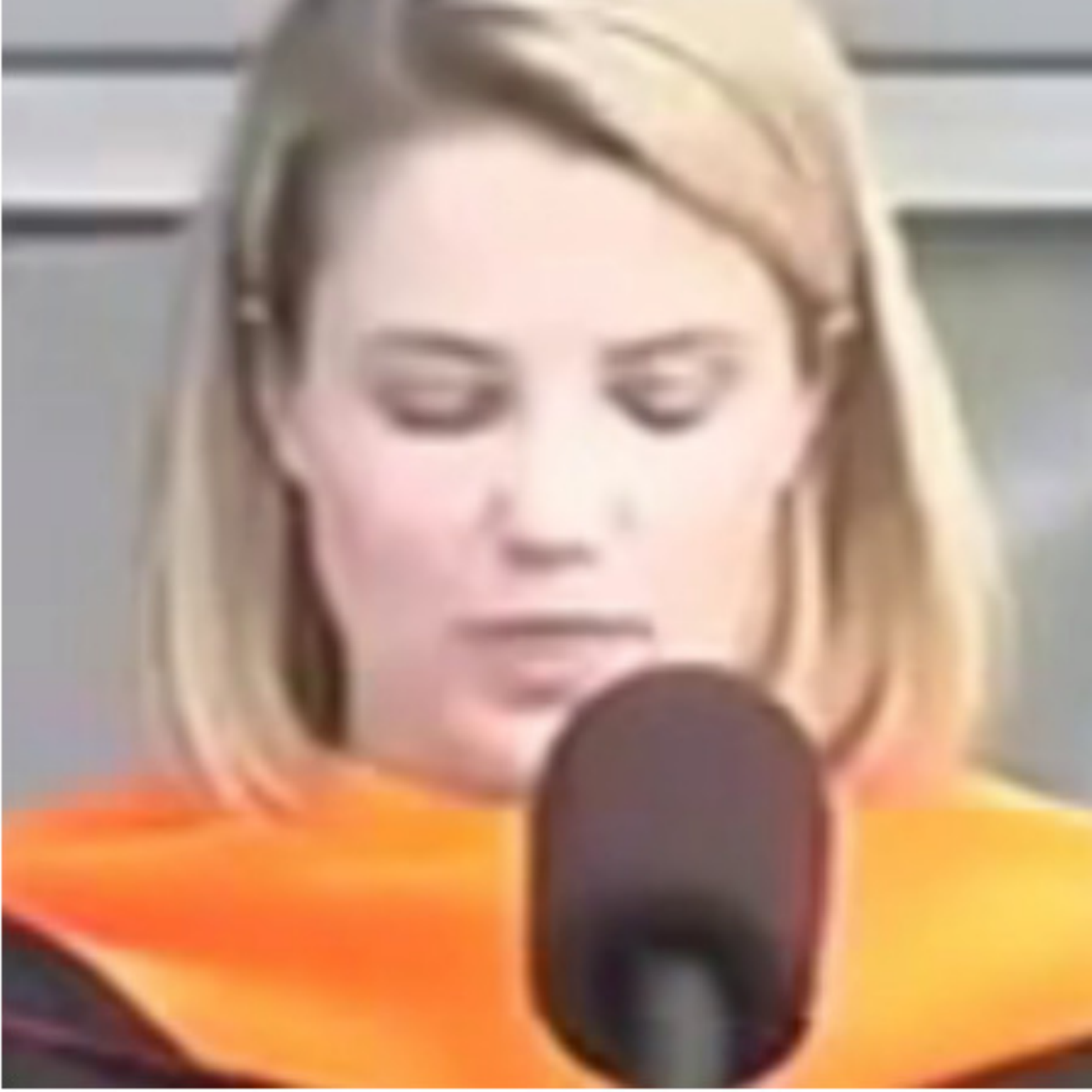}
    \includegraphics[width=0.09\linewidth]{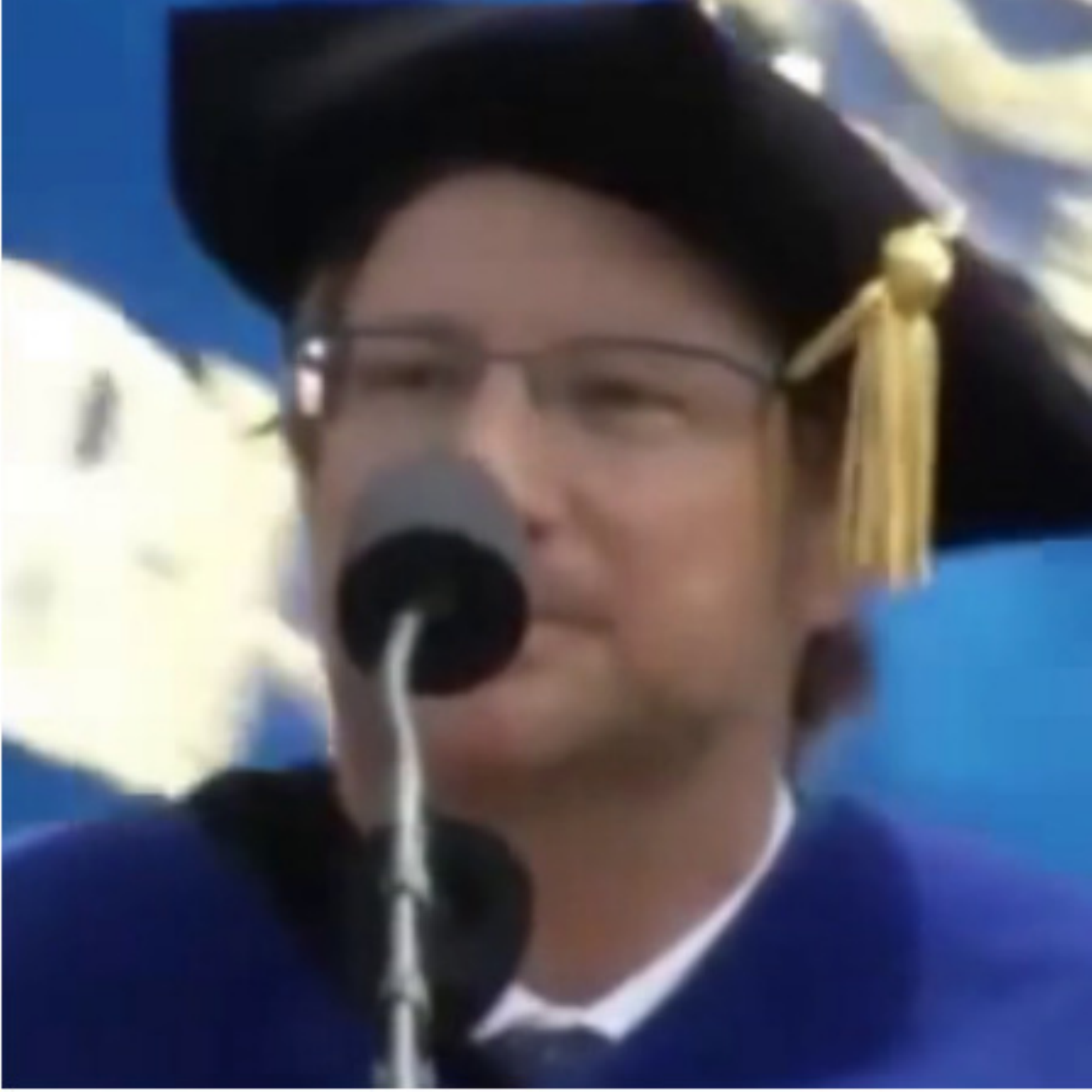}
    \includegraphics[width=0.09\linewidth]{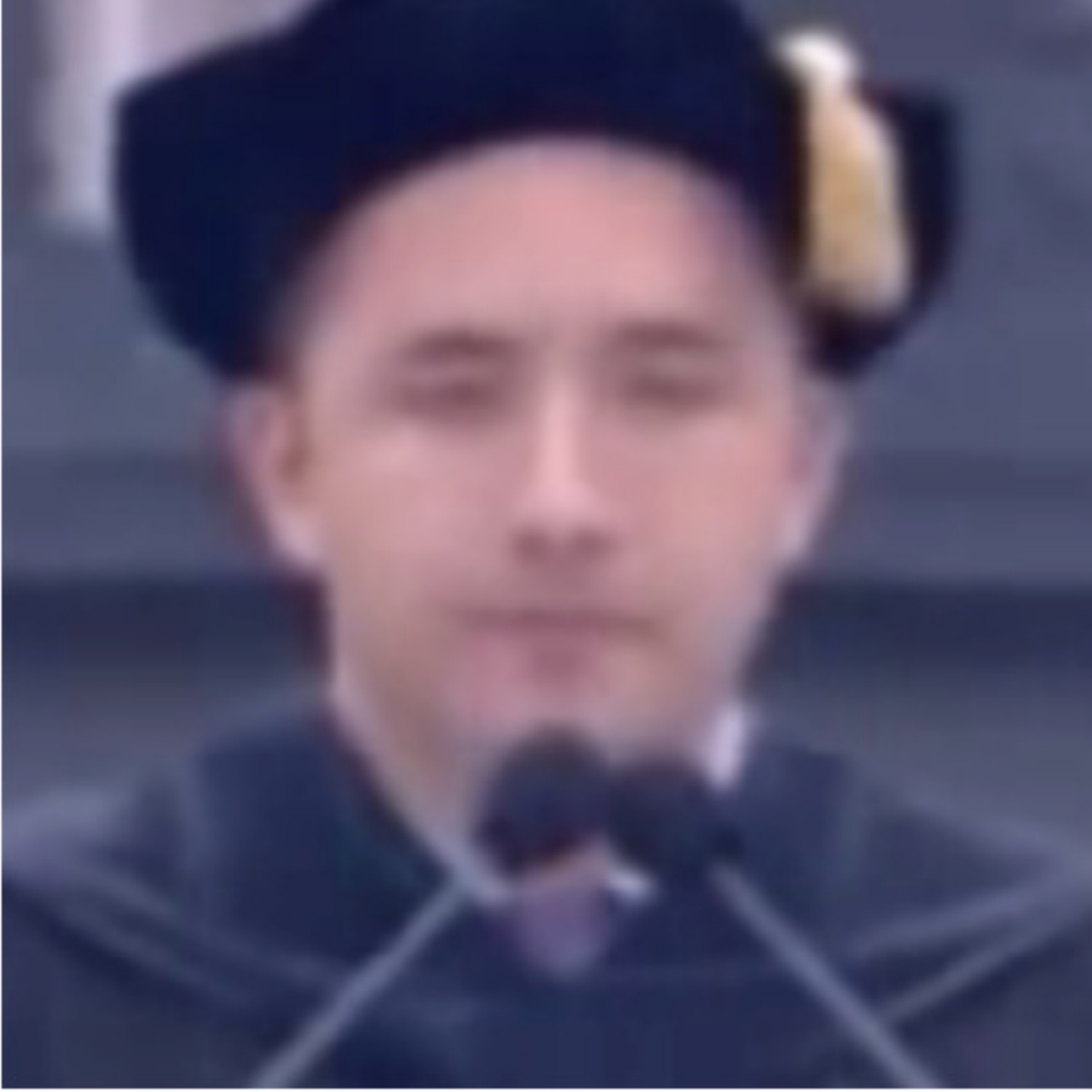}
    \includegraphics[width=0.09\linewidth]{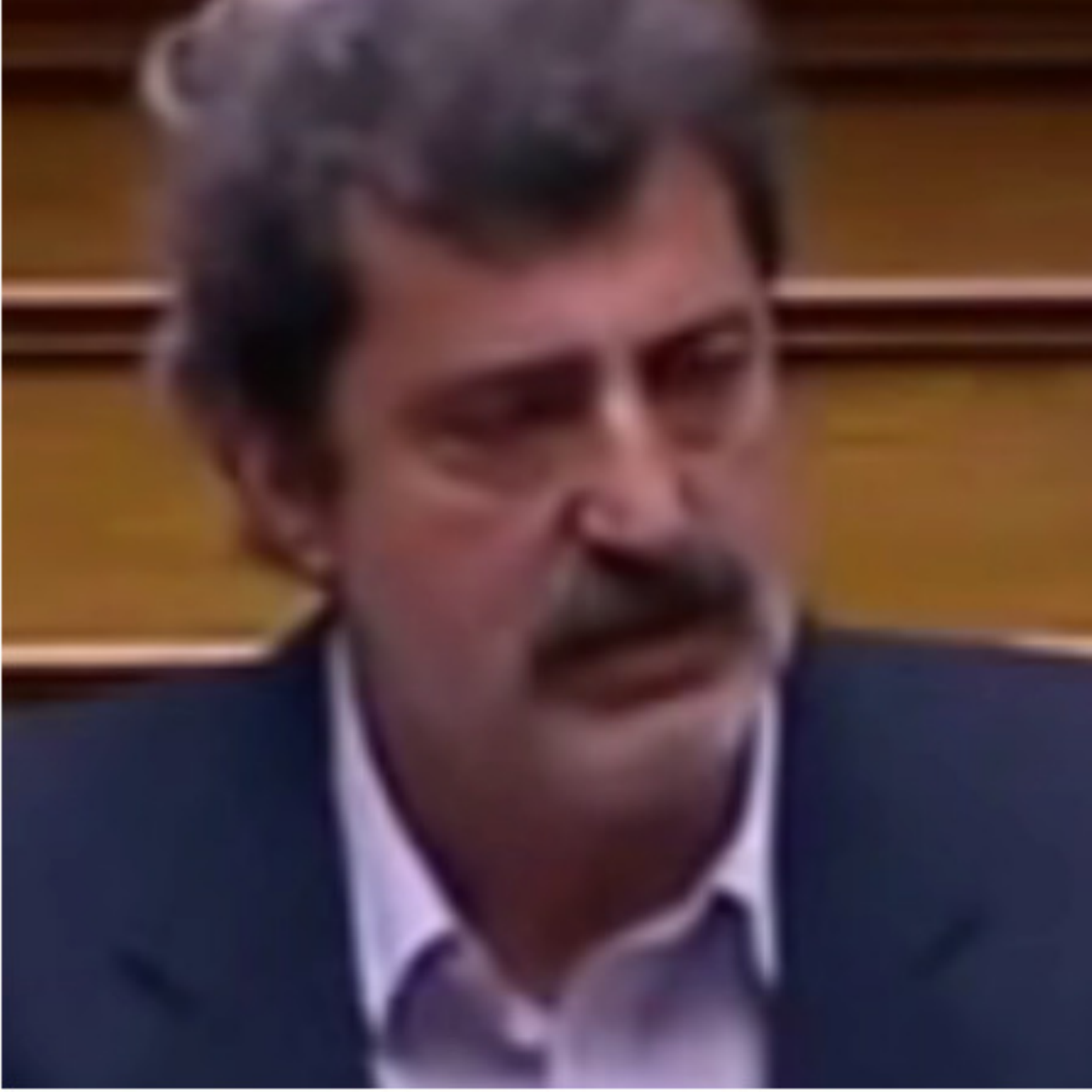}
    \includegraphics[width=0.09\linewidth]{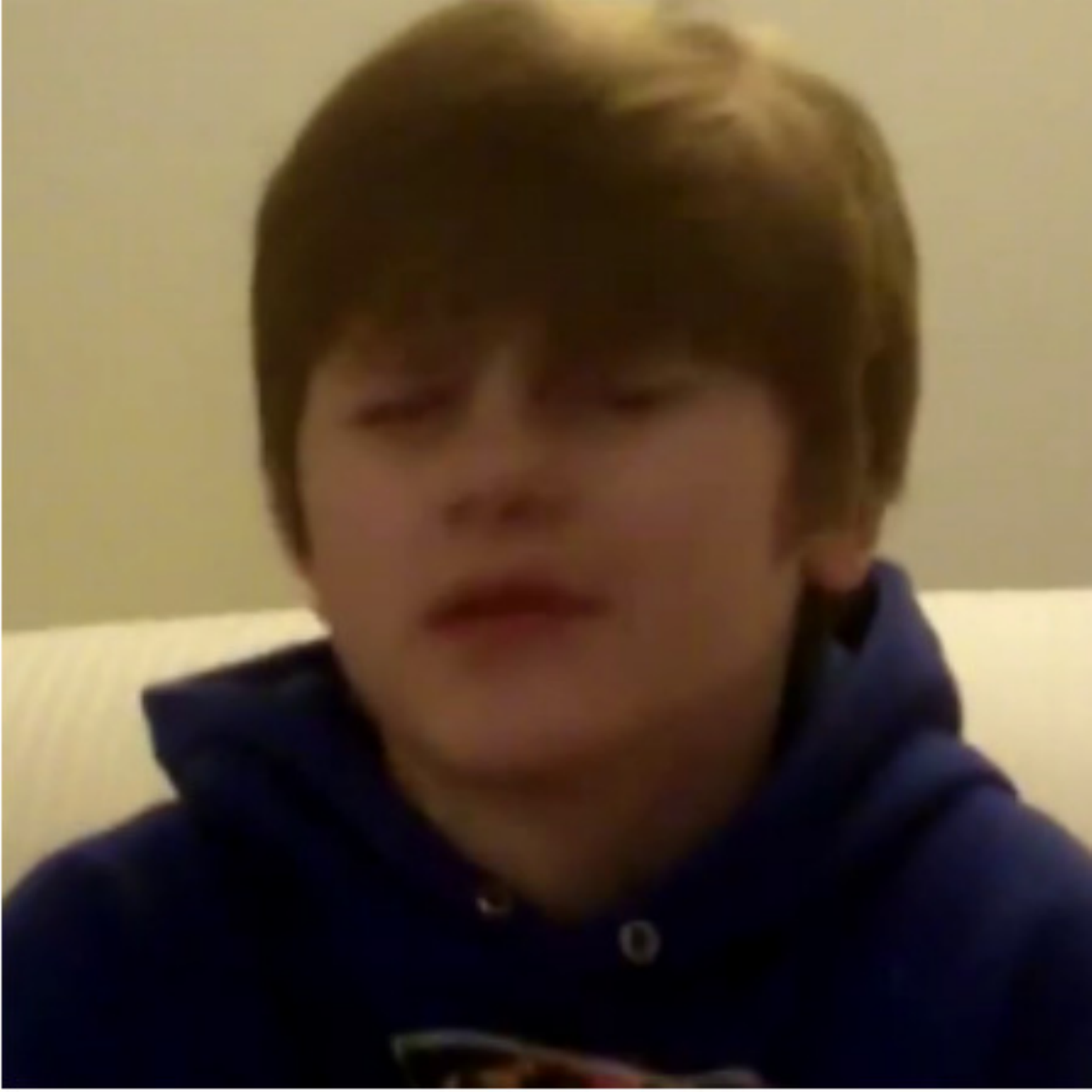}
    \includegraphics[width=0.09\linewidth]{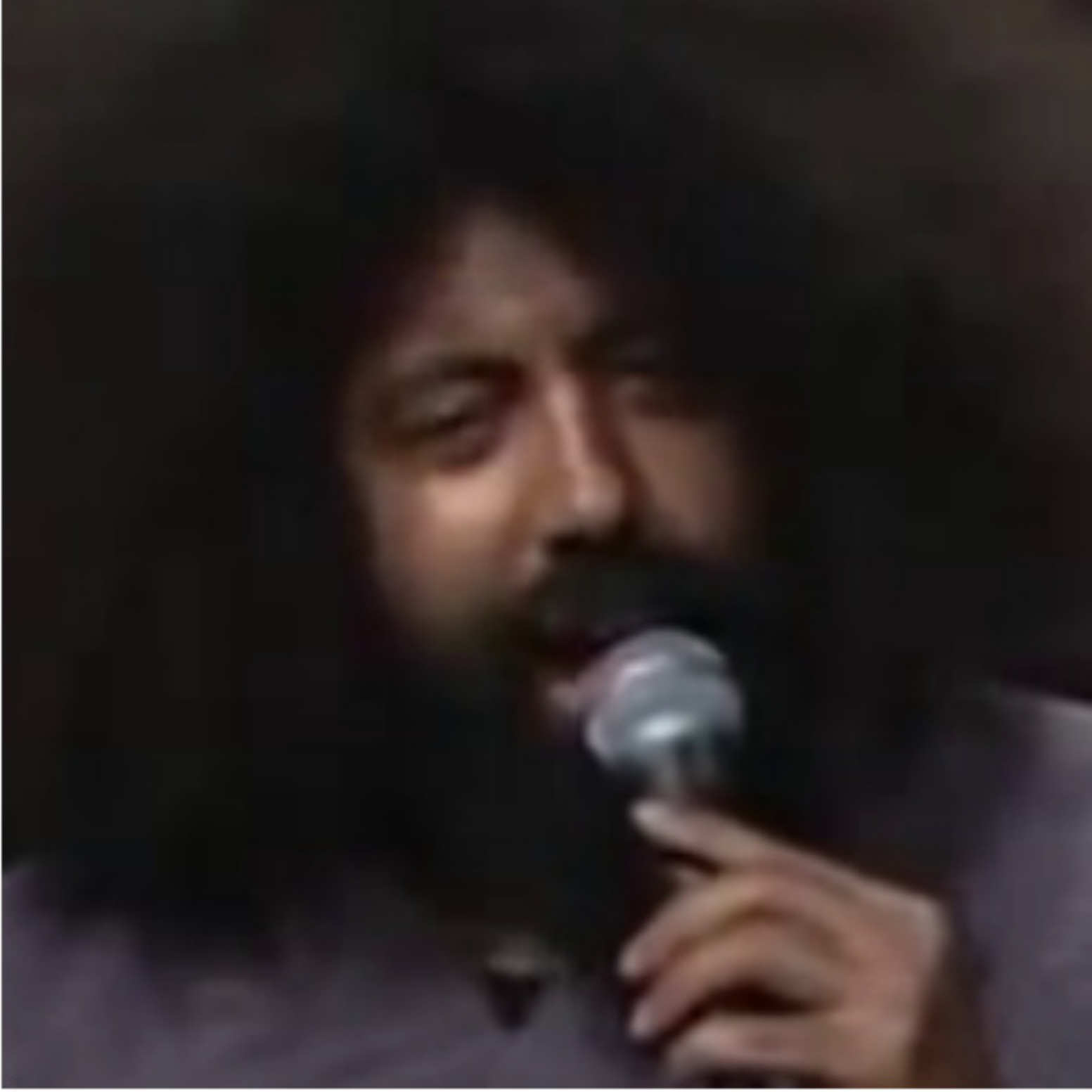}
    \includegraphics[width=0.09\linewidth]{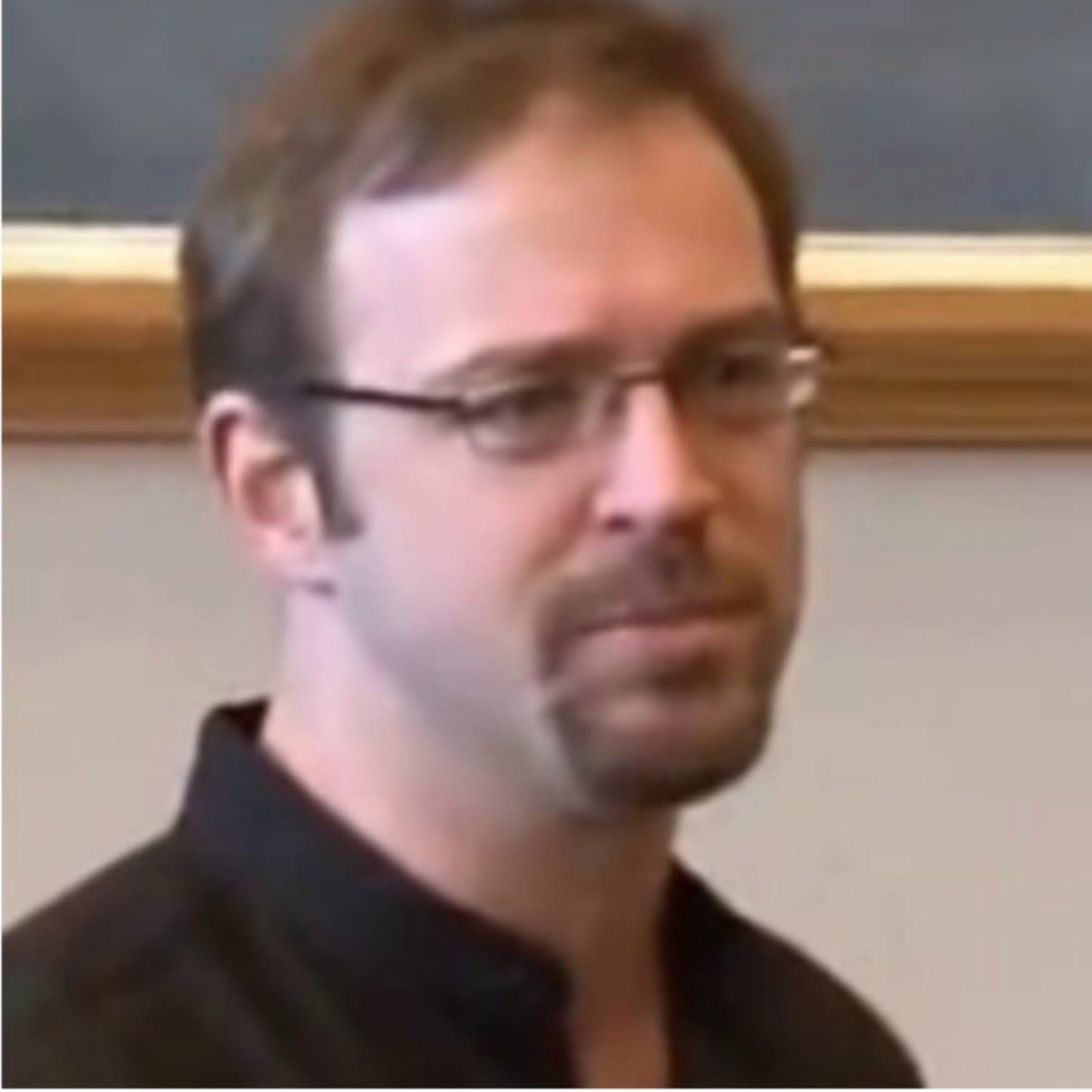}
    \includegraphics[width=0.09\linewidth]{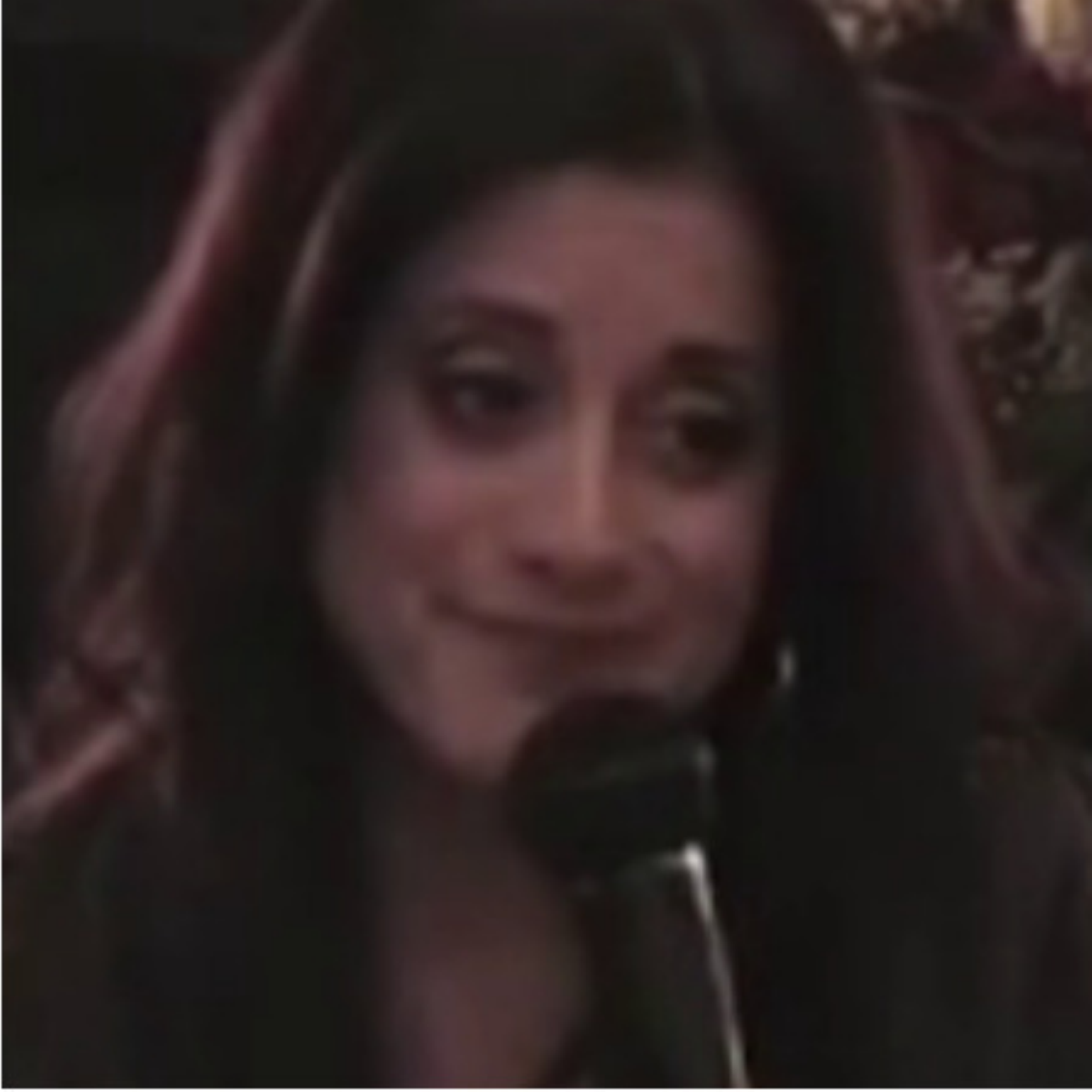}
}
\caption{Example frames from the 300VW dataset by \cite{shen2015first}. Each row contains 10 exemplar images from each category, that are indicative of the challenges that characterise the videos of the category.}
\label{fig:300vw-samples}
\end{figure*}
%%%%%%%%%%%%%%%%%%%%%%%%%%%%%%%%%%%%%%%%%%%%%%%%%%%%%%%%%
\subsection{Dataset}\label{sec:exp_datasets}
%%%%%%%%%%%%%%%%%%%%%%%%%%%%%%%%%%%%%%%%%%%%%%%%%%%%%%%%%
All the comparisons are conducted in the testset of the 300VW dataset collected by \cite{shen2015first}. This recently introduced dataset contains 114 videos (50 for training and 64 for testing). The videos are separated into the following 3 categories:
%%%%%%%%%%%%%%%
% VIDEO CATEGORIES
%%%%%%%%%%%%%%%
\begin{itemize}
    \item \emph{Category 1}: This category is composed of videos captured in well-lit environments without any occlusions.
    \item \emph{Category 2}: The second category includes videos captured in unconstrained illumination conditions.
    \item \emph{Category 3}: The final category consists of video sequences captured in totally arbitrary conditions (including severe occlusions and extreme illuminations).
\end{itemize}
%%%%%%%%%%%%%%%
Each video includes only one person and is annotated using the 68 point mark-up employed by \cite{gross2010multi} and \cite{sagonas2015300} for Multi-PIE and 300W databases, respectively. All videos are between 1500 frames and 3000 frames with a large variety of expressions, poses and capturing conditions, which makes the dataset very challenging for deformable facial tracking. A number of exemplar images, which are indicative of the challenges of each category, are provided in Figure~\ref{fig:300vw-samples}. We note that, in contrast to the results of \cite{shen2015first} in the original 300VW competition, we used the most recently provided annotations\textsuperscript{\ref{300VW_foot}} which have been corrected and do not contain missing frames. Therefore, we also provide updated results following the participants of the 300VW competition.

The public datasets of IBUG (\cite{sagonas2013300}), HELEN (\cite{le2012interactive}), AFW (\cite{zhu2012face}) and LFPW (\cite{belhumeur2013localizing}) are employed for training all the landmark localisation methods. This is further explained in Section~\ref{subsubsec:landmark_localisation_training} below.

%%%%%%%%%%%%%%%%%%%%%%%%%%%%%%%%%%%%%%%%%%%%%%%%%%%%%%%%%
\subsection{Implementation Details}\label{sec:exp_implementation}
%%%%%%%%%%%%%%%%%%%%%%%%%%%%%%%%%%%%%%%%%%%%%%%%%%%%%%%%%
The authors' implementations are utilised for the trackers, as outlined in Table~\ref{tbl:trackers}. Similarly, the face detectors' implementations are outlined in Table~\ref{tbl:detectors}. HOG+SVM was provided by the Dlib project of \cite{king2015max,king2009dlib}, the Weakly Supervised DPM (DPM) (\cite{felzenszwalb2010object}) was the model provided by \cite{mathias2014face} and the code of \cite{dubout2012exact,dubout2013deformable} was used to perform the detection. Moreover, the Strongly Supervised DPM (SS-DPM) of \cite{zhu2012face} was provided by the authors and, finally, the OpenCV implementation by \cite{opencv_library} was used for the VJ detector (\cite{viola2004robust}). The default parameters were used in all cases.

For face alignment, as outlined in Table~\ref{tbl:alignment}, the implementation of CFSS provided by \cite{zhu2015face} is adopted, while the implementations provided by \cite{menpo14} in the Menpo Project are employed for the patch-based AAM of \cite{tzimiropoulos2014gauss} and the SDM of \cite{xiong2013supervised}. Lastly, the implementation of ERT (\cite{kazemi2014one}) is provided by \cite{king2009dlib} in the Dlib library. For the three latter methods, following the original papers and the code's documentation, several parameters were validated and chosen based on the results in a validation set that consisted of a few videos from the 300VW training set.

The details of the parameters utilised for the patch-based AAM, SDM and ERT are mentioned below. For AAM, we used the algorithm of \cite{tzimiropoulos2014gauss} and applied a 2-level Gaussian pyramid with 4 and 10 shape components, and 60 and 150 appearance components in each scale, respectively. For the SDM, a 4-level Gaussian pyramid was employed. SIFT (\cite{lowe1999object}) feature vectors of length 128 were extracted at the first 3 scales, using RootSIFT by \cite{arandjelovic2012three}. Raw pixel intensities were used at the highest scale. Finally, part of the experiments were conducted on the cloud software of \cite{koukis2013okeanos}. 

%%%%%%%%%%%%%%%%%%%%%%%%%%%%%%%%%%%%%%%%%%%%%%%%%%%%%%%%%
\subsubsection{Landmark Localisation Training}\label{subsubsec:landmark_localisation_training}
%%%%%%%%%%%%%%%%%%%%%%%%%%%%%%%%%%%%%%%%%%%%%%%%%%%%%%%%%
All the landmark localisation methods were trained with respect to the 68 facial points mark-up employed by \cite{sagonas2013300,sagonas2015300} in 300W, while the rest of the parameters were determined via cross-validation. Again, this validation set consisted of frames from the 300VW trainset, as well as 60 privately collected images with challenging poses.
All of the discriminative landmark localisation methods (SDM, ERT, CFSS) were trained from images in the public datasets of IBUG (\cite{sagonas2013300}), HELEN (\cite{le2012interactive}), AFW (\cite{zhu2012face}) and LFPW (\cite{belhumeur2013localizing}). The generative AAM was trained on less data, since generative methods do not benefit as strongly from large training datasets. The training data used for the AAM was the recently released 300 images from the 600W dataset (\cite{sagonas2015300}), 500 challenging images from LFPW (\cite{belhumeur2013localizing}) and the 135 images of the IBUG dataset (\cite{sagonas2013300}).

Discriminative landmark localisation methods are tightly coupled with the initialisation statistics, as they learn to model a given variance of initialisations. Therefore, it is necessary to re-train each discriminative method for each face detection method employed. This allows the landmark localisation methods to correctly model the large amount of variance present between detectors. On aggregate 5 different detector and landmark localisation models are trained. One for each detector and landmark localisation pair (totalling 4) and a single model trained using a validation set that estimates the variance of the ground truth bounding box throughout the sequences. This model is used for all trackers.

%%%%%%%%%%%%%%%%%%%%%%%%%%%%%%%%%%%%%%%%%%%%%%%%%%%%%%%%%
%%%%%%%%%%%%% EXEMPLAR ERRORS PER CATEGORY %%%%%%%%%%%%%%
%%%%%%%%%%%%%%%%%%%%%%%%%%%%%%%%%%%%%%%%%%%%%%%%%%%%%%%%%
\begin{table*}[!t]
\centering
\begin{tabular}{c c @{\hspace{0.01cm}} c @{\hspace{0.01cm}} c @{\hspace{0.01cm}} c @{\hspace{0.01cm}} c @{\hspace{0.01cm}} c @{\hspace{0.01cm}} c @{\hspace{0.01cm}} c @{\hspace{0.01cm}} c}
\toprule
\multirow{2}{*}{\emph{Category}} & \multicolumn{9}{c}{\emph{Error}}\\
\cmidrule(lr){2-10}
& $0.01$ & $0.02$ & $0.03$ & $0.04$ & $0.05$ & $0.06$ & $0.07$ & $0.08$ & $0.09$\\
\midrule[\heavyrulewidth]
1 & \includegraphics[valign=m,width=0.09\linewidth]{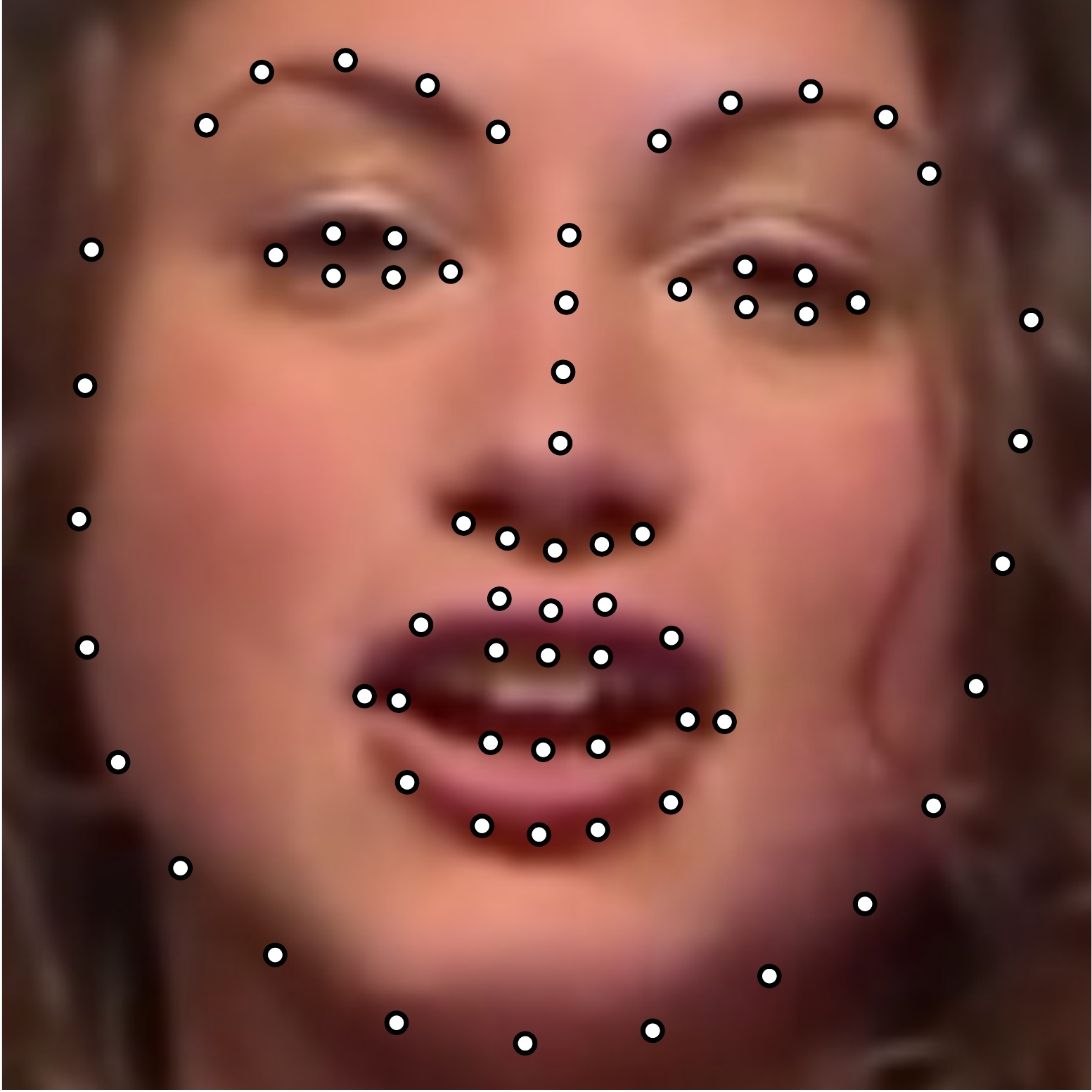} &
    \includegraphics[valign=m,width=0.09\linewidth]{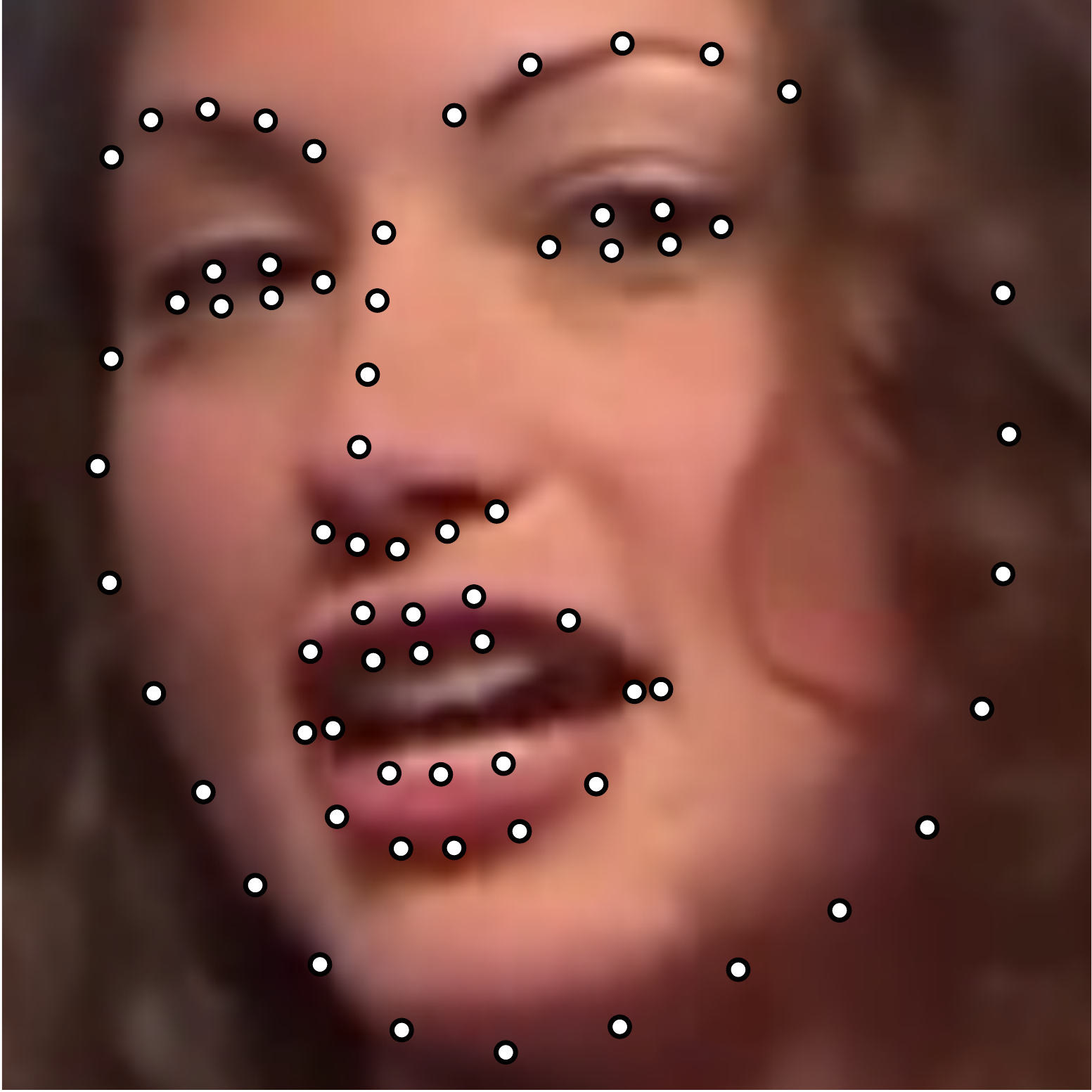} &
    \includegraphics[valign=m,width=0.09\linewidth]{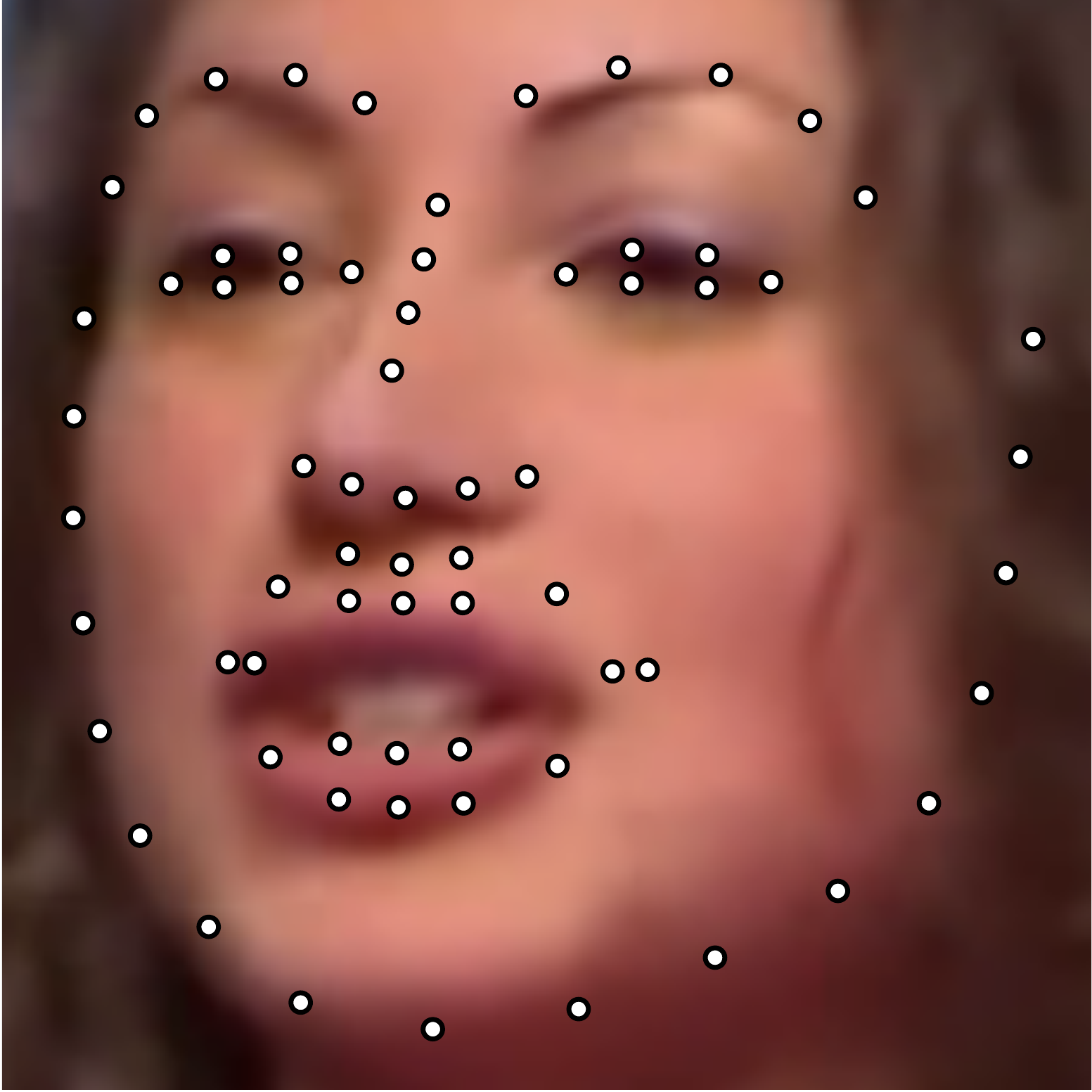} &
    \includegraphics[valign=m,width=0.09\linewidth]{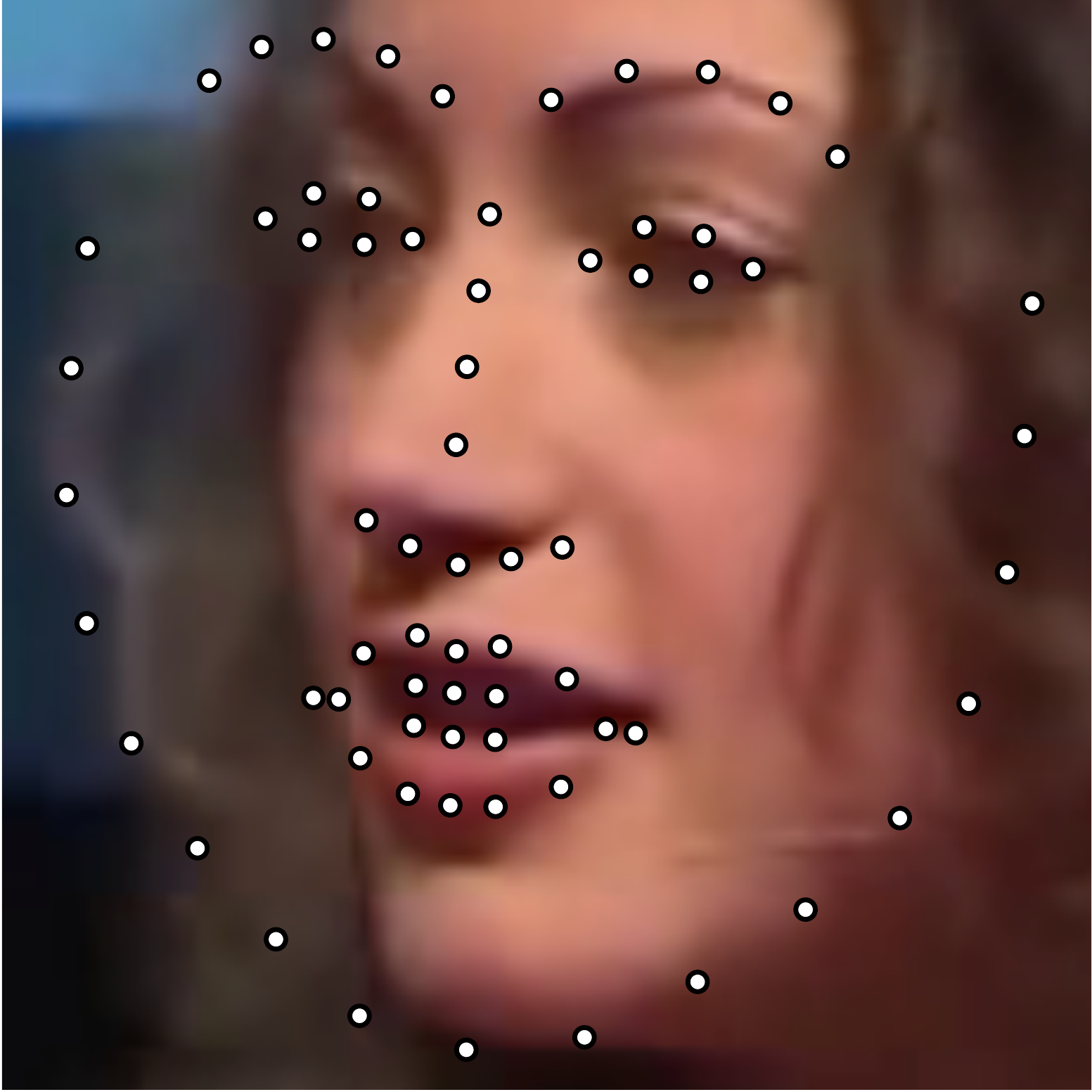} &
    \includegraphics[valign=m,width=0.09\linewidth]{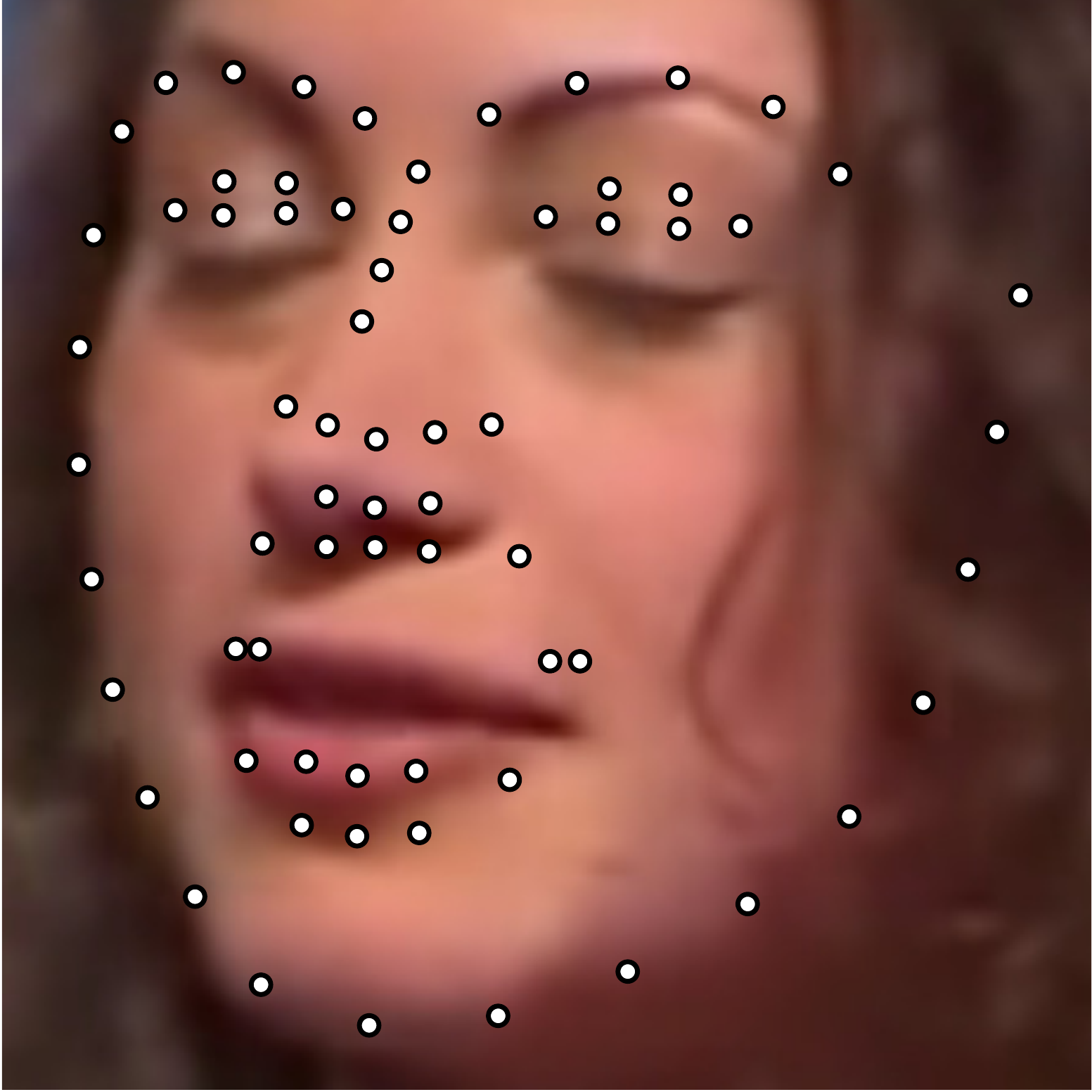} &
    \includegraphics[valign=m,width=0.09\linewidth]{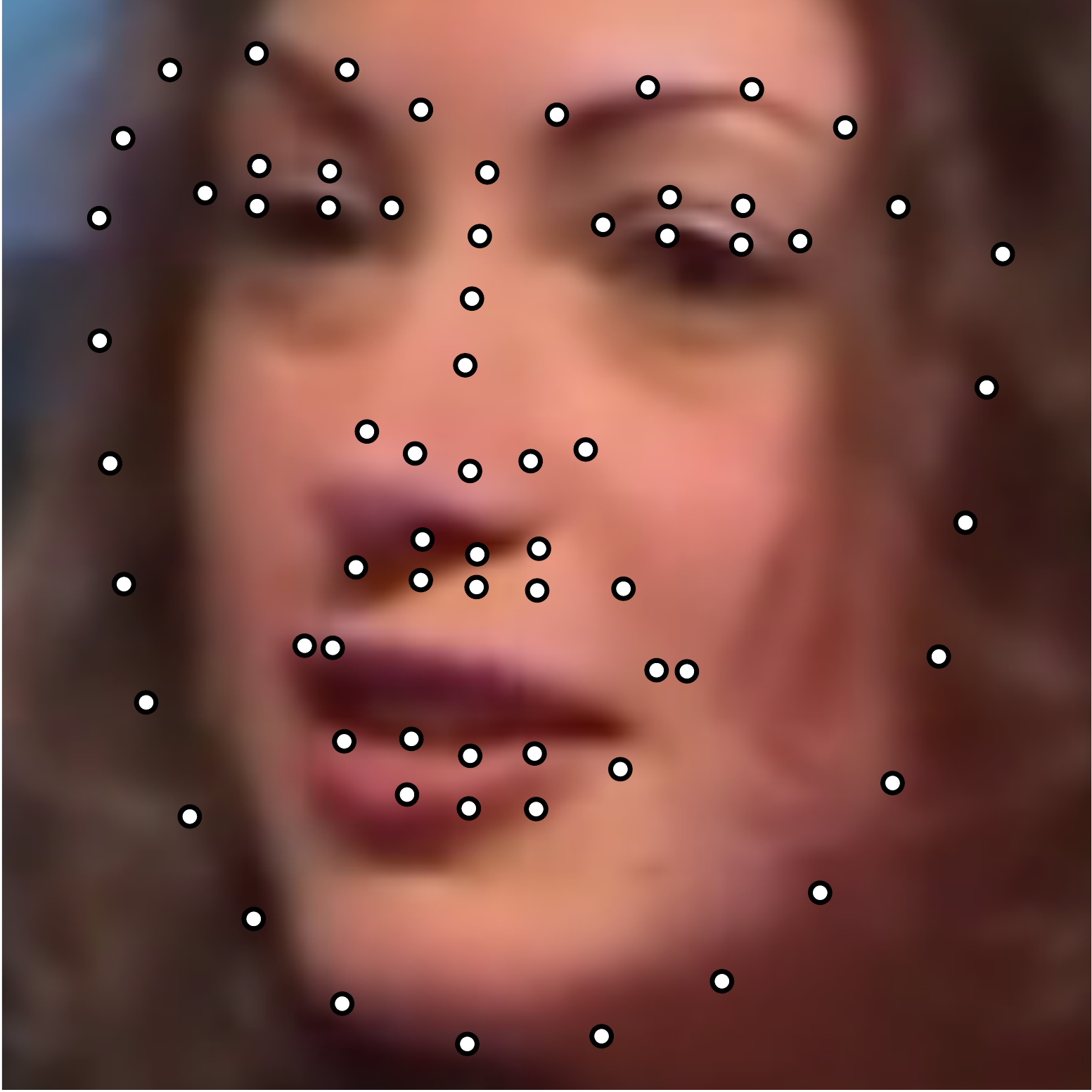} &
    \includegraphics[valign=m,width=0.09\linewidth]{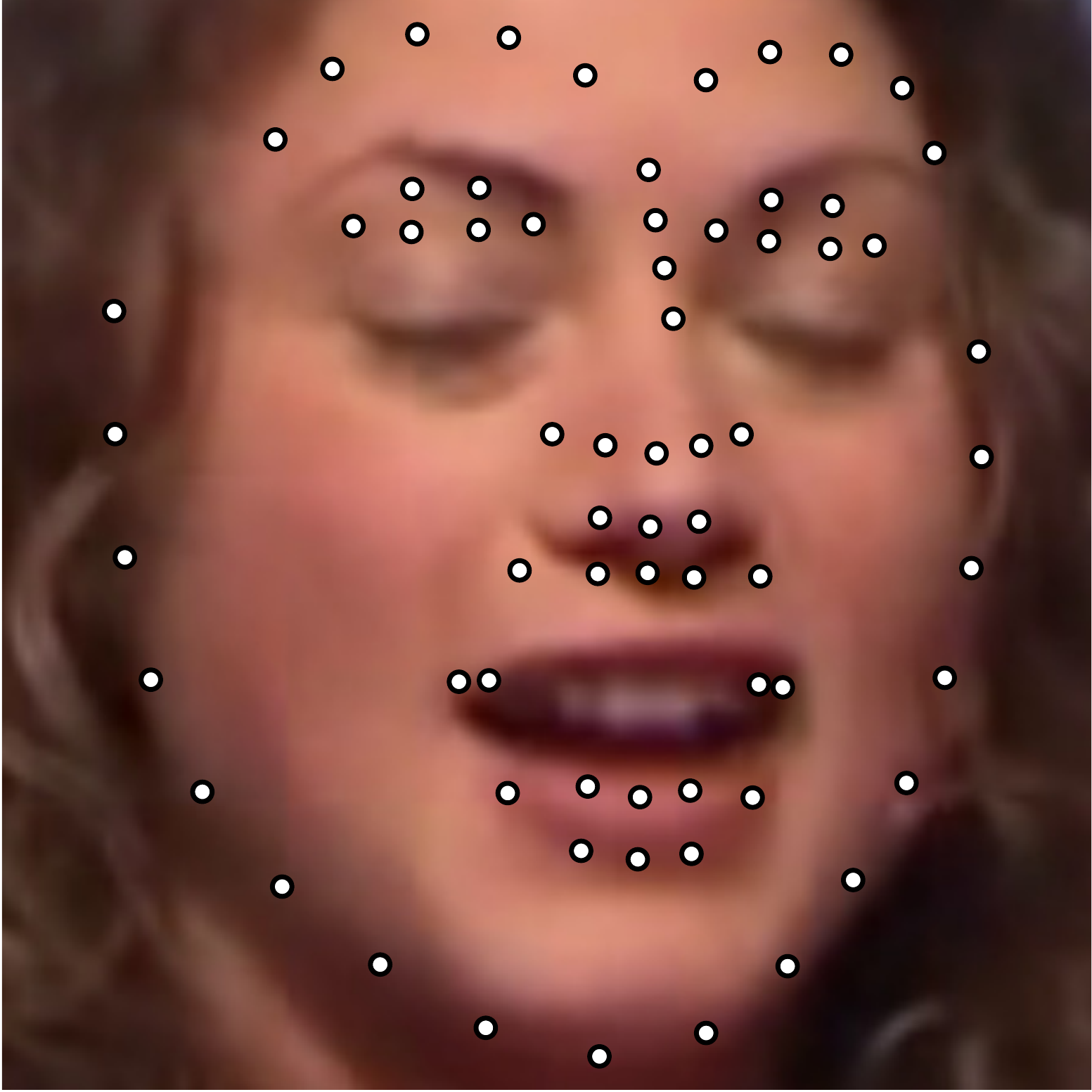} &
    \includegraphics[valign=m,width=0.09\linewidth]{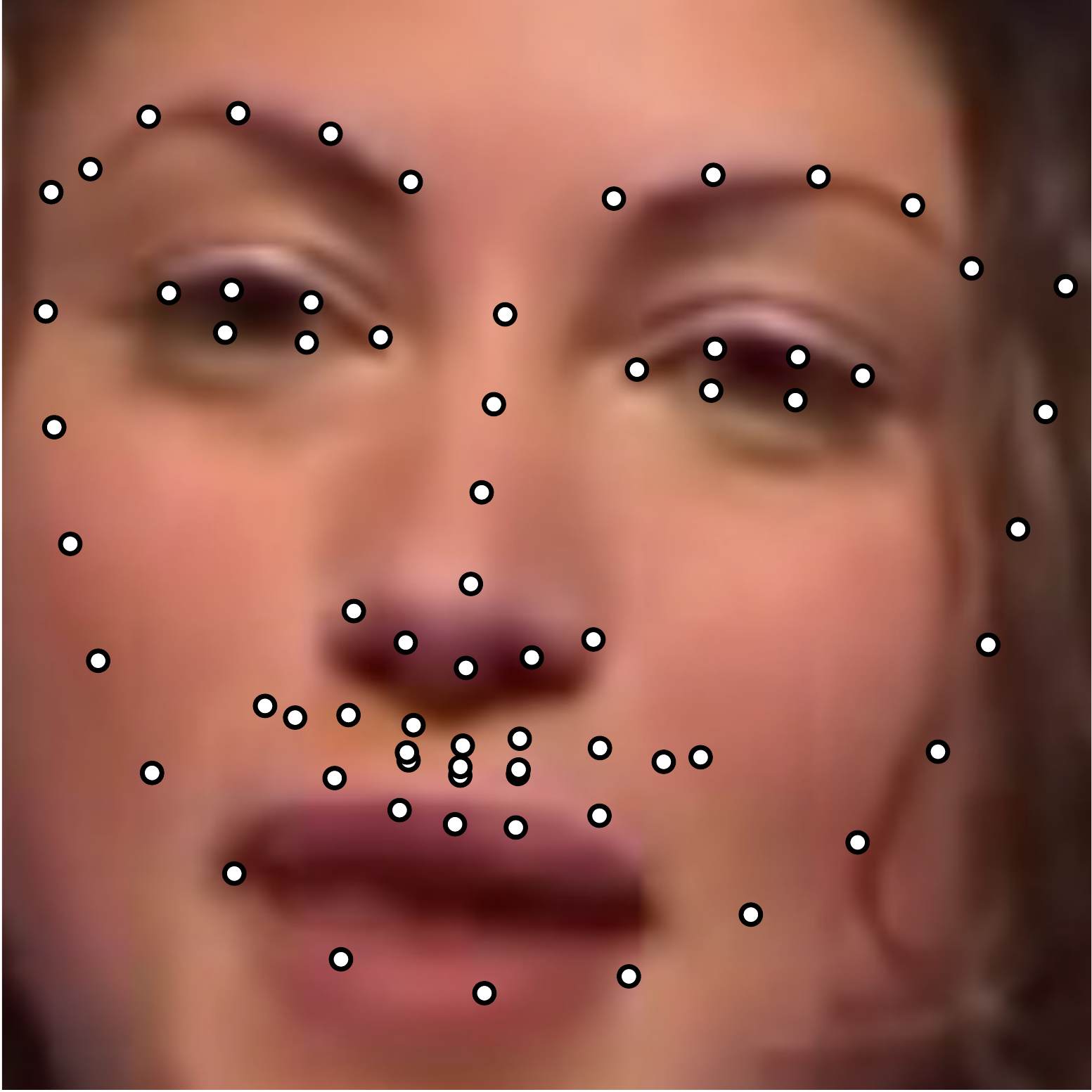} &
    \includegraphics[valign=m,width=0.09\linewidth]{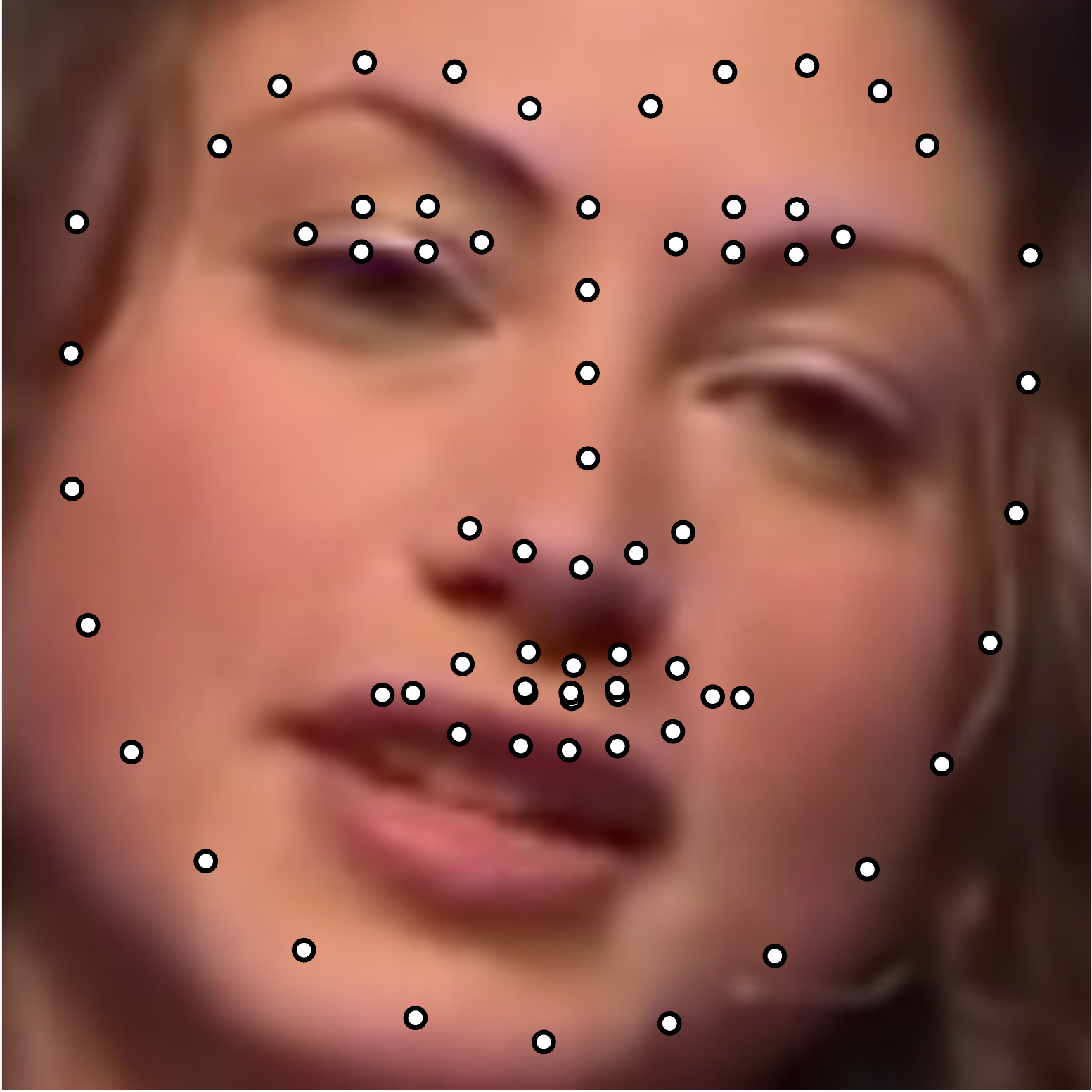}\\
2 & \includegraphics[valign=m,width=0.09\linewidth]{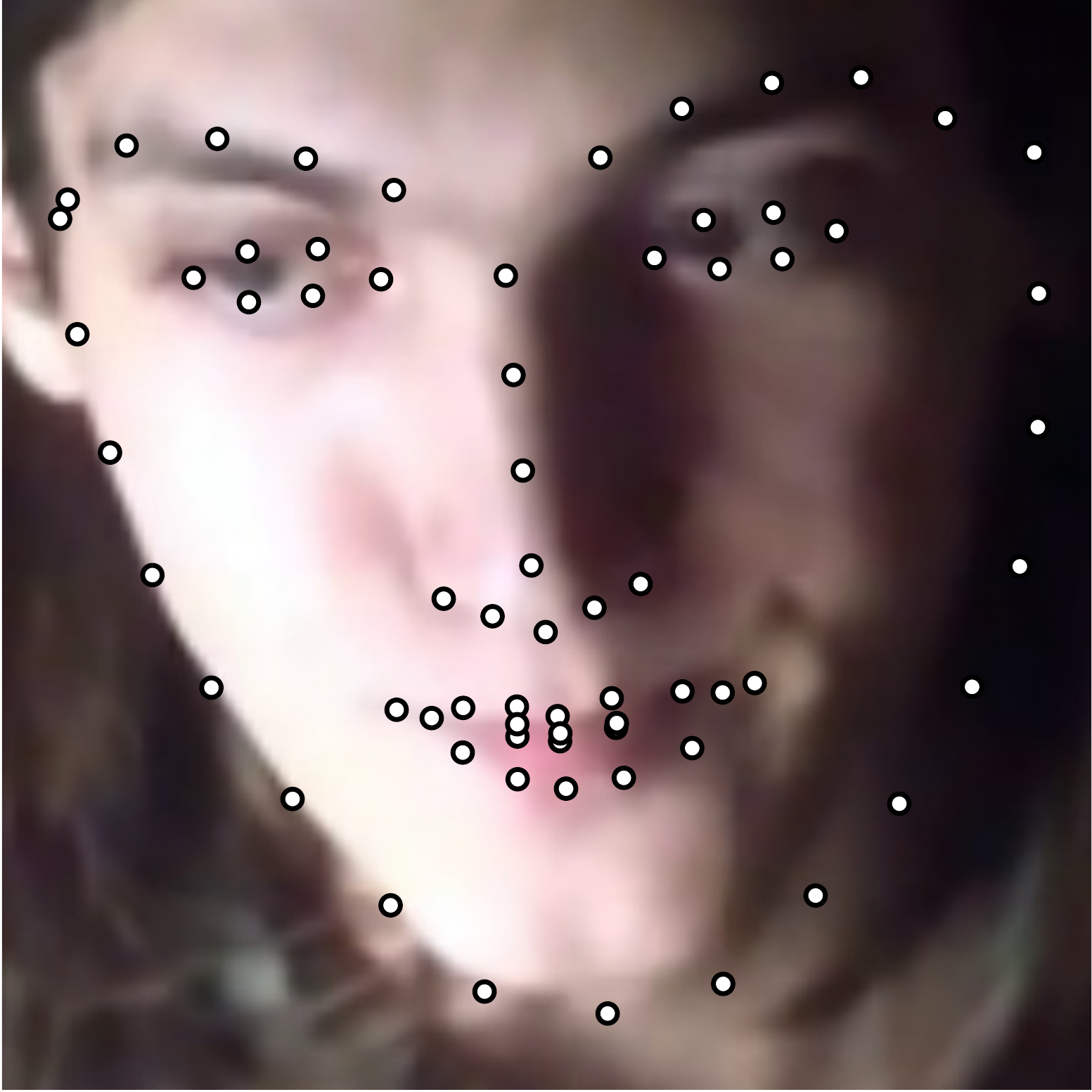} &
    \includegraphics[valign=m,width=0.09\linewidth]{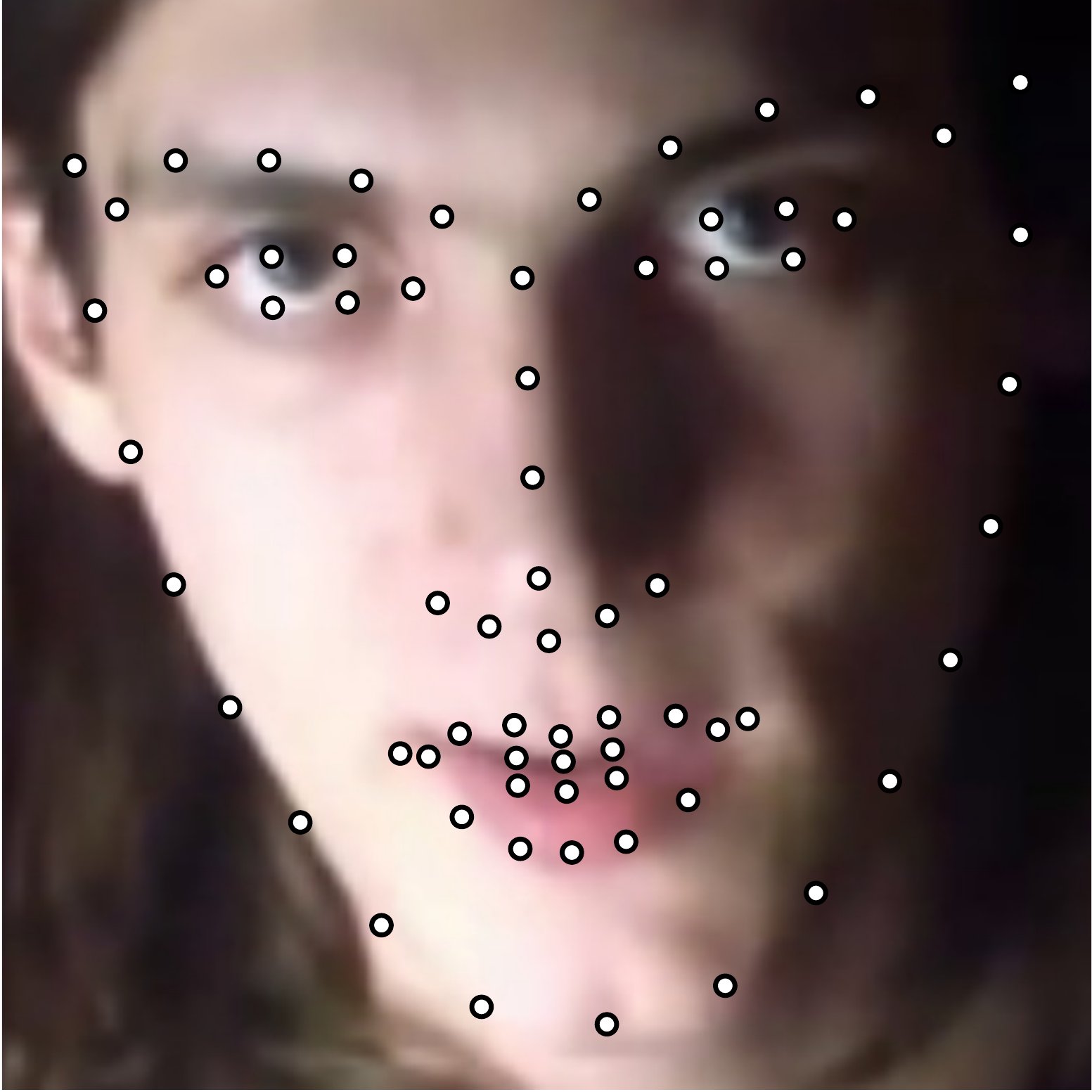} &
    \includegraphics[valign=m,width=0.09\linewidth]{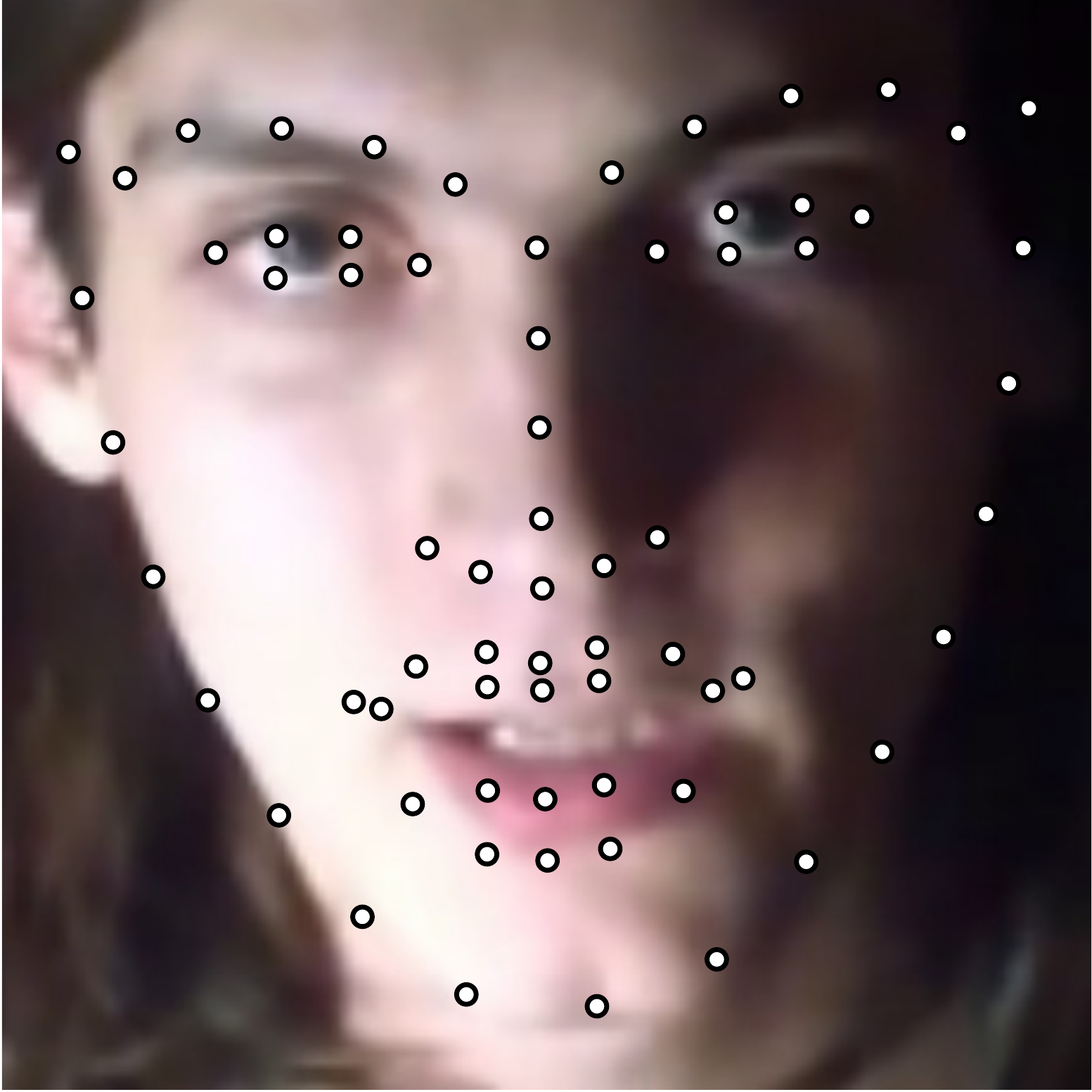} &
    \includegraphics[valign=m,width=0.09\linewidth]{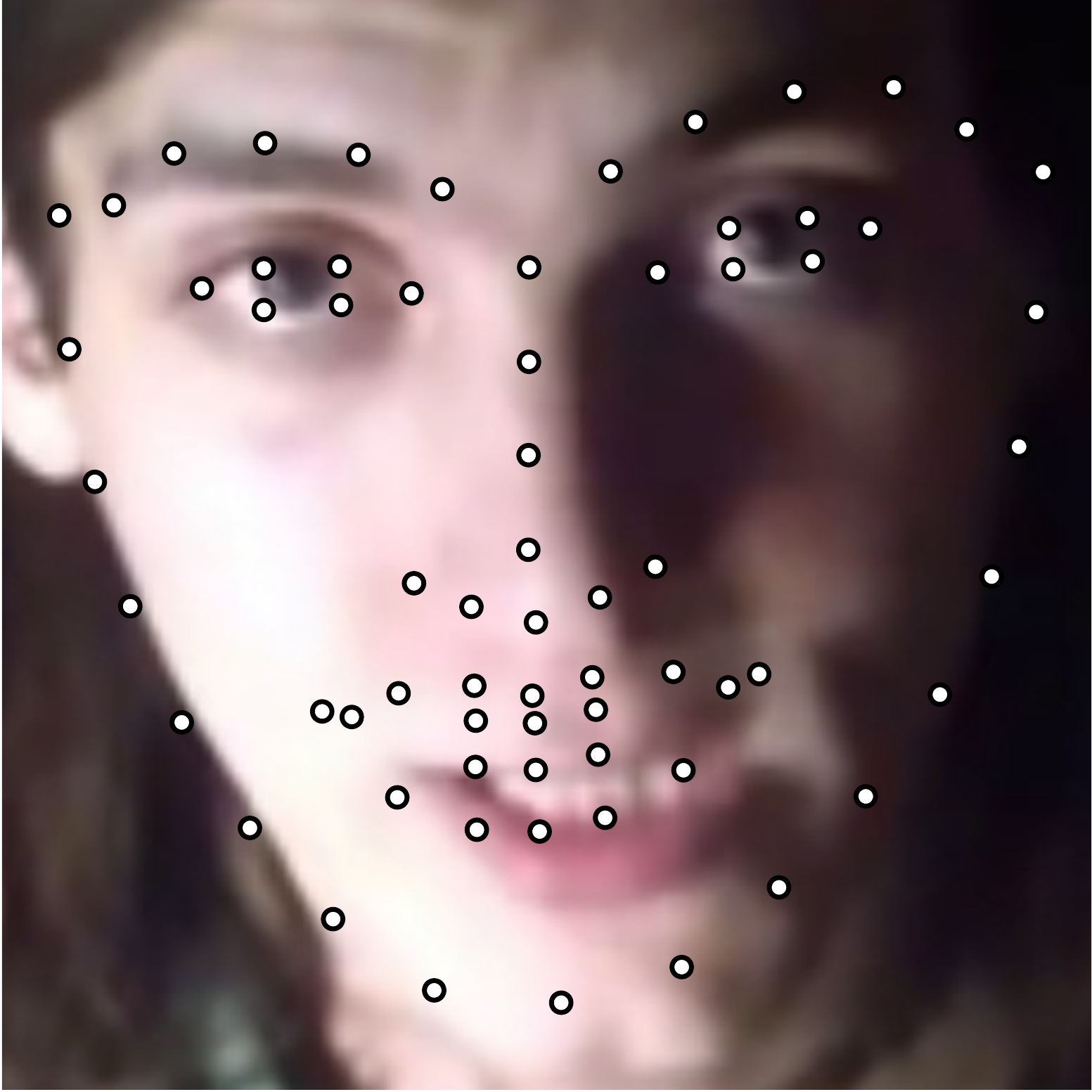} &
    \includegraphics[valign=m,width=0.09\linewidth]{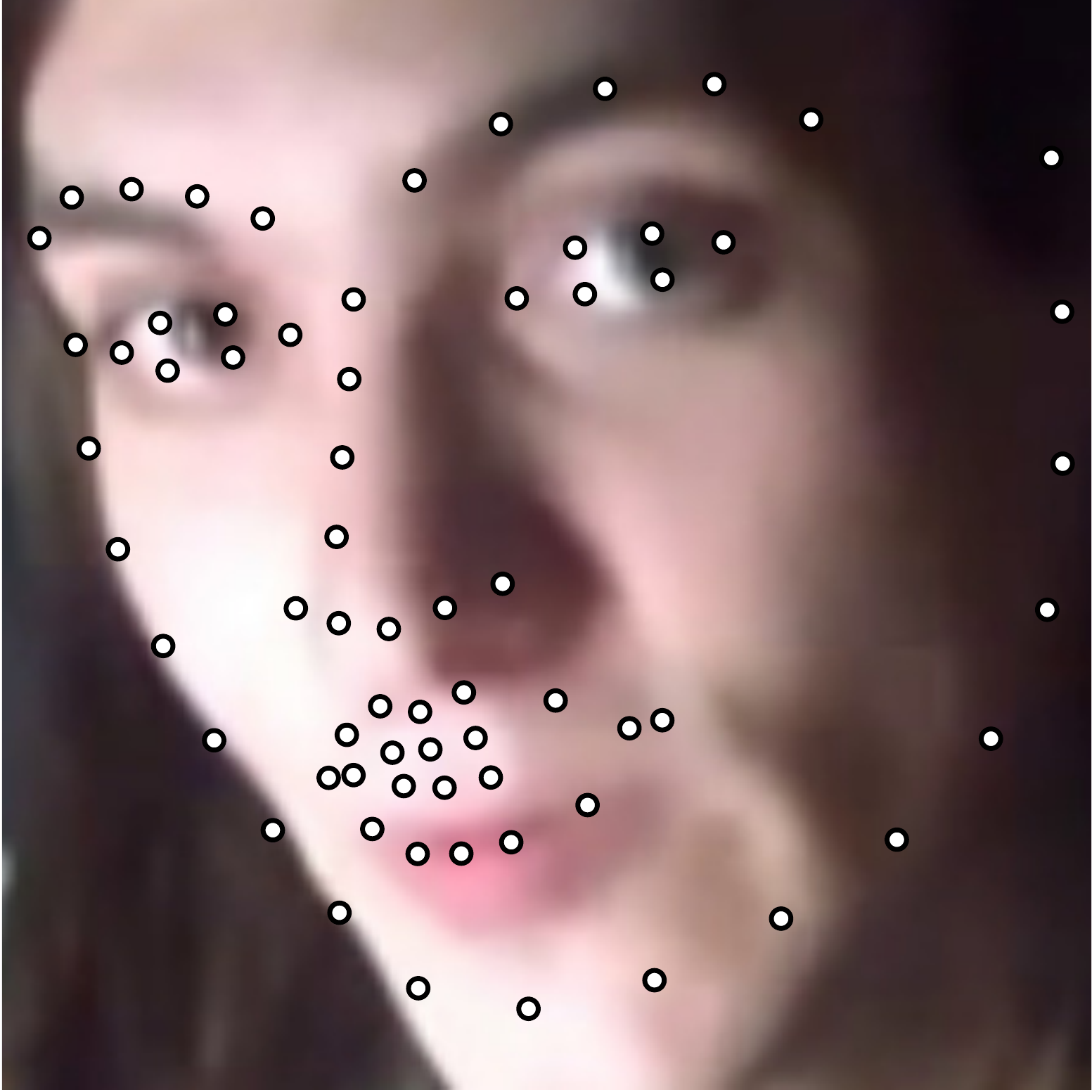} &
    \includegraphics[valign=m,width=0.09\linewidth]{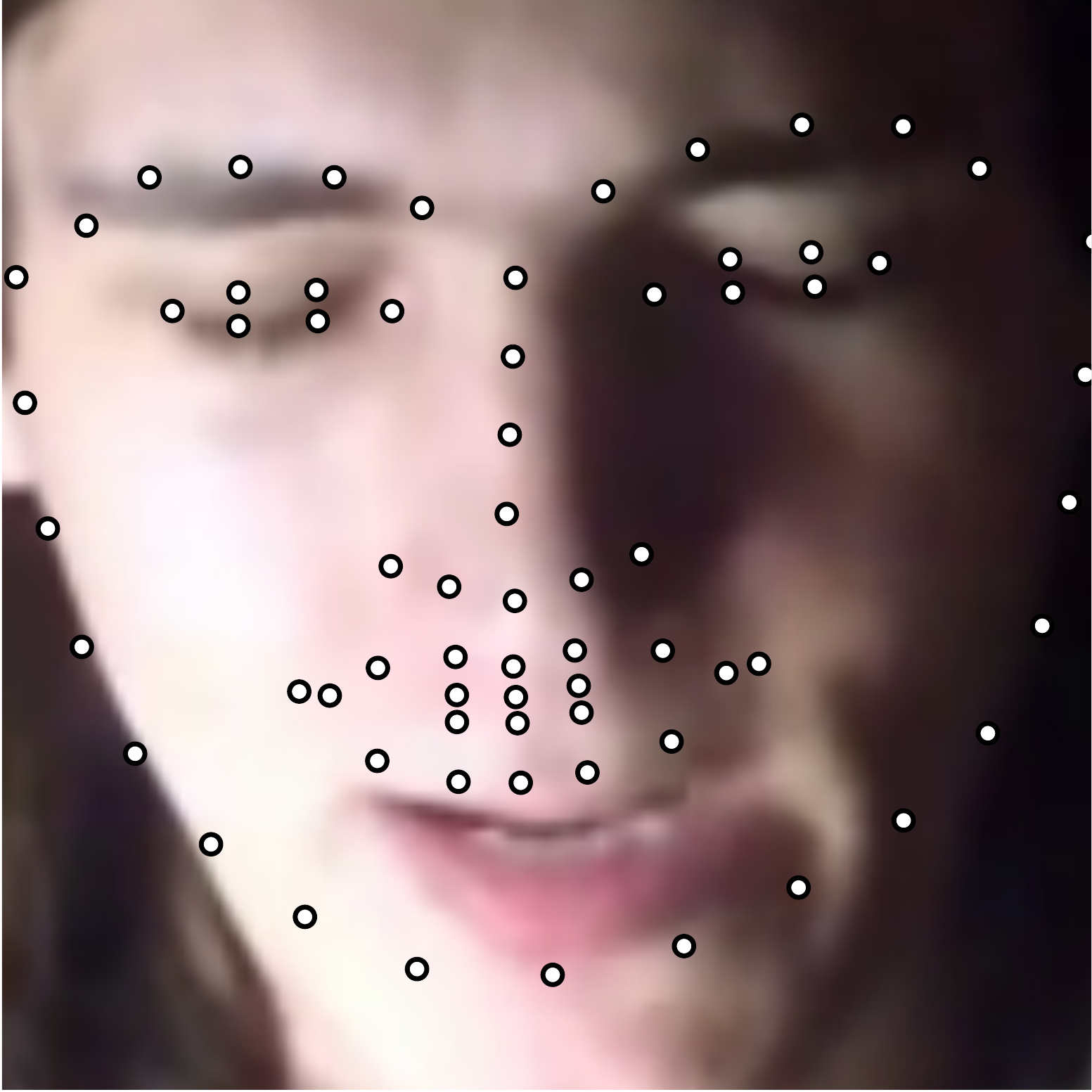} &
    \includegraphics[valign=m,width=0.09\linewidth]{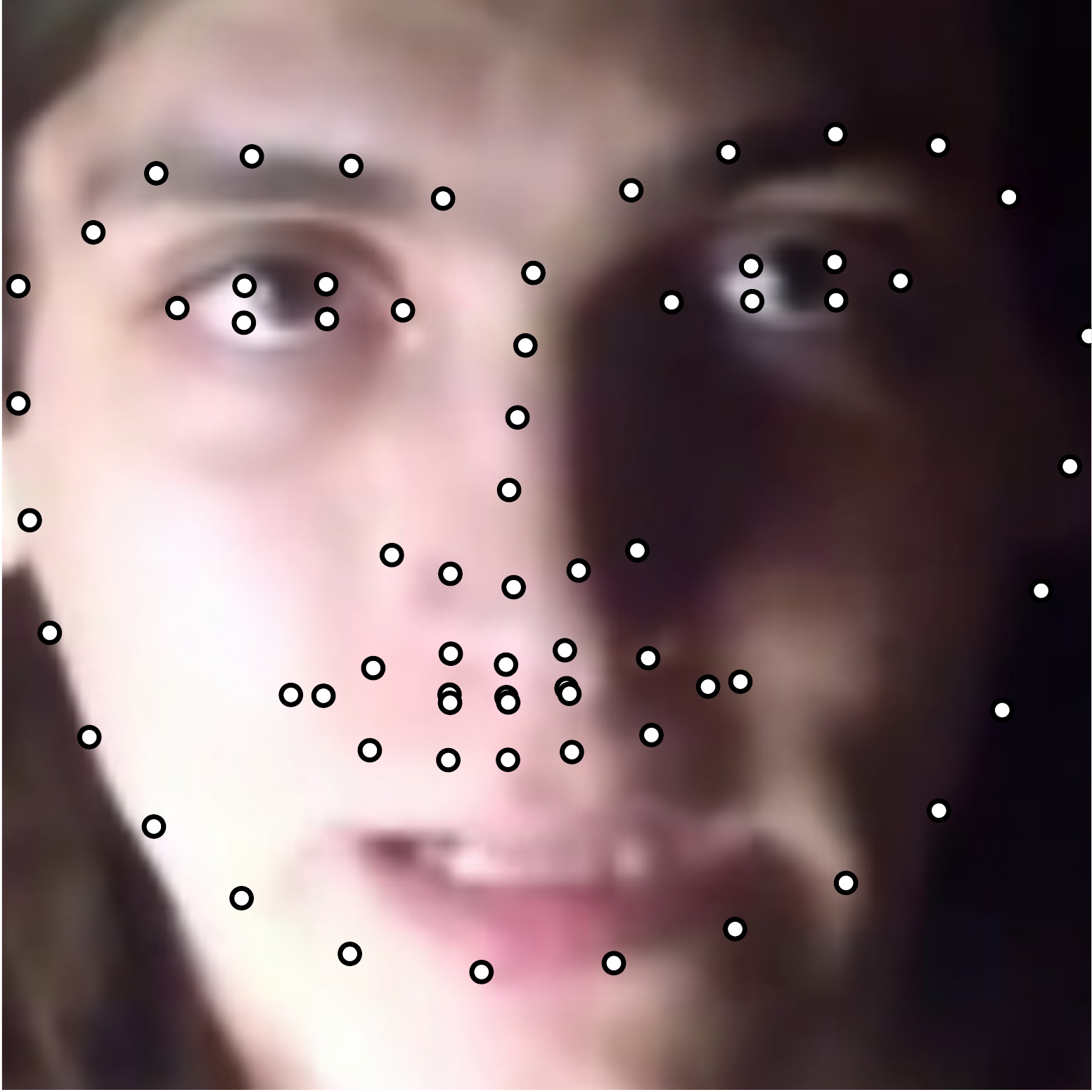} &
    \includegraphics[valign=m,width=0.09\linewidth]{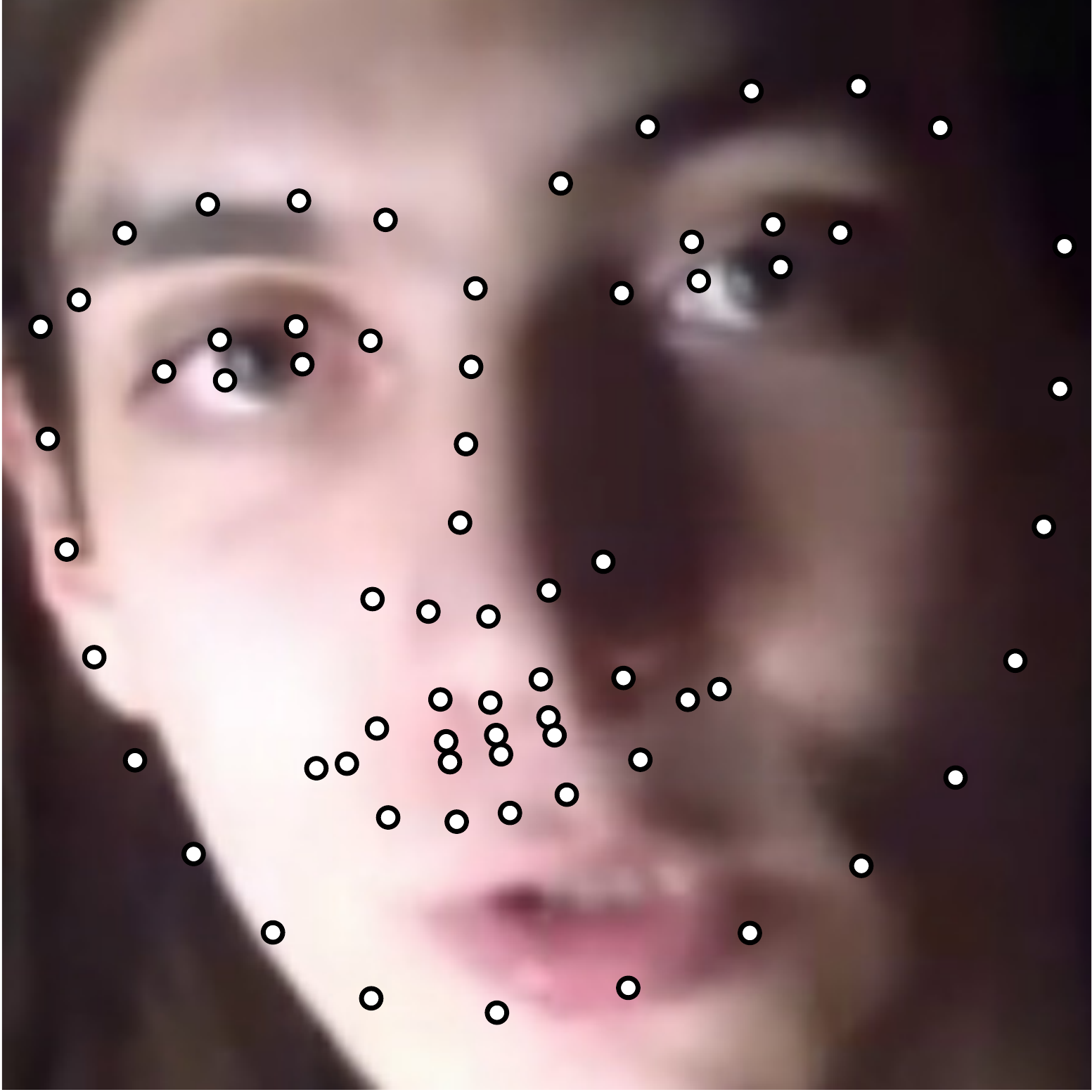} &
    \includegraphics[valign=m,width=0.09\linewidth]{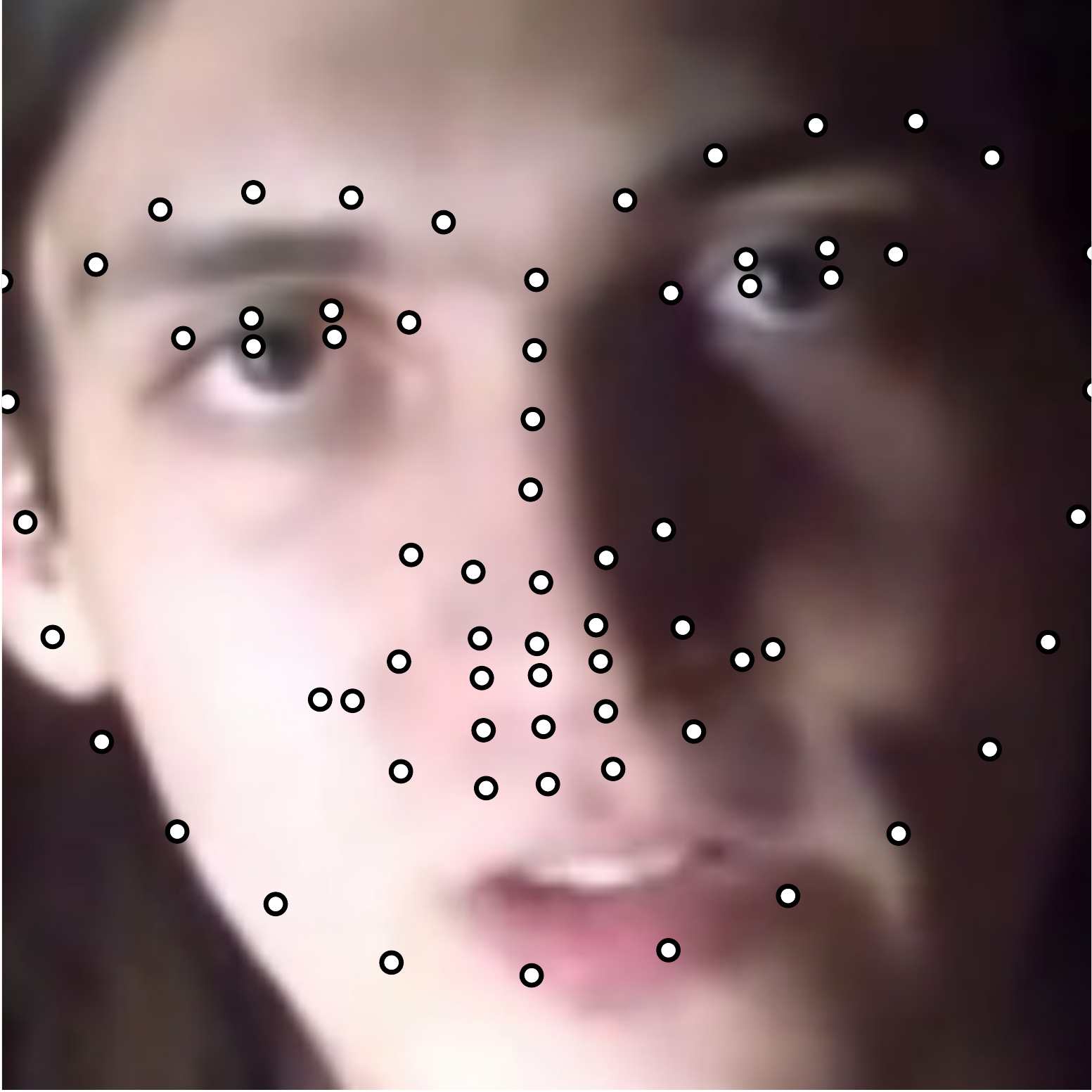}\\
3 & \includegraphics[valign=m,width=0.09\linewidth]{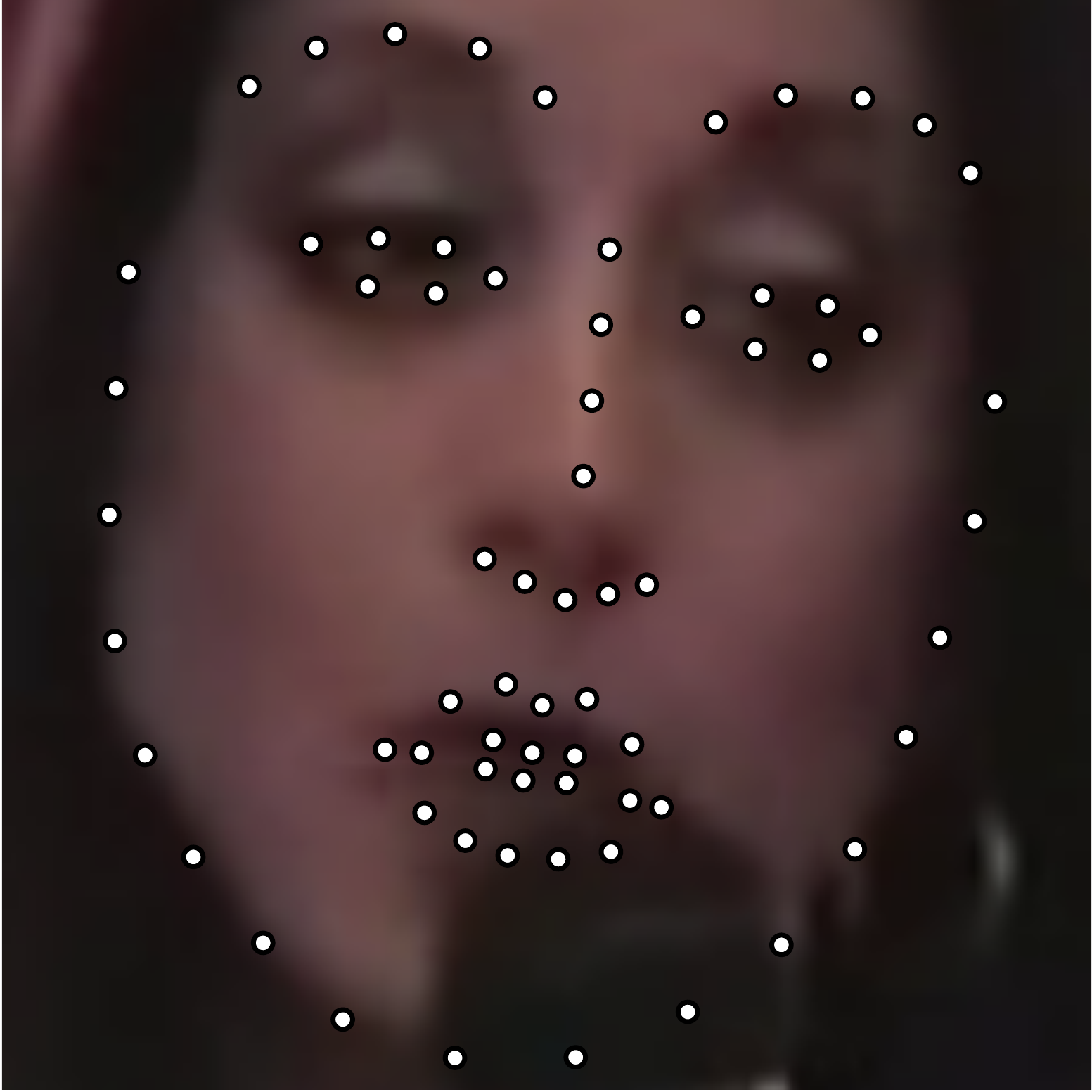} &
    \includegraphics[valign=m,width=0.09\linewidth]{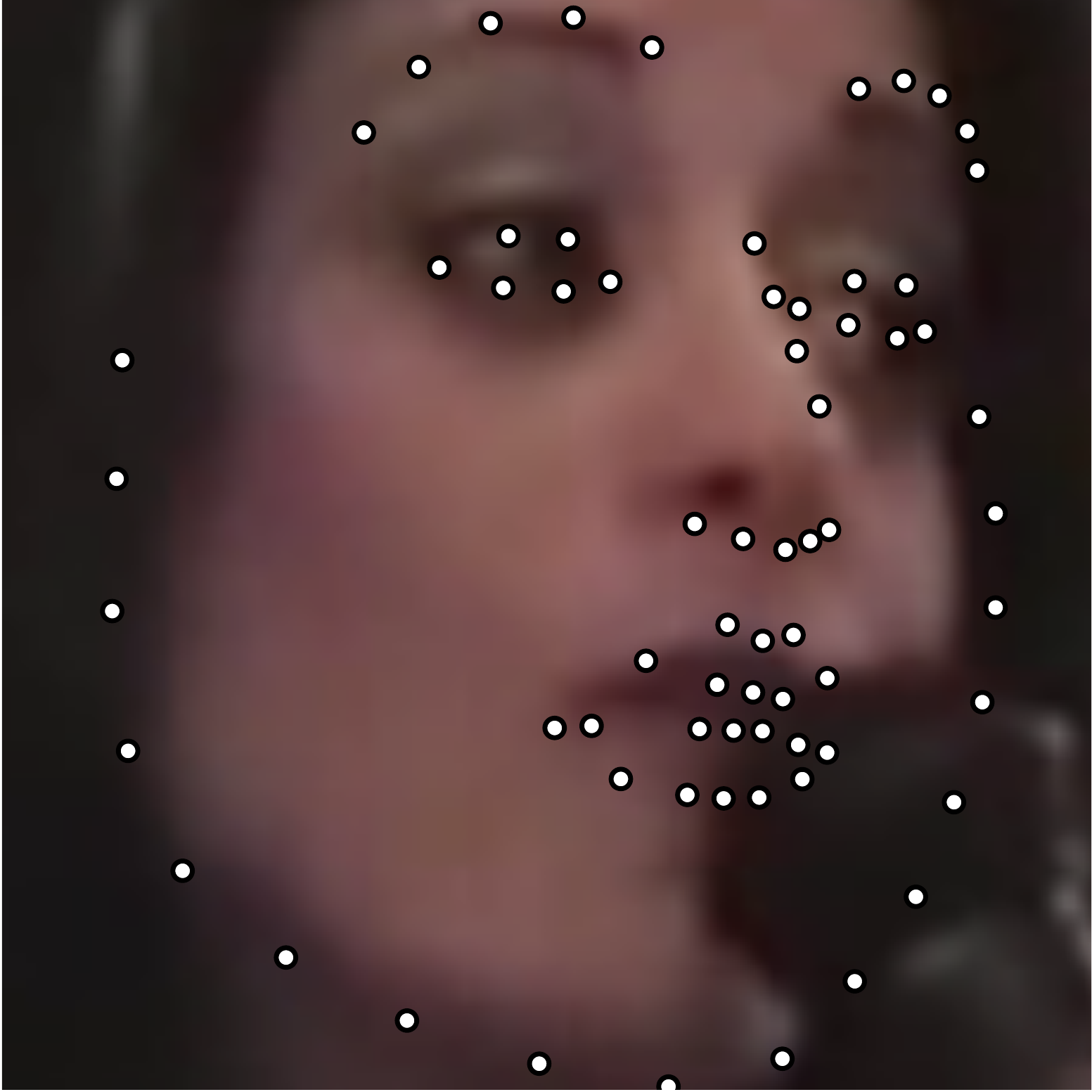} &
    \includegraphics[valign=m,width=0.09\linewidth]{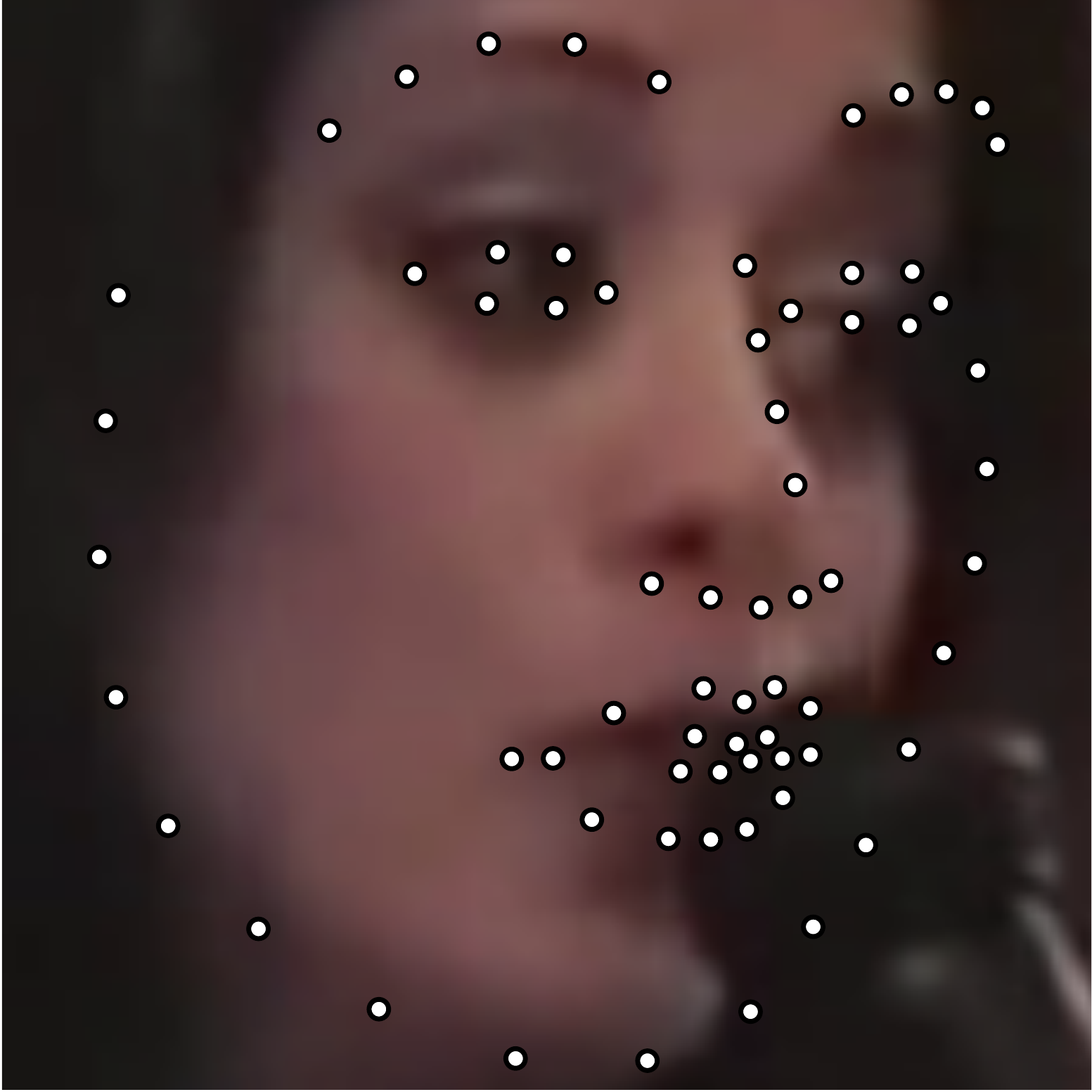} &
    \includegraphics[valign=m,width=0.09\linewidth]{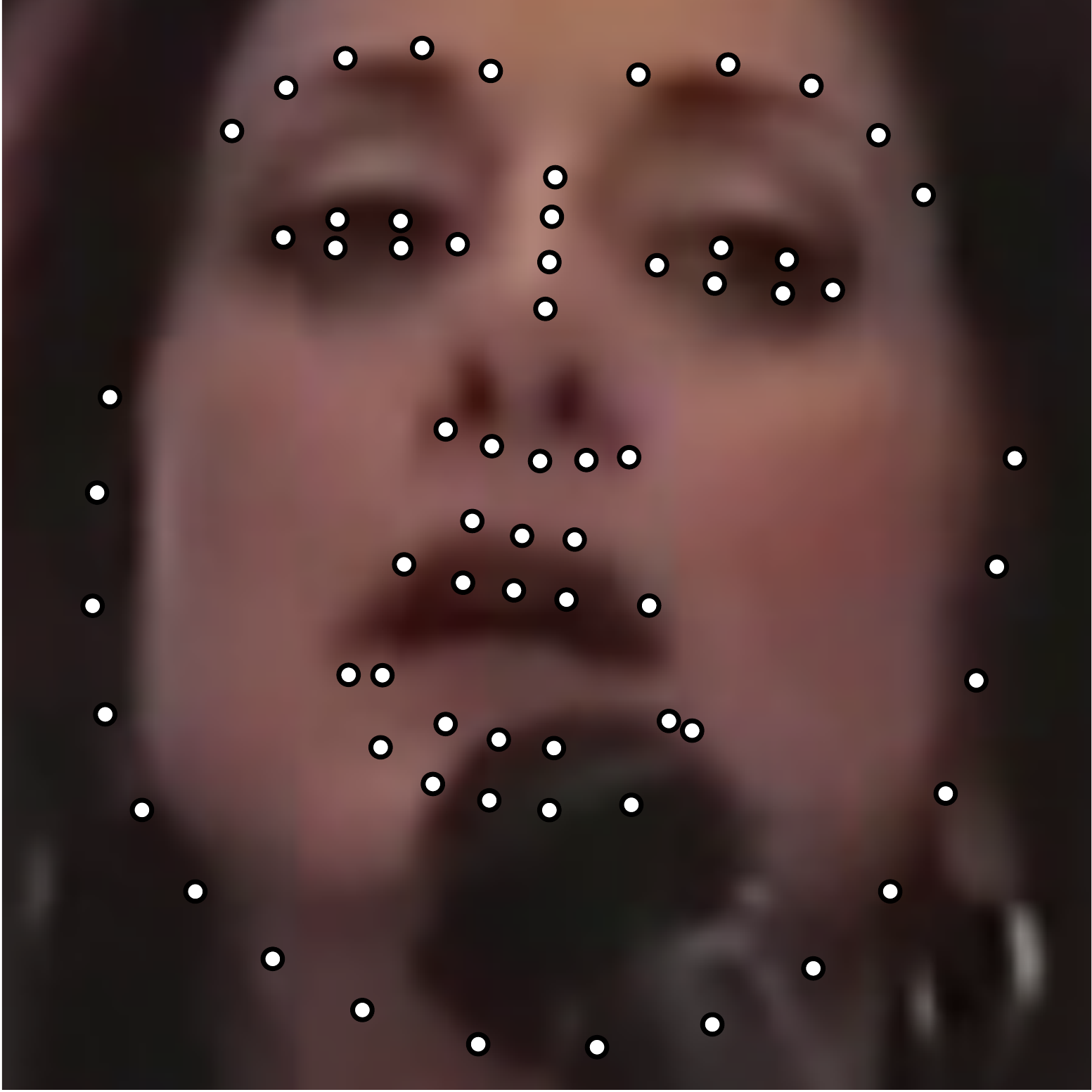} &
    \includegraphics[valign=m,width=0.09\linewidth]{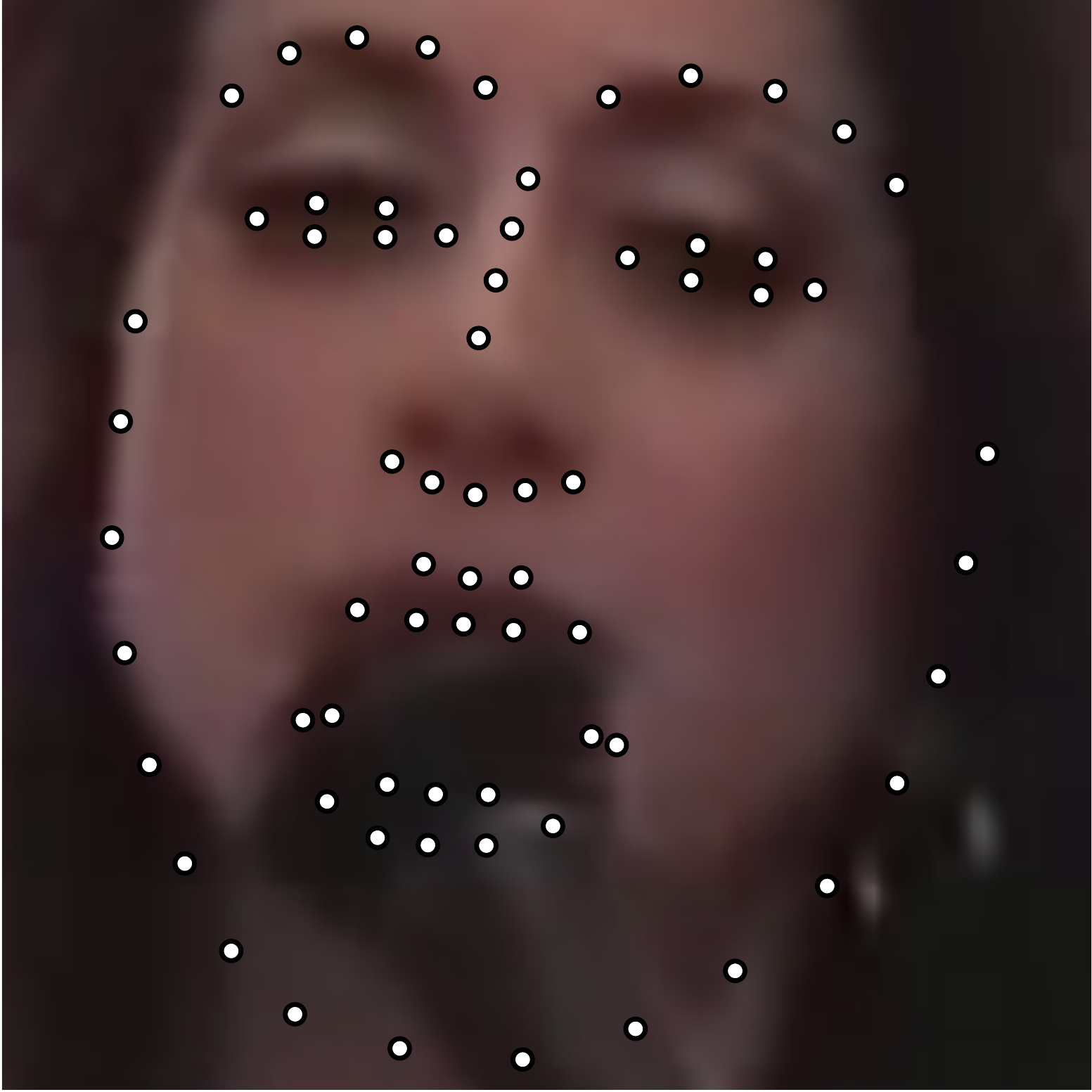} &
    \includegraphics[valign=m,width=0.09\linewidth]{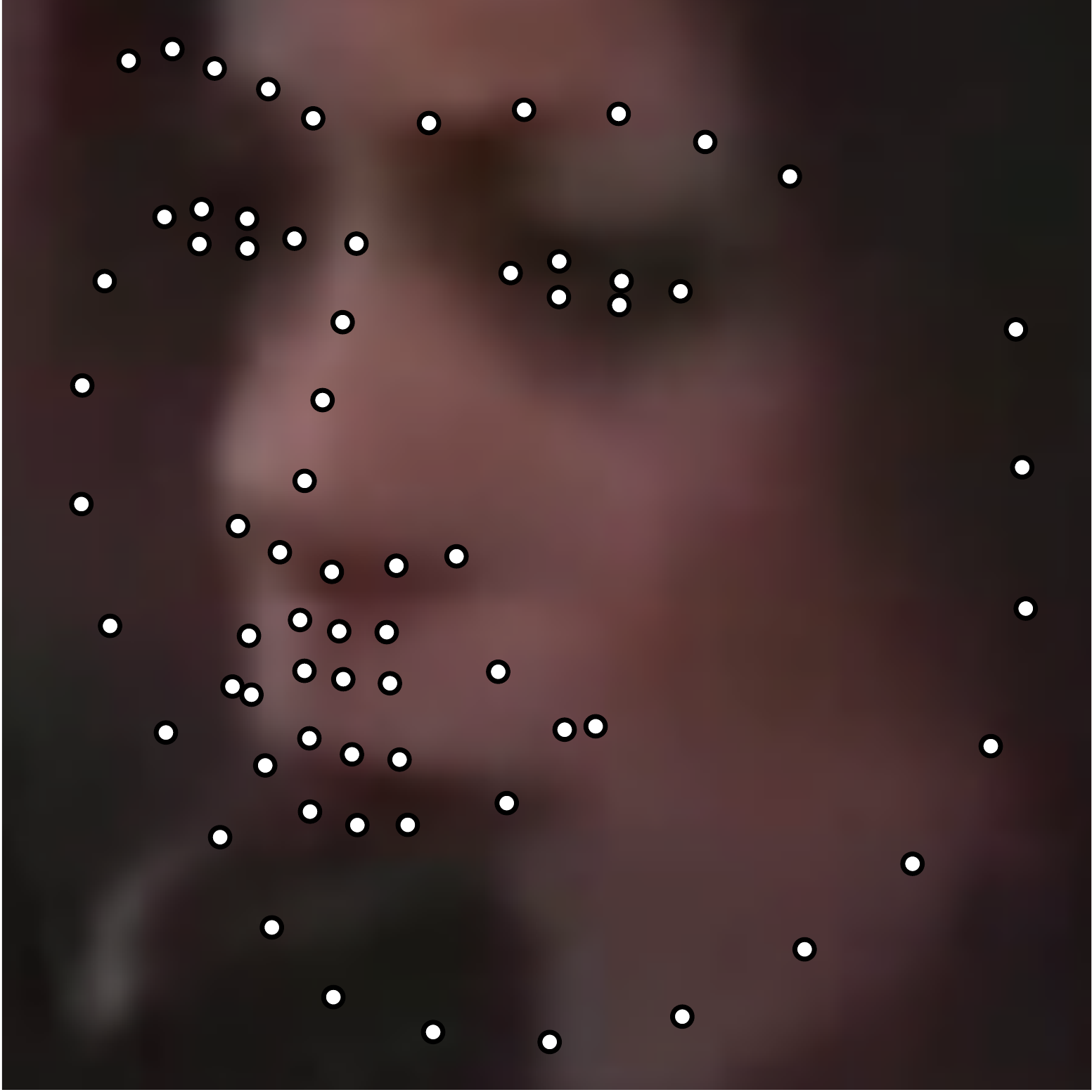} &
    \includegraphics[valign=m,width=0.09\linewidth]{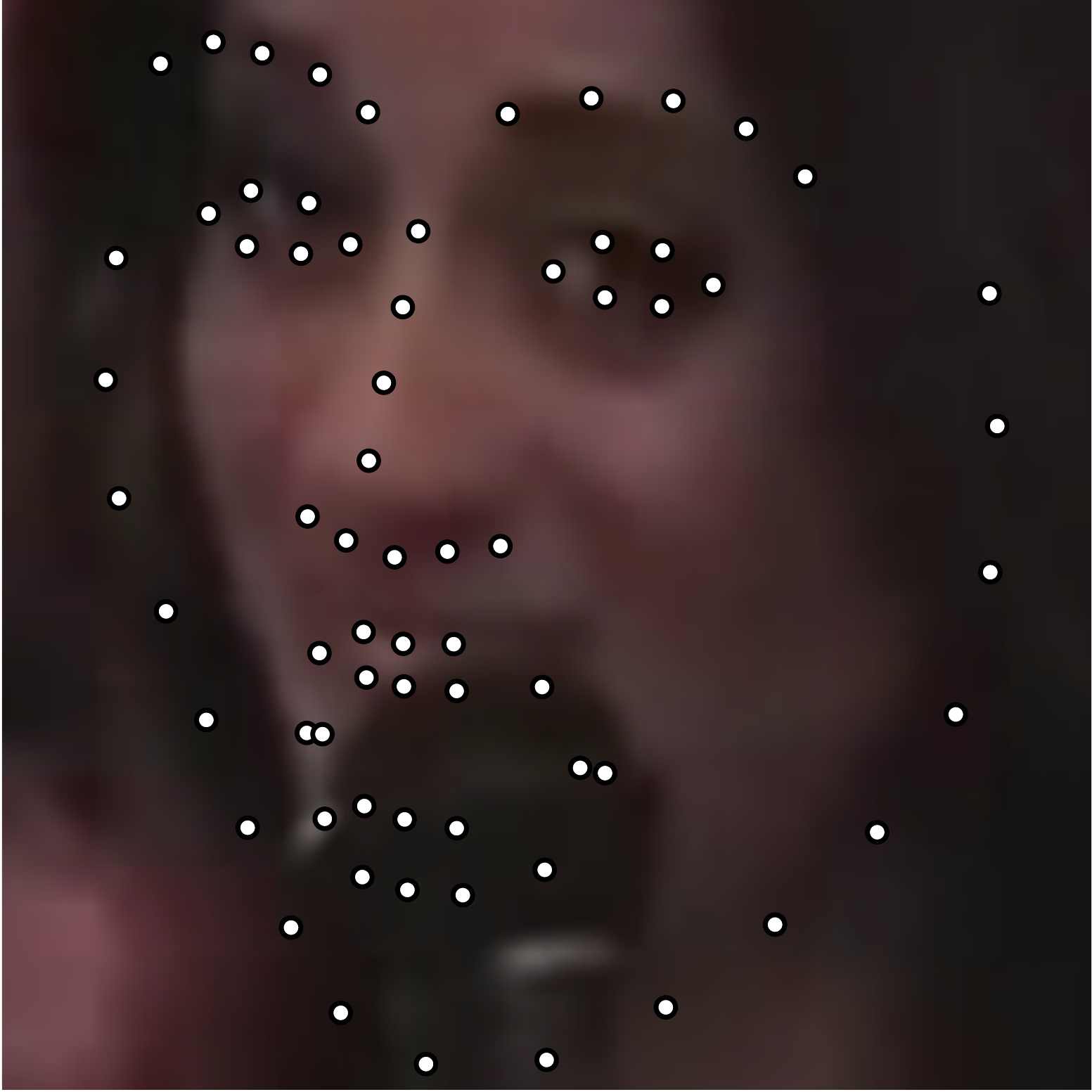} &
    \includegraphics[valign=m,width=0.09\linewidth]{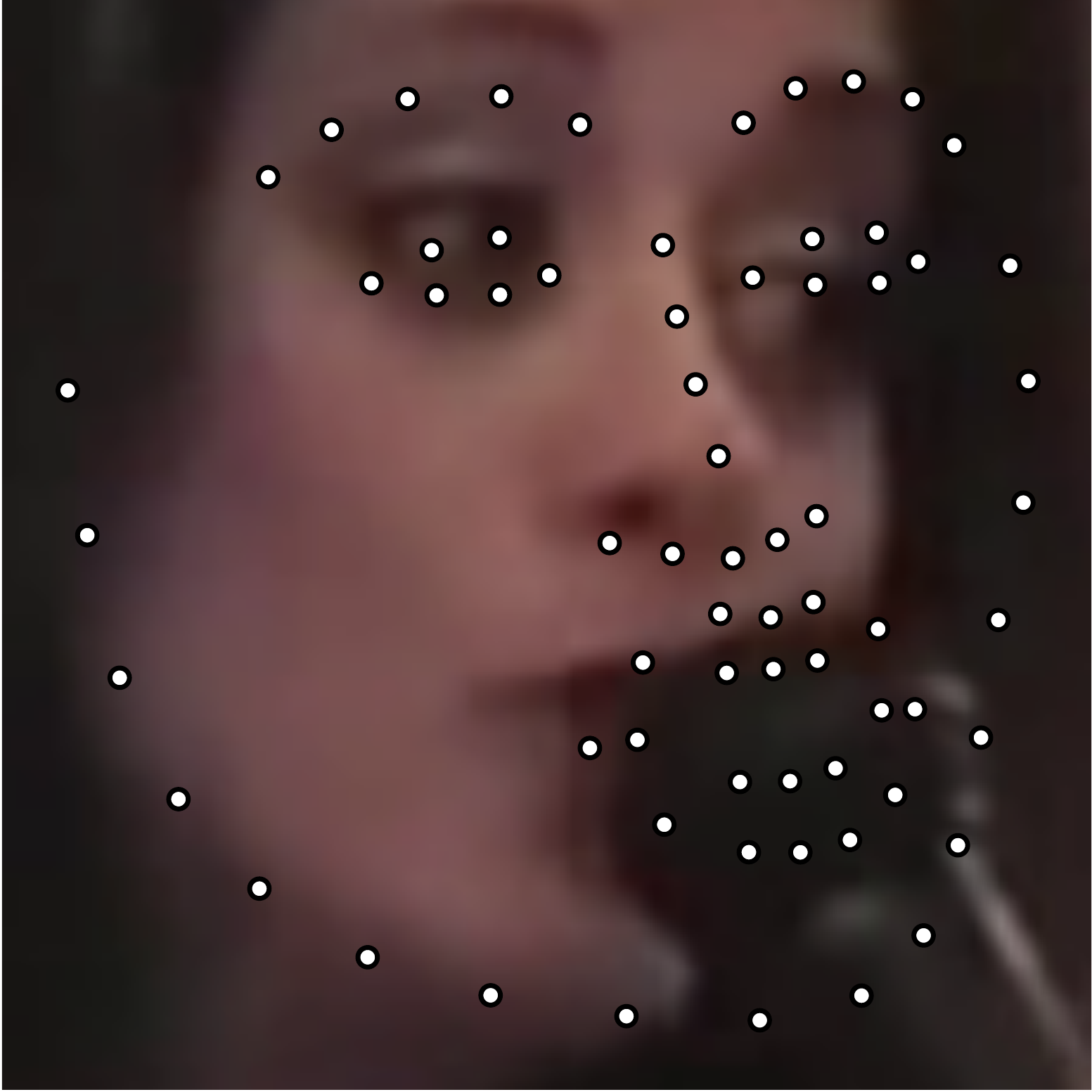} &
    \includegraphics[valign=m,width=0.09\linewidth]{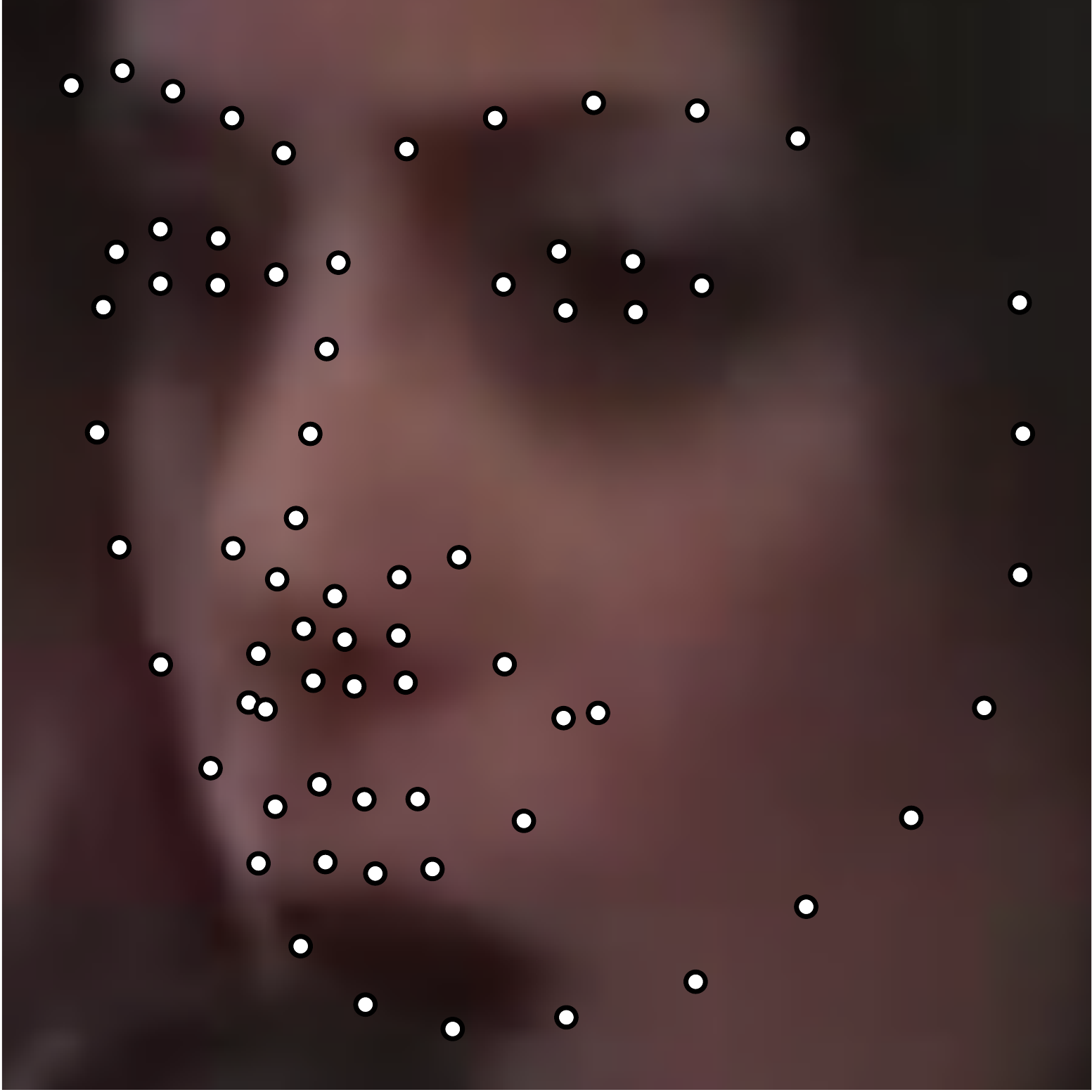}\\
\bottomrule
\end{tabular}
\caption{Exemplar deformable tracking results that are indicative of the fitting quality that corresponds to each error value for all video categories. The Area Under the Curve (AUC) and Failure Rate for all the experiments are computed based on the Cumulative Error Distributions (CED) limited at maximum error of $0.08$.}
\label{tbl:errors}
\end{table*}
%%%%%%%%%%%%%%%%%%%%%%%%%%%%%%%%%%%%%%%%%%%%%%%%%%%%%%%%%
%%%%%%%%%%%%%%%%%%%%%%%%%%%%%%%%%%%%%%%%%%%%%%%%%%%%
%%%% [TAB]: DETECTION + LANDMARK_LOCALISATION %%%%%%
%%%%%%%%%%%%%%%%%%%%%%%%%%%%%%%%%%%%%%%%%%%%%%%%%%%%
\begin{table*}[!b]
\centering
\begin{tabular}{cc r cc r cc r cc}
\toprule
   \multicolumn{2}{c}{Method} & & \multicolumn{2}{c}{Category 1} & & \multicolumn{2}{c}{Category 2} & & \multicolumn{2}{c}{Category 3} \\
   \cmidrule(lr){1-2}\cmidrule(lr){4-5}\cmidrule(lr){7-8}\cmidrule(lr){10-11}
   \multirow{2}{*}{\emph{Detection}} & \emph{Landmark} & & \multirow{2}{*}{\emph{AUC}} & \emph{Failure} & & \multirow{2}{*}{\emph{AUC}} & \emph{Failure} & & \multirow{2}{*}{\emph{AUC}} & \emph{Failure} \\
   & \emph{Localisation} & & & \emph{Rate (\%)} & & & \emph{Rate (\%)} & & & \emph{Rate (\%)} \\
   \cmidrule[\heavyrulewidth](){1-2}\cmidrule[\heavyrulewidth](){4-11}
   \multirow{4}{*}{DPM} & AAM & & 0.447 & 29.445 & & 0.466 & 21.158 & & 0.376 & 33.261 \\
   &                     CFSS & & \cellcolor{colour2}\textbf{0.764} &  \cellcolor{colour2}\textbf{3.789} & & \cellcolor{colour1}\textbf{0.767} &  \cellcolor{colour1}\textbf{1.363} & & \cellcolor{colour1}\textbf{0.717} &  \cellcolor{colour1}\textbf{5.259} \\
   &                      ERT & & \cellcolor{colour1}\textbf{0.772} & \cellcolor{colour1}\textbf{3.493} & & \cellcolor{colour2}\textbf{0.765} & \cellcolor{colour2}\textbf{1.558} & & \cellcolor{colour2}\textbf{0.714} & \cellcolor{colour2}\textbf{6.100}   \\
   &                      SDM & & 0.673 &    3.800 & & 0.646 &  1.369 & & 0.585 &  5.880 \\
   \cmidrule(lr){1-2}\cmidrule(lr){4-5}\cmidrule(lr){7-8}\cmidrule(lr){10-11}
   \multirow{4}{*}{SS-DPM} & AAM & & 0.474 & 37.473 & & 0.502 & 33.807 & & 0.161 & 77.932 \\
   &                        CFSS & & 0.609 & 21.773 & & 0.566 & 24.261 & & 0.244 & 65.926 \\
   &                         ERT & & 0.635 & 21.445 & & 0.608 & 21.638 & & 0.243 & 67.407 \\
   &                         SDM & & 0.582 & 21.225 & & 0.537 & 21.748 & & 0.217 & 67.602 \\
   \cmidrule(lr){1-2}\cmidrule(lr){4-5}\cmidrule(lr){7-8}\cmidrule(lr){10-11}
   \multirow{4}{*}{SVM-HOG} & AAM & & 0.493 & 25.891 & & 0.487 & 22.414 & & 0.380 & 36.728\\
   &                         CFSS & & \cellcolor{colour3}\textbf{0.707} & \cellcolor{colour3}\textbf{12.953} & & \cellcolor{colour3}\textbf{0.663} &  \cellcolor{colour3}\textbf{16.318} & & 0.579 & 21.422 \\
   &                          ERT & & \cellcolor{colour4}\textbf{0.705} & \cellcolor{colour4}\textbf{13.285} & & \cellcolor{colour4}\textbf{0.653} & \cellcolor{colour4}\textbf{16.500} & & 0.570 & 22.303 \\
   &                          SDM & & 0.654 & 13.252 & & 0.619 & 16.312 & & 0.480 & 21.367 \\
   \cmidrule(lr){1-2}\cmidrule(lr){4-5}\cmidrule(lr){7-8}\cmidrule(lr){10-11}
   \multirow{4}{*}{VJ} & AAM & & 0.453 & 24.277 & & 0.532 & 19.500   & & 0.413 & 25.640  \\
   &                    CFSS & & 0.660 & 18.986 & & 0.651 & 17.805 & & \cellcolor{colour4}\textbf{0.641} & \cellcolor{colour4}\textbf{15.061} \\
   &                     ERT & & 0.658 & 19.292 & & 0.646 & 17.839 & & \cellcolor{colour3}\textbf{0.653} & \cellcolor{colour3}\textbf{14.942} \\
   &                     SDM & & 0.524 & 19.249 & & 0.548 & 17.769 & & 0.505 & 15.347 \\
\midrule[\heavyrulewidth]
   \multicolumn{11}{l}{\scriptsize Colouring denotes the methods' performance ranking per category:\hspace{0.2cm}$\color{colour1}\blacksquare$~first\hspace{0.2cm}$\color{colour2}\blacksquare$~second\hspace{0.2cm}$\color{colour3}\blacksquare$~third\hspace{0.2cm}$\color{colour4}\blacksquare$~fourth}\\
\bottomrule
\end{tabular}
\caption{Results for Experiment 1 of Section~\ref{exp:detection} (Detection + Landmark Localisation). The Area Under the Curve (AUC) and Failure Rate are reported. The top 4 performing curves are highlighted for each video category.}
\label{tab:exp_detection}
\end{table*}
%%%%%%%%%%%%%%%%%%%%%%%%%%%%%%%%%%%%%%%%%%%%%%%%%%%%
%%%%%%%%%%%%%%%%%%%%%%%%%%%%%%%%%%%%%%%%%%%%%%%%%%%%
%%%% [FIG]: DETECTION + LANDMARK_LOCALISATION %%%%%%
%%%%%%%%%%%%%%%%%%%%%%%%%%%%%%%%%%%%%%%%%%%%%%%%%%%%
\begin{figure*}[!t]
\subfloat[][Category 1]{
   \begin{minipage}{0.323\linewidth}
   \hspace{0.71cm}\includegraphics[height=0.90cm]{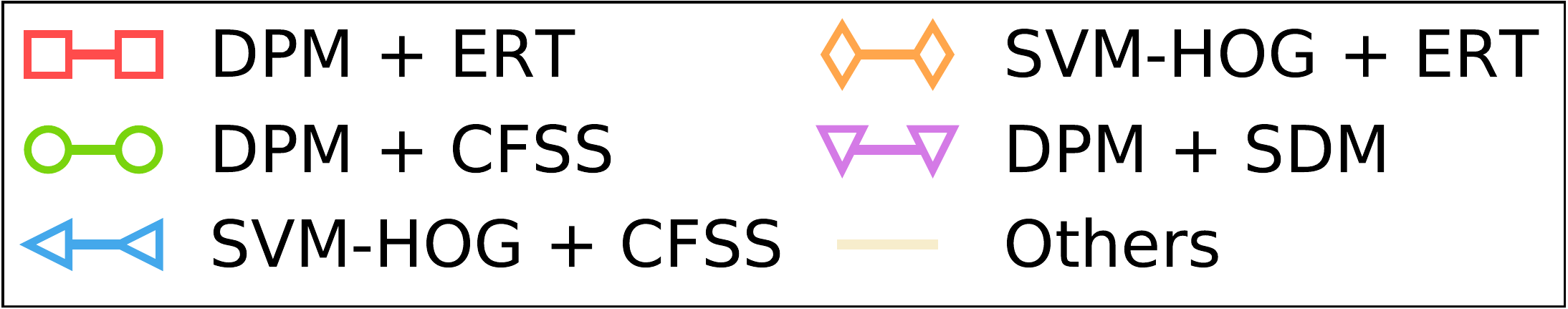}\\
   \includegraphics[width=\linewidth]{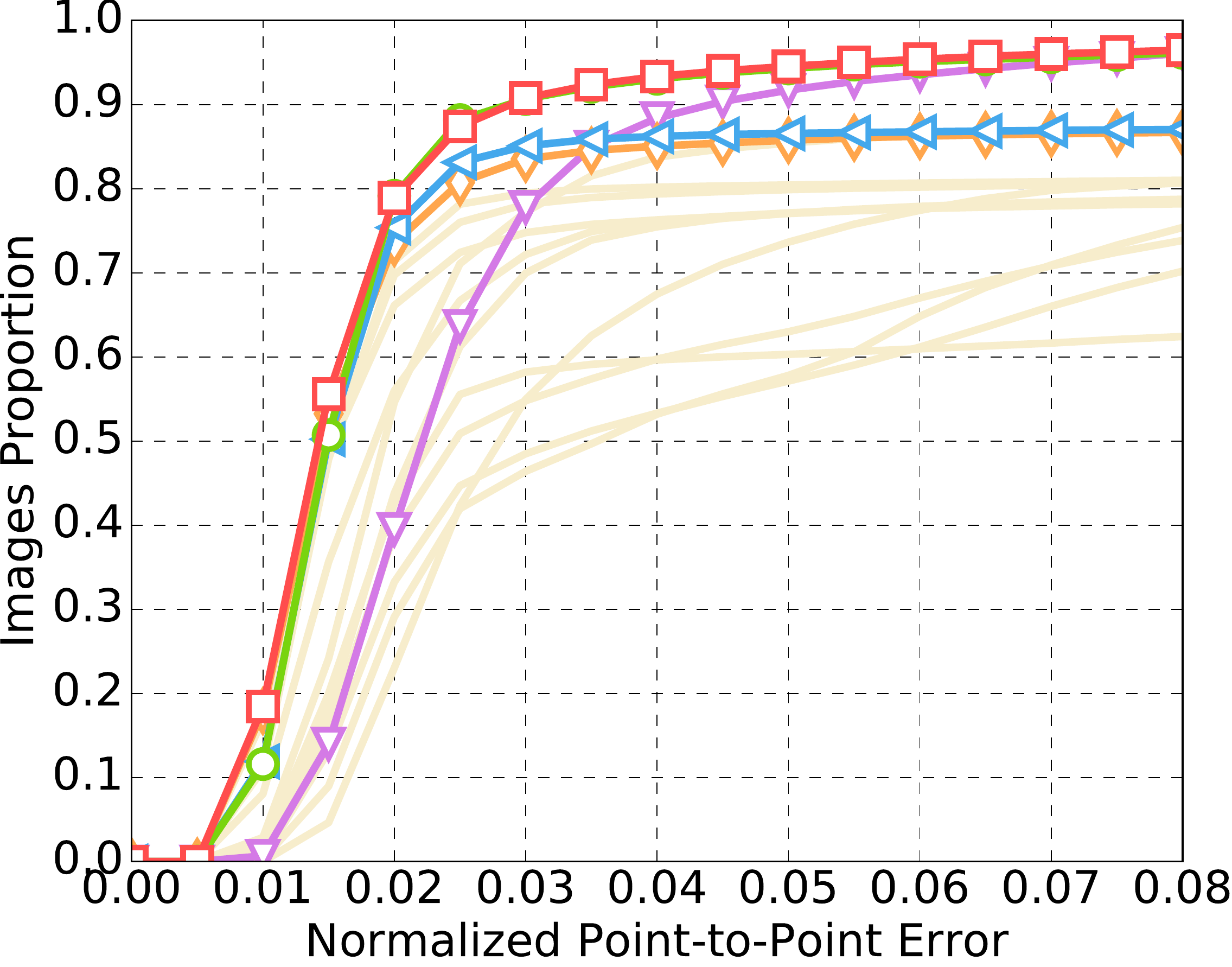}
   \end{minipage}
   }
\subfloat[][Category 2]{
   \begin{minipage}{0.323\linewidth}
   \hspace{0.71cm}\includegraphics[height=0.90cm]{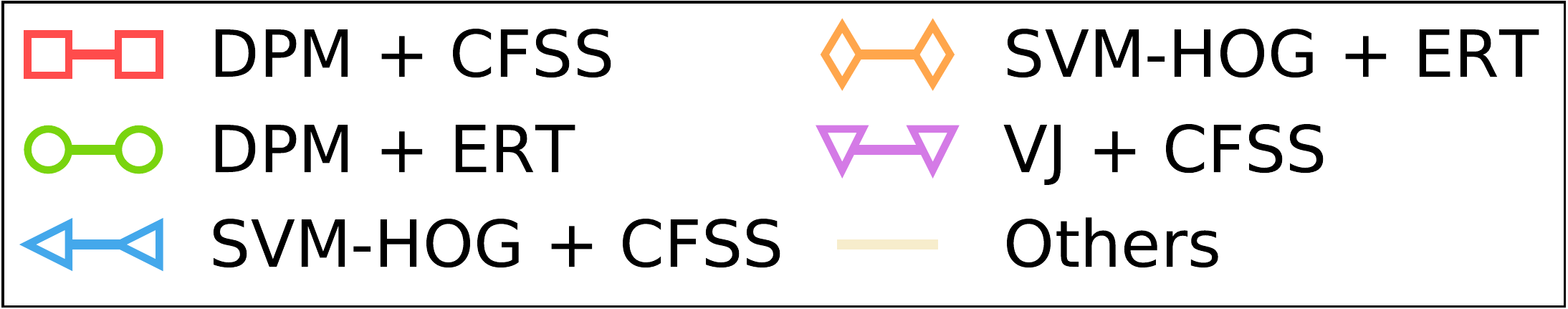}\\
   \includegraphics[width=\linewidth]{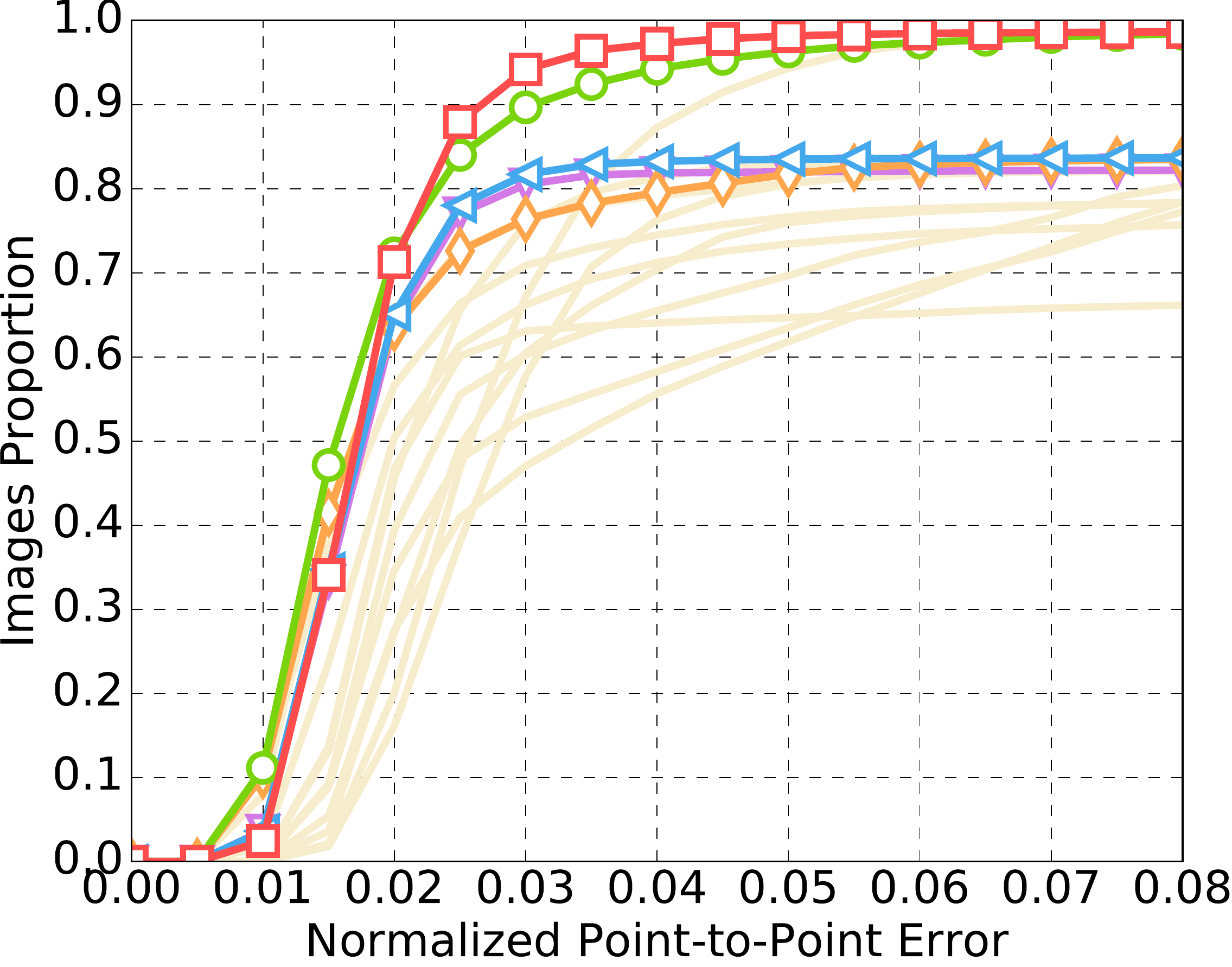}
   \end{minipage}
   }
\subfloat[][Category 3]{
   \begin{minipage}{0.323\linewidth}
   \hspace{1.175cm}\includegraphics[height=0.90cm]{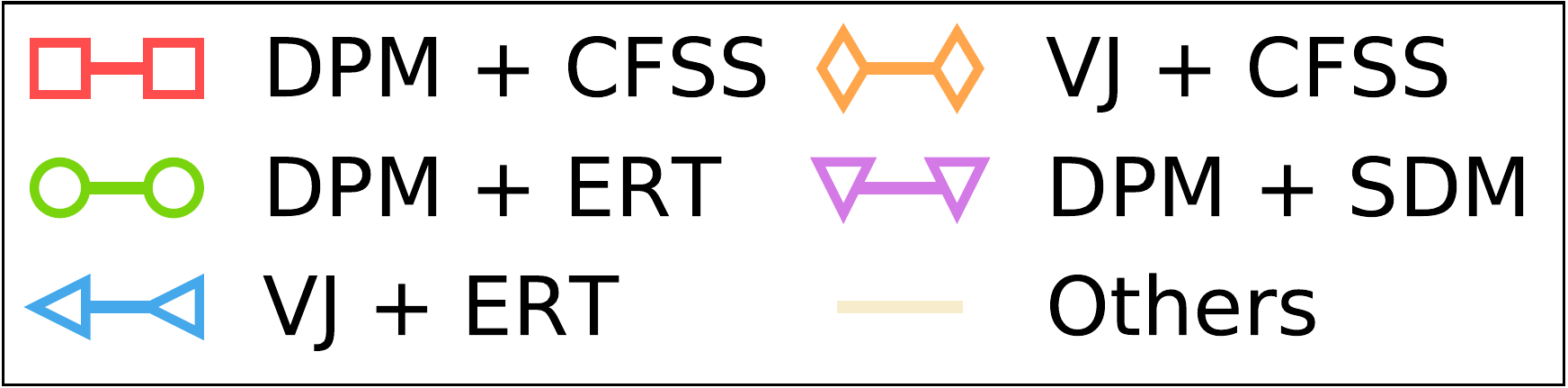}\\
   \includegraphics[width=\linewidth]{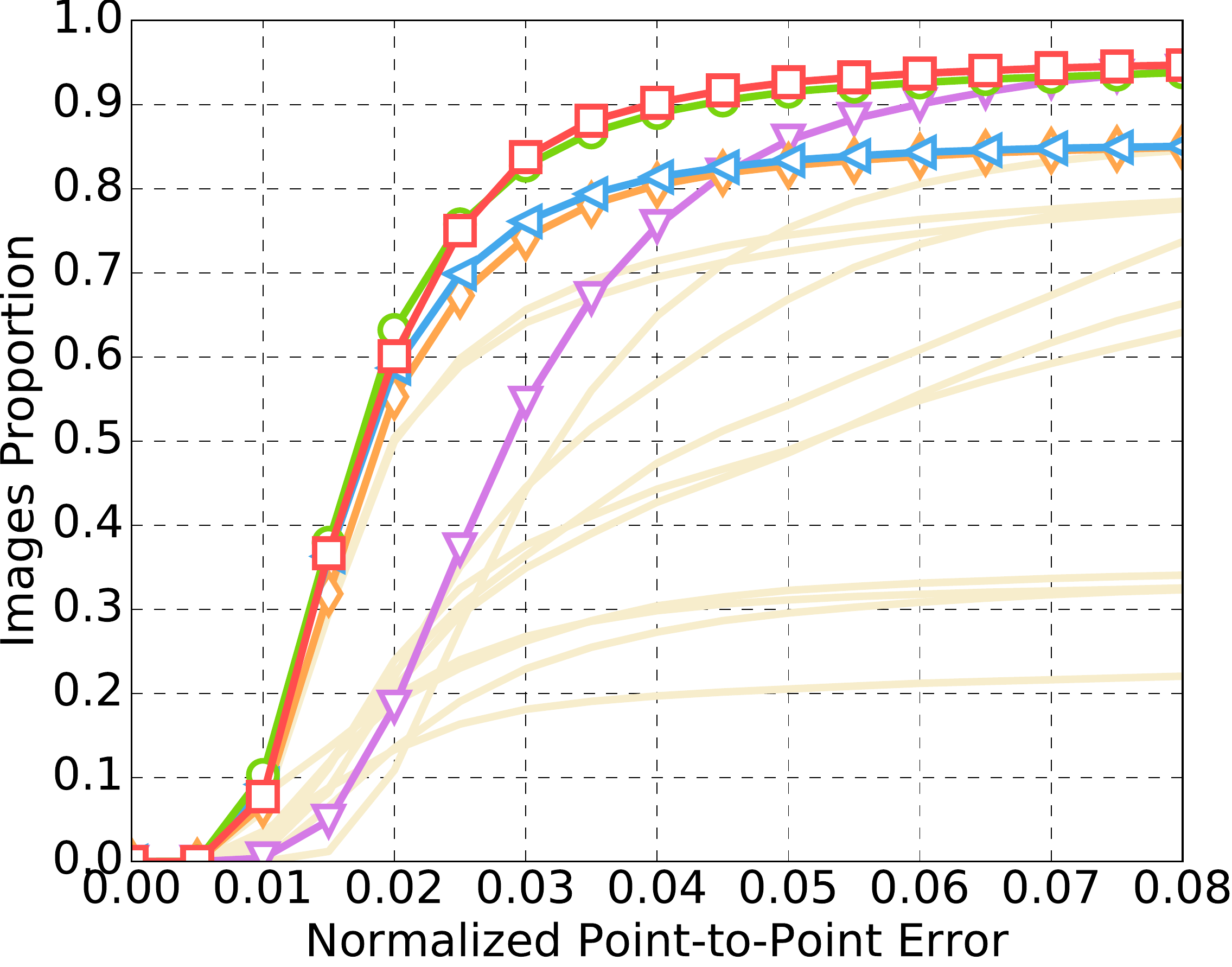}
   \end{minipage}
}
\caption{Results for Experiment 1 of Section~\ref{exp:detection} (Detection + Landmark Localisation). The top 5 performing curves are highlighted in each legend. Please see Table~\ref{tab:exp_detection} for a full summary.}
\label{fig:exp_detection}
\end{figure*}
%%%%%%%%%%%%%%%%%%%%%%%%%%%%%%%%%%%%%%%%%%%%%%%%%%%%
%%%%%%%%%%%%%%%%%%%%%%%%%%%%%%%%%%%%%%%%
%%%% DETECTION INIT FROM PREVIOUS %%%%%%
%%%%%%%%%%%%%%%%%%%%%%%%%%%%%%%%%%%%%%%%
\begin{figure*}[!b]
    \centering
    \includegraphics[width=0.8\textwidth]{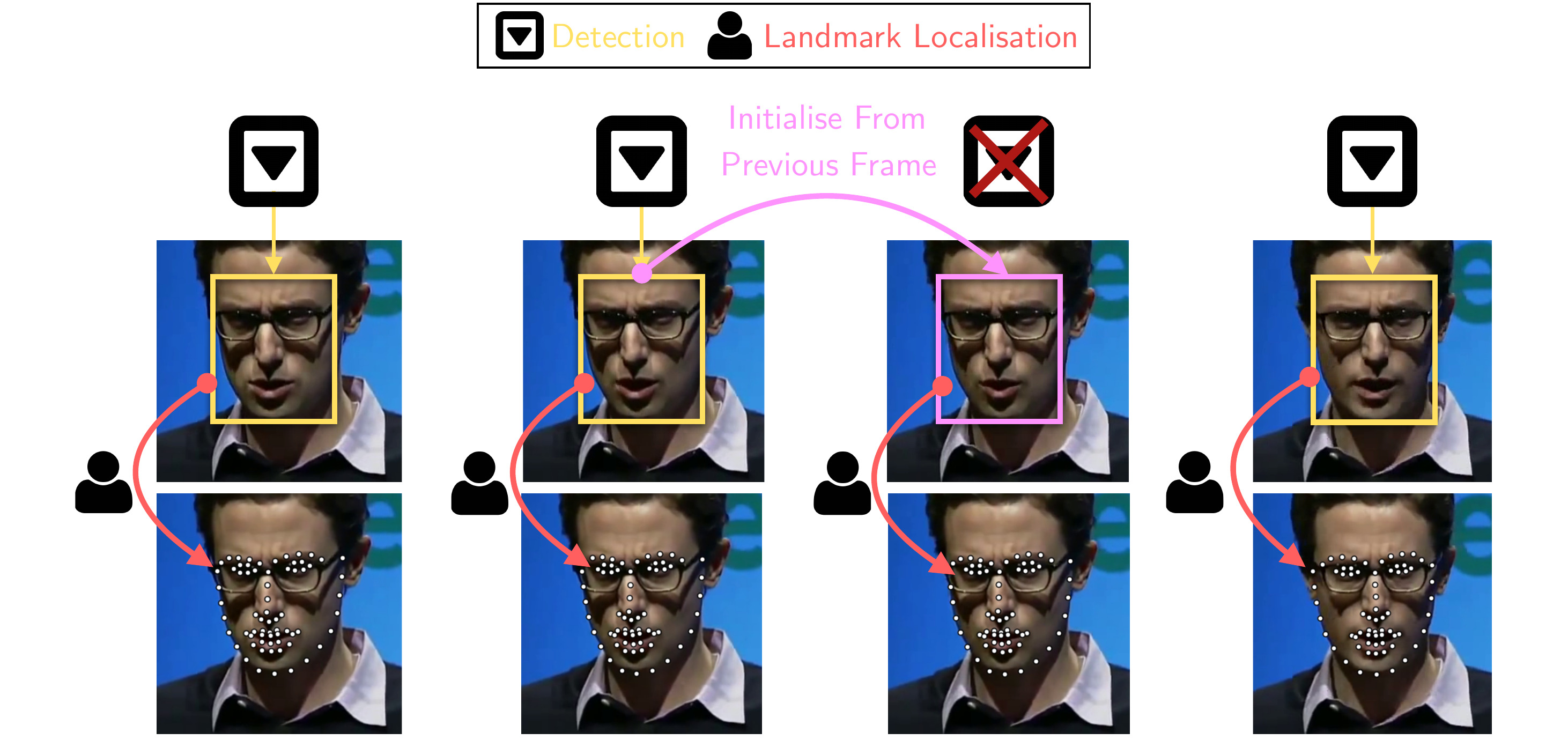}
    \caption{This figure gives a diagram of the reinitialisation scheme proposed in Section~\ref{exp:detection_init_from_previous}. Specifically, in case the face detector does not return a bounding box for a frame, the bounding box of the previous frame is used as a successful detection for the missing frame.}
    \label{fig:detection_failure}
\end{figure*}
%%%%%%%%%%%%%%%%%%%%%%%%%%%%%%%%%%%%%

%%%%%%%%%%%%%%%%%%%%%%%%%%%%%%%%%%%%%%%%%%%%%%%%%%%%%%%%%
\subsubsection{Quantitative Metrics}\label{subsubsec:metrics}
%%%%%%%%%%%%%%%%%%%%%%%%%%%%%%%%%%%%%%%%%%%%%%%%%%%%%%%%%
The errors reported for all the following experiments are with respect to the landmark localisation error. The error metric employed is the mean Euclidean distance of the 68 points, normalised by the diagonal of the ground truth bounding box ($\sqrt{width^2 + height^2}$). This metric was chosen as it is robust to changes in head pose which are frequent within the 300VW sequences. The graphs that are shown are cumulative error distribution (CED) plots that provide the proportion of images less than or equal to a particular error. We also provide summary tables with respect to the Area Under the Curve (AUC) of the CED plots, considered up to a maximum error. Errors above this maximum threshold, which is fixed to $0.08$, are considered failures to accurately localise the facial landmarks. Therefore, we also report the failure rate, as a percentage, which marks the proportion of images that are not considered within the CED plots. Table~\ref{tbl:errors} shows some indicative examples of the deformable fitting quality that corresponds to each error value for all video categories. When ranking methods, we consider the AUC as the primary statistic and only resort to considering the failure rate in cases where there is little distinction between methods' AUC values.

%%%%%%%%%%%%%%%%%%%%%%%%%%%%%%%%%%%%%%%%%%%%%%%%%%%%%%%%%%%%%%%
%%%% [TAB]: DETECTION + LANDMARK_LOCALISATION + PREVIOUS %%%%%%
%%%%%%%%%%%%%%%%%%%%%%%%%%%%%%%%%%%%%%%%%%%%%%%%%%%%%%%%%%%%%%%
\begin{table*}[!t]
\centering
\begin{tabular}{cc r cc r cc r cc}
\toprule
   \multicolumn{2}{c}{Method} & & \multicolumn{2}{c}{Category 1} & & \multicolumn{2}{c}{Category 2} & & \multicolumn{2}{c}{Category 3} \\
   \cmidrule(lr){1-2}\cmidrule(lr){4-5}\cmidrule(lr){7-8}\cmidrule(lr){10-11}
   \multirow{2}{*}{\emph{Detection}} & \emph{Landmark} & & \multirow{2}{*}{\emph{AUC}} & \emph{Failure} & & \multirow{2}{*}{\emph{AUC}} & \emph{Failure} & & \multirow{2}{*}{\emph{AUC}} & \emph{Failure} \\
   & \emph{Localisation} & & & \emph{Rate (\%)} & & & \emph{Rate (\%)} & & & \emph{Rate (\%)} \\
   \cmidrule[\heavyrulewidth](){1-2}\cmidrule[\heavyrulewidth](){4-11}
   \multirow{4}{*}{DPM} & AAM & & 0.572 & 18.840 & & 0.621 & 10.617 & & 0.493 & 21.711 \\
                       & CFSS & & \cellcolor{colour2}\textbf{0.765} & \cellcolor{colour2}\textbf{3.415} & & \cellcolor{colour1}\textbf{0.769} & \cellcolor{colour1}\textbf{0.815} & & \cellcolor{colour1}\textbf{0.720} & \cellcolor{colour1}\textbf{4.786} \\
                        & ERT & & \cellcolor{colour1}\textbf{0.773} & \cellcolor{colour1}\textbf{3.221} & & \cellcolor{colour2}\textbf{0.767} & \cellcolor{colour2}\textbf{1.156} & & \cellcolor{colour2}\textbf{0.716} & \cellcolor{colour2}\textbf{5.620} \\
                        & SDM & & 0.674 &  3.727 & & 0.654 & 1.129  & & 0.579 & 6.006 \\
   \cmidrule(lr){1-2}\cmidrule(lr){4-5}\cmidrule(lr){7-8}\cmidrule(lr){10-11}
   \multirow{4}{*}{SS-DPM} & AAM & & 0.507 & 32.867 & & 0.526 & 28.781 & & 0.175 & 75.646 \\
                          & CFSS & & 0.609 & 21.734 & & 0.576 & 22.070 & & 0.248 & 65.421 \\
                           & ERT & & 0.636 & 21.397 & & 0.622 & 18.459 & & 0.246 & 66.905 \\
                           & SDM & & 0.594 & 21.306 & & 0.569 & 18.444 & & 0.227 & 67.653 \\
   \cmidrule(lr){1-2}\cmidrule(lr){4-5}\cmidrule(lr){7-8}\cmidrule(lr){10-11}
   \multirow{4}{*}{SVM-HOG} & AAM & & 0.627 & 13.770 & & 0.643 & 11.210 & & 0.526 & 20.215 \\
                           & CFSS & & \cellcolor{colour3}\textbf{0.759} & \cellcolor{colour3}\textbf{5.009} & & \cellcolor{colour3}\textbf{0.747} & \cellcolor{colour3}\textbf{4.186} & & 0.632 & 12.179 \\
                            & ERT & & \cellcolor{colour4}\textbf{0.750} & \cellcolor{colour4}\textbf{6.002} & & \cellcolor{colour4}\textbf{0.717} & \cellcolor{colour4}\textbf{6.428} & & 0.615 & 13.963 \\
                            & SDM & & 0.685 & 6.218 & & 0.676 & 6.325 & & 0.522 & 13.234 \\
   \cmidrule(lr){1-2}\cmidrule(lr){4-5}\cmidrule(lr){7-8}\cmidrule(lr){10-11}
   \multirow{4}{*}{VJ} & AAM & & 0.570 & 18.339 & & 0.593 & 15.612 & & 0.546 & 16.831 \\
                      & CFSS & & 0.685 & 14.945 & & 0.686 & 12.619 & & \cellcolor{colour4}\textbf{0.660} & \cellcolor{colour4}\textbf{11.612} \\
                       & ERT & & 0.679 & 15.783 & & 0.675 & 12.862 & & \cellcolor{colour3}\textbf{0.672} & \cellcolor{colour3}\textbf{11.543} \\
                       & SDM & & 0.536 & 16.452 & & 0.573 & 13.175 & & 0.530 & 12.779 \\
\midrule[\heavyrulewidth]
   \multicolumn{11}{l}{\scriptsize Colouring denotes the methods' performance ranking per category:\hspace{0.2cm}$\color{colour1}\blacksquare$~first\hspace{0.2cm}$\color{colour2}\blacksquare$~second\hspace{0.2cm}$\color{colour3}\blacksquare$~third\hspace{0.2cm}$\color{colour4}\blacksquare$~fourth}\\
\bottomrule
\end{tabular}
\caption{Results for Experiment 2 of Section~\ref{exp:detection_init_from_previous} (Detection + Landmark Localisation + Initialisation From Previous Frame). The Area Under the Curve (AUC) and Failure Rate are reported. The top 4 performing curves are highlighted for each video category.}
\label{tab:exp_detection_init_from_previous}
\end{table*}
%%%%%%%%%%%%%%%%%%%%%%%%%%%%%%%%%%%%%%%%%%%%%%%%%%%%%%%%%%%%%%%
\subsection{Experiment 1: Detection and Landmark Localisation}\label{exp:detection}
In this experiment, we validate the most frequently used facial deformable tracking strategy, i.e. performing face detection followed by landmark localisation \emph{on each frame independently}. If a detector fails to return a frame, that frame is considered as having infinite error and thus will appear as part of the failures in Table~\ref{tab:exp_detection}. Note that the AUC is robust to the use of infinite errors. In frames where multiple bounding boxes are returned, the box with the highest confidence is kept, limiting the results of the detectors to a single bounding box per image. A high level diagram explaining the detection procedure for this experiment is given by Figure~\ref{fig:overview}.

Specifically, in this experiment we consider the 4 face detectors of Table~\ref{tbl:detectors} (DPM, SS-DPM, HOG+SVM, VJ) with the 4 landmark localisation techniques of Table~\ref{tbl:alignment} (AAM, CFSS, ERT, SDM), for a total of 16 results. The results of the experiment are given in Table~\ref{tab:exp_detection} and Figure~\ref{fig:exp_detection}. 
The results indicate that the AAM performs poorly as it achieves the lowest performance across all face detectors. The discriminative CFSS and ERT landmark localisation methods consistently outperform SDM. From the detectors point of view, it seems that the strongly supervised DPM (SS-DPM) is the worst and provides the highest failure rates. On the other hand, the weakly supervised DPM (DPM) outperforms the rest of the detectors for all video categories in terms of both accuracy (i.e. AUC) and robustness (i.e. Failure Rate). For the graphs that correspond to all 16 methods, as well as a video with the results of the top 5 methods\textsuperscript{\ref{foot:detection}}, please refer to the supplementary material.

%%%%%%%%%%%%%%%%%%%%%%%%%%%%%%%%%%%%%%%%%%%%%%%%%%%%%%%%%%%%%%%
%%%% [FIG]: DETECTION + LANDMARK_LOCALISATION + PREVIOUS %%%%%%
%%%%%%%%%%%%%%%%%%%%%%%%%%%%%%%%%%%%%%%%%%%%%%%%%%%%%%%%%%%%%%%
\begin{figure*}[!t]
\subfloat[][Category 1]{
   \begin{minipage}{0.323\linewidth}
   \hspace{0.69cm}\includegraphics[height=0.90cm]{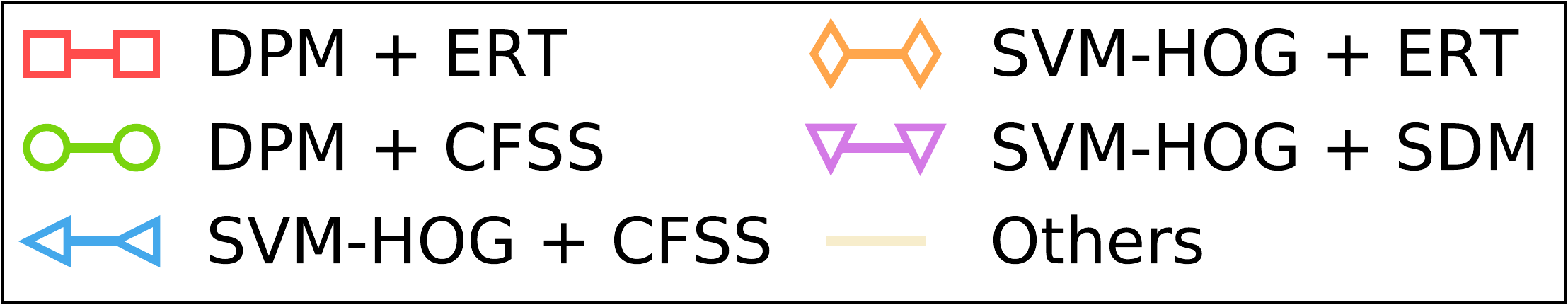}\\
   \includegraphics[width=\linewidth]{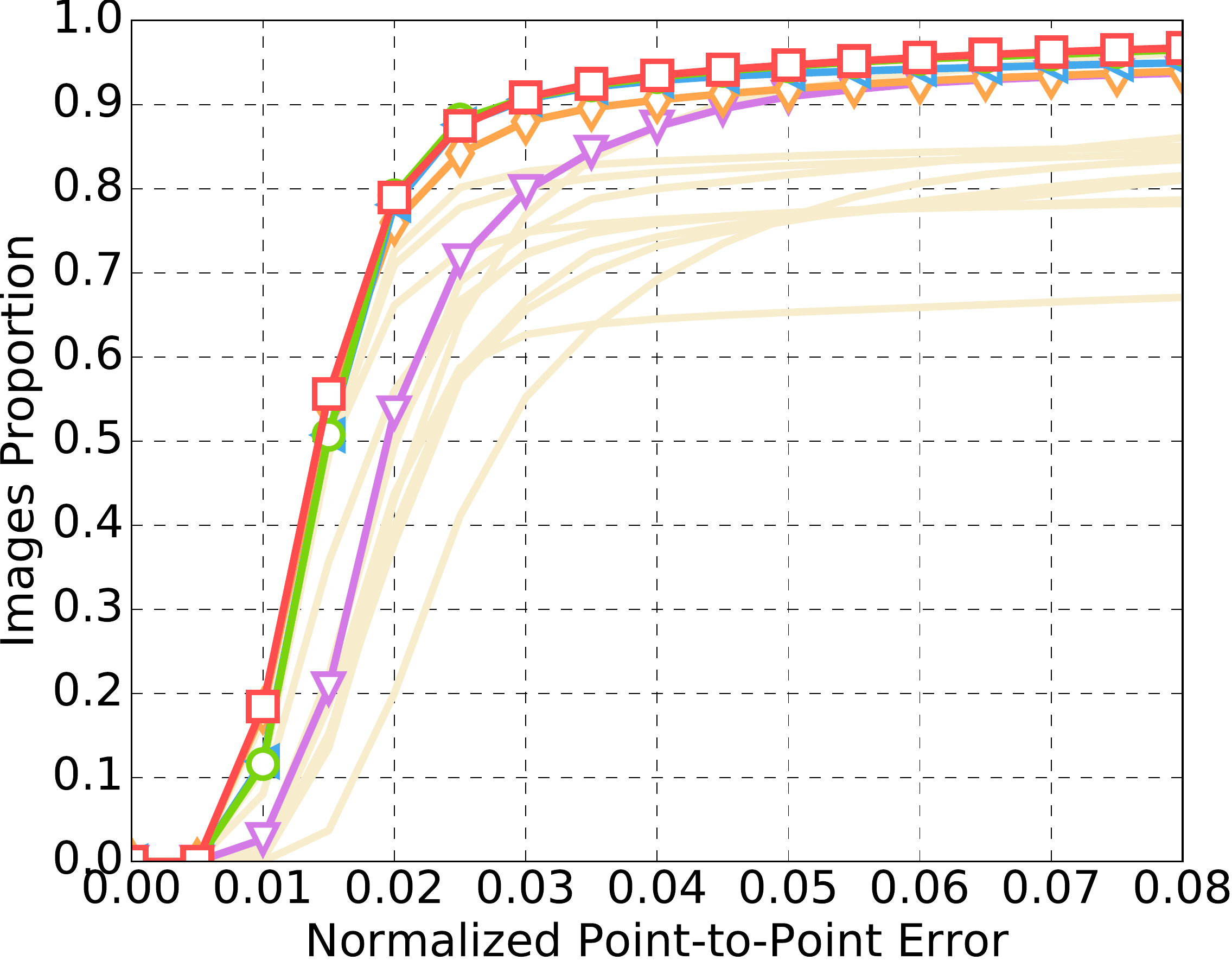}
   \end{minipage}
   }
\subfloat[][Category 2]{
   \begin{minipage}{0.323\linewidth}
   \hspace{0.71cm}\includegraphics[height=0.90cm]{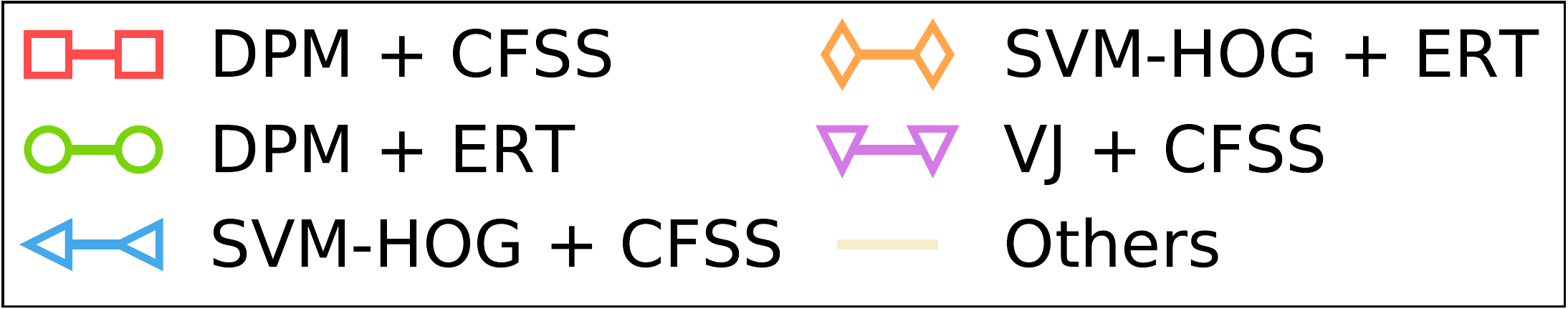}\\
   \includegraphics[width=\linewidth]{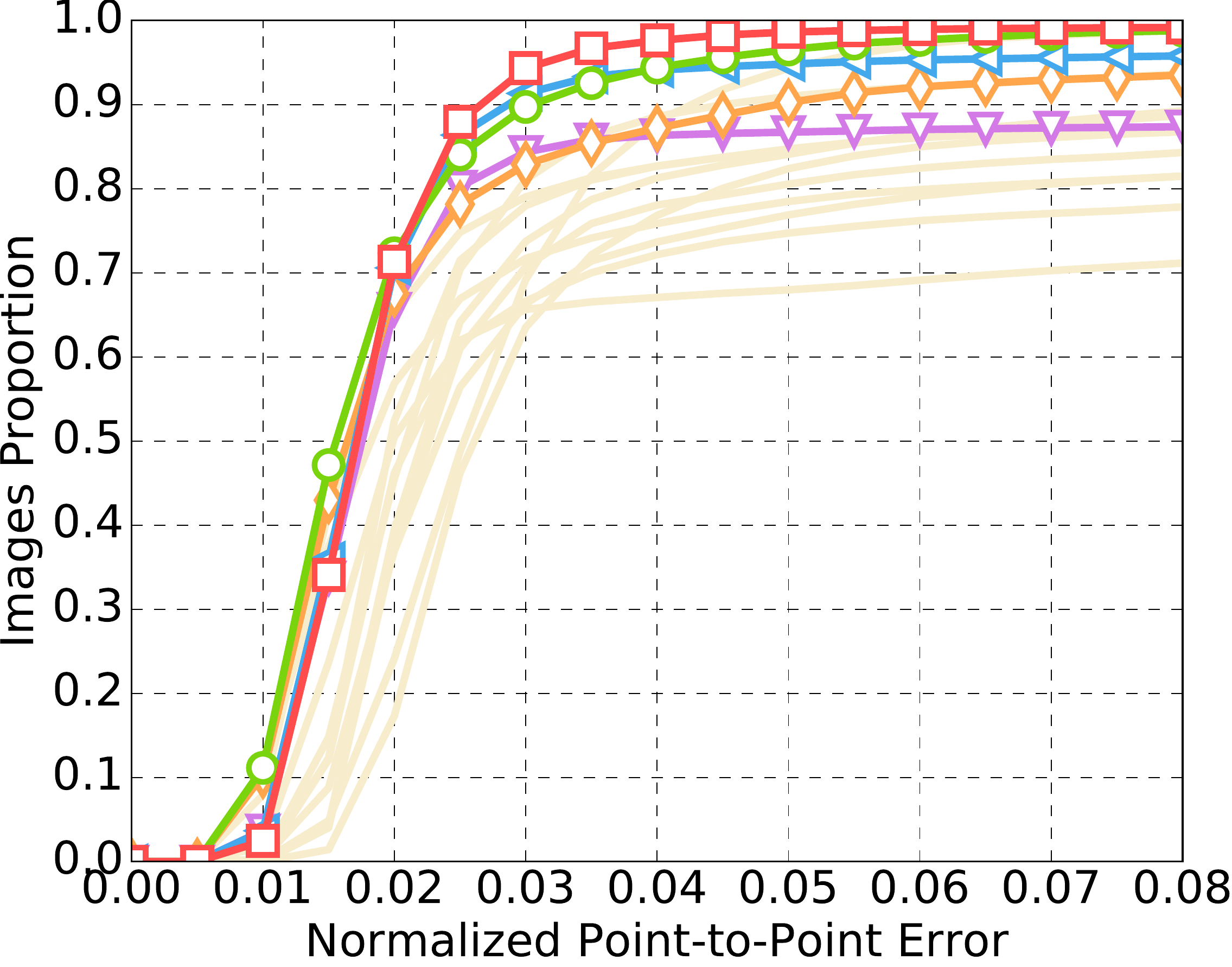}
   \end{minipage}
   }
\subfloat[][Category 3]{
   \begin{minipage}{0.323\linewidth}
   \hspace{0.9cm}\includegraphics[height=0.90cm]{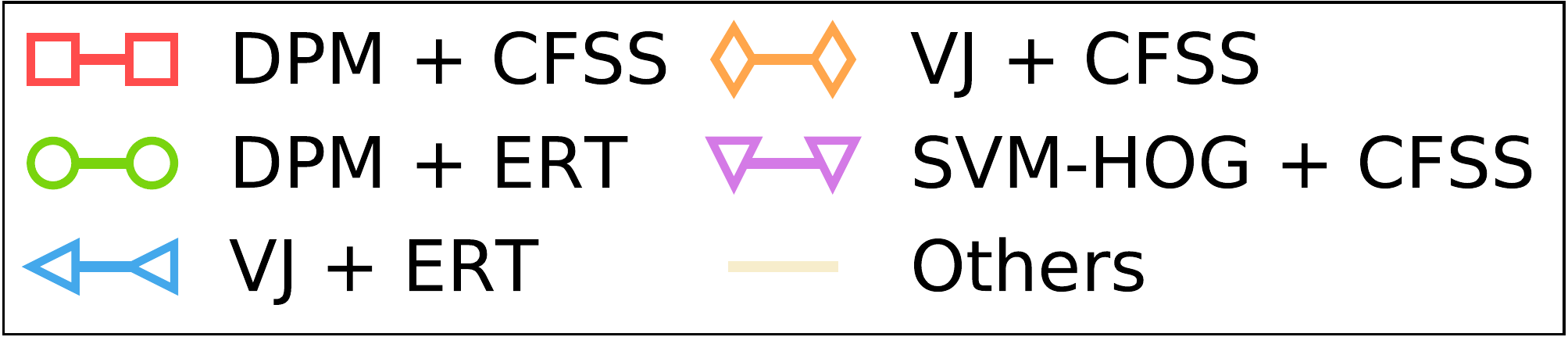}\\
   \includegraphics[width=\linewidth]{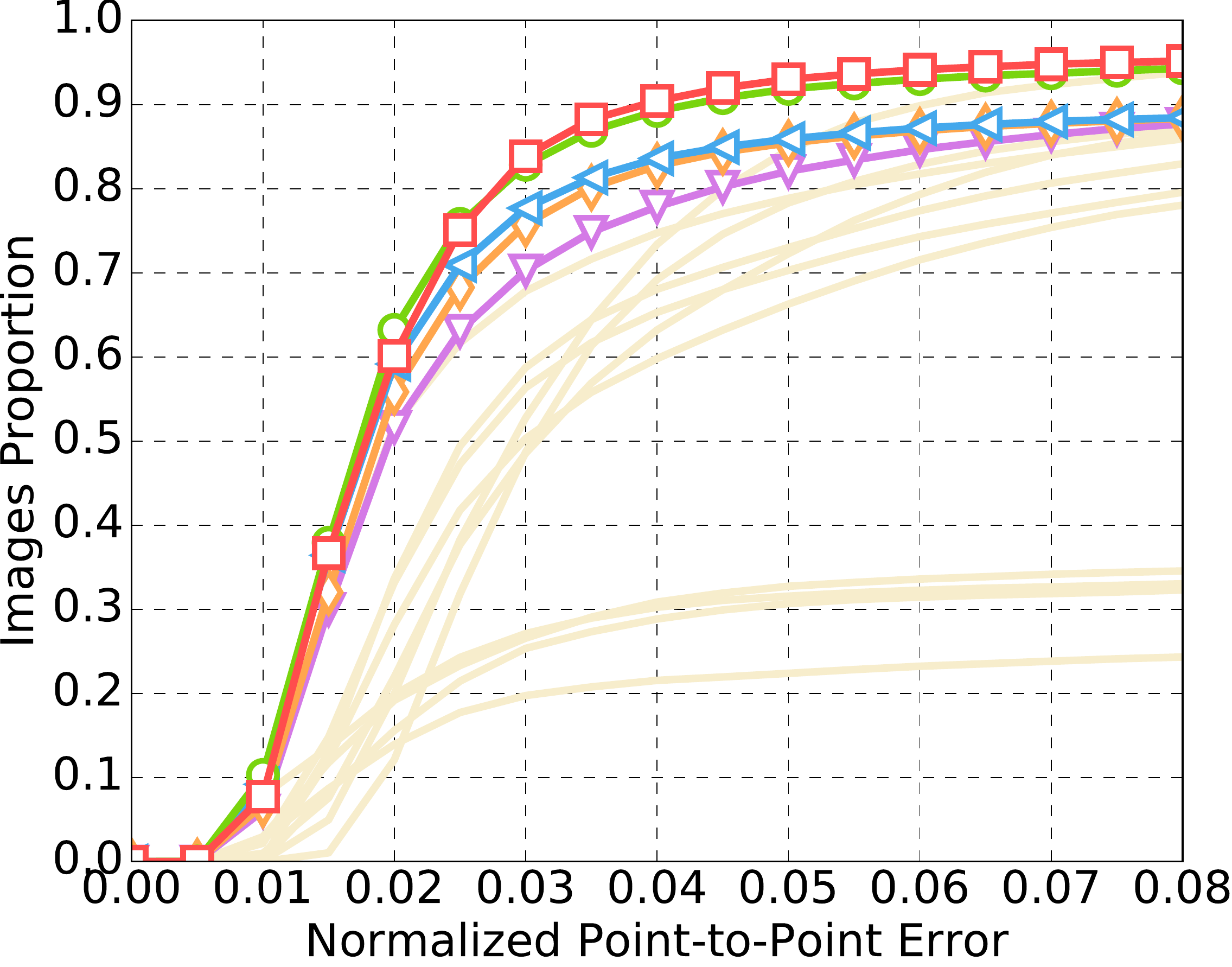}
   \end{minipage}
}
\caption{Results for Experiment 2 of Section~\ref{exp:detection_init_from_previous} (Detection + Landmark Localisation + Initialisation From Previous Frame). The top 5 performing curves are highlighted in each legend. Please see Table~\ref{tab:exp_detection_init_from_previous} for a full summary.}
\label{fig:exp_detection_init_from_previous}
\end{figure*}
%%%%%%%%%%%%%%%%%%%%%%%%%%%%%%%%%%%%%%%%%%%%%%%%%%%%%%%%%%%%%%%
%%%%%%%%%%%%%%%%%%%%%%%%%%%%%%%%%%%%%%%%%%%%%%%%%%%%%%%%%%%%%%%
%%%% [FIG]: DETECTION + LANDMARK_LOCALISATION + PREVIOUS %%%%%%
%%%%%%%%%%%%%%%%%%%%%%%%%%%%%%%%%%%%%%%%%%%%%%%%%%%%%%%%%%%%%%%
\begin{figure*}[!t]
\subfloat[][Category 1]{
   \begin{minipage}{0.323\linewidth}
   \centering
   \includegraphics[height=1.43cm]{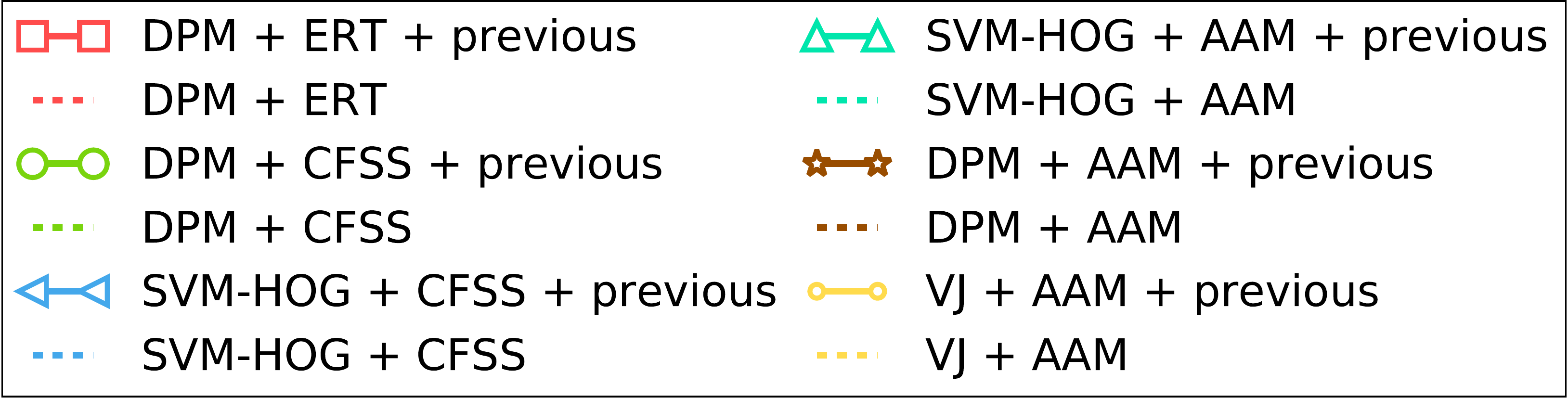}\\
   \includegraphics[width=\linewidth]{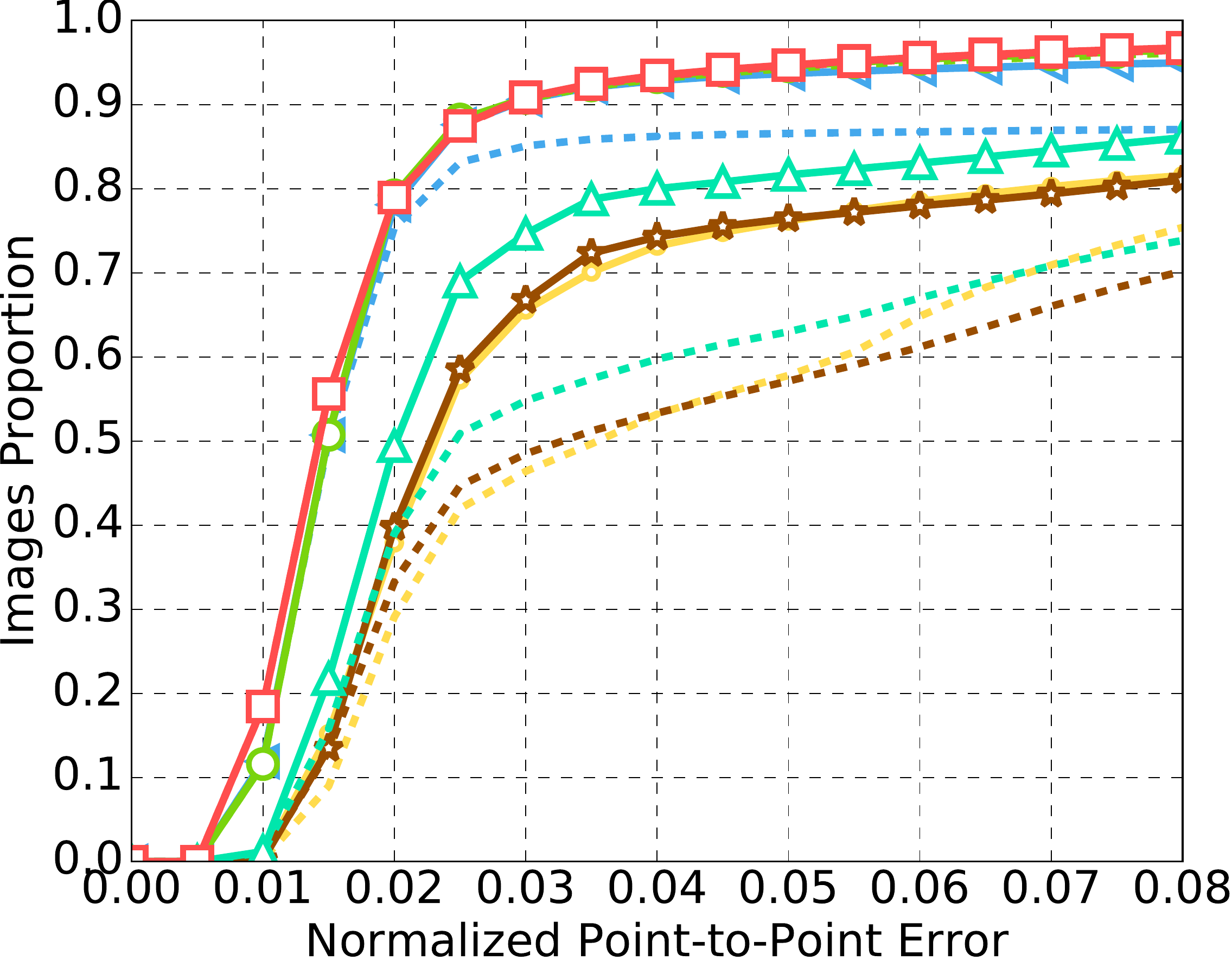}
   \end{minipage}
   }
\subfloat[][Category 2]{
   \begin{minipage}{0.323\linewidth}
   \centering
   \includegraphics[height=1.43cm]{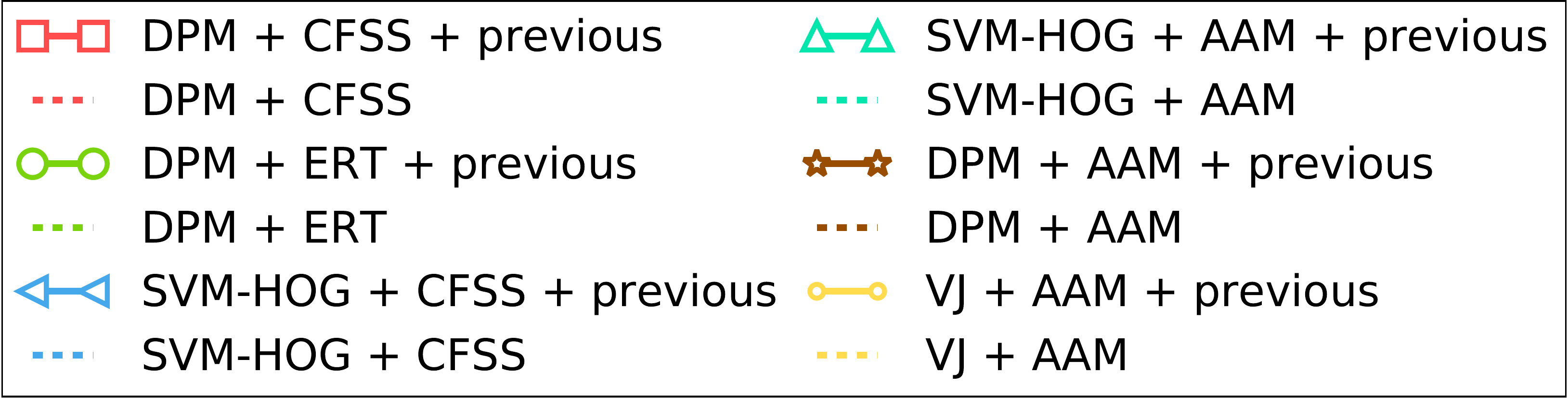}\\
   \includegraphics[width=\linewidth]{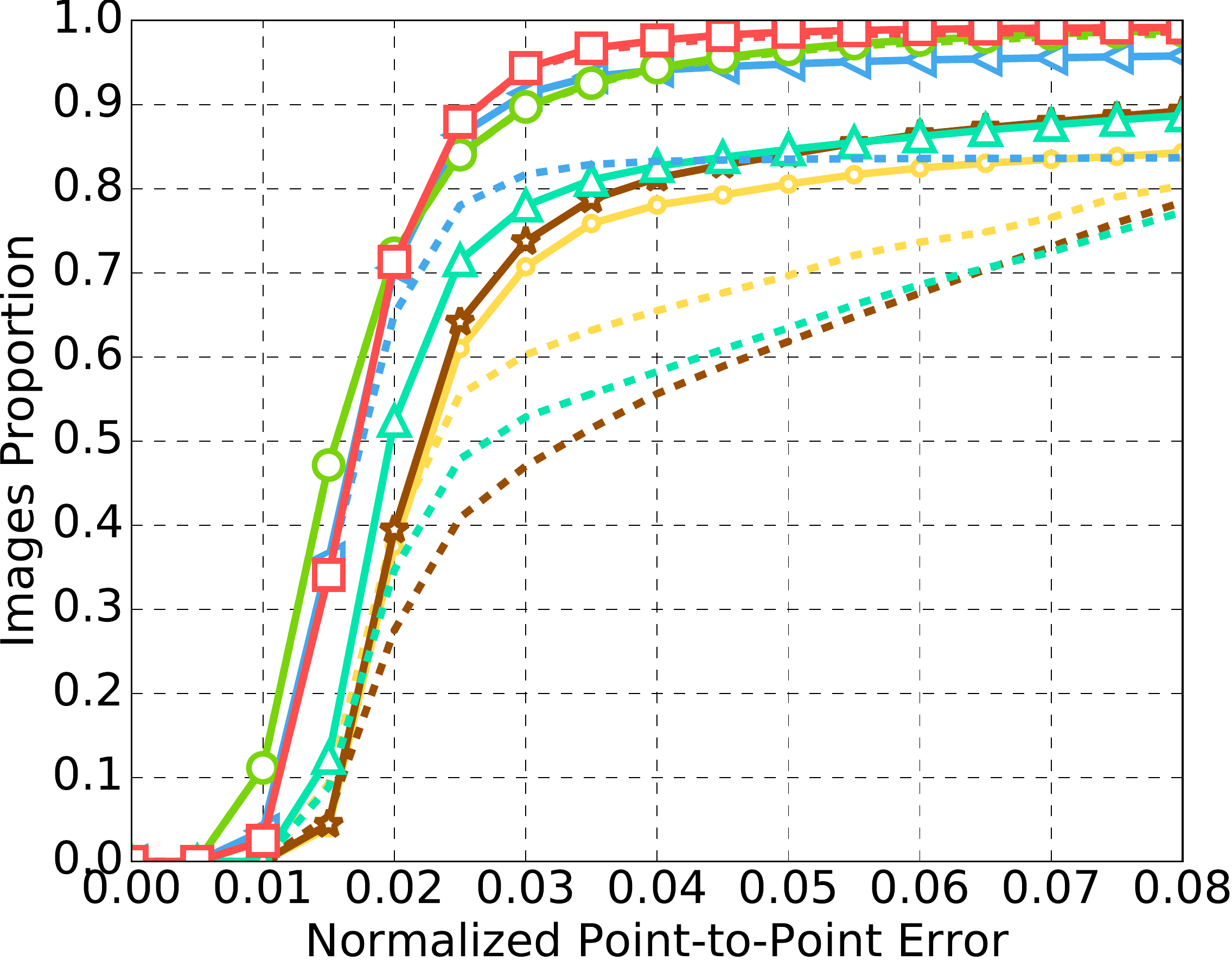}
   \end{minipage}
   }
\subfloat[][Category 3]{
   \begin{minipage}{0.323\linewidth}
   \centering
   \includegraphics[height=1.43cm]{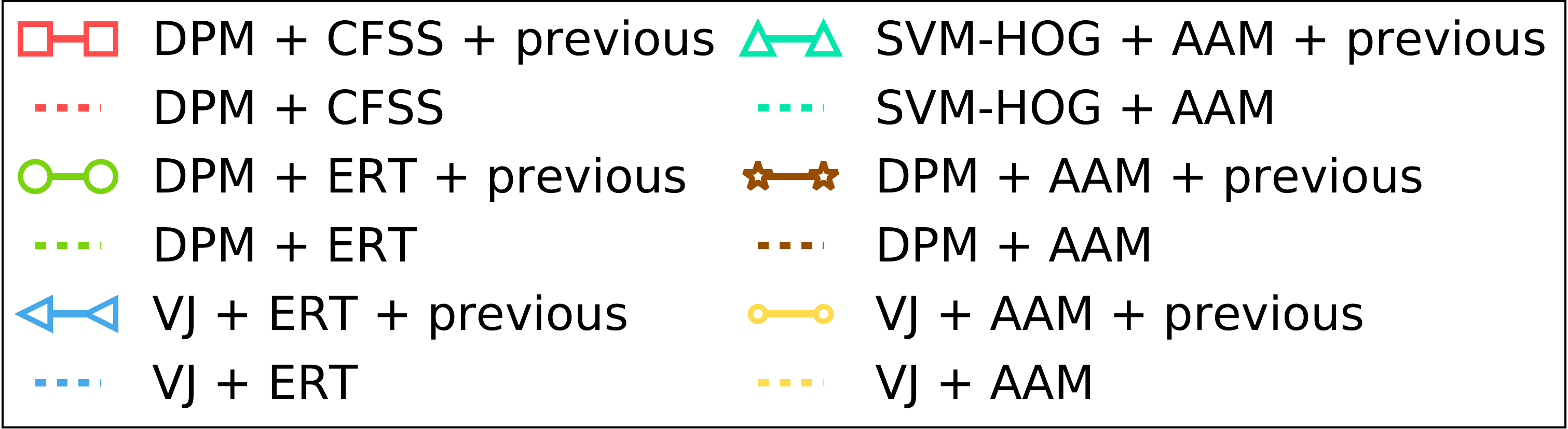}\\
   \includegraphics[width=\linewidth]{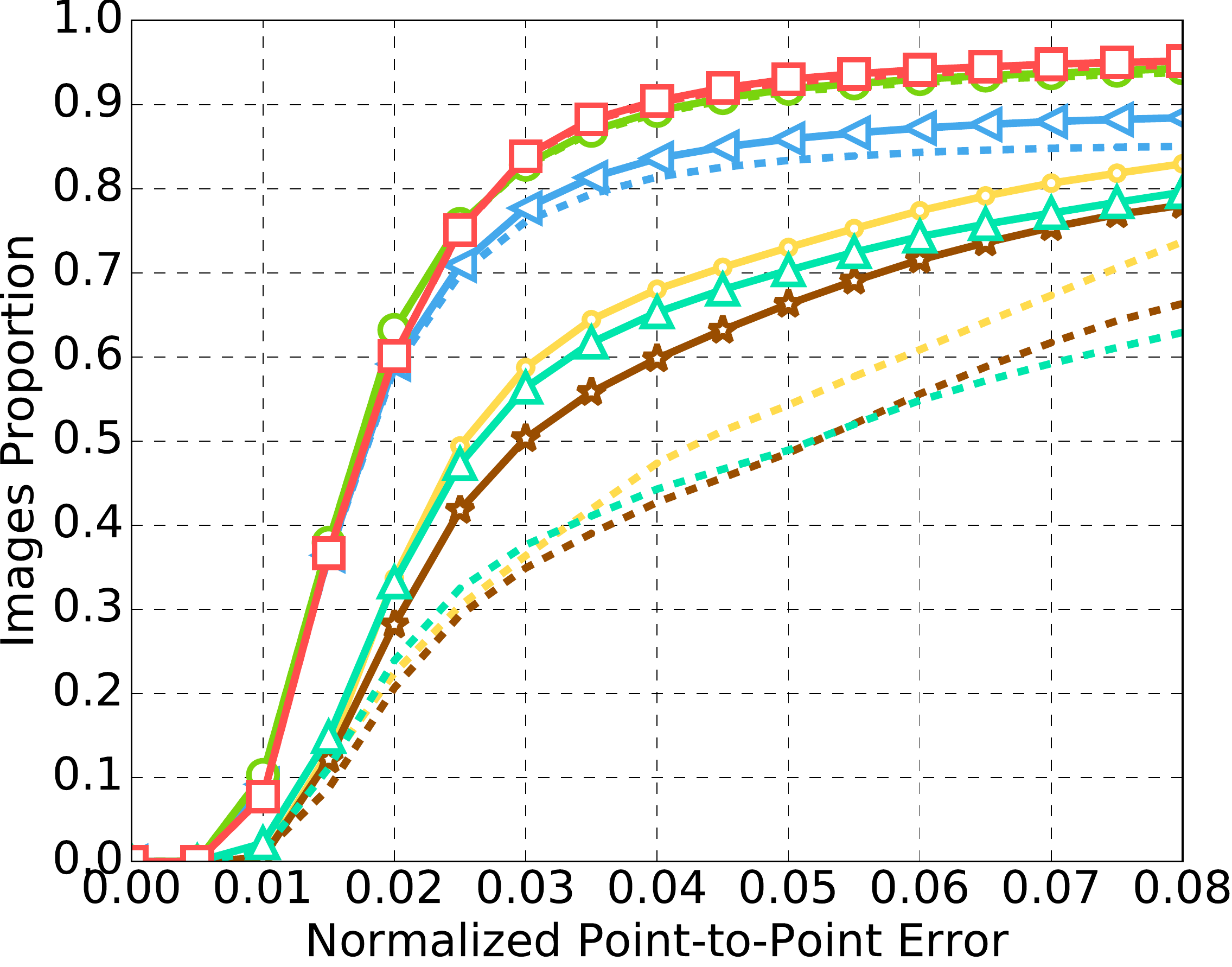}
   \end{minipage}
}
\caption{Results for Experiment 2 of Section~\ref{exp:detection_init_from_previous} (Detection + Landmark Localisation + Initialisation From Previous Frame). These results show the effect of initialisation from the previous frame, in comparison to missing detections. The top 3 performing results are given in red, green and blue, respectively, and the top 3 most improved are given in cyan, yellow and brown, respectively. The dashed lines represent the results before the reinitialisation strategy is applied, solid lines are after.}
\label{fig:exp_detection_init_from_previous_difference}
\end{figure*}
%%%%%%%%%%%%%%%%%%%%%%%%%%%%%%%%%%%%%%%%%%%%%%%%%%%%%%%%%
\subsection{Experiment 2: Detection and Landmark Localisation with Reinitialisation}\label{exp:detection_init_from_previous}
Complementing the experiments of Section~\ref{exp:detection}, the same set-up was utilised to study the effect of missed frames by assuming a first order Markov dependency. If the detector does not return a bounding box in a frame, the bounding box of the previous frame is used as a successful detection for the missing frame. This procedure is depicted in Figure~\ref{fig:detection_failure}. Given that the frame rate of the input videos is adequately high (over 20fps), this assumption is a reasonable one. The results of this experiment are summarised in Table~\ref{tab:exp_detection_init_from_previous} and in Figure~\ref{fig:exp_detection_init_from_previous}. As expected, the ranking of the methods remains the same as the previous experiment of Section~\ref{exp:detection}.

In order to better investigate the effect of this reinitialisation scheme, we also provide Figure~\ref{fig:exp_detection_init_from_previous_difference} that directly shows the improvement. Specifically, we plot the CED curves with and without the reinitialisation strategy for the 3 best performing methods, as well as the 3 techniques for which the highest improvement is achieved. It becomes evident that the top performing methods from Section~\ref{exp:detection} do not benefit from reinitialisation, since the improvement is marginal. This is explained by the fact that these methods already achieve a very high true positive rate. The largest difference is observed for methods that utilise AAM. As shown by \cite{antonakos2015feature}, AAMs are very sensitive to initialisation, due to the nature of Gauss-Newton optimisation. Additionally, note that we have not attempted to apply any kind of greedy approach for improving the detectors' bounding boxes in order to provide a better AAM initialisation. Since the initialisation of a frame with failed detection is achieved by the bounding box of the previous frame's landmarks, it is highly likely that its area will be well constrained to include only the facial parts and not the forehead or background. This kind of initialisation is very beneficial for AAMs, which justifies the large improvements that are shown in Figure~\ref{fig:exp_detection_init_from_previous_difference}. For the graphs that correspond to all 16 methods as well as a video with the results of the top 5 methods\textsuperscript{\ref{foot:detection_init_from_previous}}, please refer to the supplementary material.

\subsection{Experiment 3: Model-free Tracking and Landmark Localisation}\label{exp:tracking}
In this section, we provide, to the best of our knowledge, the first detailed analysis of the performance of model free trackers for tracking ``in-the-wild'' facial sequences. For this reason, we have considered a large number of trackers in order to attempt to give a balanced overview of the performance of modern model trackers for deformable face alignment. The 14 trackers considered in this section are summarised in Table~\ref{tbl:trackers}. To initialise all trackers, the tightest possible bounding box of the ground truth facial landmarks is provided as the initial tracker state. We also include a baseline method, which appears in results Table~\ref{tab:exp_tracking}, referred to as PREV, which is defined as applying the landmark localisation methods initialised from the bounding box of the result in the previous frame. Obviously this scheme is highly sensitive to drifting and therefore we have included it as a basic baseline that does not include any model free tracking. A high level diagram explaining the detection procedure for this experiment is given by Figure~\ref{fig:overview}.

%%%%%%%%%%%%%%%%%%%%%%%%%%%%%%%%%%%%%%%%%%%%%%%%%%%
%%%% [TAB]: TRACKING + LANDMARK_LOCALISATION %%%%%%
%%%%%%%%%%%%%%%%%%%%%%%%%%%%%%%%%%%%%%%%%%%%%%%%%%%
\begin{table*}[htp]
\centering
\begin{tabular}{cc r cc r cc r cc}
\toprule
   \multicolumn{2}{c}{Method} & & \multicolumn{2}{c}{Category 1} & & \multicolumn{2}{c}{Category 2} & & \multicolumn{2}{c}{Category 3} \\
   \cmidrule(lr){1-2}\cmidrule(lr){4-5}\cmidrule(lr){7-8}\cmidrule(lr){10-11}
   \emph{Rigid} & \emph{Landmark} & & \multirow{2}{*}{\emph{AUC}} & \emph{Failure} & & \multirow{2}{*}{\emph{AUC}} & \emph{Failure} & & \multirow{2}{*}{\emph{AUC}} & \emph{Failure} \\
   \emph{Tracking} & \emph{Localisation} & & & \emph{Rate (\%)} & & & \emph{Rate (\%)} & & & \emph{Rate (\%)} \\
   \cmidrule[\heavyrulewidth](){1-2}\cmidrule[\heavyrulewidth](){4-11}
   \multirow{4}{*}{PREV} & AAM & & 0.375 & 50.652 & & 0.465 & 38.273 & & 0.095 & 87.734 \\
                        & CFSS & & 0.545 & 27.358 & & 0.618 & 19.865 & & 0.199 & 72.991 \\
                         & ERT & & 0.340 & 57.266 & & 0.438 & 42.011 & & 0.073 & 89.959 \\
                         & SDM & & 0.497 & 36.606 & & 0.505 & 32.843 & & 0.194 & 74.111 \\
    \cmidrule(lr){1-2}\cmidrule(lr){4-5}\cmidrule(lr){7-8}\cmidrule(lr){10-11}
   \multirow{4}{*}{CMT} & AAM & & 0.574 & 20.323 & & 0.691 & 8.424 & & 0.478 & 26.334 \\
                       & CFSS & & \cellcolor{colour2}\textbf{0.748} & \cellcolor{colour2}\textbf{2.635} & & 0.758 & 1.871 & & \cellcolor{colour4}\textbf{0.595} & \cellcolor{colour4}\textbf{16.506} \\
                        & ERT & & 0.653 & 6.950 & & 0.716 & 2.847 & & 0.498 & 21.136 \\
                        & SDM & & 0.669 & 3.808 & & 0.706 & 2.184 & & 0.529 & 18.427 \\
    %                    
    % \cmidrule(lr){1-2}\cmidrule(lr){4-5}\cmidrule(lr){7-8}\cmidrule(lr){10-11}
    % \multirow{4}{*}{DF} & AAM & &      0.270 &     60.722 & &      0.290 &      57.404 & &     0.224 &      67.165 \\
    %     & CFSS & &     0.467 &     38.756 & &      0.460 &      35.465 & &     0.348 &      51.761 \\
    %     & ERT &  &     0.337 &     48.838 & &      0.344 &      46.094 & &     0.246 &      59.526 \\
    %     & SDM & &      0.358 &     47.286 & &      0.365 &      43.672 & &     0.275 &      57.901 \\
    %
    \cmidrule(lr){1-2}\cmidrule(lr){4-5}\cmidrule(lr){7-8}\cmidrule(lr){10-11}
   \multirow{4}{*}{DSST} & AAM & & 0.510 & 28.620 & & 0.675 & 8.442 & & 0.246 & 59.761 \\
                        & CFSS & & 0.670 & 13.018 & & \cellcolor{colour4}\textbf{0.764} & \cellcolor{colour4}\textbf{0.605} & & 0.380 & 44.205 \\
                         & ERT & & 0.549 & 17.341 & & 0.686 & 2.434 & & 0.286 & 48.893 \\
                         & SDM & & 0.552 & 14.509 & & 0.686 & 1.558 & & 0.304 & 46.433 \\
    \cmidrule(lr){1-2}\cmidrule(lr){4-5}\cmidrule(lr){7-8}\cmidrule(lr){10-11}
   \multirow{4}{*}{FCT} & AAM & & 0.341 & 51.592 & & 0.549 & 20.288 & & 0.148 & 76.888 \\
                       & CFSS & & 0.527 & 29.347 & & 0.706 & 9.409 & & 0.319 & 53.043 \\
                        & ERT & & 0.384 & 40.603 & & 0.619 & 11.989 & & 0.187 & 65.215 \\
                        & SDM & & 0.418 & 38.522 & & 0.627 & 12.524 & & 0.203 & 63.803 \\
    \cmidrule(lr){1-2}\cmidrule(lr){4-5}\cmidrule(lr){7-8}\cmidrule(lr){10-11}
    \multirow{4}{*}{IVT} & AAM & &       0.429 &      40.724 & &     0.424 &      42.699 & &     0.245 &      61.675 \\
       & CFSS & &      0.580 &      28.005 & &     0.533 &      28.225 & &     0.423 &      42.244 \\
       & ERT & &       0.507 &      31.802 & &     0.477 &      32.773 & &     0.329 &      47.033 \\
       & SDM & &       0.517 &      30.971 & &     0.464 &      33.706 & &     0.348 &      45.664 \\
    \cmidrule(lr){1-2}\cmidrule(lr){4-5}\cmidrule(lr){7-8}\cmidrule(lr){10-11}
   \multirow{4}{*}{KCF} & AAM & & 0.550 & 25.025 & & 0.672 & 8.731 & & 0.376 & 39.221 \\
                       & CFSS & & 0.693 & 11.221 & & 0.741 & 2.847 & & 0.554 & 16.889 \\
                        & ERT & & 0.642 & 13.318 & & 0.716 & 3.714 & & 0.438 & 24.838 \\
                        & SDM & & 0.626 & 12.119 & & 0.694 & 3.069 & & 0.444 & 22.686 \\
    \cmidrule(lr){1-2}\cmidrule(lr){4-5}\cmidrule(lr){7-8}\cmidrule(lr){10-11}
   \multirow{4}{*}{LRST} & AAM & & 0.537 & 26.997 & & 0.633 & 13.419 & & 0.426 & 32.878 \\
                        & CFSS & & 0.704 & 10.873 & & \cellcolor{colour5}\textbf{0.759} & \cellcolor{colour5}\textbf{1.600} & & \cellcolor{colour2}\textbf{0.649} & \cellcolor{colour2}\textbf{13.526} \\
                         & ERT & & 0.629 & 13.191 & & 0.698 & 4.429 & & 0.531 & 16.712 \\
                         & SDM & & 0.643 & 12.730 & & 0.696 & 4.040 & & 0.580 & 15.249 \\
    \cmidrule(lr){1-2}\cmidrule(lr){4-5}\cmidrule(lr){7-8}\cmidrule(lr){10-11}
   \multirow{4}{*}{MIL} & AAM & & 0.445 & 32.327 & & 0.544 & 21.654 & & 0.185 & 67.093 \\
                       & CFSS & & 0.683 & 11.420 & & 0.710 & 4.128 & & 0.380 & 45.910 \\
                        & ERT & & 0.536 & 16.881 & & 0.603 & 10.413 & & 0.237 & 57.771 \\
                        & SDM & & 0.589 & 14.693 & & 0.626 & 8.746 & & 0.268 & 56.023 \\
    \cmidrule(lr){1-2}\cmidrule(lr){4-5}\cmidrule(lr){7-8}\cmidrule(lr){10-11}
   \multirow{4}{*}{RPT} & AAM & & 0.477 & 32.206 & & 0.617 & 12.181 & & 0.379 & 39.640 \\
                       & CFSS & & \cellcolor{colour5}\textbf{0.725} & \cellcolor{colour5}\textbf{5.751} & & \cellcolor{colour3}\textbf{0.768} & \cellcolor{colour3}\textbf{0.271} & & \cellcolor{colour3}\textbf{0.627} & \cellcolor{colour3}\textbf{13.324} \\
                        & ERT & & 0.587 & 12.897 & & 0.709 & 2.388 & & 0.506 & 18.698 \\
                        & SDM & & 0.620 & 9.191 & & 0.708 & 0.925 & & 0.538 & 17.539 \\
    \cmidrule(lr){1-2}\cmidrule(lr){4-5}\cmidrule(lr){7-8}\cmidrule(lr){10-11}
   \multirow{4}{*}{SPOT} & AAM & & 0.535 & 25.227 & & 0.680 & 7.058 & & 0.253 & 57.121 \\
                        & CFSS & & \cellcolor{colour1}\textbf{0.769} & \cellcolor{colour1}\textbf{2.330} & & \cellcolor{colour2}\textbf{0.774} & \cellcolor{colour2}\textbf{0.435} & & 0.546 & 27.414 \\
                         & ERT & & 0.638 & 6.809 & & 0.728 & 1.095 & & 0.411 & 30.458 \\
                         & SDM & & 0.679 & 3.244 & & 0.715 & 0.532 & & 0.472 & 28.562 \\
    \cmidrule(lr){1-2}\cmidrule(lr){4-5}\cmidrule(lr){7-8}\cmidrule(lr){10-11}
   \multirow{4}{*}{SRDCF} & AAM & & 0.545 & 26.056 & & 0.675 & 7.824 & & 0.437 & 31.827 \\
                         & CFSS & & \cellcolor{colour3}\textbf{0.731} & \cellcolor{colour3}\textbf{6.810} & & \cellcolor{colour1}\textbf{0.779} & \cellcolor{colour1}\textbf{0.155} & & \cellcolor{colour1}\textbf{0.687} & \cellcolor{colour1}\textbf{8.145} \\
                          & ERT & & 0.636 & 11.251 & & 0.743 & 0.980 & & 0.544 & 11.666 \\
                          & SDM & & 0.650 & 7.929 & & 0.726 & 0.435 & & \cellcolor{colour5}\textbf{0.587} & \cellcolor{colour5}\textbf{10.788} \\
    \cmidrule(lr){1-2}\cmidrule(lr){4-5}\cmidrule(lr){7-8}\cmidrule(lr){10-11}
    \multirow{4}{*}{STRUCK} & AAM &  &    0.543 &      25.041 & &     0.648 &      13.282 & &      0.360 &      42.496 \\
             & CFSS & &    \cellcolor{colour4}\textbf{0.728} &       \cellcolor{colour4}\textbf{7.741} & &     0.741 &       4.411 & &     0.585 &       21.050 \\
             & ERT & &     0.596 &      11.148 & &     0.685 &       5.528 & &     0.430 &      27.139 \\
             & SDM & &     0.643 &       8.866 & &     0.681 &       4.965 & &     0.488 &      25.156 \\
    \cmidrule(lr){1-2}\cmidrule(lr){4-5}\cmidrule(lr){7-8}\cmidrule(lr){10-11}
   \multirow{4}{*}{TLD} & AAM & & 0.373 & 42.618 & & 0.507 & 18.837 & & 0.269 & 55.885 \\
                       & CFSS & & 0.622 & 14.940 & & 0.678 & 7.502 & & 0.469 & 29.592 \\
                        & ERT & & 0.410 & 30.337 & & 0.544 & 14.952 & & 0.302 & 38.877 \\
                        & SDM & & 0.456 & 25.006 & & 0.564 & 11.676 & & 0.333 & 37.440 \\
    %
    %\cmidrule(lr){1-2}\cmidrule(lr){4-5}\cmidrule(lr){7-8}\cmidrule(lr){10-11}
    %\multirow{4}{*}{ORIA} & AAM & &      0.364 &      48.718 & &     0.566 &       21.17 & &     0.128 &      77.014 \\
    %  & CFSS & &     0.501 &      34.015 & &     0.665 &      10.617 & &     0.273 &      60.909 \\
    %  & ERT & &      0.436 &      38.251 & &     0.640 &      12.491 & &     0.227 &      61.343 \\
    %  & SDM & &      0.395 &      43.986 & &     0.634 &      12.144 & &     0.188 &      66.970 \\
    
\midrule[\heavyrulewidth]
   \multicolumn{11}{l}{\scriptsize Colouring denotes the methods' performance ranking per category:\hspace{0.15cm}$\color{colour1}\blacksquare$~first\hspace{0.15cm}$\color{colour2}\blacksquare$~second\hspace{0.15cm}$\color{colour3}\blacksquare$~third\hspace{0.15cm}$\color{colour4}\blacksquare$~fourth\hspace{0.15cm}$\color{colour5}\blacksquare$~fifth}\\
\bottomrule
\end{tabular}
\caption{Results for Experiment 3 of Section~\ref{exp:tracking} (Model Free Tracking + Landmark Localisation).}
\label{tab:exp_tracking}
\end{table*}

%%%%%%%%%%%%%%%%%%%%%%%%%%%%%%%%%%%%%%%%%%%%%%%%%%%
%%%%%%%%%%%%%%%%%%%%%%%%%%%%%%%%%%%%%%%%%%%%%%%%%%%
%%%% [FIG]: TRACKING + LANDMARK_LOCALISATION %%%%%%
%%%%%%%%%%%%%%%%%%%%%%%%%%%%%%%%%%%%%%%%%%%%%%%%%%%
\begin{figure*}[!t]
\subfloat[][Category 1]{
   \begin{minipage}{0.323\linewidth}
   \hspace{0.89cm}\includegraphics[height=0.90cm]{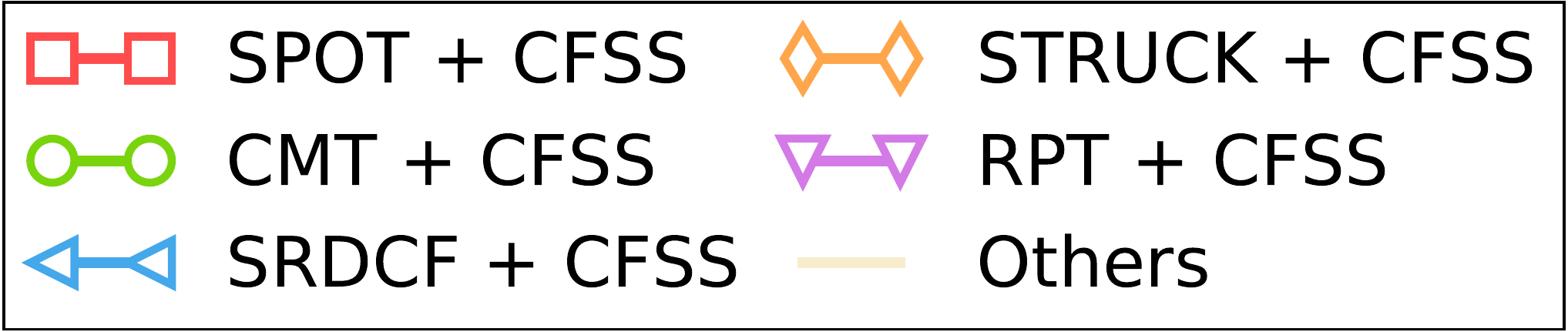}\\
   \includegraphics[width=\linewidth]{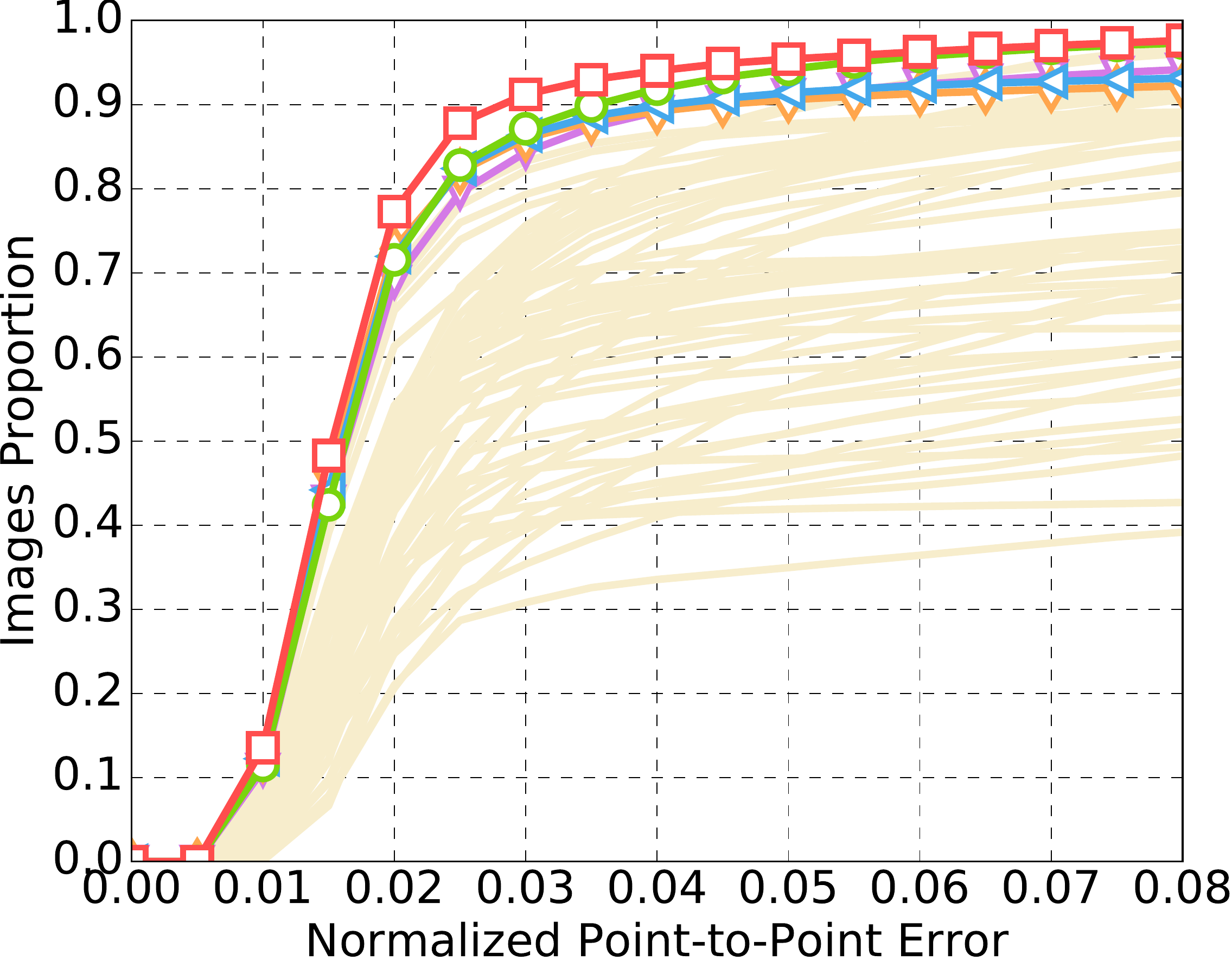}
   \end{minipage}
   }
\subfloat[][Category 2]{
   \begin{minipage}{0.323\linewidth}
   \hspace{1.0cm}\includegraphics[height=0.90cm]{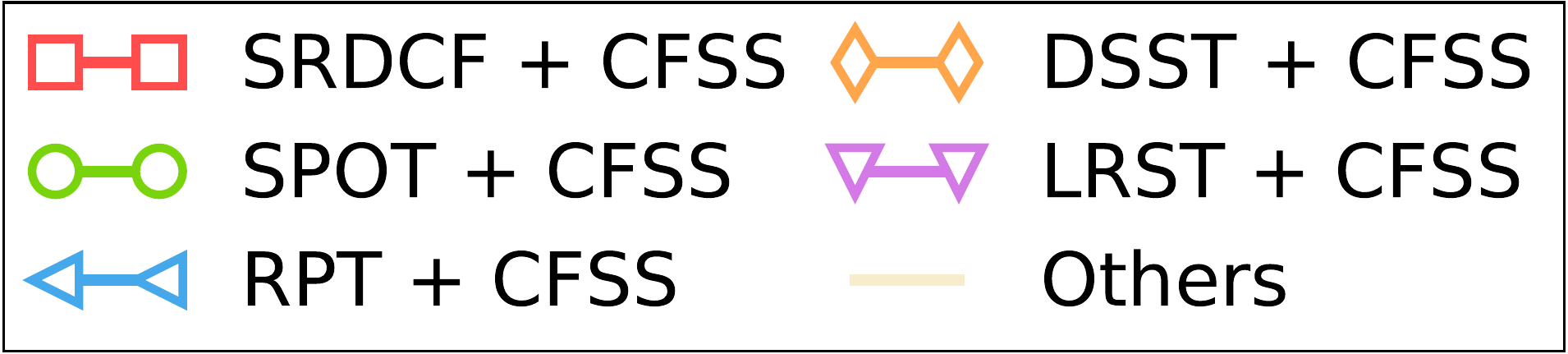}\\
   \includegraphics[width=\linewidth]{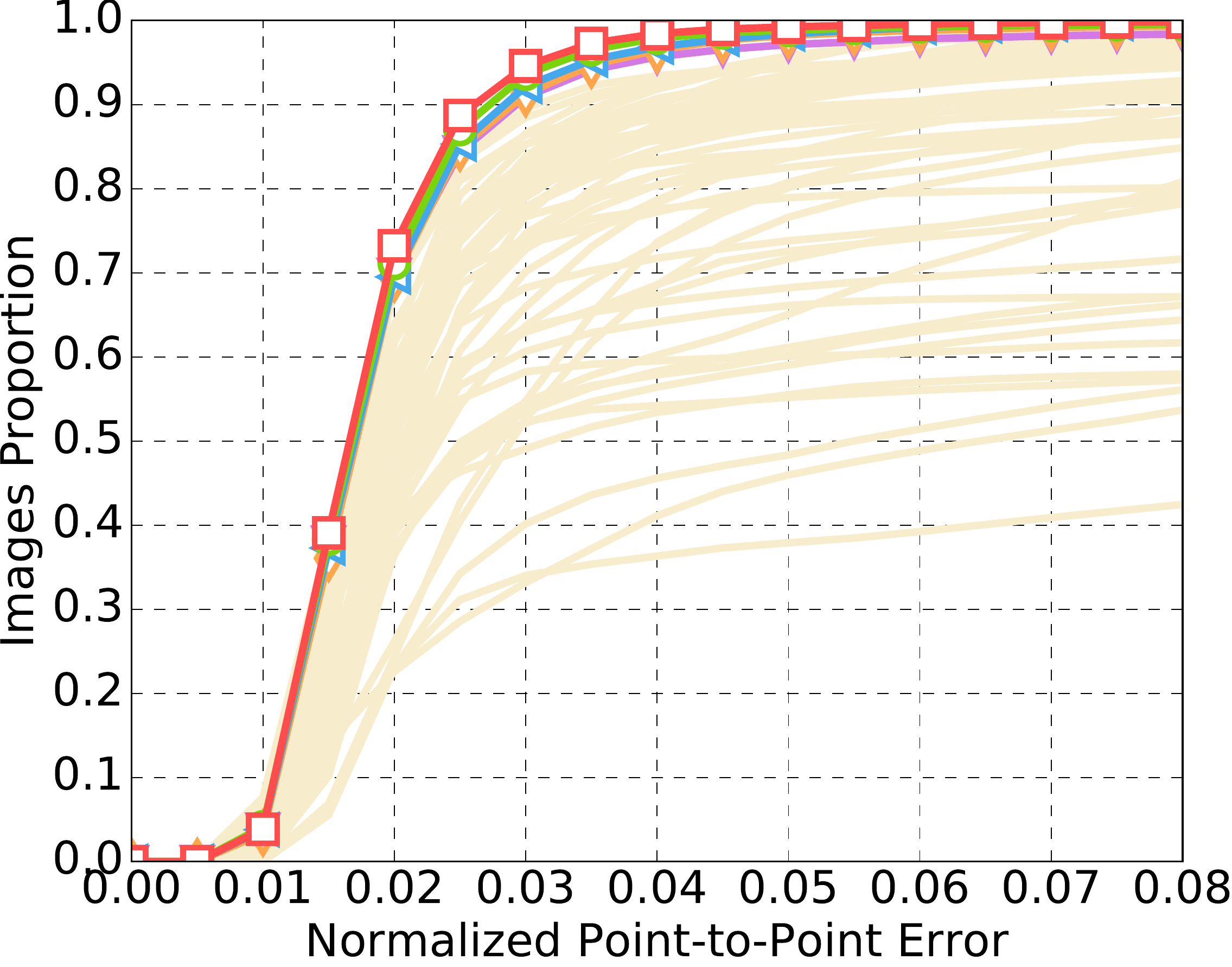}
   \end{minipage}
   }
\subfloat[][Category 3]{
   \begin{minipage}{0.323\linewidth}
   \hspace{0.97cm}\includegraphics[height=0.90cm]{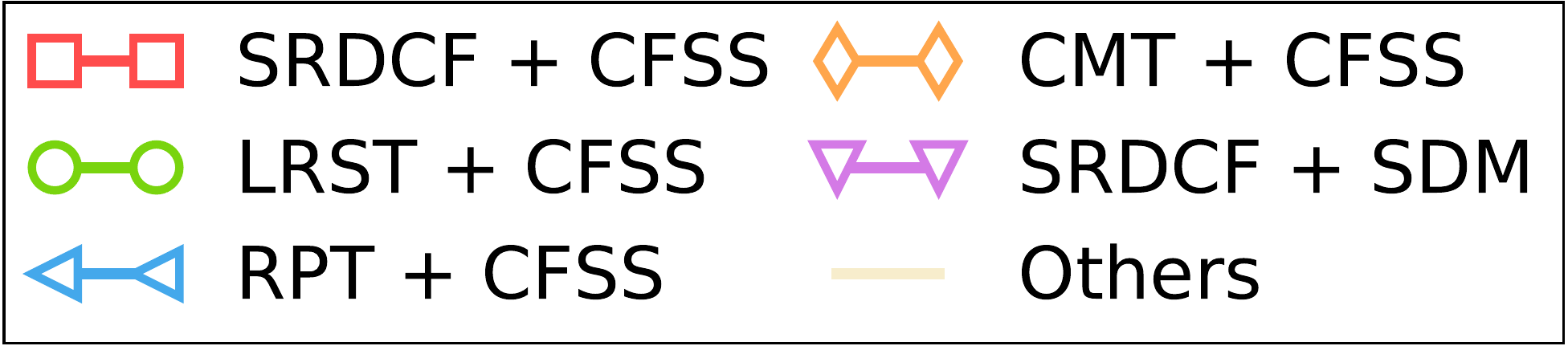}\\
   \includegraphics[width=\linewidth]{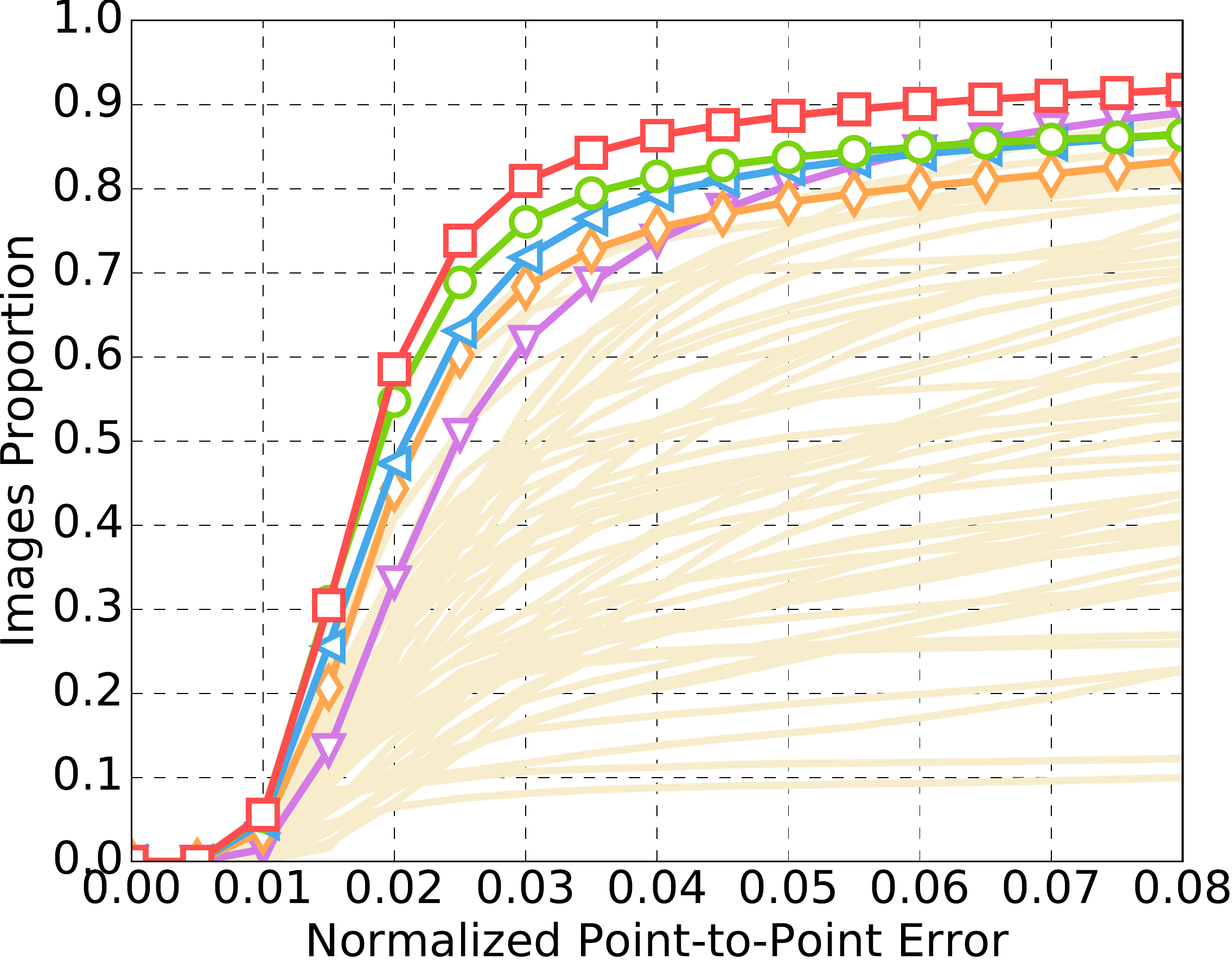}
   \end{minipage}
}
\caption{Results for Experiment 3 of Section~\ref{exp:tracking} (Model Free Tracking + Landmark Localisation). The top 5 performing curves are highlighted in each legend. Please see Table~\ref{tab:exp_tracking} for a full summary.}
\label{fig:exp_tracking}
\end{figure*}
%%%%%%%%%%%%%%%%%%%%%%%%%%%%%%%%%%%%%%%%%%%%%%%%%%%

Specifically, in this experiment we consider the 14 model free trackers of Table~\ref{tbl:trackers}, plus the PREV baseline, with the 4 landmark localisation techniques of Table~\ref{tbl:alignment} (AAM, CFSS, ERT, SDM), for a total of 60 results. The results of the experiment are given in Table~\ref{tab:exp_tracking} and Figure~\ref{fig:exp_tracking}. Note that the results for ORIA (\cite{wu2012online}) and DF (\cite{sevilla2012distribution}) do not appear in Table~\ref{tab:exp_tracking} due to lack of space and the fact that they did not perform well in comparison to PREV. Please see the supplementary material for full statistics.

By inspecting the results, we can firstly notice that most generative trackers perform poorly (i.e. ORIA, DF, FCT, IVT), except LRST which achieves the second best performance for the most challenging video category. On the other hand, the discriminative approaches of SRDCF and SPOT are consistently performing very well. Additionally, similar to the face detection experiments, the combination of all trackers with CFSS returns the best result, whereas AAM constantly demonstrates the poorest performance. Finally, it becomes evident that a straightforward application of the simplistic baseline approach (PREV) is not suitable for deformable tracking, even though it is surprisingly outperforming some model free trackers, such as DF, ORIA and FCT. For the curves that correspond to all 60 methods as well as a video with the tracking result of the top 5 methods\textsuperscript{\ref{foot:tracking}}, please refer to the supplementary material.

%%%%%%%%%%%%%%%%%%%%%%%%%%%%%%%%%%%%%
%%%% FAILURE CHECKING OVERVIEW %%%%%%
%%%%%%%%%%%%%%%%%%%%%%%%%%%%%%%%%%%%%
\begin{figure*}[!t]
    \centering
    \includegraphics[width=0.8\textwidth]{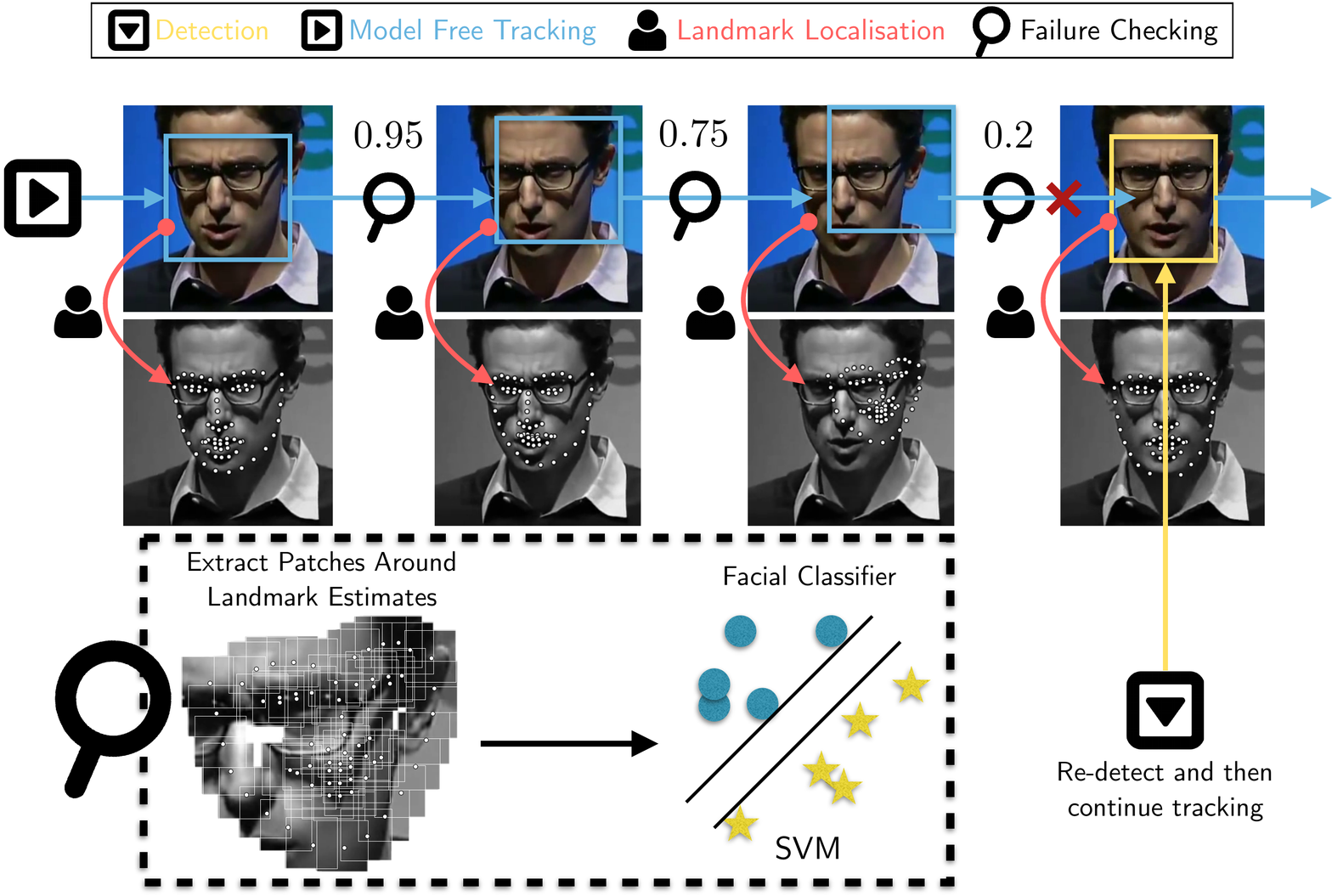}
    \caption{This figure gives a diagram of the reinitialisation scheme proposed in Section~\ref{exp:tracking_restart} for tracking with failure detection. For all frames after the first, the result of the current landmark localisation is used to decide whether or not a face is still being tracked. If the classification fails, a re-detection is performed and the tracker is reinitialised with the bounding box returned by the detector.}
    \label{fig:tracking_failure}
\end{figure*}
%%%%%%%%%%%%%%%%%%%%%%%%%%%%%%%%%%%%%
%%%%%%%%%%%%%%%%%%%%%%%%%%%%%%%%%%%%%%%%%%%%%%%%%%%%%%%%%%%%%%%%%%%%%%
%%%% [TAB]: TRACKING + LANDMARK_LOCALISATION + FAILURE CHECKING %%%%%%
%%%%%%%%%%%%%%%%%%%%%%%%%%%%%%%%%%%%%%%%%%%%%%%%%%%%%%%%%%%%%%%%%%%%%%
\begin{table*}[!b]
\centering
\begin{tabular}{cc r cc r cc r cc}
\toprule
   \multicolumn{2}{c}{Method} & & \multicolumn{2}{c}{Category 1} & & \multicolumn{2}{c}{Category 2} & & \multicolumn{2}{c}{Category 3} \\
   \cmidrule(lr){1-2}\cmidrule(lr){4-5}\cmidrule(lr){7-8}\cmidrule(lr){10-11}
   \emph{Rigid} & \emph{Landmark} & & \multirow{2}{*}{\emph{AUC}} & \emph{Failure} & & \multirow{2}{*}{\emph{AUC}} & \emph{Failure} & & \multirow{2}{*}{\emph{AUC}} & \emph{Failure} \\
   \emph{Tracking} & \emph{Localisation} & & & \emph{Rate (\%)} & & & \emph{Rate (\%)} & & & \emph{Rate (\%)} \\
   \cmidrule[\heavyrulewidth](){1-2}\cmidrule[\heavyrulewidth](){4-11}
   FCT & \multirow{4}{*}{CFSS} & & \cellcolor{colour3}\textbf{0.693} & \cellcolor{colour3}\textbf{13.414} & & \cellcolor{colour3}\textbf{0.763} & \cellcolor{colour3}\textbf{1.661} & & 0.516 & 32.376 \\
   RPT & & & \cellcolor{colour2}\textbf{0.745} & \cellcolor{colour2}\textbf{6.239} & & \cellcolor{colour2}\textbf{0.769} & \cellcolor{colour2}\textbf{0.697} & & \cellcolor{colour1}\textbf{0.704} & \cellcolor{colour1}\textbf{6.108} \\
   SPOT & & & 0.688 & 13.342 & & 0.751 & 2.896 & & \cellcolor{colour3}\textbf{0.570} & \cellcolor{colour3}\textbf{22.913} \\
   SRDCF & & & \cellcolor{colour1}\textbf{0.748} & \cellcolor{colour1}\textbf{5.999} & & \cellcolor{colour1}\textbf{0.772} & \cellcolor{colour1}\textbf{0.505} & & \cellcolor{colour2}\textbf{0.698} & \cellcolor{colour2}\textbf{6.657} \\
\midrule[\heavyrulewidth]
   \multicolumn{11}{l}{\scriptsize Colouring denotes the methods' performance ranking per category:\hspace{0.2cm}$\color{colour1}\blacksquare$~first\hspace{0.2cm}$\color{colour2}\blacksquare$~second\hspace{0.2cm}$\color{colour3}\blacksquare$~third}\\
\bottomrule
\end{tabular}
\caption{Results for Experiment 4 of Section~\ref{exp:tracking_restart} (Model Free Tracking + Landmark Localisation + Failure Checking). The Area Under the Curve (AUC) and Failure Rate are reported. The top 3 performing curves are highlighted for each video category.}
\label{tab:exp_tracking_restart}
\end{table*}
%%%%%%%%%%%%%%%%%%%%%%%%%%%%%%%%%%%%%%%%%%%%%%%%%%%%%%%%%%%%%%%%%%%%%%
%%%%%%%%%%%%%%%%%%%%%%%%%%%%%%%%%%%%%%%%%%%%%%%%%%%%%%%%%%%%%%%%%%%%%%
%%%% [FIG]: TRACKING + LANDMARK_LOCALISATION + FAILURE CHECKING %%%%%%
%%%%%%%%%%%%%%%%%%%%%%%%%%%%%%%%%%%%%%%%%%%%%%%%%%%%%%%%%%%%%%%%%%%%%%
\begin{figure*}[!t]
\subfloat[][Category 1]{
   \begin{minipage}{0.323\linewidth}
   \hspace{0.92cm}\includegraphics[height=0.65cm]{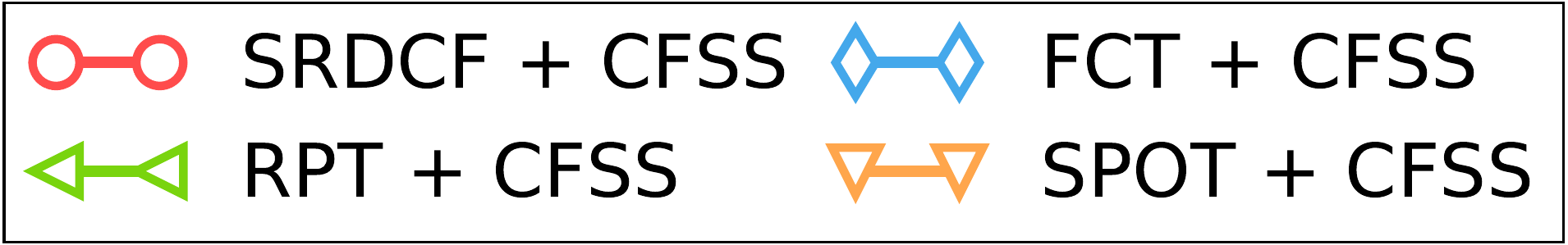}\\
   \includegraphics[width=\linewidth]{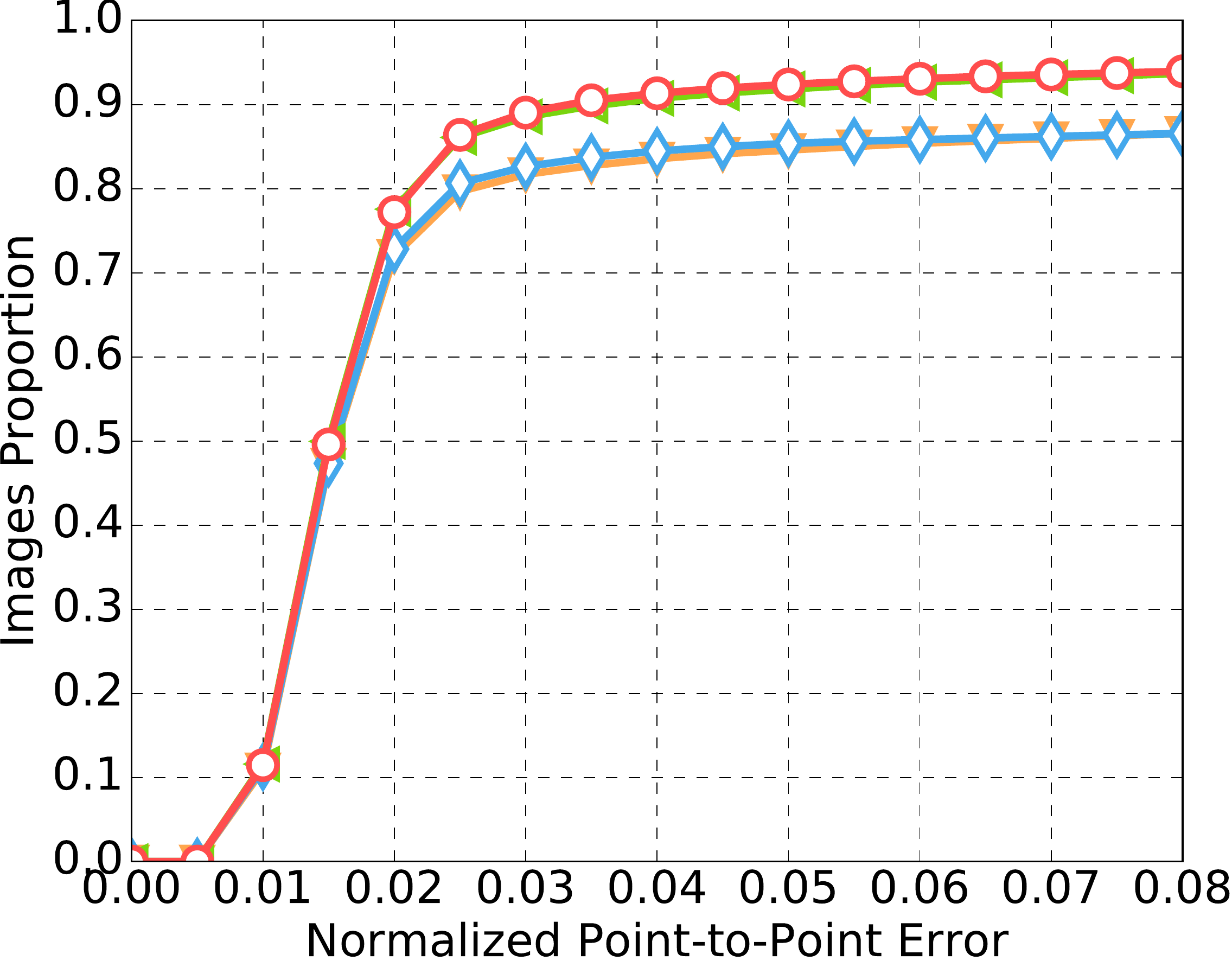}
   \end{minipage}
   }
\subfloat[][Category 2]{
   \begin{minipage}{0.323\linewidth}
   \hspace{0.92cm}\includegraphics[height=0.65cm]{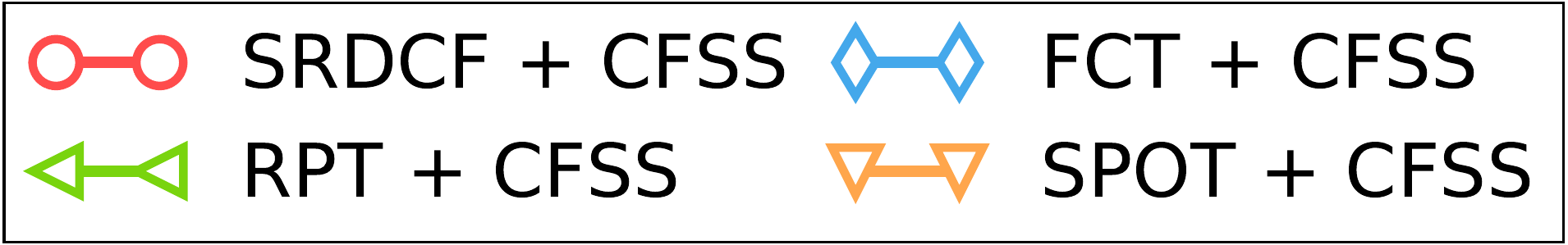}\\
   \includegraphics[width=\linewidth]{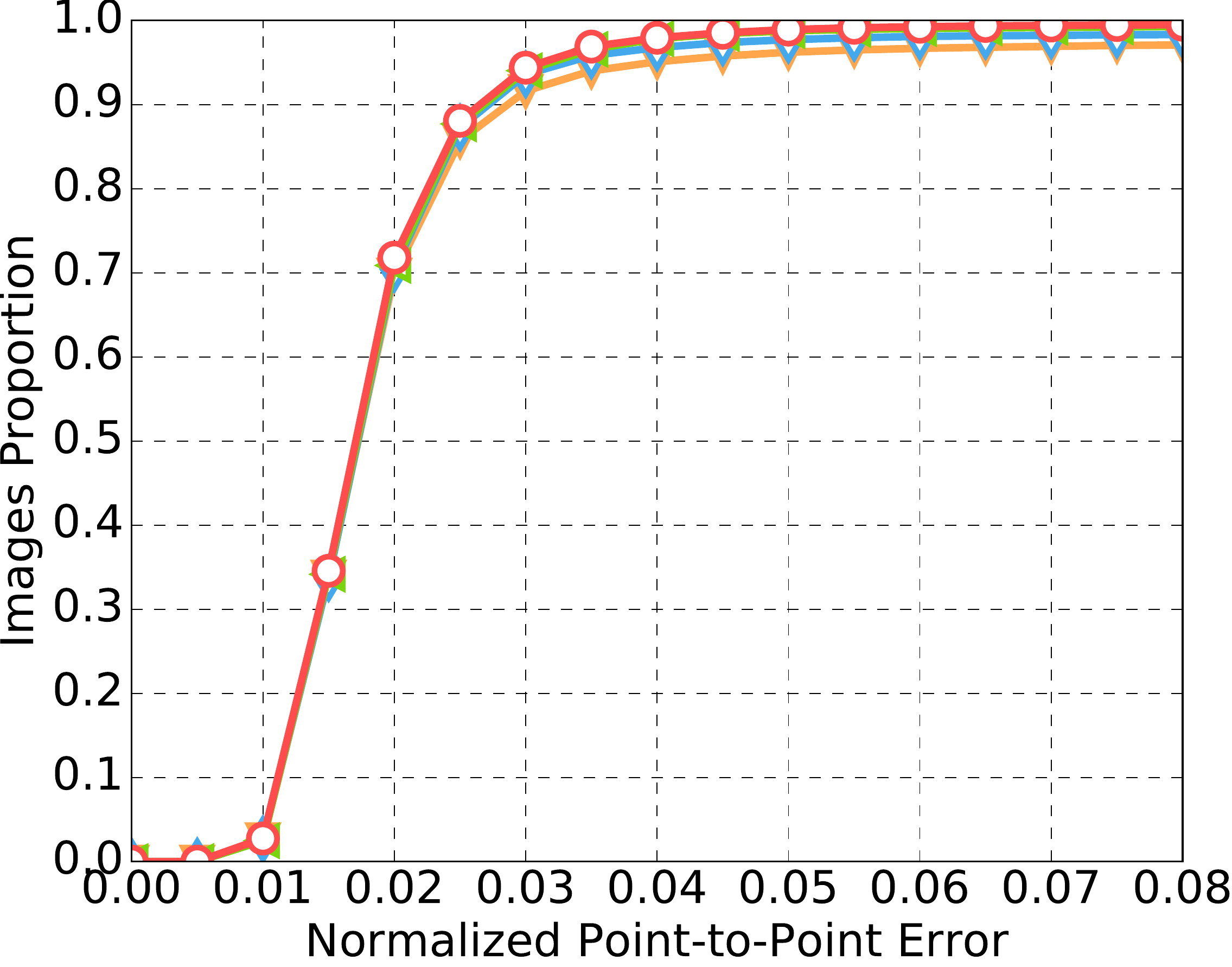}
   \end{minipage}
   }
\subfloat[][Category 3]{
   \begin{minipage}{0.323\linewidth}
   \hspace{0.92cm}\includegraphics[height=0.65cm]{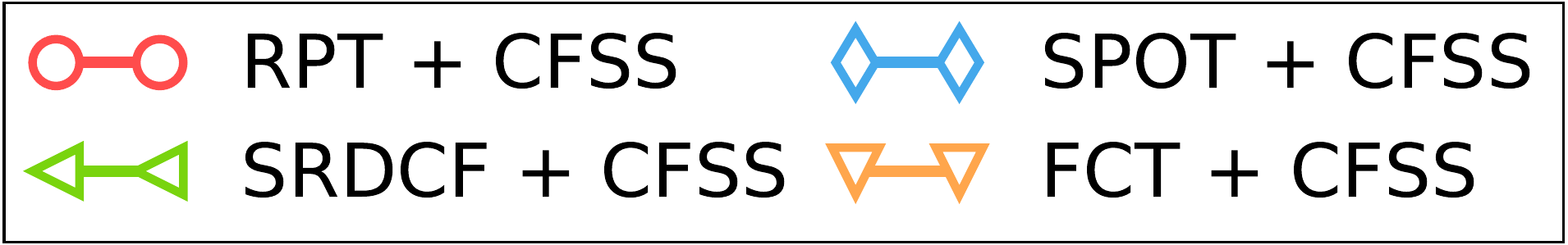}\\
   \includegraphics[width=\linewidth]{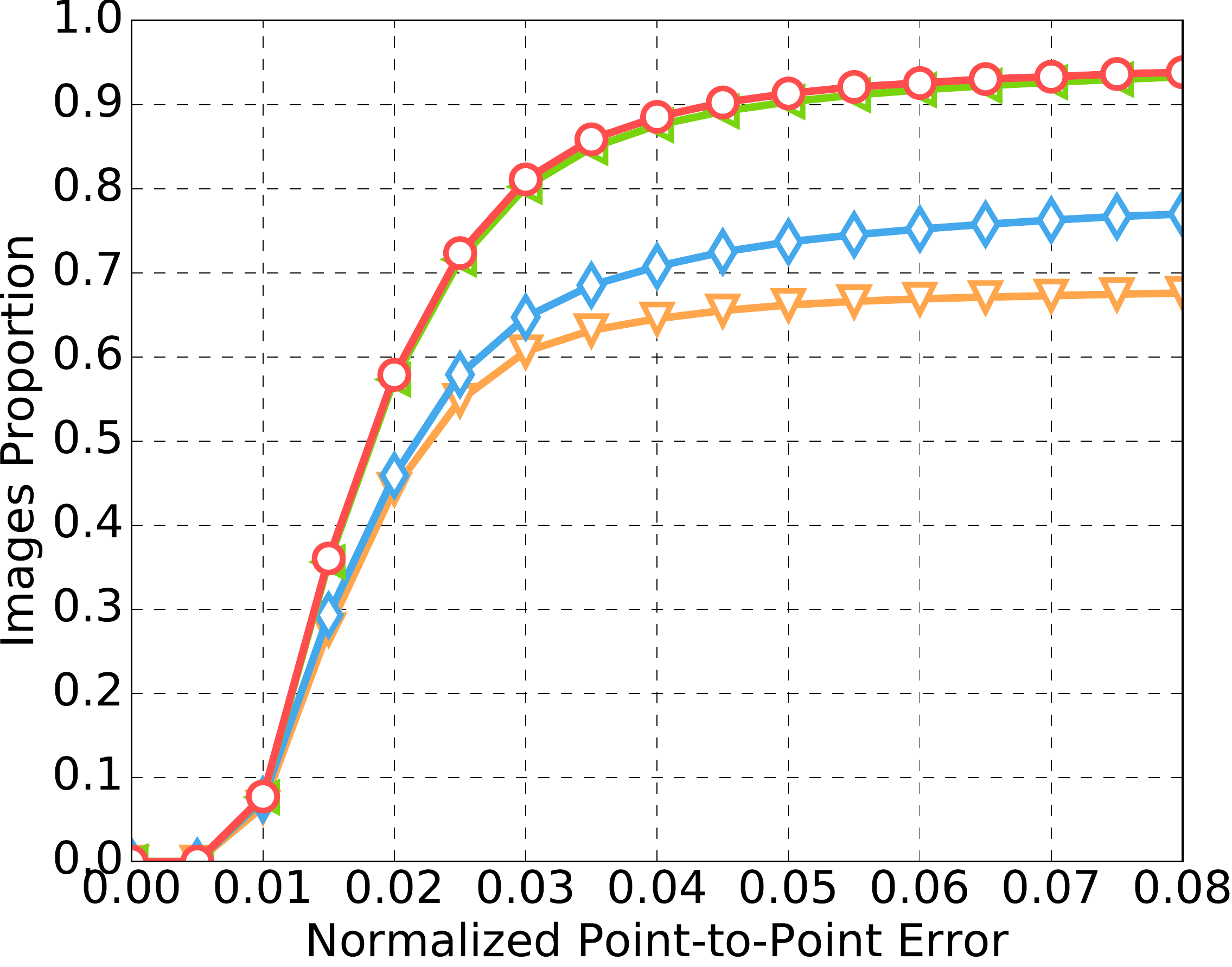}
   \end{minipage}
}
\caption{Results for Experiment 4 of Section~\ref{exp:tracking_restart} (Model Free Tracking + Landmark Localisation + Failure Checking). The top 5 performing curves are highlighted in each legend. Please see Table~\ref{tab:exp_tracking_restart} for a full summary.}
\label{fig:exp_tracking_restart}
\end{figure*}
%%%%%%%%%%%%%%%%%%%%%%%%%%%%%%%%%%%%%%%%%%%%%%%%%%%%%%%%%%%%%%%%%%%%%%
%%%%%%%%%%%%%%%%%%%%%%%%%%%%%%%%%%%%%%%%%%%%%%%%%%%%%%%%%%%%%%%%%%%%%%
%%%% [FIG]: TRACKING + LANDMARK_LOCALISATION + FAILURE CHECKING %%%%%%
%%%%%%%%%%%%%%%%%%%%%%%%%%%%%%%%%%%%%%%%%%%%%%%%%%%%%%%%%%%%%%%%%%%%%%
\begin{figure*}[!t]
\subfloat[][Category 1]{
   \begin{minipage}{0.323\linewidth}
   \hspace{0.43cm}\includegraphics[height=1.2cm]{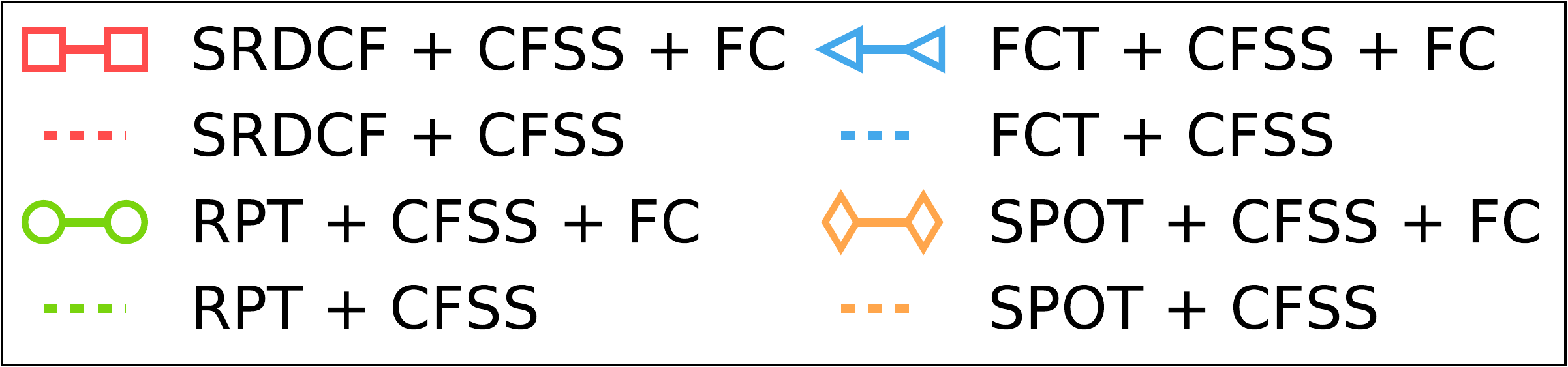}\\
   \includegraphics[width=\linewidth]{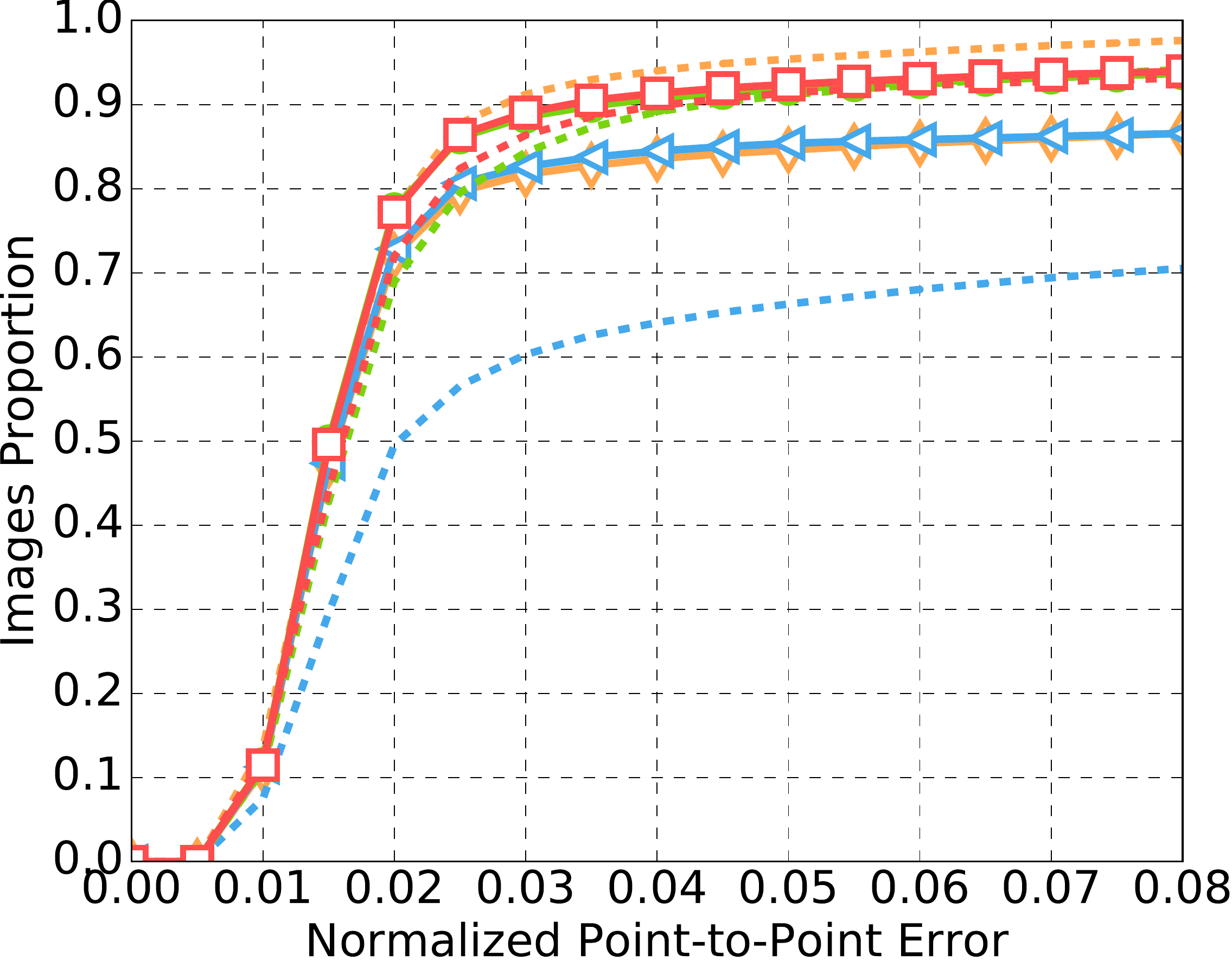}
   \end{minipage}
   }
\subfloat[][Category 2]{
   \begin{minipage}{0.323\linewidth}
   \hspace{0.43cm}\includegraphics[height=1.2cm]{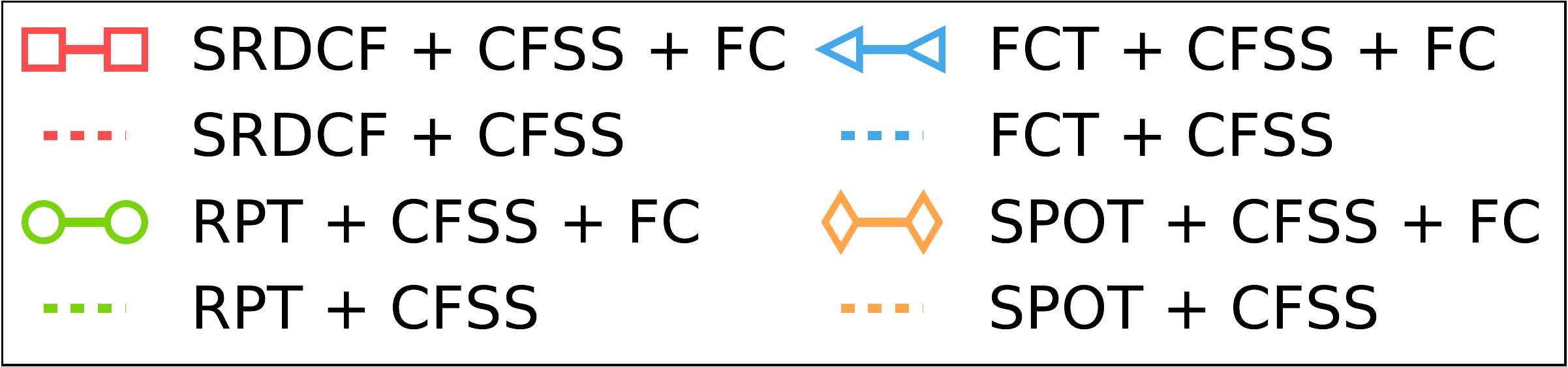}\\
   \includegraphics[width=\linewidth]{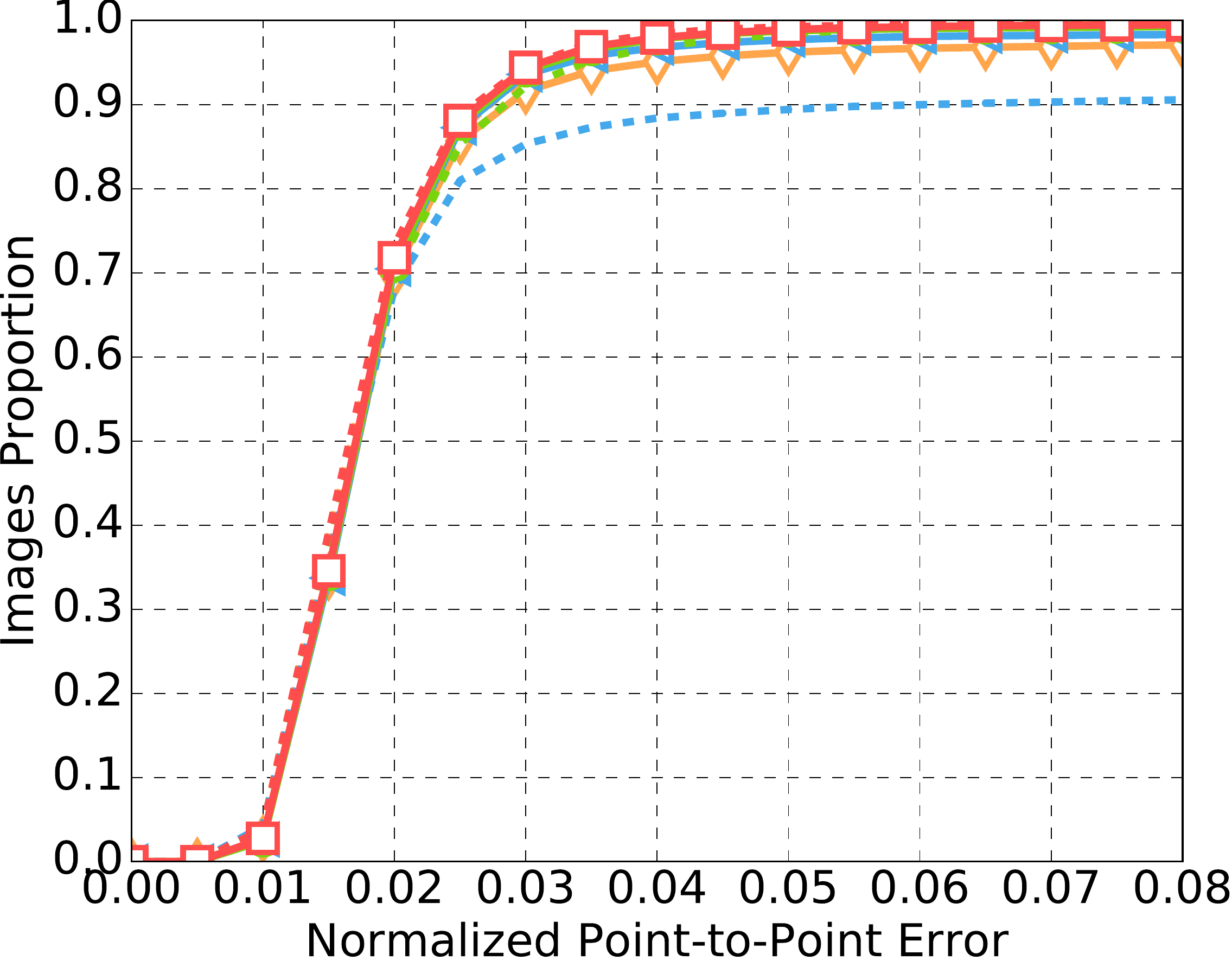}
   \end{minipage}
   }
\subfloat[][Category 3]{
   \begin{minipage}{0.323\linewidth}
   \hspace{0.43cm}\includegraphics[height=1.2cm]{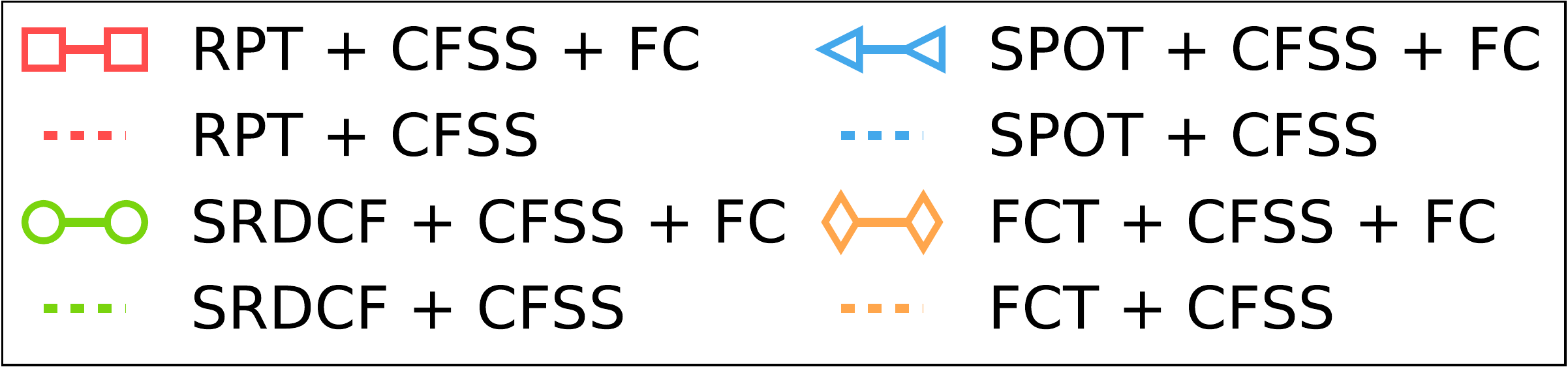}\\
   \includegraphics[width=\linewidth]{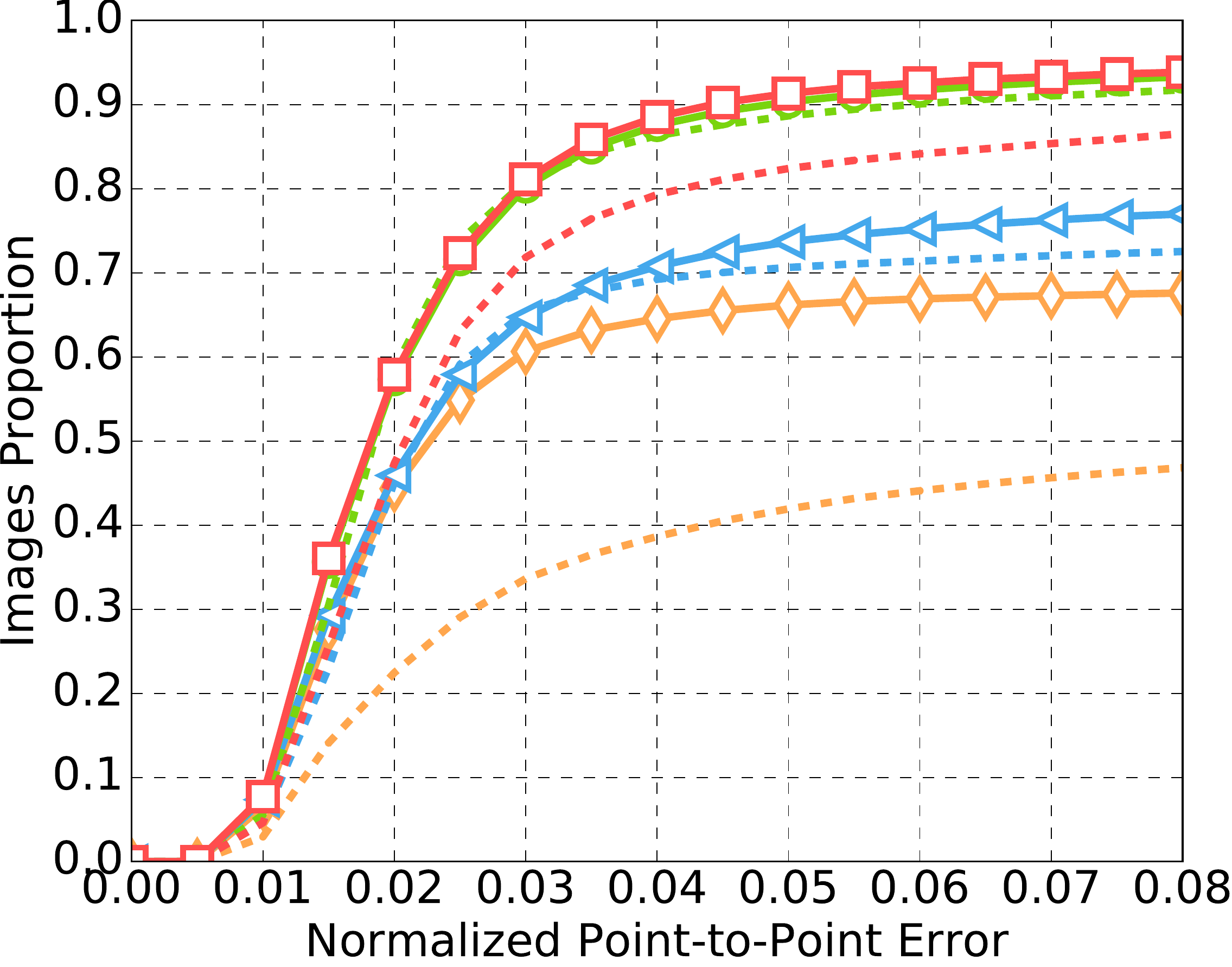}
   \end{minipage}
}
\caption{Results for Experiment 4 of Section~\ref{exp:tracking_restart} (Model Free Tracking + Landmark Localisation + Failure Checking). These results show the effect of the failure checking, in comparison to only tracking. The results are coloured by their performance red, green, blue and orange, respectively. The dashed lines represent the results before the reinitialisation strategy is applied, solid lines are after.}
\label{fig:exp_tracking_restart_difference}
\end{figure*}
%%%%%%%%%%%%%%%%%%%%%%%%%%%%%%%%%%%%%%%%%%%%%%%%%%%%%%%%%%%%%%%%%%%%%%

\subsection{Experiment 4: Failure Checking and Tracking Reinitialisation}\label{exp:tracking_restart}
Complementing the experiments of Section~\ref{exp:tracking}, we investigate the improvement in performance of performing failure checking during tracking. Here we define failure checking as the process of determining whether or not the currently tracked object is a face. Given that we have prior knowledge of the class of object we are tracking, namely faces, this enables us to train an offline classifier that attempts to determine whether a given input is a face or not. Furthermore, since we are also applying landmark localisation, we can perform a strong classification by using the facial landmarks as position priors when extracting features for the failure checking. To train the failure checking classifier, we perform the following methodology:
%%%%%%%%%%%%%%%%%%%%%%%%%
\begin{enumerate}
    \item For all images in the Landmark Localisation training set, extract a fixed sized patch around each of the 68 landmarks and compute HOG (\cite{dalal2005histograms}) features for each patch. These patches are the positive training samples.
    \item Generate negative training samples by perturbing the ground truth bounding box, extracting fixed size patches and computing HOG.
    \item Train an SVM classifier using the positive and negative samples.
\end{enumerate}
%%%%%%%%%%%%%%%%%%%%%%%%%
For the experiments in this section, we use a fixed patch size of $18 \times 18$ pixels, with 100 negative patches sampled for each positive patch. The failure checking classification threshold is chosen via cross-validation on two sequences from the 300VW training videos. Any hyper-parameters of the SVM are also trained using these two validation videos.

Given the failure detector, our restart procedure, is as follows:
%%%%%%%%%%%%%%%%%%%%%%%%%
\begin{itemize}
    \item Classify the current frame to determine if the tracking has failed. If a failure is verified, perform a restart, otherwise continue.
    \item Following the convention of the VOT challenges by \cite{Kristan2013a,Kristan2014a,Kristan2015a}, we attempt to reduce the probability that poor trackers will overly rely on the output of the failure detection system. In the worst case, a very poor tracker would fail on most frames and thus the accuracy of the detector would be validated rather than the tracker itself. Therefore, when a failure is identified, the tracker is allowed to continue for 10 more frames. The results from the drifting tracker are used in these 10 frames in order reduce the affect of the detector. The tracker is then reinitialised at the frame it was first detected as failing at. The next 10 frames, as previously described, already have results computed and therefore no landmark localisation or failure checking is performed in these frames. At the 11th frame, the tracker continues as normal, with landmark localisation and failure checking.
    \item In the unlikely event that the detector fails to detect the face, the previous frame is used as described in Section~\ref{exp:detection_init_from_previous}.
\end{itemize}
%%%%%%%%%%%%%%%%%%%%%%%%%
The diagram given in Figure~\ref{fig:tracking_failure} gives a pictorial representation of this scheme.

The results of this experiment are given in Table~\ref{tab:exp_tracking_restart} and Figure~\ref{fig:exp_tracking_restart}. In contrast to Section~\ref{exp:tracking}, we only perform the experiments on a subset of the total trackers using CFSS. We use the top 3 performing trackers (SRDCF, RPT, SPOT) as well as FCT which had mediocre performance in Section~\ref{exp:tracking}. The results indicate that SRDCF is the best model free tracking methodology for the task.

In order to better investigate the effect of this failure checking scheme, we also provide Figure~\ref{fig:exp_detection_init_from_previous_difference} which shows the differences between the initial tracking results of Section~\ref{exp:tracking} and the results after applying failure detection. The performance of the top trackers (i.e. SRDCF, SPOT, RPT) does not improve much, which is expected since they are already able to return a robust tracking result. However, FCT benefits from the failure checking process, which apparently minimises its drifting issues.

%%%%%%%%%%%%%%%%%%%%%%%%%%%%%%%%%%
%%%% [TAB]: KALMAN SMOOTHING %%%%%
%%%%%%%%%%%%%%%%%%%%%%%%%%%%%%%%%%
\begin{table*}[!t]
\centering
\begin{tabular}{cc r cc r cc r cc}
\toprule
   \multicolumn{2}{c}{Method} & & \multicolumn{2}{c}{Category 1} & & \multicolumn{2}{c}{Category 2} & & \multicolumn{2}{c}{Category 3} \\
   \cmidrule(lr){1-2}\cmidrule(lr){4-5}\cmidrule(lr){7-8}\cmidrule(lr){10-11}
   \emph{Detection or} & \emph{Landmark} & & \multirow{2}{*}{\emph{AUC}} & \emph{Failure} & & \multirow{2}{*}{\emph{AUC}} & \emph{Failure} & & \multirow{2}{*}{\emph{AUC}} & \emph{Failure} \\
   \emph{Tracking} & \emph{Localisation} & & & \emph{Rate (\%)} & & & \emph{Rate (\%)} & & & \emph{Rate (\%)} \\
   \cmidrule[\heavyrulewidth](){1-2}\cmidrule[\heavyrulewidth](){4-11}
   \multirow{3}{*}{DPM} & CFSS & & \cellcolor{colour3}\textbf{0.766} & \cellcolor{colour3}\textbf{3.741} & & 0.770 & 1.317 & & \cellcolor{colour1}\textbf{0.724} & \cellcolor{colour1}\textbf{5.234} \\
                         & ERT & & \cellcolor{colour1}\textbf{0.777} & \cellcolor{colour1}\textbf{3.442} & & \cellcolor{colour3}\textbf{0.772} & \cellcolor{colour3}\textbf{1.509} & & \cellcolor{colour2}\textbf{0.721} & \cellcolor{colour2}\textbf{6.082} \\
                         & SDM & & 0.678 & 3.728 & & 0.652 & 1.354 & & 0.592 & 5.786 \\
   \cmidrule(lr){1-2}\cmidrule(lr){4-5}\cmidrule(lr){7-8}\cmidrule(lr){10-11}
   \multirow{4}{*}{FCT} & AAM & & 0.342 & 51.503 & & 0.552 & 20.172 & & 0.149 & 76.765 \\
                       & CFSS & & 0.529 & 29.283 & & 0.709 & 9.358 & & 0.320 & 53.061 \\
                        & ERT & & 0.386 & 40.506 & & 0.623 & 11.937 & & 0.188 & 65.121 \\
                        & SDM & & 0.419 & 38.506 & & 0.629 & 12.515 & & 0.204 & 63.730 \\
   \cmidrule(lr){1-2}\cmidrule(lr){4-5}\cmidrule(lr){7-8}\cmidrule(lr){10-11}
   \multirow{3}{*}{RPT} & CFSS & & 0.727 & 5.722 & & \cellcolor{colour4}\textbf{0.772} & \cellcolor{colour4}\textbf{0.252} & & \cellcolor{colour4}\textbf{0.632} & \cellcolor{colour4}\textbf{13.331} \\
                         & ERT & & 0.589 & 12.765 & & 0.713 & 2.303 & & 0.507 & 18.687 \\
                         & SDM & & 0.622 & 9.169 & & 0.710 & 0.888 & & 0.539 & 17.535 \\
   \cmidrule(lr){1-2}\cmidrule(lr){4-5}\cmidrule(lr){7-8}\cmidrule(lr){10-11}
   \multirow{4}{*}{SPOT} & AAM & & 0.536 & 24.998 & & 0.682 & 6.957 & & 0.254 & 56.803 \\
                        & CFSS & & \cellcolor{colour2}\textbf{0.773} & \cellcolor{colour2}\textbf{2.237} & & \cellcolor{colour2}\textbf{0.777} & \cellcolor{colour2}\textbf{0.417} & & 0.551 & 27.323 \\
                         & ERT & & 0.640 & 6.745 & & 0.731 & 1.074 & & 0.412 & 30.296 \\
                         & SDM & & 0.681 & 3.194 & & 0.717 & 0.508 & & 0.474 & 28.548 \\
   \cmidrule(lr){1-2}\cmidrule(lr){4-5}\cmidrule(lr){7-8}\cmidrule(lr){10-11}
   \multirow{4}{*}{SRDCF} & AAM & & 0.546 & 25.988 & & 0.676 & 7.697 & & 0.440 & 31.499 \\
                         & CFSS & & \cellcolor{colour4}\textbf{0.734} & \cellcolor{colour4}\textbf{6.815} & & \cellcolor{colour1}\textbf{0.783} & \cellcolor{colour1}\textbf{0.131} & & \cellcolor{colour3}\textbf{0.693} & \cellcolor{colour3}\textbf{8.134} \\
                          & ERT & & 0.637 & 11.145 & & 0.746 & 0.922 & & 0.544 & 11.572 \\
                          & SDM & & 0.652 & 7.905 & & 0.729 & 0.414 & & 0.588 & 10.774 \\
   \cmidrule(lr){1-2}\cmidrule(lr){4-5}\cmidrule(lr){7-8}\cmidrule(lr){10-11}
   \multirow{2}{*}{TLD} & CFSS & & 0.624 & 14.827 & & 0.681 & 7.477 & & 0.473 & 29.548 \\
                         & SDM & & 0.457 & 24.965 & & 0.566 & 11.645 & & 0.335 & 37.389 \\
\midrule[\heavyrulewidth]
   \multicolumn{11}{l}{\scriptsize Colouring denotes the methods' performance ranking per category:\hspace{0.2cm}$\color{colour1}\blacksquare$~first\hspace{0.2cm}$\color{colour2}\blacksquare$~second\hspace{0.2cm}$\color{colour3}\blacksquare$~third\hspace{0.2cm}$\color{colour4}\blacksquare$~fourth}\\
\bottomrule
\end{tabular}
\caption{Results for Experiment 5 of Section~\ref{exp:kalman} (Kalman Smoothing). The Area Under the Curve (AUC) and Failure Rate are reported. The top 4 performing curves are highlighted for each video category.}
\label{tab:exp_kalman}
\end{table*}
%%%%%%%%%%%%%%%%%%%%%%%%%%%%%%%%%%
%%%%%%%%%%%%%%%%%%%%%%%%%%%%%%%%%%
%%%% [FIG]: KALMAN SMOOTHING %%%%%
%%%%%%%%%%%%%%%%%%%%%%%%%%%%%%%%%%
\begin{figure*}[!b]
\subfloat[][Category 1]{
   \begin{minipage}{0.323\linewidth}
   \hspace{1.0cm}\includegraphics[height=0.90cm]{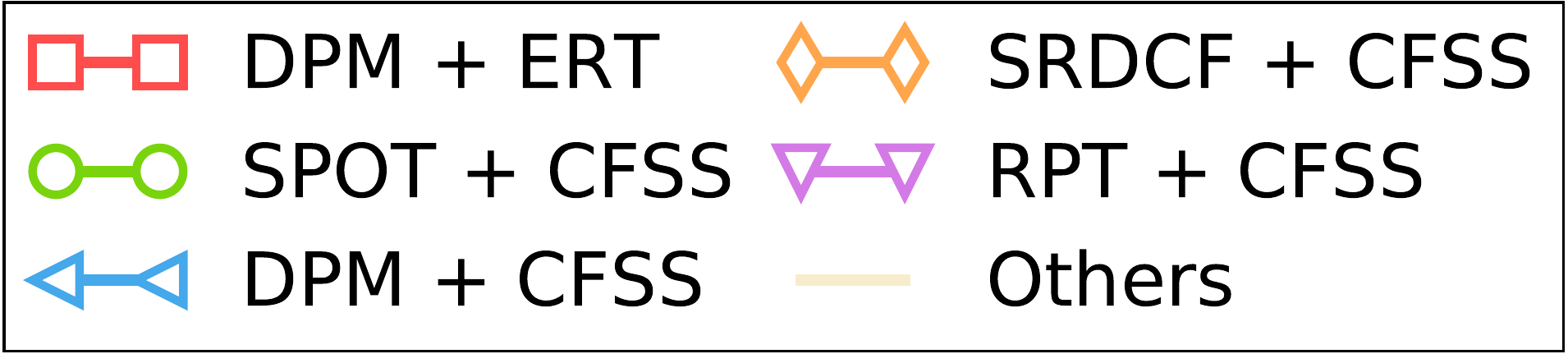}\\
   \includegraphics[width=\linewidth]{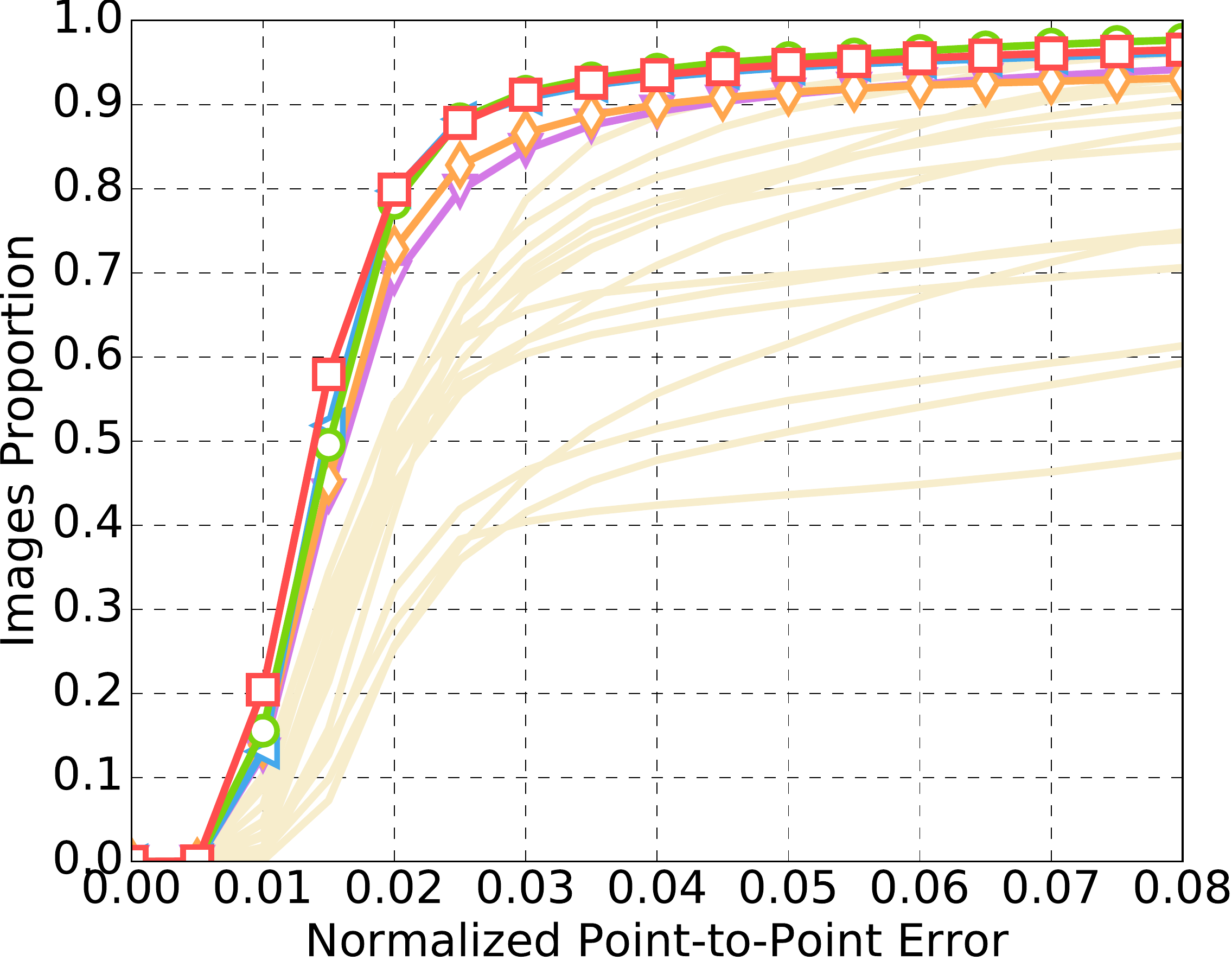}
   \end{minipage}
   }
\subfloat[][Category 2]{
   \begin{minipage}{0.323\linewidth}
   \hspace{1.04cm}\includegraphics[height=0.90cm]{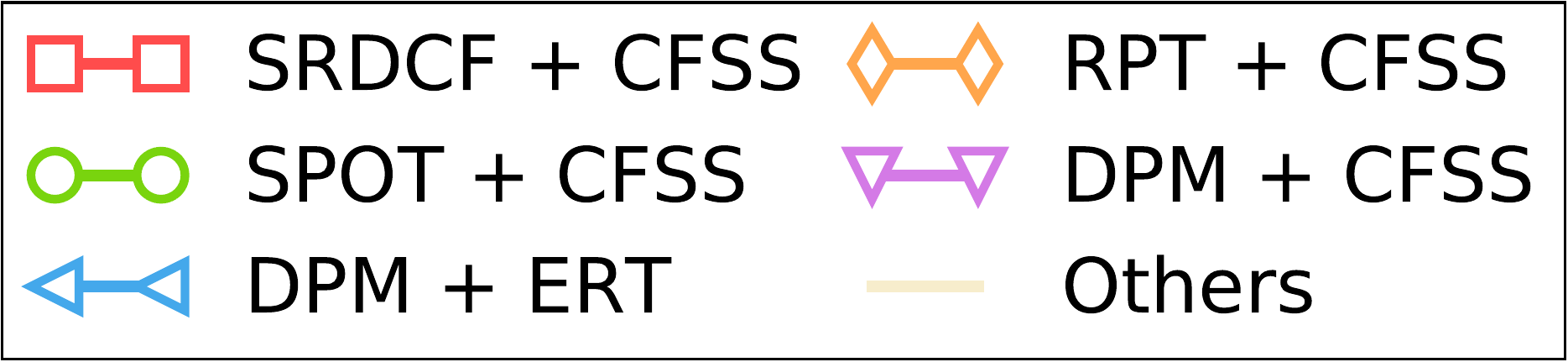}\\
   \includegraphics[width=\linewidth]{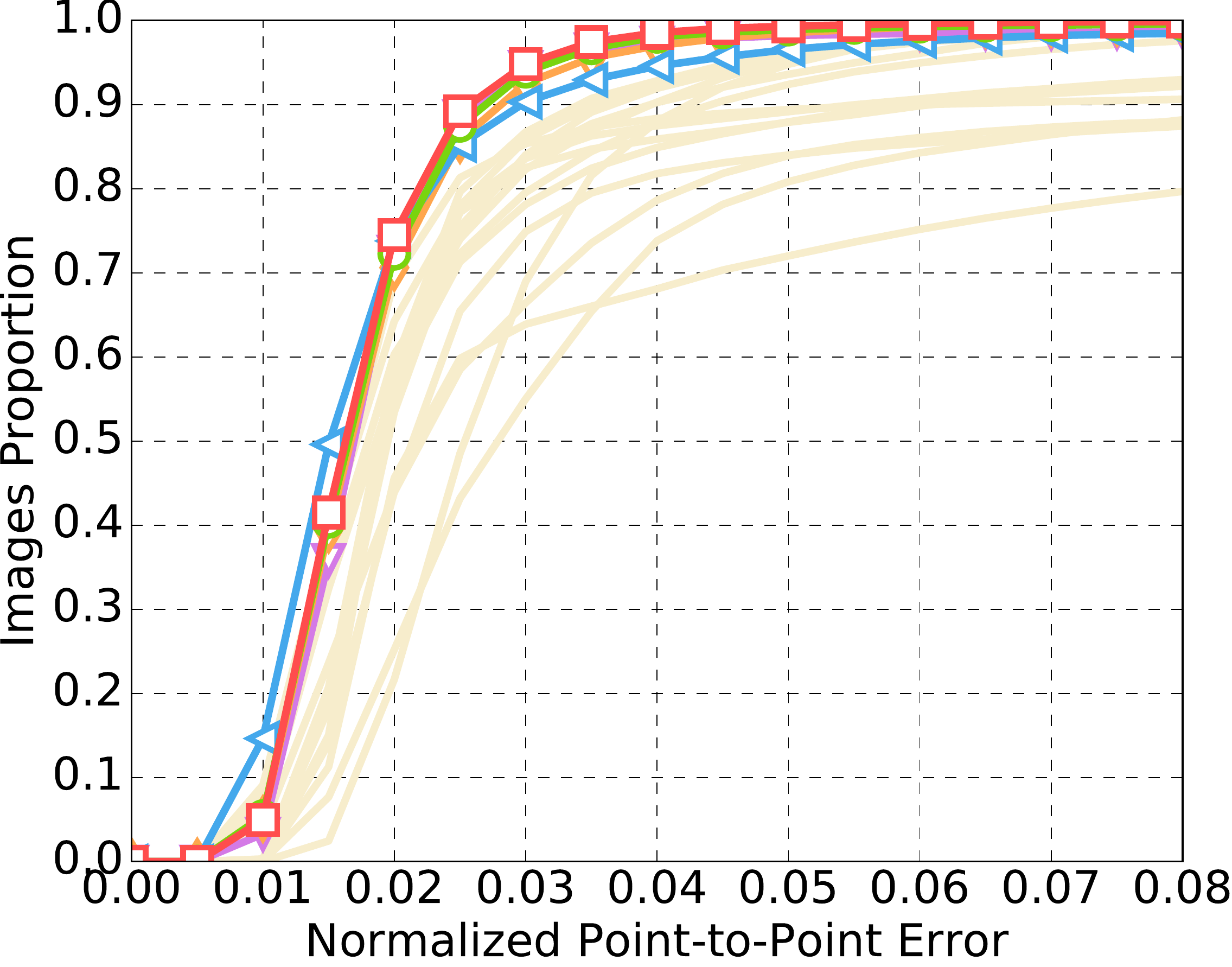}
   \end{minipage}
   }
\subfloat[][Category 3]{
   \begin{minipage}{0.323\linewidth}
   \hspace{1.06cm}\includegraphics[height=0.90cm]{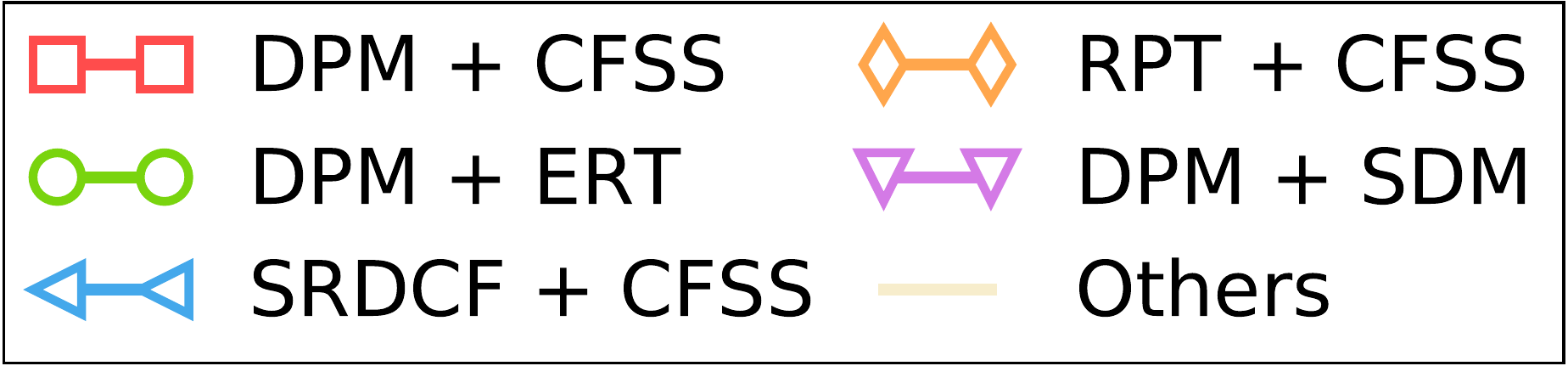}\\
   \includegraphics[width=\linewidth]{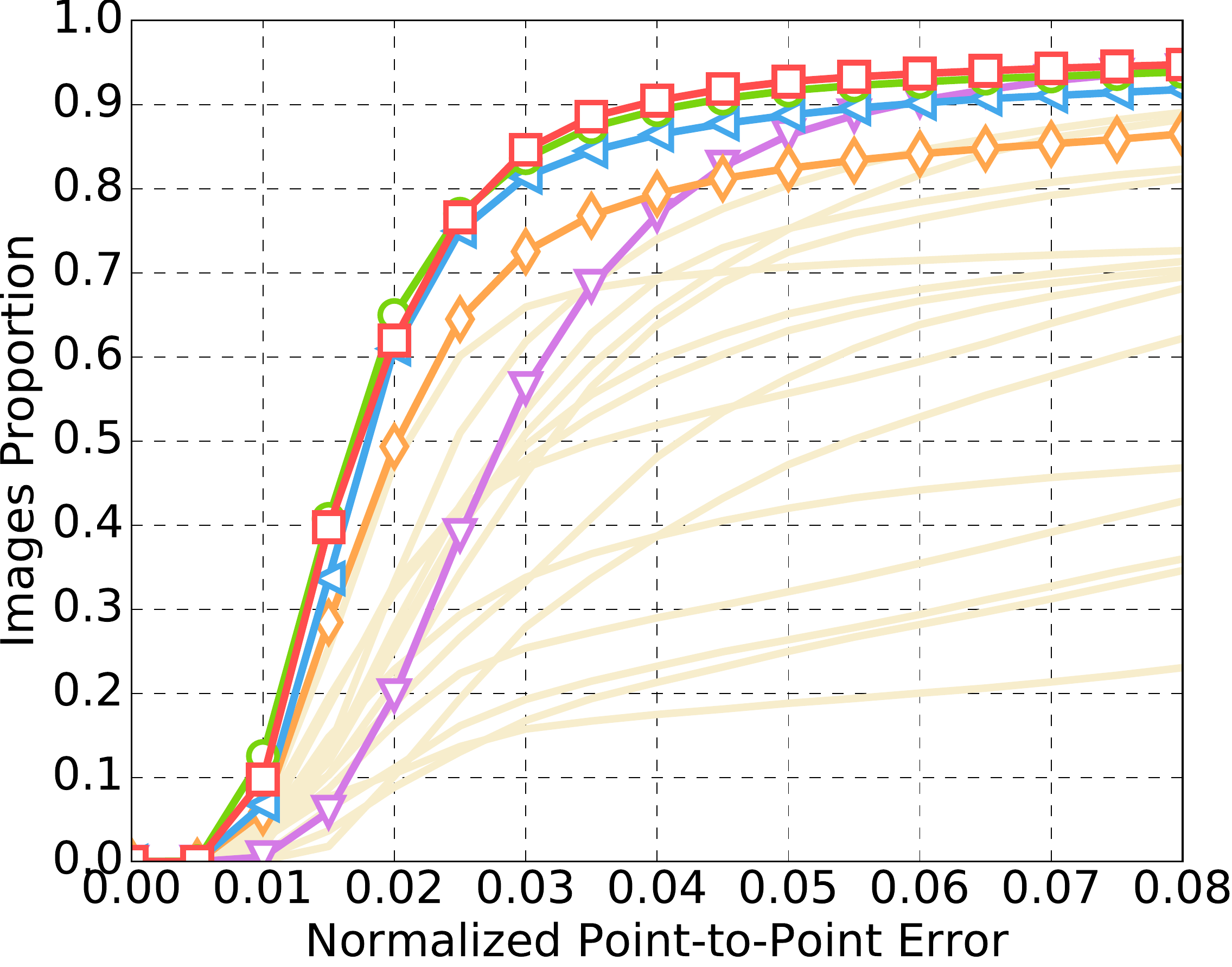}
   \end{minipage}
}
\caption{Results for Experiment 5 of Section~\ref{exp:kalman} (Kalman Smoothing). The top 5 performing curves are highlighted in each legend. Please see Table~\ref{tab:exp_kalman} for a full summary.}
\label{fig:exp_kalman}
\end{figure*}
%%%%%%%%%%%%%%%%%%%%%%%%%%%%%%%%%%
%%%%%%%%%%%%%%%%%%%%%%%%%%%%%%%%%%
%%%% [FIG]: KALMAN SMOOTHING %%%%%
%%%%%%%%%%%%%%%%%%%%%%%%%%%%%%%%%%
\begin{figure*}[!t]
\subfloat[][Category 1]{
   \begin{minipage}{0.323\linewidth}
   \centering
   \includegraphics[width=\linewidth]{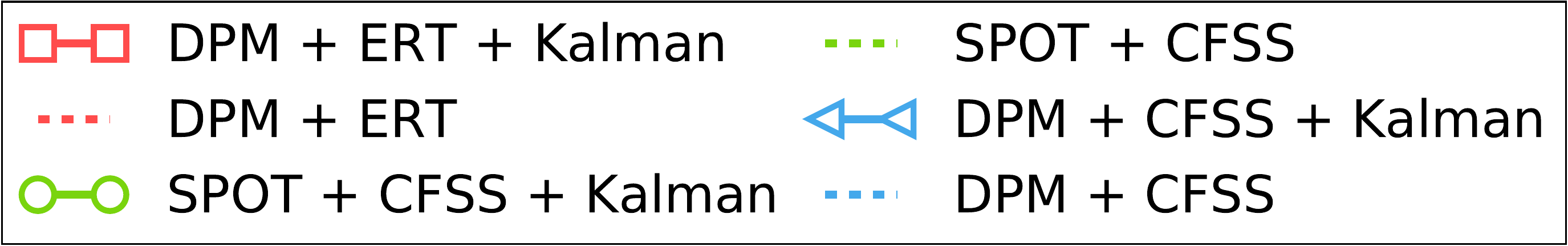}\\
   \includegraphics[width=\linewidth]{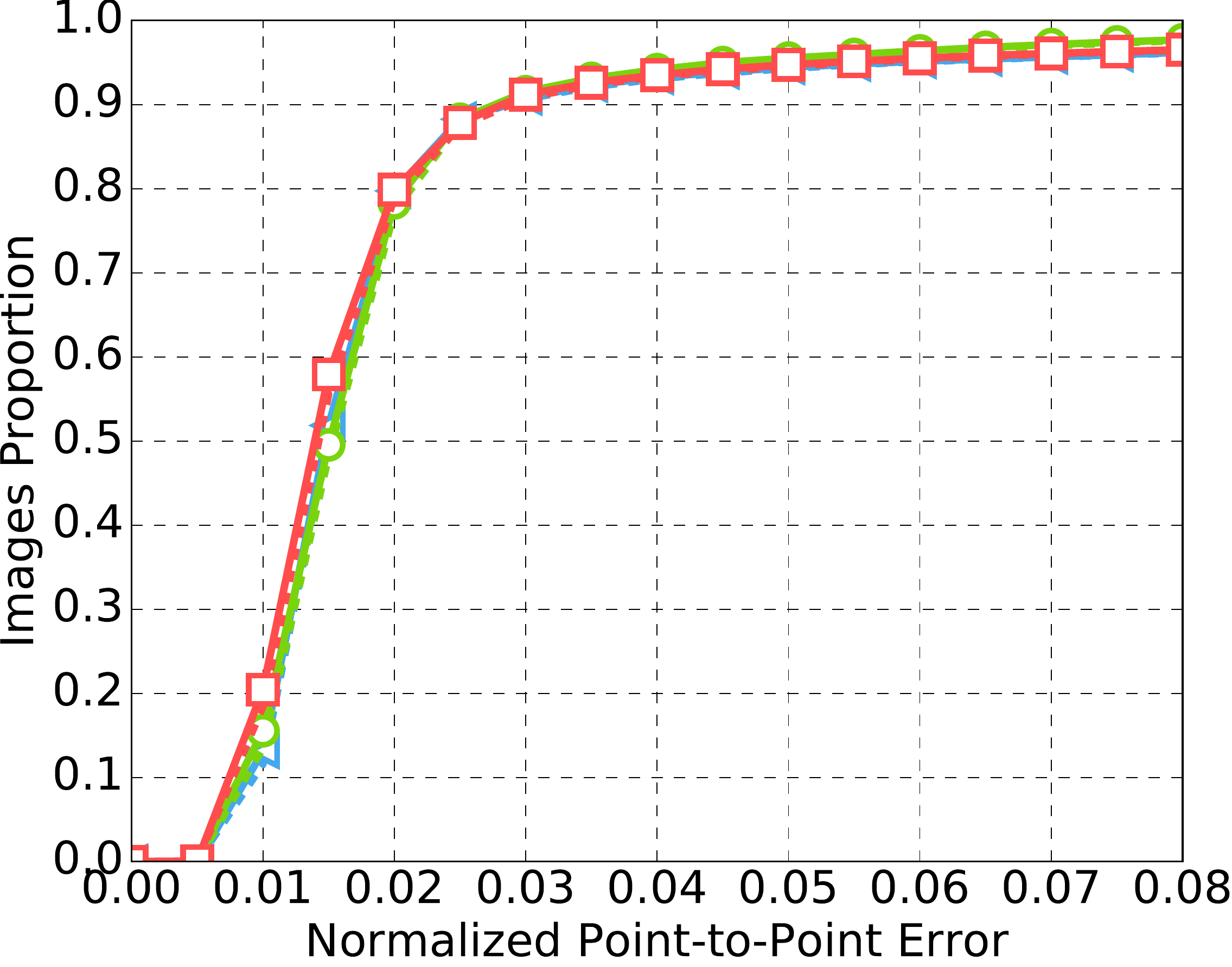}
   \end{minipage}
   }
\subfloat[][Category 2]{
   \begin{minipage}{0.323\linewidth}
   \centering
   \includegraphics[width=\linewidth]{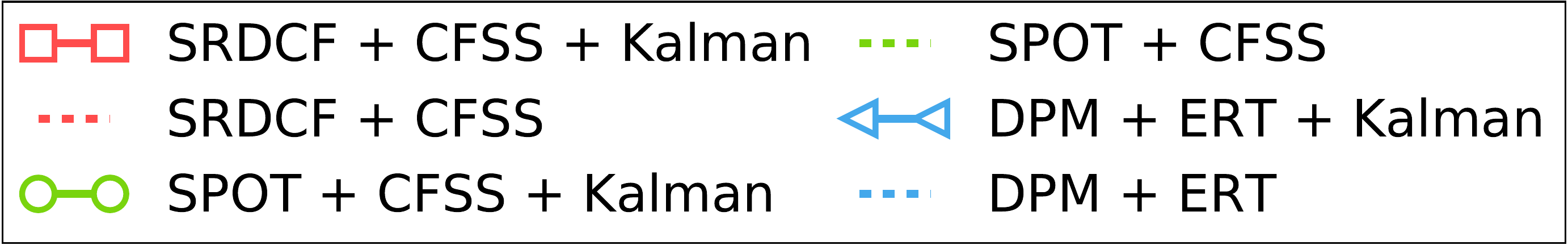}\\
   \includegraphics[width=\linewidth]{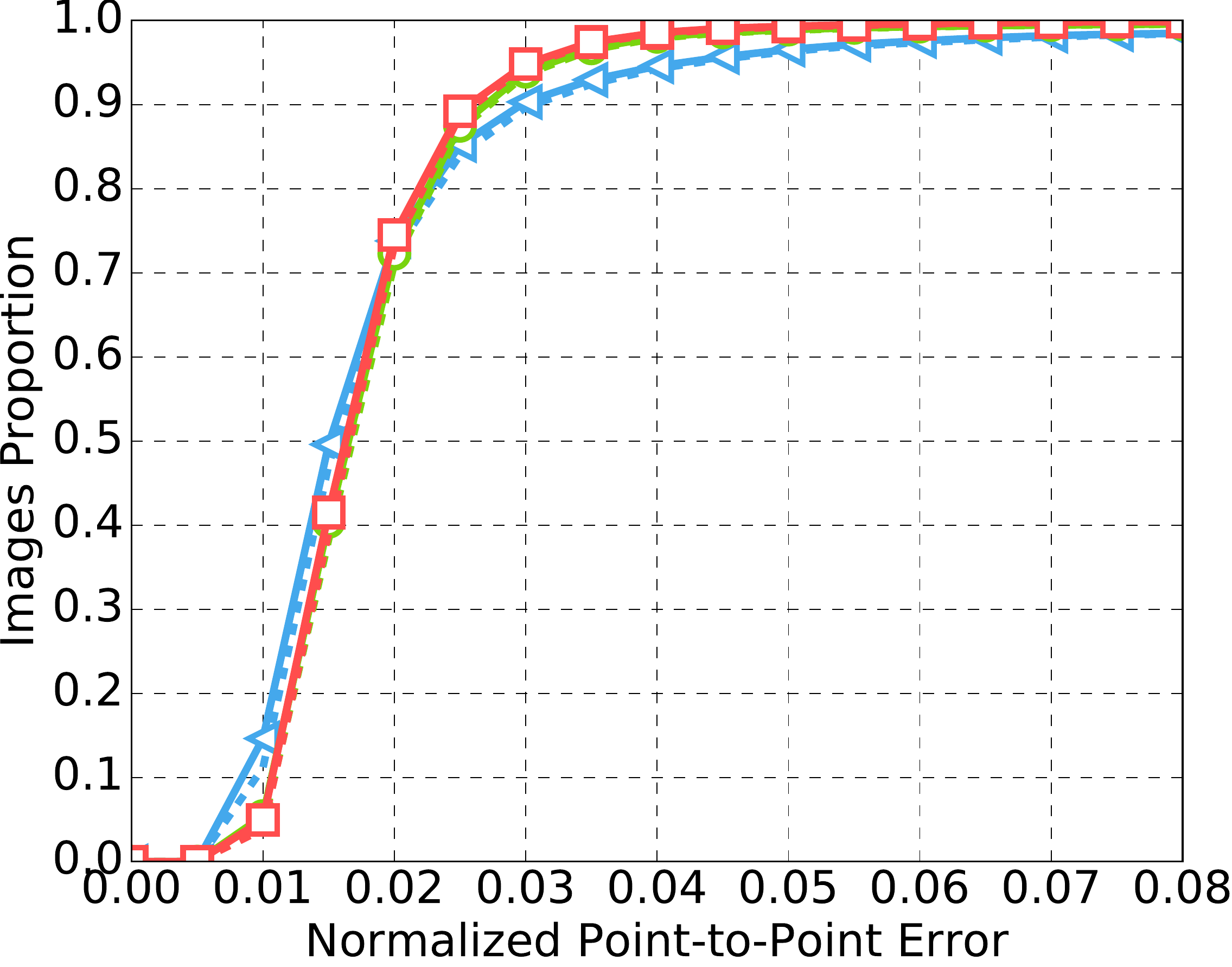}
   \end{minipage}
   }
\subfloat[][Category 3]{
   \begin{minipage}{0.323\linewidth}
   \centering
   \includegraphics[width=\linewidth]{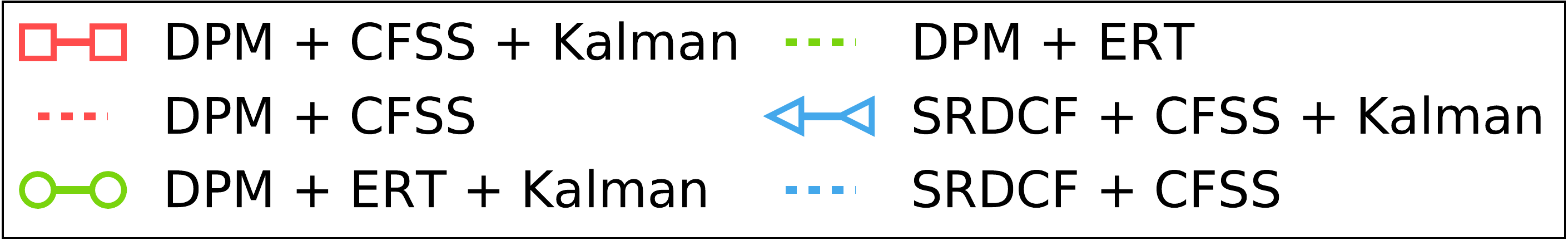}\\
   \includegraphics[width=\linewidth]{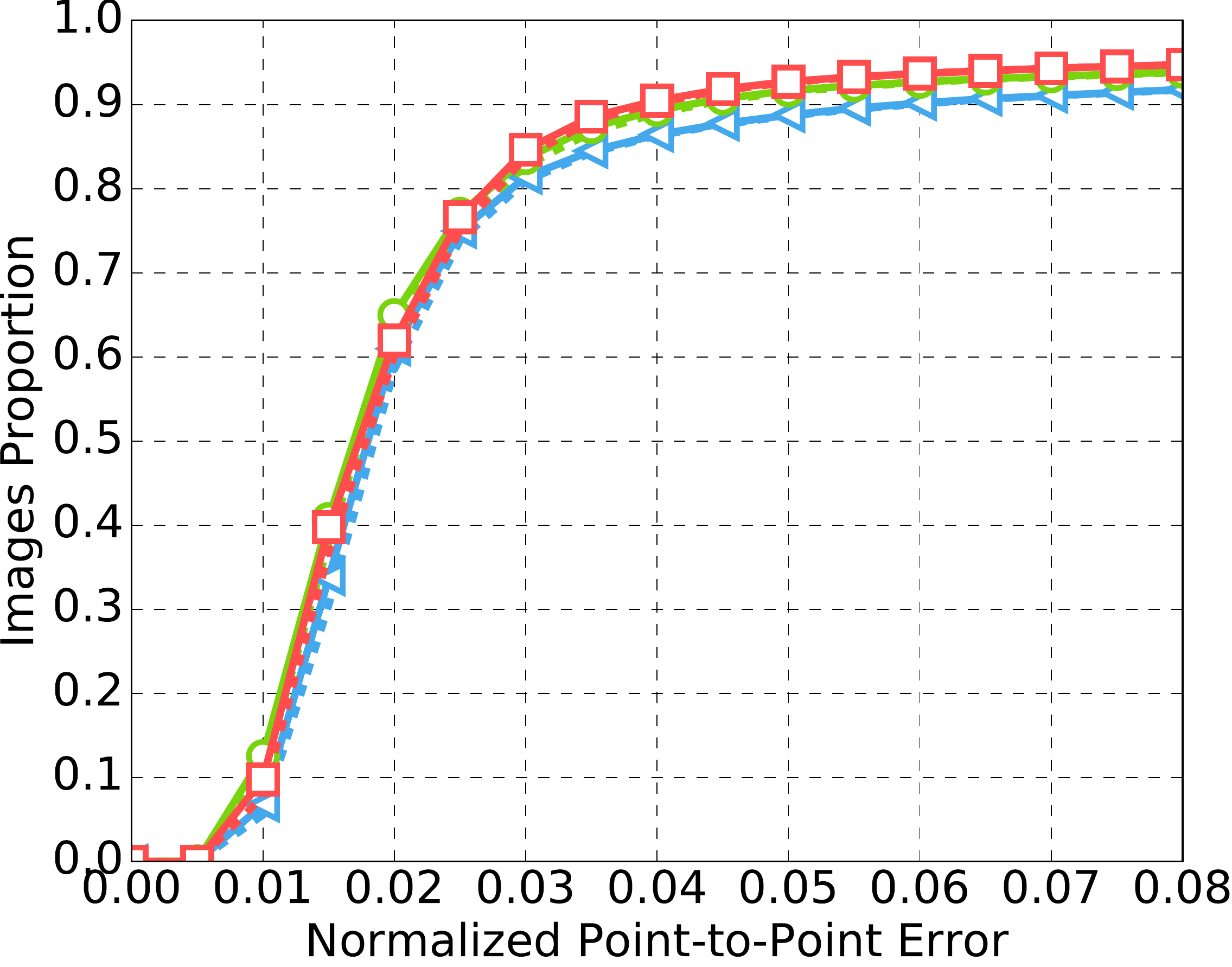}
   \end{minipage}
}
\caption{Results for Experiment 5 of Section~\ref{exp:kalman} (Kalman Smoothing). These results show the effect of Kalman smoothing on the final landmark localisation results. The top 3 performing results are given in red, green and blue, respectively, and the top 3 most improved are given in cyan, yellow and brown, respectively. The dashed lines represent the results before the smoothing is applied, solid lines are after.}
\label{fig:exp_kalman_difference}
\end{figure*}
%%%%%%%%%%%%%%%%%%%%%%%%%%%%%%%%%%

\subsection{Experiment 5: Kalman Smoothing}\label{exp:kalman}
In this section, we report the effect of performing Kalman Smoothing (\cite{kalman1960new}) on the results of the detectors of Section~\ref{exp:detection} and the trackers of Section~\ref{exp:tracking}. This experiment is designed to highlight the stability of the current landmark localisation methods with respect to noisy movement between frames (or jittering as it often known). However, when attempting to smooth the trajectories of the tracked bounding boxes themselves, we found an extremely negative effect on the results. Therefore, to remove jitter from the results we perform Kalman smoothing on the landmarks themselves. To robustly smooth the landmark trajectories, a generic facial shape model is constructed in a similar manner as described in the AAM literature by \cite{cootes2001active}.
Specifically, given the sparse shape of the face consisting of $n$ landmark points, we denote the coordinates
of the $i$-th landmark point within the Cartesian space of the image $\mathbf{I}$ as $\mathbf{x}_i=[x_i,y_i]^T$. Then a \emph{shape instance} of the face is
given by the $2n\times 1$ vector
$\mathbf{s} = \left[\mathbf{x}_1^T,\ldots,\mathbf{x}_n^T\right]^T = \left[x_1,y_1,\ldots,x_n,y_n\right]^T$.
Given a set of $N$ such shape samples $\{\mathbf{s}^1,\ldots,\mathbf{s}^N\}$,
a parametric statistical subspace of the object's shape variance can be retrieved
by first applying Generalised Procrustes Analysis on the shapes to normalise
them with respect to the global similarity transform (i.e., scale, in-plane
rotation and translation) and then using Principal Component Analysis (PCA).
The resulting \emph{shape model}, denoted as $\{\mathbf{U}_s,\bar{\mathbf{s}}\}$, consists of the orthonormal basis
$\mathbf{U}_s\in\mathbb{R}^{2n\times n_s}$ with $n_s$ eigenvectors
and the mean shape vector $\bar{\mathbf{s}}\in\mathbb{R}^{2n}$. This parametric
model can be used to generate new shape instances as
$\mathbf{s}(\mathbf{p})=\bar{\mathbf{s}} + \mathbf{U}_s\mathbf{p}$
where $\mathbf{p}=[p_1,\ldots,p_{n_s}]^T$ is the $n_s\times1$ vector of
\emph{shape parameters} that control the linear combination of the eigenvectors.
The Kalman smoothing is thus learnt via Expectation-Maximisation (EM) for the 
parameters $\mathbf{p}$ of each shape within a sequence.

The results of this experiment are given in Table~\ref{tab:exp_kalman} and Figure~\ref{fig:exp_kalman}. These experiments also provide a direct comparison between the best detection and model free tracking based techniques. For the videos of categories 1 and 3, the Kalman smoothing applied on DPM followed by a discriminative landmark localisation method (CFSS, ERT) outperforms all the combinations that involve model free rigid tracking. The combination of SRDCF with CFSS with Kalman smoothing achieves the best performance for Category 2.

In order to better investigate the effect of the smoothing, we also provide Figure~\ref{fig:exp_kalman_difference} which shows the differences between the initial tracking results and the results after applying Kalman smoothing. This comparison is shown for the best methods of Table~\ref{tab:exp_kalman}. It becomes obvious that the improvement introduced by Kalman smoothing is marginal.

%%%%%%%%%%%%%%%%%%%%%%%%%%%%%
%%%% [TAB]: COMPETITION %%%%%
%%%%%%%%%%%%%%%%%%%%%%%%%%%%%
\begin{table*}[!t]
\centering
\begin{tabular}{c r cc r cc r cc}
\toprule
   \multirow{3}{*}{Method} & & \multicolumn{2}{c}{Category 1} & & \multicolumn{2}{c}{Category 2} & & \multicolumn{2}{c}{Category 3} \\
   \cmidrule(lr){3-4}\cmidrule(lr){6-7}\cmidrule(lr){9-10}
   & & \multirow{2}{*}{\emph{AUC}} & \emph{Failure} & & \multirow{2}{*}{\emph{AUC}} & \emph{Failure} & & \multirow{2}{*}{\emph{AUC}} & \emph{Failure} \\
   & & & \emph{Rate (\%)} & & & \emph{Rate (\%)} & & & \emph{Rate (\%)} \\
   \cmidrule[\heavyrulewidth](){1-1}\cmidrule[\heavyrulewidth](){3-10}
   DPM + ERT + Kalman    & & \cellcolor{colour2}\textbf{0.775} & \cellcolor{colour2}\textbf{3.472} & & \cellcolor{colour5}\textbf{0.770} & \cellcolor{colour5}\textbf{1.527} & & \cellcolor{colour2}\textbf{0.719} &  \cellcolor{colour2}\textbf{6.111} \\
   DPM + ERT + previous  & & \cellcolor{colour3}\textbf{0.771} & \cellcolor{colour3}\textbf{3.262} & & 0.764 & 1.205 & & \cellcolor{colour3}\textbf{0.714} & \cellcolor{colour3}\textbf{5.692} \\
   DPM + CFSS + Kalman   & & \cellcolor{colour4}\textbf{0.764} & \cellcolor{colour4}\textbf{3.784} & & 0.767 & 1.326 & & \cellcolor{colour1}\textbf{0.721} & \cellcolor{colour1}\textbf{5.255} \\
   SRDCF + CFSS + Kalman & & 0.732 &  6.847 & & \cellcolor{colour2}\textbf{0.780} & \cellcolor{colour2}\textbf{0.131} & & 0.690 &  8.206 \\
   SRDCF + CFSS          & & 0.729 &  6.849 & & \cellcolor{colour3}\textbf{0.777} & \cellcolor{colour3}\textbf{0.167} & & 0.684 &  8.242 \\
   \cmidrule(lr){1-1}\cmidrule(lr){3-4}\cmidrule(lr){6-7}\cmidrule(lr){9-10}
   \cite{yang2015facial} & & \cellcolor{colour1}\textbf{0.791} & \cellcolor{colour1}\textbf{2.400} & & \cellcolor{colour1}\textbf{0.788} & \cellcolor{colour1}\textbf{0.322} & & \cellcolor{colour4}\textbf{0.710} & \cellcolor{colour4}\textbf{4.461} \\
   \cite{uricar2015real} & & 0.657 &  7.622 & & 0.677 & 4.131 & & 0.574 &  7.957 \\
   \cite{xiao2015facial} & & \cellcolor{colour5}\textbf{0.760} & \cellcolor{colour5}\textbf{5.899} & & \cellcolor{colour4}\textbf{0.782} & \cellcolor{colour4}\textbf{3.845} & & \cellcolor{colour5}\textbf{0.695} & \cellcolor{colour5}\textbf{7.379} \\
   \cite{rajamanoharan2015multi} & & 0.735 &  6.557 & & 0.717 & 3.906 & & 0.659 &  8.289 \\
   \cite{yue2015shape} & & 0.674 & 13.925 & & 0.732 & 5.601 & & 0.602 & 13.161 \\
\midrule[\heavyrulewidth]
   \multicolumn{10}{l}{\scriptsize Colouring denotes the methods' performance ranking per category:\hspace{0.2cm}$\color{colour1}\blacksquare$~first\hspace{0.2cm}$\color{colour2}\blacksquare$~second\hspace{0.2cm}$\color{colour3}\blacksquare$~third\hspace{0.2cm}$\color{colour4}\blacksquare$~fourth\hspace{0.2cm}$\color{colour5}\blacksquare$~fifth}\\
\bottomrule
\end{tabular}
\caption{Comparison between the best methods of Sections~\ref{exp:detection}-\ref{exp:kalman} and the participants of the 300VW challenge by \cite{shen2015first}. The Area Under the Curve (AUC) and Failure Rate are reported. The top 5 performing curves are highlighted for each video category.}
\label{tab:exp_competition}
\end{table*}
%%%%%%%%%%%%%%%%%%%%%%%%%%%%%
%%%%%%%%%%%%%%%%%%%%%%%%%%%%%
%%%% [FIG]: COMPETITION %%%%%
%%%%%%%%%%%%%%%%%%%%%%%%%%%%%
\begin{figure*}[!t]
\subfloat[][Category 1]{
   \begin{minipage}{0.323\linewidth}
   \centering
   \includegraphics[height=0.86cm]{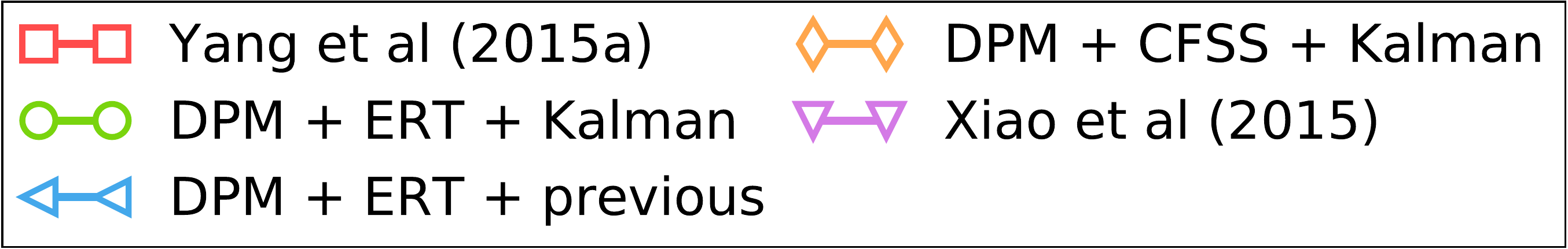}\\
   \includegraphics[width=\linewidth]{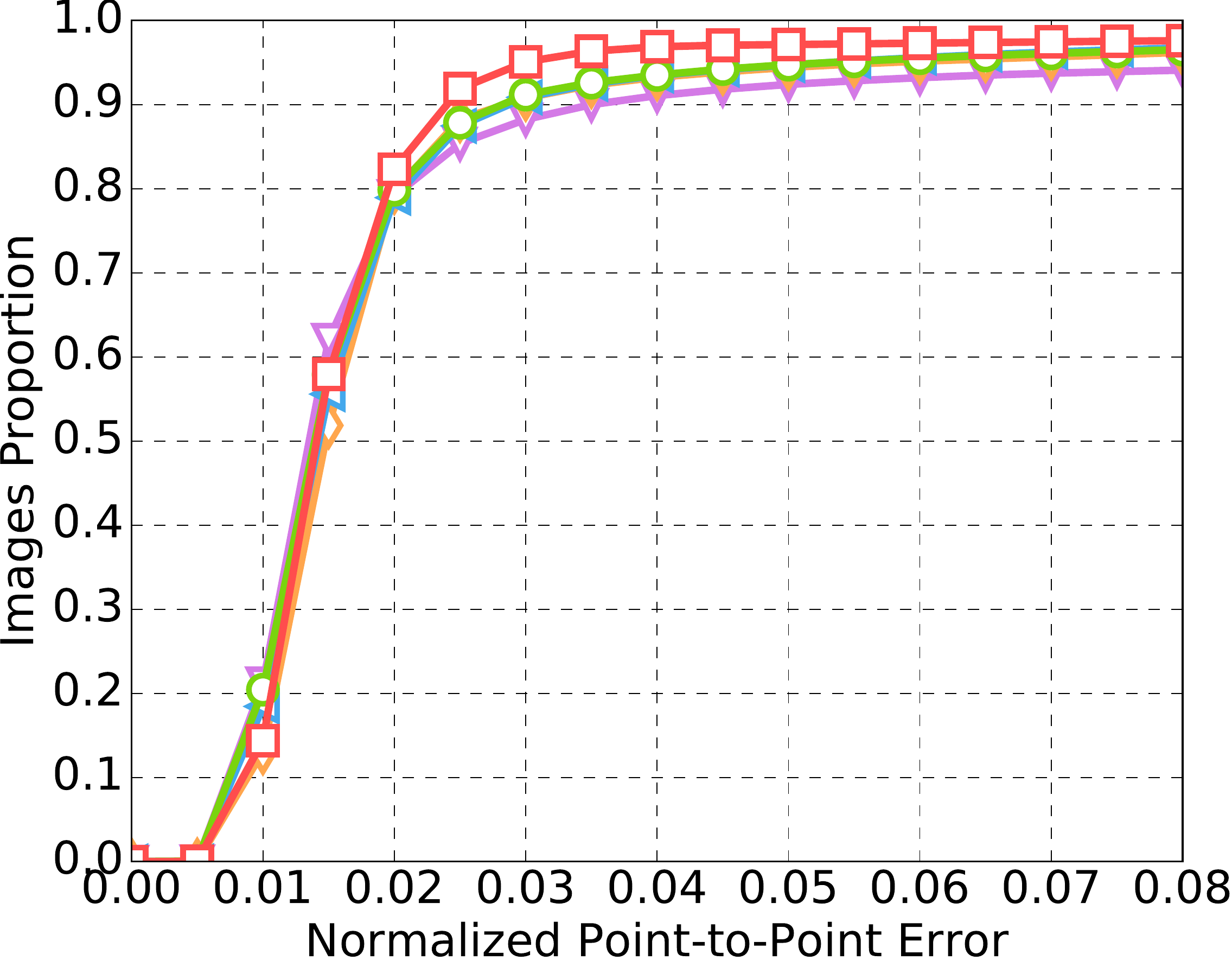}
   \end{minipage}
   }
\subfloat[][Category 2]{
   \begin{minipage}{0.323\linewidth}
   \centering
   \includegraphics[height=0.86cm]{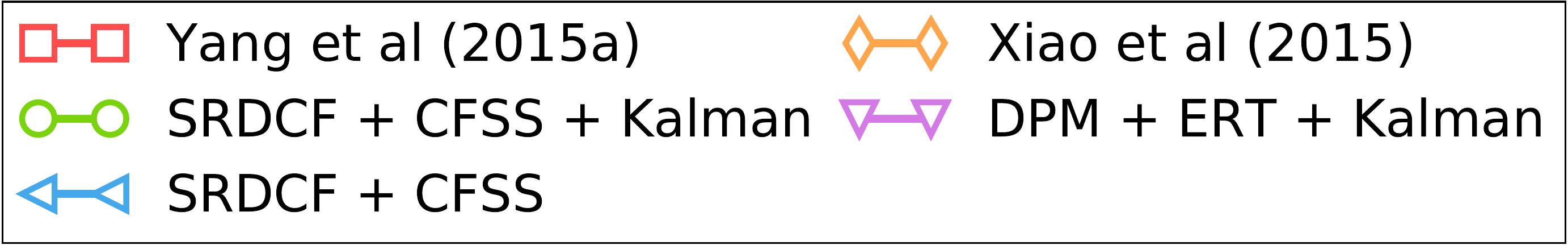}\\
   \includegraphics[width=\linewidth]{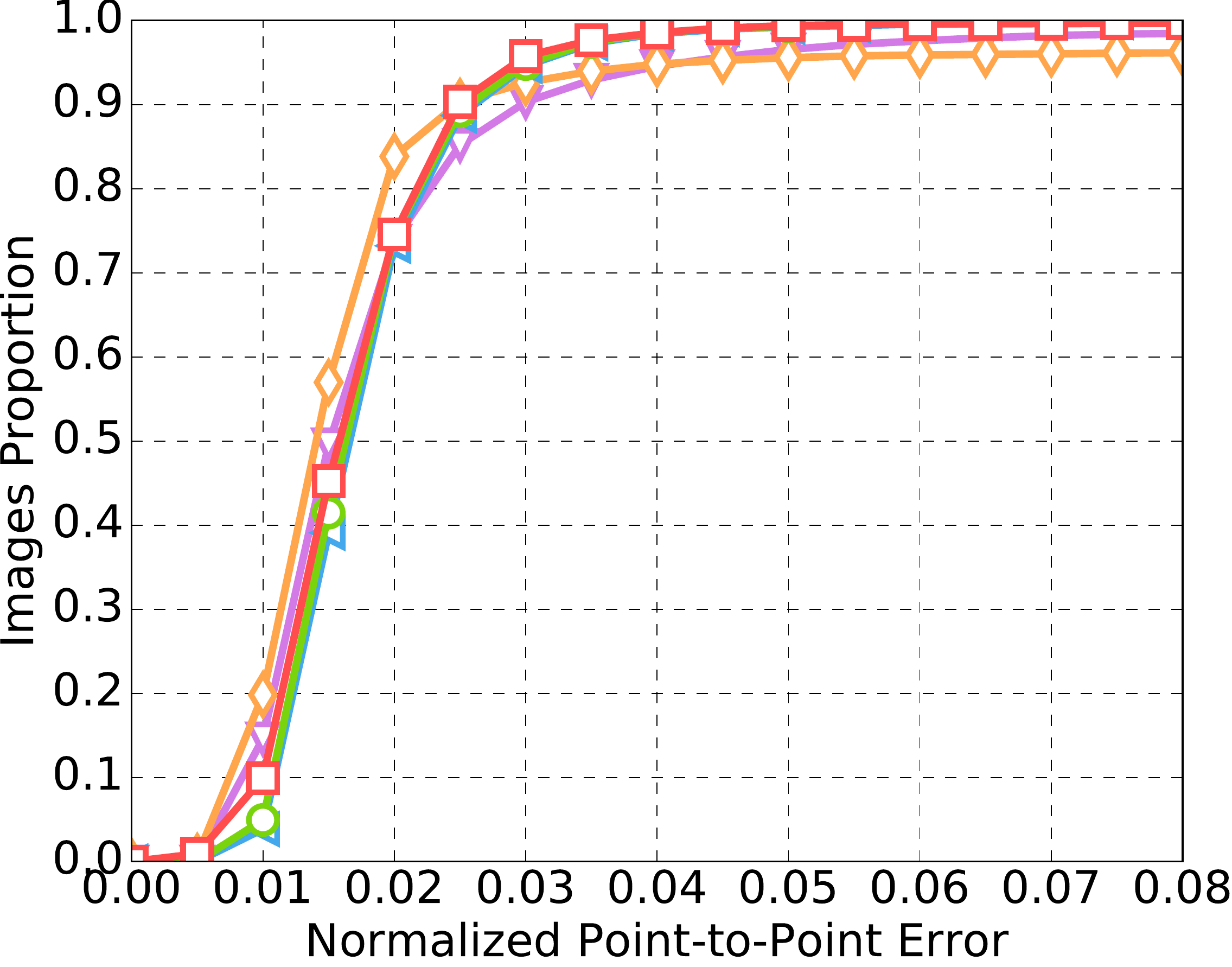}
   \end{minipage}
   }
\subfloat[][Category 3]{
   \begin{minipage}{0.323\linewidth}
   \centering
   \includegraphics[height=0.86cm]{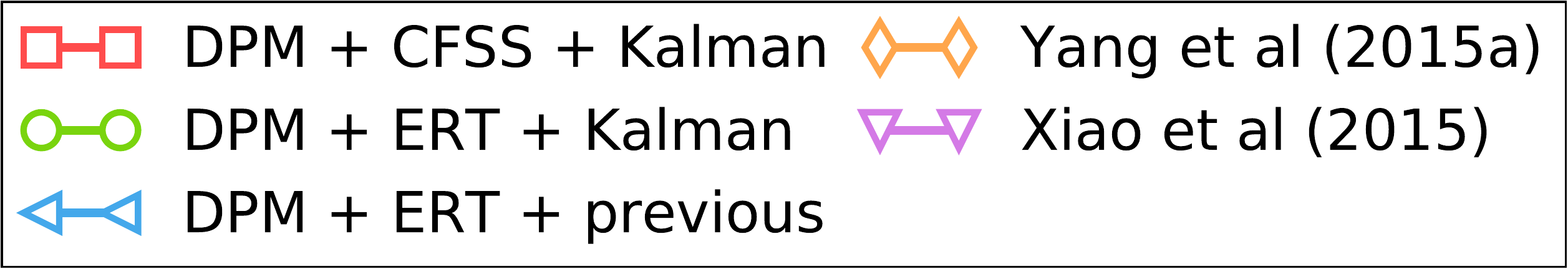}\\
   \includegraphics[width=\linewidth]{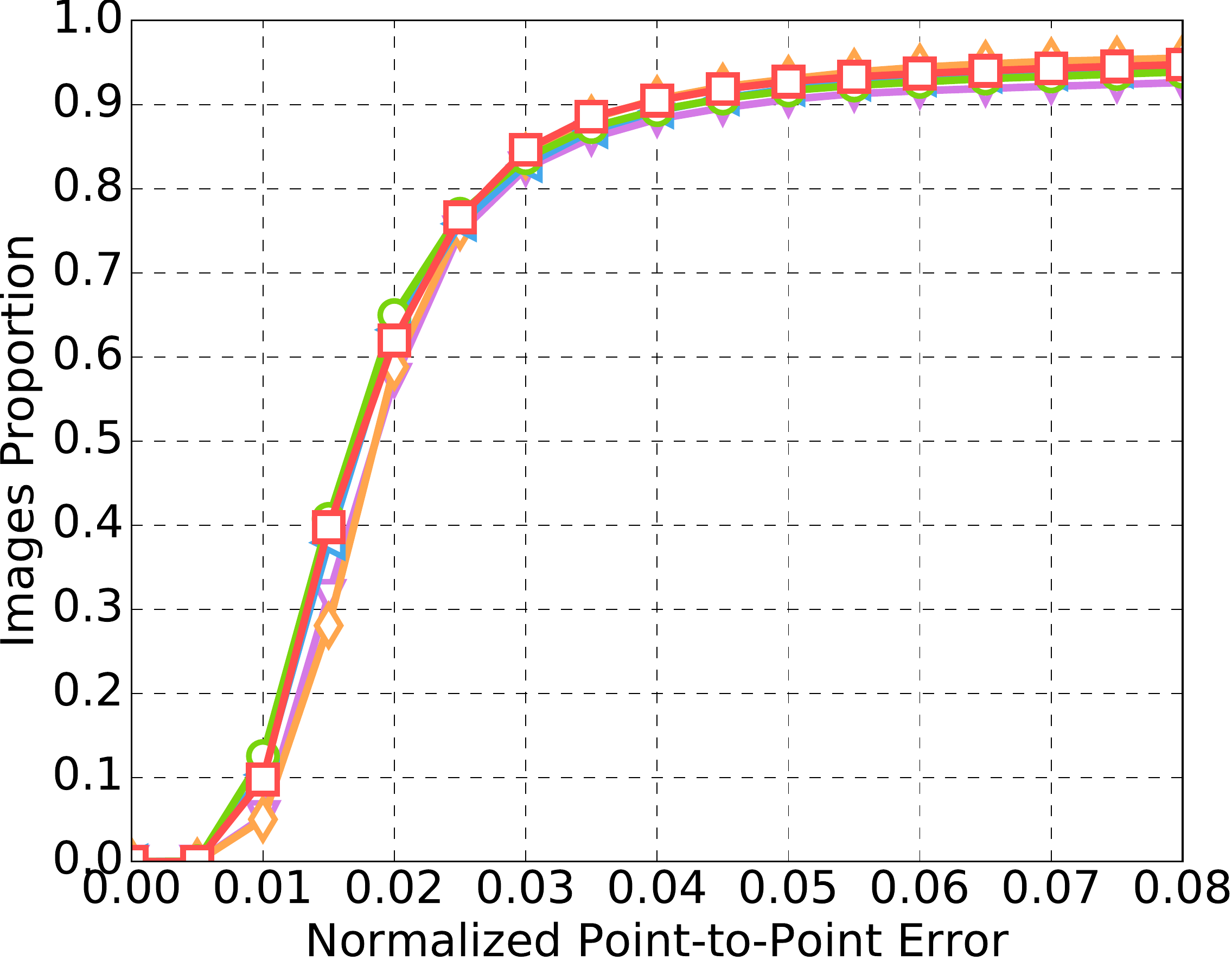}
   \end{minipage}
}
\caption{Comparison between the best methods of Sections~\ref{exp:detection}-\ref{exp:kalman} and the participants of the 300VW challenge by \cite{shen2015first}. The top 5 methods are shown and are coloured red, blue, green, orange and purple, respectively. Please see Table~\ref{tab:exp_competition} for a full summary.}
\label{fig:exp_competition}
\end{figure*}
%%%%%%%%%%%%%%%%%%%%%%%%%%%%%
\subsection{300VW Comparison}\label{exp:competition}
In this section we provide results that compare the best performing methods of the previous sections (\ref{exp:detection}-\ref{exp:kalman}) to the participants of the 300VW challenge by \cite{shen2015first}. The challenge had 5 competitors. \cite{rajamanoharan2015multi} employ a multi-view Constrained Local Model (CLM) with a global shape model and different response maps per pose and explore shape-space clustering strategies to determine the optimal pose-specific CLM. \cite{uricar2015real} apply a DPM at each frame as well as Kalman smoothing on the face positions. \cite{yue2015shape} utilise a shape augmented regression model, where the regression function is automatically selected based on the facial shape. \cite{xiao2015facial} propose a multi-stage regression-based approach that progressively provides initialisations for ambiguous landmarks such as boundary and eyebrows, based on landmarks with semantically strong meaning such as eyes and mouth corners. Finally, \cite{yang2015facial} employ a multi-view spatio-temporal cascade shape regression model along with a novel reinitialisation mechanism.

The results are summarised in Table~\ref{tab:exp_competition} and Figure~\ref{fig:exp_competition}. Note that the error metric considered in this paper (as described in Section~\ref{subsubsec:metrics}) differs from that of the original competition. This was intended to improve the robustness of the results with respect to variation in pose. Also, as noted in Section~\ref{sec:exp_implementation}, the 300VW annotations have been corrected and thus this experiment represents updated results for the 300VW competitors. The results indicate that \cite{yang2015facial} outperform the rest of the methods for the videos of Categories 1 and 2, whereas a weakly supervised DPM combined with CFSS and Kalman smoothing is the top performing for the challenging videos of Category 3. Moreover, it becomes evident that methodologies which employ face detection dominate Categories 1 and 3. Category 2 is dominated by approaches that utilise a model free tracker.

\section{Discussion and Conclusions}\label{sec:discussion}
In Section~\ref{sec:experiments} we presented a number of experiments on deformable tracking of sequences containing a single face. We investigated the performance of state-of-the-art face detectors and model free trackers on the recently released 300VW dataset\textsuperscript{\ref{300VW_foot}}. We also devised a number of hybrid systems that attempt to improve the performance of both detectors and trackers with respect to tracking failures. A summary of the proposed experiments are given in Table~\ref{tbl:experiments_summary}.

Overall, it appears that modern detectors are capable of handling videos of the complexity provided by the 300VW dataset. This supports the most commonly proposed deformable face tracking methodology that couples a detector with a landmark localisation algorithm. More interestingly, it appears that modern model free trackers are also highly capable of tracking videos that contain variations in pose, expression and illumination. This is particularly evident in the videos of Category 2 where the model free trackers perform the best. The performance on the videos of Category 2 is likely due to the decreased amount of pose variation in comparison to the other two categories. Category 2 contains many illumination variations which model free trackers appear invariant to. Our work also supports the most recent model free tracking benchmarks (\cite{Kristan2015a} and \cite{wu2015object}) which have demonstrated that DCF-based trackers are currently the most competitive. However, the performance of the trackers does deteriorate significantly in Category 3 which supports the categorisation of these videos in the 300VW as the most difficult category. The difficulty in the videos of Category 3 largely stems from the amount of pose variation present, which both detectors and model free trackers struggle with.

The DPM detector provided by \cite{mathias2014face} is very robust across a variety of poses and illumination conditions. Overall, it outperformed the other methods by a fairly significant margin, particularly when failure rate is considered. Even in the most challenging videos of Category 3, the failure rate of DPM is only approximately $5\%$, which is over $50\%$ less than the next best performing method, SRDCF, at $8\%$. The CFSS landmark localisation method of \cite{zhu2015face} outperforms all other considered landmark localisation methods, although the random forest based ERT method of \cite{kazemi2014one} also performed very well. The difference between CFSS and SDM supports the findings of \cite{zhu2015face} as the videos contain very challenging pose variations. 

The stable performance of both the best model free trackers and detectors on these videos is further demonstrated by the minimal improvement gained from the proposed hybrid systems. Neither reinitialisation from the previous frame (Section~\ref{exp:detection_init_from_previous}), nor the failure detection methodology proposed (Section~\ref{exp:tracking_restart}) improved the best performing methods with any significance. Furthermore, Kalman smoothing the facial shapes across the sequences also had a very minimal positive improvement.

In comparison to the recent results of the 300VW competition (\cite{shen2015first}), our review of combinations of modern state-of-the-art detectors and trackers found that very strong performance can be obtained through fairly simple deformable tracking schemes. In fact, only the work of \cite{yang2015facial} outperforms our best performing method and the difference shown by Figure~\ref{fig:exp_competition} appears to be marginal, particular in Category 3. However, the overall results show that, particularly for videos that contain significant pose, there are still improvements to be made.

To summarise, there are a number of important issues that must be tackled in order to improve deformable face tracking:
\begin{enumerate}
    \item Pose is still a challenging issue for landmark localisation methods. In fact, the videos of 300VW do not even exhibit the full range of possible facial pose as they do not contain profile faces. The challenges of considering profile faces have yet to be adequately addressed and have not be verified with respect to current state-of-the-art benchmarks.
    \item In this work, we only consider videos that contain a single visible face. However, there are many scenarios in which multiple faces may be present and this represents further challenges to deformable tracking. Detectors for example, are particularly vulnerable to multi-object tracking scenarios as they require extending with the ability to determine whether the object being localised is the same as in the previous frame.
    \item It is very common for objects to leave the frame of the camera during a sequence, and then reappear. Few model free trackers are robust to reinitialisation after an object has disappeared and then reappeared. When combined with multiple objects, this scenario becomes particularly challenging as it requires a re-identification step in order to verify whether the object to be tracked is one that was seen before.
\end{enumerate}

We believe that deformable face tracking is a very exciting line of research and future advances on the field can have an important impact on several areas of Computer Vision.

\bibliographystyle{spbasic}
\bibliography{egbib.bib}

\end{document}